\ifdefined\pdfminorversion
\fi

\documentclass[
  12pt,
  a4paper,
  twoside,
  openright,
  BCOR=12mm,        
  DIV=16,           
  bibliography=totoc,
  listof=totoc,
  headings=normal,
  chapterprefix=true,   
  appendixprefix=true,
]{scrbook}

\newcommand{\thesisTitle}{Learning Reachability-Based Safety Representations for Real-Time Robot Navigation in Unknown and Dynamic Environments}
\newcommand{\thesisAuthor}{Bojan Derajić}
\newcommand{\thesisAuthorDegree}{M.\,Sc.}     
\newcommand{\thesisORCID}{0009-0000-6175-535X}                    

\newcommand{\thesisFacultyEN}{Faculty IV -- Electrical Engineering and Computer Science}

\newcommand{\thesisDoctoralDegreeEN}{Doctor of Engineering Sciences}
\newcommand{\thesisDoctoralDegreeAbbrev}{Dr.-Ing.}
\newcommand{\thesisYear}{2026}
\newcommand{\thesisKeywords}{safe robot navigation, Hamilton--Jacobi reachability, model predictive control, control barrier functions, neural safety representations}

\usepackage[T1]{fontenc}       
\usepackage[utf8]{inputenc}
\usepackage[main=english, ngerman]{babel}

\usepackage{lmodern}
\usepackage{microtype}

\usepackage{silence} 
\usepackage{amsmath}
\usepackage{amssymb}
\usepackage{mathtools}

\usepackage{graphicx}
\usepackage{subcaption}
\graphicspath{{figures/}}

\usepackage{tikz}
\usetikzlibrary{positioning, arrows.meta, fit, decorations.markings, backgrounds, calc}

\usepackage{booktabs}
\usepackage{array}
\usepackage{longtable}
\usepackage{multirow}
\newcolumntype{R}[1]{>{\raggedright\arraybackslash}p{#1}}

\usepackage{siunitx}

\usepackage[
  backend=biber,
  style=numeric,
  sorting=none,
  maxbibnames=999,       
  minbibnames=999,
  doi=true,
  isbn=false,
  url=false,
  eprint=true,        
]{biblatex}
\appto{\bibsetup}{\emergencystretch=3em}
\usepackage{csquotes}

\usepackage[
  bookmarks=true,
  bookmarksopen=true,
  bookmarksopenlevel=1,
  pdfpagelayout=SinglePage,
  colorlinks=true,
  linkcolor=black,
  citecolor=black,
  urlcolor=black,
]{hyperref}
\usepackage{hyperxmp}
\usepackage{bookmark}   

\makeatletter
\renewcommand*{\@pnumwidth}{2.2em}  
\makeatother

\usepackage[capitalise, noabbrev]{cleveref}
\hypersetup{
  pdftitle    = {\thesisTitle},
  pdfauthor   = {\thesisAuthor},
  pdfkeywords = {\thesisKeywords},
  pdflang     = {en},
}

\begin{document}

\frontmatter


\begin{titlepage}
  \centering
  \null\vfill

  {\LARGE\bfseries \thesisTitle \par}

  \vfill

  submitted by\\[0.6em]
  \thesisAuthor, \thesisAuthorDegree
  \ifx\thesisORCID\empty\else\\[0.3em]ORCID: \thesisORCID\fi

  \vspace{2em}

  to \thesisFacultyEN\\
  of Technische Universität Berlin\\
  in fulfillment of the requirements for the academic degree of\\[0.6em]
  \thesisDoctoralDegreeEN\\
  --\,\thesisDoctoralDegreeAbbrev\,--\\[1.2em]
  submitted dissertation

  \vfill


  \vfill

  Berlin \thesisYear

\end{titlepage}











\cleardoublepage

\addchap{Zusammenfassung}

Für eine sichere Roboternavigation in unbekannten und dynamischen Umgebungen reicht eine rein geometrische Kollisionsprüfung nicht aus. Ein Roboter kann noch Abstand zu einem Hindernis haben und dennoch eine Kollision nicht mehr vermeiden können, weil er zu schnell fährt, ungünstig ausgerichtet oder einem bewegten Objekt zu nahe ist. Ob ein Zustand eine sichere Fortsetzung erlaubt, hängt daher von der Roboterdynamik, den Stellgrenzen, dem Wahrnehmungshorizont und der Bewegung der Hindernisse ab und nicht allein vom Abstand zu den Hindernissen. Die Erreichbarkeitsanalyse nach Hamilton-Jacobi (HJ) beantwortet diese Frage mit einer Wertfunktion, deren Vorzeichen Zustände mit einer sicheren Fortsetzung von jenen trennt, in denen eine Kollision unter den angenommenen Modellen unvermeidbar ist. Die Berechnung dieser Wertfunktion ist jedoch zu aufwendig, um sie online für jede neu beobachtete lokale Umgebung zu wiederholen.

Diese Dissertation entwickelt neuronale implizite Repräsentationen, die offline auf Erreichbarkeitslösungen trainiert werden und die gewonnene Sicherheitsinformation optimierungsbasierten Reglern in Echtzeit zugänglich machen. Zur Laufzeit bildet das trainierte Modell die aktuelle Beobachtung oder die Menge der erfassten Hindernisse auf eine differenzierbare Sicherheitsfunktion ab, die mit der Abtastrate des Reglers ausgewertet und differenziert werden kann. Für die hier betrachteten Vermeidungsprobleme übersteigt die Wertfunktion den momentanen Wert der Nebenbedingung nie, sodass sie als Differenz aus einer direkt berechneten Nebenbedingung, etwa einer Signed-Distance-Funktion, und einer nichtnegativen Korrektur dargestellt werden kann, die die Roboterdynamik berücksichtigt. Die hier entwickelten residualen Repräsentationen lernen nur diese Korrektur, wodurch die vorhergesagte sichere Menge unabhängig vom Approximationsfehler nie in die repräsentierten Hindernisse hineinreicht.

Auf dieser Grundlage werden fünf Methoden entwickelt. Die ersten vier bilden zwei Paare und beziehen ihre Trainingsdaten aus numerisch gelösten HJ-Problemen. Das erste Paar nutzt gelernte Erreichbarkeitsinformationen als terminale Nebenbedingung in der modellprädiktiven Regelung (MPC), für lokal beobachtete statische Umgebungen und für vorhergesagte Hindernisbewegungen. Das zweite Paar nutzt gelernte Erreichbarkeitsfunktionen als Barrierekandidaten in Sicherheitsfiltern auf Basis von Control Barrier Functions (CBFs), für lokal beobachtete statische Umgebungen und, über paarweise in Relativkoordinaten kombinierte Barrieren, für eine variable Anzahl bewegter Hindernisse. In den untersuchten Szenarien zeigen Simulationen und Hardwareexperimente auf Boden- und Flugrobotern eine bessere Kollisionsvermeidung als geometrische und handentworfene Vergleichsverfahren, insbesondere bei kurzen Prädiktionshorizonten und bewegten Hindernissen.

Die fünfte Methode betrifft die Offline-Phase selbst, deren gitterbasierter Löser die ersten vier Methoden auf niedrigdimensionale Robotermodelle beschränkt. Für steuerungsaffine Dynamiken und eine polytopische Stellgrößenmenge wird der HJ-Hamiltonian in einer Ecke maximiert. Darauf aufbauend approximiert eine endliche Baumsuche über Stellgrößen an den Ecken der zulässigen Menge die Sicherheitswerte und lässt sich auf moderner Hardware effizient parallelisieren. Die dabei gefundenen Sicherheitswerte dienen als Trainingsdaten. Zusammen mit einem Verlustterm aus der HJ-Variationsungleichung ersetzen sie den gitterbasierten Löser und liefern Barrierekandidaten für Systeme mit bis zu 13 Zuständen. Insgesamt machen die fünf Methoden dynamikbewusste, erreichbarkeitsbasierte Sicherheitsinformationen für die Echtzeitnavigation von Robotern in unbekannten und dynamischen Umgebungen praktisch nutzbar.

\cleardoublepage

\addchap{Abstract}

Safe robot navigation in unknown and dynamic environments requires more than geometric collision checking. A robot may still be at a distance from an obstacle yet be unable to avoid a collision because of its speed, orientation, or proximity to a moving object. A safe continuation therefore depends on the robot dynamics, actuation limits, sensing horizon, and obstacle motion. Hamilton--Jacobi (HJ) reachability analysis encodes this information in a value function whose sign separates states that admit a safe continuation from those in which collision is unavoidable under the assumed models. However, computing this function is too expensive to repeat online for every newly observed local environment.

This dissertation develops neural implicit representations trained offline on reachability solutions to provide safety information to optimization-based controllers in real time. At runtime, the model maps the current observation or detected obstacles to a differentiable safety function. For the avoidance problems considered here, the value function never exceeds the instantaneous constraint value, so it can be written as a directly computed constraint, such as a signed-distance function, minus a nonnegative correction that accounts for the robot dynamics. The residual representations developed here learn only this correction, which excludes represented obstacles from the predicted safe set regardless of approximation error.

Five methods are developed on this basis. The first four form two pairs and take their training data from numerically solved HJ problems. The first pair uses learned reachability information as a terminal constraint in model predictive control (MPC), for locally observed static environments and for predicted obstacle motion. The second pair uses learned reachability functions as barrier candidates in safety filters based on control barrier functions (CBFs), for locally observed static environments and, through pairwise barriers composed in relative coordinates, for a varying number of moving obstacles. In the evaluated scenarios, simulations and hardware experiments on ground and aerial robots show better collision avoidance than geometric and hand-designed baselines, in particular for short prediction horizons and moving obstacles.

The fifth method addresses the offline phase itself, whose grid-based solver restricts the first four methods to low-dimensional robot models. For control-affine dynamics and a polytopic input set, the HJ Hamiltonian is maximized at a vertex. Building on this property, a finite tree search over vertex inputs approximates safety values and can be efficiently parallelized on modern hardware. The safety values found by this search serve as training labels. Together with a loss derived from the HJ variational inequality, they replace the grid-based solver and yield barrier candidates for systems with up to 13 states. Taken together, the five methods make dynamics-aware, reachability-based safety information practical for real-time robot navigation in unknown and dynamic environments.

\cleardoublepage

\addchap{Acknowledgments}

\begingroup
\setlength{\parindent}{0pt}
\setlength{\parskip}{0.8\baselineskip}

I would like to thank my supervisor, Prof.\ Wolfgang Hönig, for his guidance and support. He was always kind, patient, and approachable, and I particularly appreciated his availability and responsiveness, even outside regular working hours. I am grateful for his trust, his openness to my ideas, and the many discussions that helped shape this dissertation.

I am also very grateful to Prof.\ Sebastian Bernhard, who mentored me during my time in industry and has been involved in this research from the very beginning. He offered guidance and encouragement at every stage, from the initial ideas to the final revisions. I greatly appreciate his steady support and the time and energy he invested in our collaboration.

I am thankful to my family and friends for their patience, encouragement, and support throughout these years. They were there for me during the more demanding periods, and spending time with them helped me recharge and regain perspective.

Finally, I would like to thank my colleagues at the AUMOVIO AI Lab in Berlin. They made my time at the lab fun and interesting, and I learned from them about a wide range of topics, often far beyond my own research. I appreciated their different perspectives, as well as the thoughtful discussions and humor that made everyday work enjoyable.

\bigskip
\begin{flushright}
  Bojan Derajić\\
  Berlin, October 2026
\end{flushright}
\endgroup

\cleardoublepage

\tableofcontents
\listoffigures
\listoftables

\mainmatter

\chapter{Introduction}
\label{ch:introduction}

\section{Motivation}
\label{sec:motivation}

Autonomous mobile robots increasingly operate in spaces shared with people and other agents. Self-driving cars use public roads, delivery robots navigate sidewalks and hospital corridors, and cleaning and intralogistics robots work alongside people in public buildings and warehouses. These applications promise economic and societal value, but they share the difficulty that the environment is not fully known at deployment and changes over time. Because mistakes in such environments can harm people or damage property, the motion planning and control systems of these robots need explicit safety mechanisms before they can be deployed.

\begin{figure}[tb]
  \centering
  \begin{subfigure}[t]{0.49\textwidth}
    \centering
    \includegraphics[width=\linewidth]{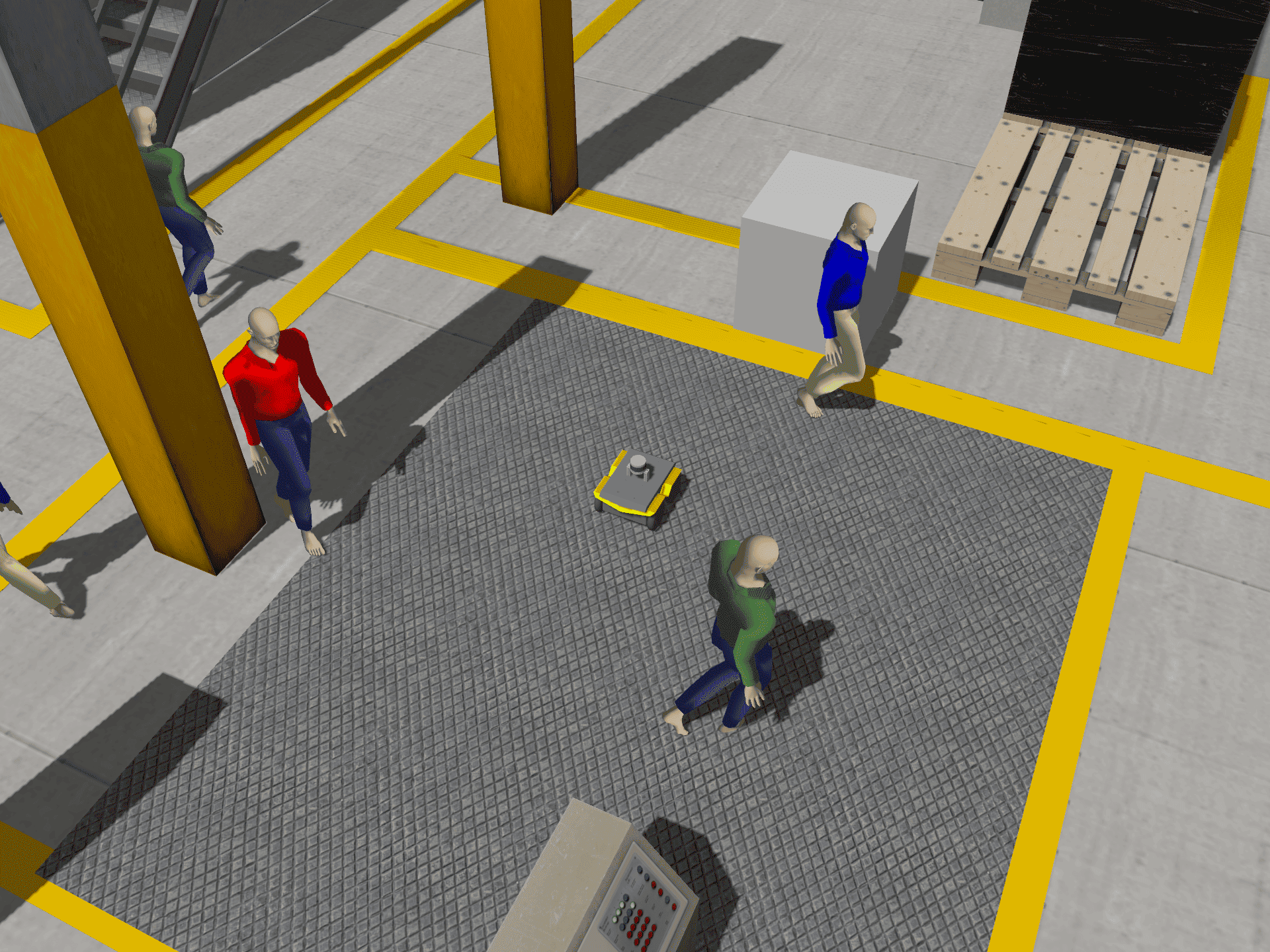}
    \caption{Sample scene from the Gazebo simulator.}
    \label{fig:intro-illustr-gazebo}
  \end{subfigure}\hfill
  \begin{subfigure}[t]{0.49\textwidth}
    \centering
    \includegraphics[width=\linewidth]{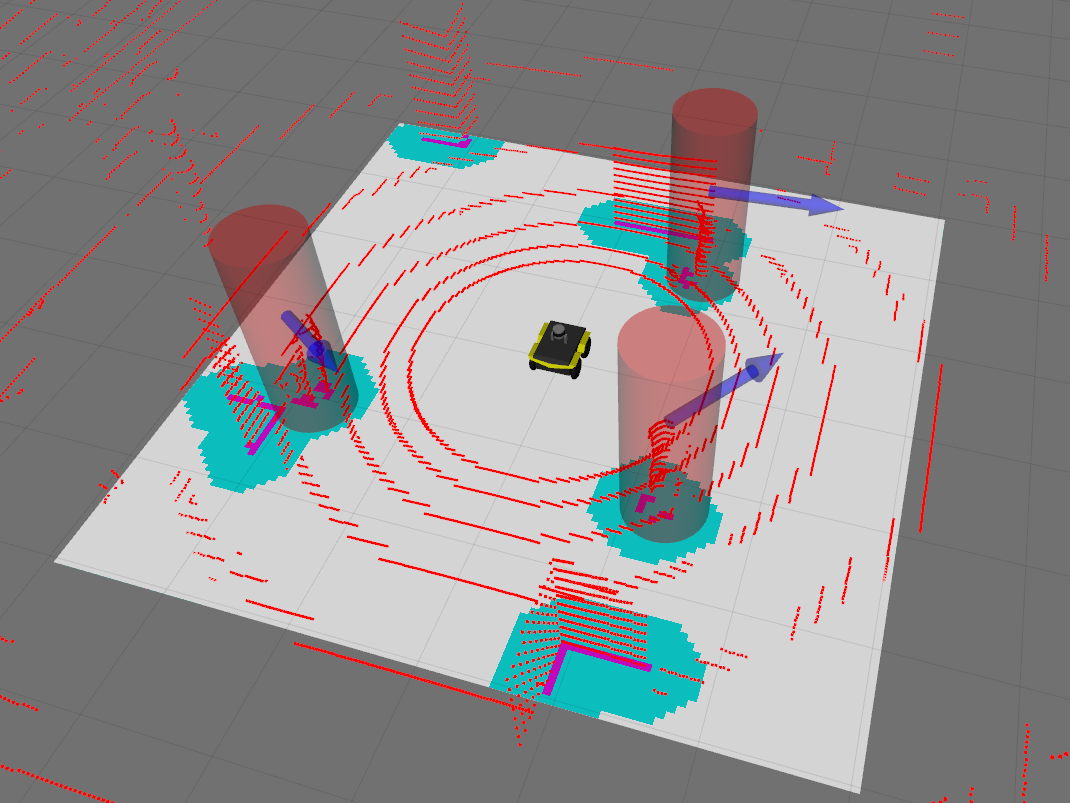}
    \caption{Visualized LiDAR point cloud, detected objects, and estimated agent motion.}
    \label{fig:intro-illustr-rviz}
  \end{subfigure}
  \caption[Mobile robot navigation in an unknown warehouse environment]{Mobile robot navigation in an unknown warehouse environment shared with other agents and perceived through local onboard sensing. The space outside the perception field is unknown, and the robot must react to new observations while maintaining safety margins under its dynamics and actuation limits.}
  \label{fig:intro-illustr}
\end{figure}

Safe operation in such environments is difficult because robots often perceive their surroundings only through onboard sensors, such as cameras or laser scanners, with a limited field of view. Free space and obstacles are discovered at runtime, other agents appear and change course, and each decision must be made within fractions of a second. The challenge is not only to complete the task, but to do so with motion that is physically executable, responsive to new observations, and safe despite incomplete knowledge of the environment \cite{hwang_safe_2024,bouraine_provably_2012}. \Cref{fig:intro-illustr} shows a representative example of a mobile robot navigating an unknown warehouse environment using only local onboard perception.

In robot navigation, safety is most commonly treated as \emph{geometric collision avoidance}, meaning that the robot body and the obstacles must not overlap at any time. The perceived environment is then modeled using geometric representations such as occupancy grids, which mark occupied map cells, or distance functions, which store the distance to the nearest obstacle, possibly together with predicted future positions of moving objects. These representations are intuitive, efficient, and accurate for the current geometry, but they do not describe how the situation can evolve. Whether a collision can still be avoided depends on three properties of the robot. The first is its \emph{state}, such as position, heading, and velocity. The second is its \emph{dynamics}, which govern how that state can change. The third is its \emph{actuation limits}, such as bounded forces, accelerations, and steering rates.

A familiar example is the braking distance of a car. A vehicle approaching a wall at high speed may still be several meters away from it and thus be collision-free in a geometric sense. Nevertheless, a collision may already be unavoidable because no admissible braking or steering maneuver can prevent it. States of this kind are known as \emph{inevitable collision states} \cite{fraichard_short_2007,bautin_inevitable_2010}. They arise for robots in narrow passages, at higher speeds, and near moving obstacles.

\begin{figure}[tb]
  \centering
  \begin{subfigure}[b]{0.48\textwidth}
    \centering
    \colorlet{brtline}{red!62!black}
\colorlet{brtfill}{red!9}
\colorlet{obsfill}{black!22}
\colorlet{obsline}{black!55}

\begin{tikzpicture}[
  font=\small, >={Latex[length=2mm]},
  lab/.style={font=\scriptsize, align=center},
  unsafe/.style={fill=brtfill, draw=brtline, semithick},
  ghost/.style={densely dashed, black!55, thin},
  traj/.style={semithick, postaction={decorate, decoration={markings,
              mark=at position #1 with {\arrow{Latex[length=1.8mm]}}}}},
  robot/.pic={
    \fill[black!70, rounded corners=0.8pt] (-0.21,0.17) rectangle (0.13,0.26);
    \fill[black!70, rounded corners=0.8pt] (-0.21,-0.26) rectangle (0.13,-0.17);
    \draw[semithick, fill=white, rounded corners=1.5pt] (-0.30,-0.19) rectangle (0.30,0.19);
    \fill[black!75] (0.21,0) -- (0.03,0.10) -- (0.03,-0.10) -- cycle;
  },
  ghostrobot/.pic={
    \fill[black!35, rounded corners=0.8pt] (-0.21,0.17) rectangle (0.13,0.26);
    \fill[black!35, rounded corners=0.8pt] (-0.21,-0.26) rectangle (0.13,-0.17);
    \draw[ghost, fill=white, rounded corners=1.5pt] (-0.30,-0.19) rectangle (0.30,0.19);
    \draw[ghost] (0.21,0) -- (0.03,0.10) -- (0.03,-0.10) -- cycle;
  },
  crash/.pic={
    \foreach \a in {15,60,105,150,195,240,285,330}{
      \draw[brtline, line width=0.6pt] (\a:0.07) -- (\a:0.19);}
    \fill[brtline] (0,0) circle (1.7pt);
  },
  ]
  \path[unsafe]
  (3.0,2.4)
  .. controls (3.0,2.9) and (3.6,3.48) .. (4.6,3.6)
  -- (6.25,3.6)
  arc[start angle=90, end angle=0, radius=0.3]
  -- (6.55,1.5)
  arc[start angle=0, end angle=-90, radius=0.3]
  -- (4.6,1.2)
  .. controls (3.6,1.32) and (3.0,1.9) .. cycle;
  \draw[fill=obsfill, draw=obsline, thin, rounded corners=1.5pt] (4.75,1.5) rectangle (6.25,3.3);
  \node[lab] at (5.5,2.4) {static\\obstacle};
  \pic at (0.75,2.4) {robot};
  \draw[black!50, thin] (0.05,2.52) -- (0.36,2.52);
  \draw[black!50, thin] (0.05,2.28) -- (0.36,2.28);
  \draw[traj=0.45, densely dashed] (1.05,2.4) -- (4.45,2.4);
  \pic at (4.45,2.4) {ghostrobot};
  \draw[traj=0.6, densely dashed] (3.2,2.4) .. controls (3.6,2.44) and (4.10,2.73) .. (4.44,3.04);
  \pic[rotate=45] at (4.44,3.04) {ghostrobot};
  \draw[traj=0.6, densely dashed] (3.2,2.4) .. controls (3.6,2.36) and (4.10,2.07) .. (4.44,1.76);
  \pic[rotate=-45] at (4.44,1.76) {ghostrobot};
  \pic at (4.75,2.40) {crash};
  \pic at (4.77,3.12) {crash};
  \pic at (4.77,1.68) {crash};
  \node[lab] at (1.85,1.80) {distance check only};
  \draw[traj=0.32] (1.05,2.4) .. controls (2.2,2.55) and (2.5,3.95) .. (4.2,4.0)
  .. controls (5.7,4.15) and (6.8,3.9) .. (7.05,2.95);
  \pic[rotate=-75] at (7.05,2.95) {ghostrobot};
  \node[lab] at (3.1,4.32) {avoids unsafe states};
  \draw (7.2,2.45) circle (0.13);
  \fill (7.2,2.45) circle (0.045);
  \node[lab] at (7.2,2.08) {goal};
  \node[lab] at (1.8,0.95) {unsafe states:\\collision unavoidable};
  \draw[thin, black!60] (2.72,1.1) -- (3.6,1.72);
\end{tikzpicture}
    \caption{Static obstacle.}
    \label{fig:intro-unsafe-static}
  \end{subfigure}\hfill
  \begin{subfigure}[b]{0.48\textwidth}
    \centering
    \colorlet{brtline}{red!62!black}
\colorlet{brtfill}{red!9}
\colorlet{obsfill}{black!22}
\colorlet{obsline}{black!55}

\begin{tikzpicture}[
  font=\small, >={Latex[length=2mm]},
  lab/.style={font=\scriptsize, align=center},
  unsafe/.style={fill=brtfill, draw=brtline, semithick},
  ghost/.style={densely dashed, black!55, thin},
  traj/.style={semithick, postaction={decorate, decoration={markings,
              mark=at position #1 with {\arrow{Latex[length=1.8mm]}}}}},
  robot/.pic={
    \fill[black!70, rounded corners=0.8pt] (-0.21,0.17) rectangle (0.13,0.26);
    \fill[black!70, rounded corners=0.8pt] (-0.21,-0.26) rectangle (0.13,-0.17);
    \draw[semithick, fill=white, rounded corners=1.5pt] (-0.30,-0.19) rectangle (0.30,0.19);
    \fill[black!75] (0.21,0) -- (0.03,0.10) -- (0.03,-0.10) -- cycle;
  },
  ghostrobot/.pic={
    \fill[black!35, rounded corners=0.8pt] (-0.21,0.17) rectangle (0.13,0.26);
    \fill[black!35, rounded corners=0.8pt] (-0.21,-0.26) rectangle (0.13,-0.17);
    \draw[ghost, fill=white, rounded corners=1.5pt] (-0.30,-0.19) rectangle (0.30,0.19);
    \draw[ghost] (0.21,0) -- (0.03,0.10) -- (0.03,-0.10) -- cycle;
  },
  crash/.pic={
    \foreach \a in {15,60,105,150,195,240,285,330}{
      \draw[brtline, line width=0.6pt] (\a:0.07) -- (\a:0.19);}
    \fill[brtline] (0,0) circle (1.7pt);
  },
  ]
  \begin{scope}[shift={(4.9,4.05)}, rotate=248]
    \path[unsafe]
    (0,0.80)
    arc[start angle=90, end angle=270, radius=0.80]
    .. controls (1.53,-0.80) and (2.51,-0.60) .. (3.10,0)
    .. controls (2.51,0.60) and (1.53,0.80) .. cycle;
  \end{scope}
  \draw[densely dashed, draw=obsline, fill=obsfill!45, thin] (4.9,2.45) circle (0.52);
  \draw[fill=obsfill, draw=obsline, thin] (4.9,4.05) circle (0.52);
  \draw[semithick, ->] (4.9,4.43) -- (4.9,3.67);
  \pic at (0.75,2.4) {robot};
  \draw[black!50, thin] (0.05,2.52) -- (0.36,2.52);
  \draw[black!50, thin] (0.05,2.28) -- (0.36,2.28);
  \draw[traj=0.45, densely dashed] (1.05,2.4) -- (4.10,2.4);
  \draw[traj=0.55, densely dashed] (3.30,2.40) .. controls (3.85,2.40) and (4.069,2.790) .. (4.545,3.065);
  \draw[traj=0.55, densely dashed] (3.30,2.40) .. controls (3.85,2.40) and (4.069,2.110) .. (4.545,1.835);
  \pic at (4.10,2.40) {ghostrobot};
  \pic[rotate=30] at (4.545,3.065) {ghostrobot};
  \pic[rotate=-30] at (4.545,1.835) {ghostrobot};
  \pic at (4.38,2.42) {crash};
  \pic at (4.64,2.90) {crash};
  \pic at (4.64,2.00) {crash};
  \node[lab] at (1.85,1.88) {distance check only};
  \draw[traj=0.32] (1.05,2.4) .. controls (2.3,2.6) and (2.6,4.85) .. (4.6,5.05)
  .. controls (6.2,5.20) and (7.0,3.9) .. (7.05,2.95);
  \pic[rotate=-78] at (7.05,2.95) {ghostrobot};
  \node[lab] at (2.70,5.28) {avoids unsafe states};
  \draw (7.2,2.45) circle (0.13);
  \fill (7.2,2.45) circle (0.045);
  \node[lab] at (7.2,2.08) {goal};
  \node[lab] at (6.28,1.62) {moving\\obstacle};
  \draw[thin, black!60] (6.18,1.90) -- (5.33,3.78);
  \node[lab] at (1.8,0.95) {unsafe states:\\collision unavoidable};
  \draw[thin, black!60] (2.72,1.15) -- (3.45,1.62);
\end{tikzpicture}
    \caption{Moving obstacle.}
    \label{fig:intro-unsafe-dynamic}
  \end{subfigure}
  \caption[Unsafe states for a mobile robot near static and moving obstacles]{Unsafe states for a mobile robot avoiding a static obstacle and a dynamic obstacle crossing its path. The shaded regions are still collision-free, but neither full braking nor maximal steering to either side prevents collision. For the static obstacle, the region grows roughly with the braking distance. For the moving obstacle, it is additionally shaped by the relative motion. A controller that monitors only current distance (dashed trajectories) may enter this region, but a dynamics-aware controller (solid trajectory) evades while recovery remains possible.}
  \label{fig:intro-unsafe-states}
\end{figure}
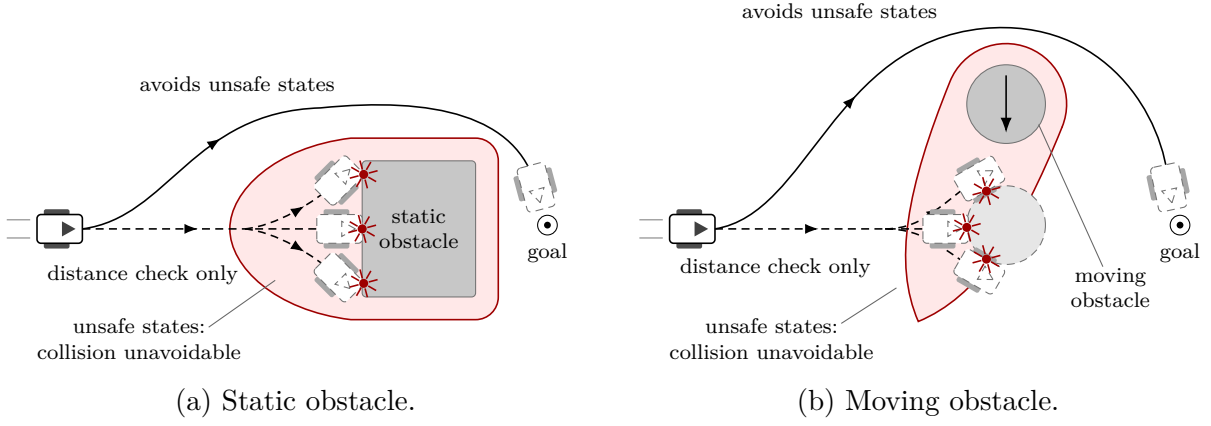

\Cref{fig:intro-unsafe-states} illustrates these \emph{unsafe states} for a mobile robot. In the shaded region, the robot is not yet in collision but can no longer brake or steer away in time. For a static obstacle, this region extends toward the approaching robot by roughly its braking distance and grows with speed. For a moving obstacle, it also reflects the obstacle's relative motion, so states at a considerable current distance from the obstacle may already be unsafe. The three illustrated extreme evasive maneuvers---full braking and maximal steering to either side---all end in collision after the boundary has been crossed. A controller that accounts for these states can intervene while recovery remains possible. A meaningful safety mechanism must therefore reason about motion under the robot dynamics and actuation limits, beyond the current distance to the obstacles \cite{borquez_safety_2024}.

Modern navigation systems often compute motion by solving optimization problems in real time. A motion planner optimizes a short trajectory toward the goal subject to safety and actuation constraints. A safety filter modifies the control command from a separate controller as little as necessary to satisfy a safety condition. Both approaches need to account for future collision risks by identifying the states from which the robot can avoid collision indefinitely, which are the unshaded states in \cref{fig:intro-unsafe-states}. If the environment were fully known and static, this long-term safe set could be computed once before deployment. In the settings considered throughout this dissertation, however, the safe set depends on obstacles that are discovered at runtime and may move, so it changes with every new observation and cannot simply be precomputed.

\emph{Hamilton--Jacobi (HJ) reachability analysis} provides a principled way to obtain this set \cite{bansal_hamilton-jacobi_2017}. For a given state, it determines whether some admissible maneuver keeps the robot out of collision over a specified future horizon, which extends to all future time in the infinite-horizon setting relevant to long-term safety. For specified dynamics, actuation limits, obstacle geometry, and disturbance assumptions, the answer is encoded by a safety value function whose sign separates the states from which collision can still be avoided from those from which it cannot. The framework can also account for worst-case disturbances and adversarial obstacle motion, resulting in a robust safety value function. However, this rigor is computationally expensive, as obtaining one solution can take seconds to minutes even for robots with only a few state variables.

The central challenge is to make reachability-based safety information available as quickly as geometric representations can be updated. This dissertation investigates learned safety representations trained offline on HJ solutions for representative environments. At runtime, the current local observation or tracked obstacle states determine a differentiable safety function without repeated reachability computation. The learned function defines safety constraints for motion planning or for a safety filter that modifies a controller's commands.

\section{Background and Related Work}
\label{sec:background-literature-overview}

This section reviews navigation architectures, local motion-generation methods, and the two safety interfaces studied here, MPC terminal constraints and safety filters. It then examines HJ reachability and learned representations, drawing on surveys of mobile robot motion planning \cite{sanchez-ibanez_path_2021,xiao_motion_2022-1} and safety-critical control \cite{hsu_safety_2024,wabersich_data-driven_2023,hwang_safe_2024}. \Cref{subsec:intro-positioning} identifies the remaining research gap. The mathematical framework follows in \cref{ch:preliminaries}, while each technical chapter reviews work specific to its method.

\subsection{General Navigation Architecture}
\label{subsec:intro-navigation-architecture}

\begin{figure}[tb]
  \centering
  \resizebox{\textwidth}{!}{%
\begin{tikzpicture}[
  font=\small,
  block/.style={draw, rounded corners=2pt, align=center, minimum height=10mm, inner xsep=2.5mm, inner ysep=1.5mm},
  arrow/.style={-{Latex[length=2.2mm]}},
  albl/.style={font=\scriptsize, align=center},
  ]
  \node[block] (global) {Global\\planner};
  \node[block, right=13mm of global] (local) {Local\\planner};
  \node[block, right=20mm of local] (ctrl) {Feedback\\controller};
  \node[block, dashed, right=20mm of ctrl] (filter) {Safety\\filter};
  \node[block, right=20mm of filter] (robot) {Robot};
  \path (global.south) -- (filter.south) coordinate[midway] (pmid);
  \node[block, below=15mm of pmid, minimum width=104mm] (percept)
  {Perception: mapping, localization, obstacle detection, motion prediction};
  \node[block, below=9mm of percept] (env) {Environment};
  \node[albl, above=5mm of global] (goal) {task goal};
  \draw[arrow] (goal) -- (global);
  \draw[arrow] (global) -- node[albl, above] {route} (local);
  \draw[arrow] (local) -- node[albl, above] {reference\\trajectory} (ctrl);
  \draw[arrow] (ctrl) -- node[albl, above] {nominal\\command} (filter);
  \draw[arrow] (filter) -- node[albl, above] {safe\\command} (robot);
  \draw[arrow] (robot.south) |- node[albl, right, pos=0.22] {motion} (env.east);
  \draw[arrow] (env) -- node[albl, right] {sensor data} (percept);
  \draw[arrow] (percept.north -| global) -- node[albl, left] {map} (global);
  \draw[arrow] (percept.north -| local) -- node[albl, left] {local observations,\\obstacle predictions} (local);
  \draw[arrow] (percept.north -| ctrl) -- node[albl, left] {state\\estimate} (ctrl);
  \draw[arrow] (percept.north -| filter) -- node[albl, right] {observations,\\state estimate} (filter);
\end{tikzpicture}%
}
  \caption[Typical mobile robot navigation architecture]{Typical mobile robot navigation architecture. Perception converts sensor data into maps, state estimates, and obstacle information. A global planner provides a route toward the task goal, a local planner converts this route and the latest observations into executable local motion, and a feedback controller tracks the resulting reference. An optional safety filter (dashed) supervises the nominal command and modifies it when it violates the specified safety condition.}
  \label{fig:nav-architecture}
\end{figure}
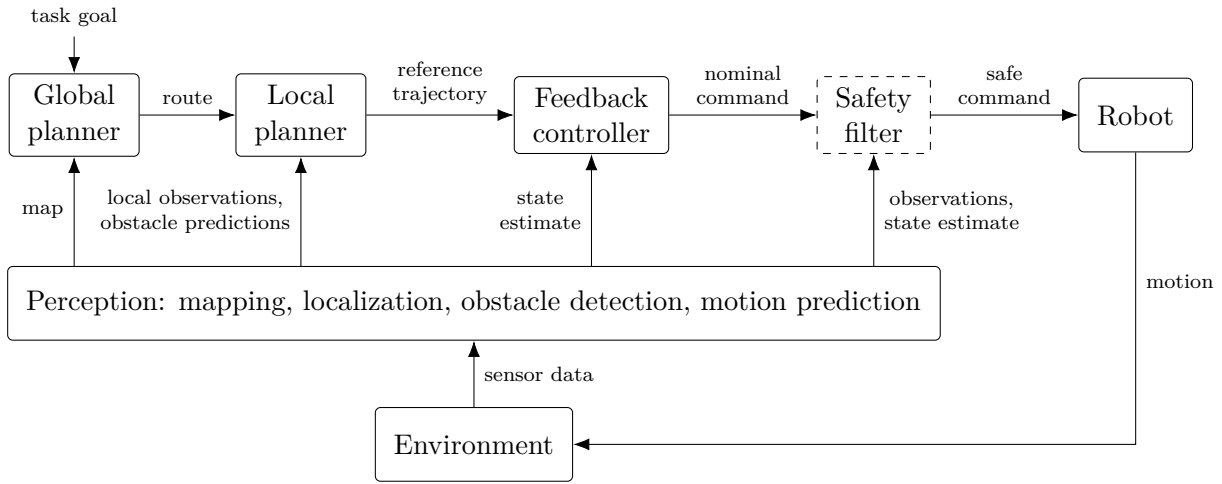

Mobile robot navigation is commonly organized as a hierarchy that connects perception, planning, and feedback control, as illustrated in \cref{fig:nav-architecture}. Perception condenses the raw sensor data into the quantities used by the other components. These are an estimate of the robot state, a local map of the surroundings, and, in dynamic settings, the positions and velocities of nearby moving objects. The most widespread map representation is the occupancy grid, which divides space into cells and marks each cell as free or occupied \cite{elfes_using_1989,hornung_octomap_2013}. From an occupancy grid, a signed-distance function (SDF) can be computed efficiently \cite{felzenszwalb_distance_2012,oleynikova_voxblox_2017}. At each position, the SDF stores the distance to the nearest obstacle and thereby provides a continuous distance measure that motion planners can use.

A global planner searches a coarse map for a route toward the task goal. In this dissertation, the term refers to geometric planners, which search for a collision-free path of a simplified robot model without considering its dynamics. Classical graph-search algorithms such as Dijkstra's algorithm and A* operate on a discretized map \cite{dijkstra_note_1959,hart_formal_1968}, while sampling-based planners such as probabilistic roadmaps and rapidly-exploring random trees (RRT), together with their asymptotically optimal variants, scale to high-dimensional configuration spaces \cite{kavraki_probabilistic_1996,lavalle_randomized_2001,karaman_sampling-based_2011}. Both families also have kinodynamic variants that plan directly with the robot dynamics \cite{lavalle_randomized_2001,dolgov_path_2010}, but these variants typically scale less well, because they search the larger state space of the robot instead of its configuration space. Because a geometric route does not account for the dynamics, it indicates where the robot should go, not how it should move. Throughout this dissertation, global planning is treated as an upstream component that supplies the local planner with information such as waypoints.

The local planner turns a segment of the route into executable motion. From the latest observations and a model of the robot, it produces one of three outputs. A \emph{path} is a sequence of positions or configurations, a \emph{trajectory} additionally assigns timing to these positions, and a \emph{control command} is applied to the robot directly. A feedback controller tracks the planned motion and compensates for deviations between the plan and the measured state. Since the local planner is the component that reacts to newly observed obstacles, it is also where collision avoidance is primarily enforced.

The boundaries between these blocks are conceptual rather than rigid. Model predictive control (MPC), for example, optimizes a short trajectory but applies its first control input directly, thereby merging local planning and feedback control. A learned policy may replace individual blocks or map sensor data directly to commands. A safety filter introduces an additional interface. It treats the output of any upstream component as a \emph{nominal command}, whether that component is a planner, a tracking controller, a learned policy, or a human operator. The filter forwards this command when it satisfies the safety condition and modifies it otherwise. This dissertation focuses on the two blocks of \cref{fig:nav-architecture} where safety constraints are enforced. Reachability-based safety information enters either as the terminal constraint of an MPC local planner or as the barrier function of a downstream safety filter.

\subsection{Local Motion Planning and Control}
\label{subsec:intro-local-motion-generation}

Local motion-planning methods differ in how they select commands. Reactive methods respond to current geometry, velocity-space methods evaluate candidate commands under a constant-velocity assumption, and optimization-based methods plan trajectories \cite{sanchez-ibanez_path_2021,xiao_motion_2022-1}.

\paragraph{Reactive methods.}
Reactive methods compute commands from the geometric relation between the robot, goal, and obstacles. In the artificial potential field \cite{khatib_real-time_1985}, the goal exerts an attractive force and obstacles exert repulsive forces. The robot follows their net force, the negative gradient of the combined potential. The vector field histogram instead selects a direction from a local polar histogram of obstacle density \cite{borenstein_vector_1991}. Both methods are inexpensive, operate at sensor rate, and require no prior map. However, potential-field forces can cancel before the goal is reached, trapping the robot in a local minimum, and narrow passages can cause oscillations. More fundamentally, a direction based on distance alone does not indicate whether the robot can still brake or turn away in time.

\paragraph{Velocity-space methods.}
Velocity-space methods plan one step further ahead by comparing candidate velocity commands. The dynamic window approach (DWA) restricts the candidates to velocities that are reachable within the next control interval under the acceleration limits \cite{fox_dynamic_1997}. It simulates a short arc for each candidate, discards those that lead to a collision or do not allow the robot to stop in time, and scores the remainder by goal progress, obstacle distance, and speed. Velocity obstacles (VOs) address moving obstacles by identifying, for each robot--obstacle pair, the robot velocities that lead to a collision if the obstacle maintains its current velocity, so that the controller can select a velocity outside this set \cite{fiorini_motion_1998}. Reciprocal formulations, most prominently optimal reciprocal collision avoidance (ORCA), divide the required evasion between the agents and produce efficient multi-robot motion \cite{van_den_berg_reciprocal_2008,van_den_berg_reciprocal_2011}. Their collision-avoidance guarantee requires every agent to apply the same rule, an assumption that pedestrians and human-driven vehicles do not satisfy.

\paragraph{Trajectory optimization and model predictive control.}
Optimization-based planners compute several future states or inputs jointly and can trade off goal progress, smoothness, obstacle distance, and control effort explicitly. An early example is the elastic band, which deforms a path like a rubber band under repulsive obstacle forces \cite{quinlan_elastic_1993}. The timed elastic band (TEB) additionally optimizes the velocity profile along the deformed path \cite{rosmann_timed-elastic-bands_2015}. General trajectory optimizers encode collision avoidance as costs or constraints on the planned states and solve the resulting problem numerically \cite{ratliff_chomp_2009,schulman_motion_2014}. The environment typically enters through convex collision-free regions or through the SDF, whose value is the distance to the nearest obstacle and whose gradient points away from it \cite{oleynikova_signed_2016-1,zhang_optimization-based_2021}. MPC embeds this trajectory optimization in a feedback loop \cite{mayne_constrained_2000,schwenzer_review_2021}. At every control cycle, it optimizes the predicted motion over a finite horizon, applies only the first input, and re-solves from the next measured state, so new observations can enter at every cycle. Sampling-based variants such as model predictive path integral control (MPPI) evaluate many randomly perturbed input sequences instead of computing gradients \cite{williams_information_2017}. Because the dynamics, actuation limits, task objectives, and collision constraints all appear explicitly in a single problem, MPC has become a principal method for local planning and is one of the two main frameworks considered in this dissertation. Its finite prediction horizon nevertheless creates a safety problem of its own, which \cref{subsec:intro-safety-mechanisms} examines.

\paragraph{Learning-based methods.}
Learning enters local planning in two different ways. End-to-end approaches use reinforcement learning (RL) or imitation learning (IL) from expert demonstrations to train a complete policy that maps sensor observations directly to commands \cite{tai_virtual--real_2017,everett_motion_2018,chen_crowd-robot_2019,xiao_motion_2022-1}. Such policies remove hand-designed planning rules, but unless a separate safety mechanism is added, their safety depends on how well the policy generalizes from the training scenarios to the deployment environment. Hybrid approaches instead retain a conventional planner and learn one of its components, such as a correction to the dynamics model, an obstacle-motion predictor, or a cost function \cite{hewing_learning-based_2020,xu_machine_2021,brunke_safe_2022}. The position of the learned component determines the consequences of an error. An inaccurate cost may only degrade efficiency, while an inaccurate prediction or constraint can make an unsafe motion appear feasible. The methods developed in this dissertation are hybrid in this sense. They learn neither the task motion nor the dynamics but the safety constraint itself.

\paragraph{Navigation among moving agents.}
Methods for navigation among moving agents differ in what they assume about the other agents. If the agents form a team of robots, they can coordinate their motion, either explicitly by communicating their planned trajectories or implicitly by following shared rules. Each robot can then reason about the behavior of the others, as in decentralized multi-robot planners such as MADER, EGO-Swarm, RLSS, and HDSM \cite{tordesillas_mader_2021,zhou_ego-swarm_2021,senbaslar_rlss_2023,toumieh_high-speed_2024}. This dissertation considers the other case, in which the agents act separately and do not coordinate with the robot. The planner then needs a model of how the environment evolves while the robot executes its motion. Prediction-based approaches extrapolate each obstacle's motion using a constant-velocity (CV) assumption, a physics-motivated interaction model such as the social force model (SFM) \cite{helbing_social_1995}, or a learned predictor that outputs several possible future trajectories and their probabilities \cite{brito_model_2019,korbmacher_review_2022}. Interaction-aware planners additionally model how the other agents react to the robot's own plan \cite{sadigh_planning_2016,zhu_learning_2021}. Accurate predictions permit efficient, minimally conservative motion, but behavior outside the model can invalidate the collision constraints. A different strategy assumes only bounds on obstacle motion and plans against the worst case within these bounds \cite{bouraine_provably_2012}. Worst-case reasoning does not require accurate behavioral predictions as long as the assumed bounds hold, but it is more conservative. In dense crowds, it is a known cause of the freezing robot problem, in which the planner finds no admissible forward motion \cite{trautman_unfreezing_2010}. The dynamic-environment chapters study these two non-interactive viewpoints, explicit motion prediction and bounded-motion models, without assuming that other agents participate in collision avoidance.

\subsection{Terminal Constraints and Safety Filters}
\label{subsec:intro-safety-mechanisms}

Model-based planners commonly enforce collision avoidance through constraints on predicted states. Because these constraints apply only within the prediction horizon, a planned trajectory may satisfy all of them and still end at high speed close to an obstacle just beyond the horizon. From such a final state, the robot can no longer brake or steer away in time, so a collision is already inevitable (\cref{fig:intro-unsafe-states}). Lengthening the horizon postpones this problem at a growing computational cost. Long-term safety requires a mechanism that addresses what happens after the final predicted state.

\paragraph{Terminal safe sets.}
MPC theory addresses the end of the prediction horizon through a \emph{terminal constraint}, which requires the final predicted state to lie in a set in which the robot can remain safe indefinitely. The required property is \emph{controlled invariance}, meaning that from every state in the set, some admissible control keeps the robot inside it. With a controlled-invariant terminal set, feasibility of the MPC problem at one control cycle implies feasibility at the next, a property known as \emph{recursive feasibility} \cite{mayne_constrained_2000,kerrigan_invariant_2000}. In a static environment, the simplest valid terminal set for a robot that can remain stopped consists of collision-free states with zero velocity. However, this choice forces every plan to nearly come to rest at the end of its horizon. Viability-based terminal sets reduce this conservatism by admitting every state from which a safe continuation exists, whether moving or stationary \cite{liniger_viability_2015,bouguerra_viability-based_2019}. The difficulty in local navigation is that this set depends on the obstacles, so a terminal set computed offline for one environment need not be valid in another. The resulting challenge is to construct a valid environment-dependent terminal set at runtime.

\paragraph{Safety filters.}
The second mechanism decouples safety from the task. A \emph{safety filter} monitors the nominal command and checks whether it is compatible with keeping the robot inside a specified safe set. If it is, the command is applied unchanged. Otherwise, the filter minimally modifies it or switches to a recovery maneuver \cite{hsu_safety_2024,wabersich_data-driven_2023}. The advantage of this architecture is its modularity, since one filter can supervise a classical planner, a learned policy, or a human operator without modifying the supervised component. Its guarantee extends only as far as the model, the uncertainty assumptions, and the safe set on which the filter relies.

\paragraph{Control barrier functions.}
The most widely used filter formulation relies on \emph{control barrier functions} (CBFs) \cite{wieland_constructive_2007,ames_control_2019}. A CBF is a scalar function of the state that is nonnegative on the safe set it represents and zero on the boundary of that set. The CBF condition limits how fast this value may decrease along the motion, so that the closer the robot comes to the boundary, the less it may move toward it. For control-affine dynamics, this condition is a linear inequality in the input. Filtering then reduces to a quadratic program (QP), known as the CBF-QP, which finds the admissible input closest to the nominal command \cite{ames_control_2014,rauscher_constrained_2016}. Classical potential fields can be recovered as a special case of such filters \cite{singletary_comparative_2021}. Numerous extensions widen the scope, including higher-order formulations for systems in which the input affects the barrier only through several integrations, robust, time-varying, and discrete-time variants, and formulations that embed CBF conditions as constraints in MPC \cite{nguyen_exponential_2016,xiao_high-order_2022,garg_advances_2024,zeng_safety-critical_2021,vulcano_safe_2022}. All variants share one premise, that a valid barrier function representing a controlled-invariant set must be available. Constructing one is straightforward for simple geometric constraints and low-order dynamics but difficult in general.

\paragraph{Predictive and reachability-based filters.}
Two further filter families replace the pointwise barrier condition. Predictive safety filters accept a command only if a feasible backup trajectory into a safe terminal set remains available after the command is applied. Otherwise, they minimally adjust the command or execute the previously verified backup \cite{wabersich_predictive_2021,gurriet_online_2018,chen_backup_2021}. The online computation resembles MPC, but its purpose is certification rather than task optimization, and it inherits the need for a valid terminal safe set. Reachability-based filters use the HJ safety value function directly. The filter remains passive while the value is sufficiently positive and applies the optimal evasive input when the state approaches the safety boundary \cite{leung_infusing_2020,borquez_safety_2024}. Such filters intervene exactly when required by the dynamics and actuation limits, but they require that the value function for the current environment is available.

\bigskip

These mechanisms all require a dynamics-aware safety representation for the current environment, which may be a controlled-invariant terminal set, a backup set, or a valid barrier or value function. Obtaining that representation is the bottleneck studied here. A geometric map or SDF can be updated rapidly, but its collision-free region need not be controlled invariant under limited braking and steering. HJ reachability computes the required sets, as the following subsection explains.

\subsection{Hamilton--Jacobi Reachability Analysis}
\label{subsec:intro-hj-reachability}

The requirement identified above can be stated precisely using two notions that recur throughout the dissertation. A state is \emph{collision-free} if the robot body does not currently intersect an obstacle. In a time-invariant setting, it is \emph{safe} if, in addition, some admissible control keeps the robot collision-free for all future time. All such states form the largest controlled-invariant subset of the collision-free states, known as the viability kernel \cite{aubin_viability_2011,blanchini_survey_1999} or, equivalently, the \emph{maximal safe set}, the term used throughout this dissertation. The collision-free states outside this set are exactly the inevitable collision states described in \cref{sec:motivation} \cite{fraichard_short_2007,bautin_inevitable_2010}. The maximal safe set is the ideal set for both safety mechanisms of the previous subsection, since it is the largest valid MPC terminal set and the largest safe set that a filter can enforce under the modeled assumptions. When the environment evolves with time, the corresponding set is instead a time- and horizon-dependent avoidance set that depends on the assumed evolution model.

HJ reachability analysis is the most general computational framework for obtaining such sets for nonlinear systems \cite{lygeros_reachability_2004,mitchell_time-dependent_2005,bansal_hamilton-jacobi_2017}. Its inputs are the system dynamics and actuation limits, a \emph{failure set} containing the states that must never be reached, such as all collision states, and optionally bounds on disturbances or obstacle actions, which are treated as adversarial within these bounds. Dynamic programming propagates the failure set backward through time and yields the \emph{backward reachable tube} (BRT) of states from which failure cannot be avoided within the considered horizon. The result is encoded by a scalar \emph{value function} whose sign separates the tube from its complement, and for a time-invariant failure set the complement of the BRT converges toward the maximal safe set as the horizon grows. The value function carries more information than the set alone, because its magnitude serves as a safety margin and its gradient indicates the locally best evasive direction. These properties make the HJ solution directly usable as an MPC terminal constraint, the switching criterion of a reachability-based filter, or a CBF candidate\footnote{The HJ value function is generally nonsmooth and, under standard regularity assumptions, only locally Lipschitz continuous, while conventional CBF formulations require a continuously differentiable barrier function. This distinction is discussed further in \cref{ch:preliminaries}.} \cite{liniger_viability_2015,leung_infusing_2020,choi_robust_2021}.

\paragraph{The computational barrier.}
This generality has a computational cost. Standard solvers discretize the state space on a grid, so memory and computation grow exponentially with the number of state variables. This curse of dimensionality typically limits practical grid-based computations to systems with roughly five or six state variables, and even small problems require computation times of seconds to minutes \cite{bansal_hamilton-jacobi_2017}. Local navigation adds a further limitation, because the failure set encodes the obstacles and each value function is thus tied to one specific environment. A robot that discovers its surroundings through onboard sensors would, strictly speaking, require a new reachability solution after every observation update, and when obstacles move, the failure set also changes over time. Grid-based HJ reachability is consequently impractical for real-time use and is generally reserved for offline analysis.

\paragraph{Reachability-augmented planning.}
Several methods integrate precomputed reachability information into online planning. FaSTrack plans quickly with a simplified model and inflates all obstacles by a worst-case tracking-error bound computed offline through HJ reachability \cite{herbert_fastrack_2017,chen_fastrack_2021}. Other approaches propagate forward reachable sets that enclose all trajectories under uncertainty \cite{althoff_set_2021} or assemble plans from motion primitives whose swept volumes were verified offline \cite{kousik_bridging_2020}. These methods demonstrate a principle on which this dissertation also builds. The expensive dynamics-aware computation runs offline, and the online planner reuses its result. The information that they reuse typically describes the robot model, such as a tracking-error bound or a verified library of motion primitives, rather than a complete safe-set field for the local scene that the robot currently observes.

\paragraph{Scalable reachability formulations.}
There is a substantial literature on reducing the computational cost of HJ reachability by exploiting system structure. Decomposition splits dynamics with separable structure into lower-dimensional subsystems \cite{chen_decomposition_2018}, and warm-starting reuses a previous solution when the problem changes only moderately \cite{herbert_reachability-based_2019,bajcsy_efficient_2019}. Parameter-conditioned formulations precompute the dependence of the value function on a small number of changing quantities, such as an obstacle position or a model parameter \cite{borquez_parameter-conditioned_2023,jeong_parameterized_2024,nakamura_online_2023}. A particularly effective approach uses a coordinate transformation. When the state of a robot--obstacle pair is expressed in relative coordinates, collision becomes a fixed region around the origin. A single pairwise solution can then cover every obstacle with the modeled dynamics and shape, although the pairwise safety conditions must be combined online \cite{chen_multi_vehicle_2016}. Finally, self-supervised neural solvers and RL formulations approximate the value function without a dense grid and thereby reach higher state dimensions \cite{bansal_deepreach_2021,fisac_bridging_2019}. Each technique removes a particular bottleneck, but decomposition requires separable structure, parameter conditioning grows with the number of changing quantities, and pairwise coordinates do not directly represent the joint geometry of a local scene.

\subsection{Learned Safety Representations}
\label{subsec:intro-learned-safety}

Learning offers a way to avoid online reachability analysis. The expensive computation is performed offline, and a network trained on its results provides an approximate solution at runtime. Early supervised approaches reproduce HJ value functions or safe-set labels for a fixed system or a low-dimensional family of settings \cite{allen_machine_2014,rubies-royo_classification-based_2019}. A related body of work learns barrier functions from expert demonstrations, sampled dynamics, or labeled safe and unsafe states \cite{robey_learning_2020,srinivasan_synthesis_2020,liu_safe_2022,dawson_safe_2023}. These networks represent safety constraints for planning or filtering control inputs, whereas the learned policies of \cref{subsec:intro-local-motion-generation} generate the commands themselves.

Navigation in previously unseen environments additionally requires the representation to depend on what the robot currently observes. Neural implicit representations provide a suitable model class for such conditioned scalar fields \cite{park_deepsdf_2019,sitzmann_implicit_2020}. Neural distance fields can be constructed from range measurements \cite{long_learning_2021}, and LiDAR-based neural barriers can operate without a consistent global map \cite{harms_neural_2024-2}. Most closely related to the reachability-supervised approach of this dissertation, an observation-conditioned reachability value network predicts HJ safety values from LiDAR and combines them with online disturbance estimates for quadrupedal safety filtering \cite{lin_one_2026}. These works establish that observation-conditioned safety prediction is feasible. They do not determine what a learned representation must provide before an optimization-based controller can rely on it.

Two practical requirements are especially important for learned safety representations. First, a controller may query the safety function and its gradient hundreds of times per second, either across the iterations of an MPC optimization or through successive safety-filter updates, while the observation changes much less frequently. This usage favors architectures that separate high-dimensional observation processing from repeated low-dimensional state queries. Hypernetworks, which generate the weights of a smaller network from an input, provide one such mechanism \cite{ha_hypernetworks_2017}. Second, a learned safety function is not automatically a certificate. A small value error near the safety boundary can change the predicted sign and label an unsafe state as safe, and an inaccurate gradient can produce an incorrect control constraint even where the sign is correct. A training or validation loss evaluated at sampled states provides no guarantee between those samples or outside the training distribution \cite{so_how_2024,dawson_safe_2023}. Beyond average approximation accuracy, this raises the open question of whether selected safety properties can be imposed by the structure of the representation itself rather than inferred from sampled data.

\subsection{Research Gap and Dissertation Positioning}
\label{subsec:intro-positioning}

\begin{figure}[tb]
  \centering
  \begin{tikzpicture}[
  font=\small,
  grp/.style={draw, rounded corners=3pt, align=center, font=\scriptsize, inner sep=3pt, fill=black!5},
  arrow/.style={-{Latex[length=2.5mm]}},
  ]
  \draw[arrow] (0,0) -- (11.4,0);
  \draw[arrow] (0,0) -- (0,7.2);
  \node[align=center] at (5.7,-0.55) {online adaptability to new observations};
  \node[align=center, rotate=90] at (-0.55,3.6) {dynamics-aware safety reasoning};
  \node[grp] (hj) at (1.5,6.5) {grid-based HJ\\reachability~\cite{bansal_hamilton-jacobi_2017,mitchell_time-dependent_2005}};
  \node[grp] (scal) at (4.2,5.0) {scalable and parameter-\\conditioned reachability~\cite{chen_decomposition_2018,borquez_parameter-conditioned_2023}};
  \node[grp] (cbf) at (6.8,3.4) {hand-designed\\safety filters~\cite{ames_control_2014,nguyen_exponential_2016}};
  \node[grp] (obs) at (8.4,4.3) {learning-based\\safety filters~\cite{robey_learning_2020,lin_one_2026}};
  \node[grp] (mpc) at (6.2,2.55) {MPC with geometric\\collision constraints~\cite{zhang_optimization-based_2021,zeng_safety-critical_2021}};
  \node[grp] (learn) at (8.6,1.7) {end-to-end learned\\motion policies~\cite{tai_virtual--real_2017,chen_crowd-robot_2019}};
  \node[grp] (geom) at (9.7,0.7) {geometric and\\reactive methods~\cite{khatib_real-time_1985,fox_dynamic_1997}};
  \node[grp, fill=black!12, thick] (thesis) at (8.6,5.4) {\textbf{this dissertation:}\\reachability-informed\\safety representations};
  \draw[arrow, dashed] (hj.east) -- node[font=\scriptsize, above, sloped] {offline supervision} (thesis.west);
\end{tikzpicture}
  \caption[Qualitative positioning of the reviewed method families and this dissertation]{Qualitative positioning of the reviewed method families and this dissertation. The horizontal axis indicates adaptation to new local observations, and the vertical axis indicates the extent to which safety reasoning accounts for robot dynamics and actuation limits. Placements represent typical formulations, and the learning-based safety filters include both learned barrier functions and learned reachability values. This dissertation studies reachability-informed safety constraints across static and dynamic environments and across MPC and safety-filter interfaces.}
  \label{fig:positioning}
\end{figure}
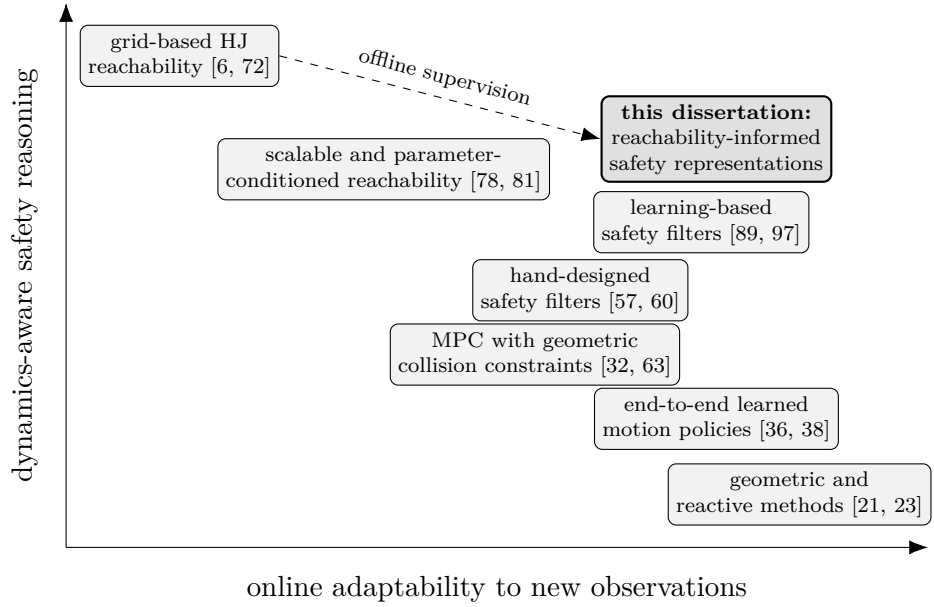

\Cref{fig:positioning} summarizes the reviewed methods along two dimensions, their ability to incorporate new observations and the extent to which their safety reasoning accounts for the robot dynamics and actuation limits. Geometric and reactive planners, MPC with geometric collision constraints, and end-to-end learned policies adapt to new observations at sensor rate, but their safety reasoning is based on the current obstacle distance, a finite prediction horizon, or empirical performance. Grid-based HJ reachability lies in the opposite corner of the figure. Within its stated model, it explicitly accounts for the dynamics and actuation limits, but each solution is tied to the environment that was fixed when it was computed. Scalable and parameter-conditioned reachability formulations relax this restriction along a small number of pre-selected parameters. Hand-designed CBF safety filters can be updated online because their barrier functions are simple enough to construct directly, but such barriers typically capture only part of the safe set or lose validity under the actuation limits. Learning-based safety filters, which include learned barrier functions and learned reachability values, come closest to combining adaptability with dynamics-aware safety reasoning and form the immediate context of this dissertation.

The reviewed work leaves five related problems. Four of them concern the representation, which must adapt reachability values to unseen geometry, support repeated value and gradient queries, handle moving obstacles and variable obstacle counts, and impose structural safeguards against approximation error. The fifth concerns the supervision. Learning from grid-based reachability solutions transfers the dimensionality limit of the solver to the offline stage, so aerial robots whose model includes the body orientation and angular rates, manipulators, and legged systems can exceed the practical range of the training-data generator itself. The following section formulates these problems as research questions.

\section{Research Problem and Questions}
\label{sec:problem-statement}

This dissertation studies how to make dynamics-aware safety information available to real-time optimization-based controllers in environments that the robot has not seen before and that may change during operation. The dissertation develops learned representations that map the current local observation, or the states of the currently tracked obstacles, to an approximate HJ safety value function. The resulting function must be differentiable and inexpensive to evaluate, because the controller queries its values and gradients repeatedly at a high rate, and it must be accurate near the safety boundary, where its sign separates safe from unsafe states. These requirements are studied for two controller interfaces. As a terminal constraint in MPC, the learned function addresses safety beyond the finite prediction horizon. As a barrier candidate in a CBF safety filter, it defines the safety condition used to modify the nominal control command.

A second part of the problem concerns the supervision on which such a representation is trained. Grid-based reachability computation is practical mainly for low-dimensional models, so the dissertation also asks how the training targets themselves can be produced for systems beyond that range.

The scope is local navigation under specified robot dynamics, actuation limits, and, where applicable, disturbance and obstacle-motion models. Perception, state estimation, behavioral prediction, and global planning are upstream components. Moving agents are modeled as non-interactive obstacles that need not participate in collision avoidance. Since learned functions do not automatically inherit HJ guarantees, the analysis separates what the exact model guarantees, how accurately the network approximates it, which properties the architecture enforces, and how the resulting controller performs. Four research questions guide the technical development:

\begin{enumerate}
  \item \label{rq:realtime} How can reachability-based safety functions be made available in real time from local observations or tracked obstacle states in previously unseen environments, as terminal constraints of MPC planners and as barriers of CBF safety filters?
  \item \label{rq:dynamic} How can such representations account for moving obstacles, whose motion makes the safe set time-varying, and for a varying number of such obstacles?
  \item \label{rq:structure} Which safety properties can the structure of a learned representation guarantee regardless of the approximation error, and which properties remain empirical when the representation is used by an optimization-based controller?
  \item \label{rq:supervision} How can reachability supervision be generated for systems whose state dimension exceeds the practical reach of grid-based HJ solvers, while keeping the accuracy of the resulting safe-set approximation?
\end{enumerate}

\section{Contributions}
\label{sec:contributions}

This dissertation makes five methodological contributions, one from each of the author's peer-reviewed publications. The first four combine MPC terminal constraints or CBF safety filters with static or dynamic environments. They learn from grid-based HJ solutions and adapt the safety function online to current observations or obstacle states. The fifth retains the residual representation and safety-filter interface but replaces grid-based supervision with generated search targets. The heading of each contribution names the venue and year of its publication, and \cref{app:publications} lists the five publications together with the contributions of each author.

\paragraph{NTC-MPC: neural terminal constraints for static environments (RA-L 2025).}
Presented in \cref{ch:ntc-mpc}, \emph{neural terminal constraint MPC} (NTC-MPC) learns observation-conditioned terminal safe sets for unknown static environments from HJ reachability supervision \cite{derajic_learning_2025}. A hypernetwork maps the local SDF observation to the weights of a compact main network that approximates the corresponding HJ value function and serves as the MPC terminal constraint. This factorization processes the observation once per planning cycle while the solver repeatedly queries only the main network. The chapter also contributes a radially weighted regression loss that concentrates accuracy near the zero level set, where misclassification changes the safe-set estimate. Experiments show improved collision avoidance at comparable online computation time, with the largest benefit at short prediction horizons.

\paragraph{RNTC-MPC: residual neural terminal constraints for dynamic environments (CoRL 2025).}
Presented in \cref{ch:rntc-mpc}, \emph{residual neural terminal constraint MPC} (RNTC-MPC) extends neural terminal constraints to moving obstacles by conditioning on predicted geometry beyond the MPC horizon \cite{derajic_residual_2025}. Its second contribution is structural, since the learned terminal value is expressed as the predicted terminal SDF minus a nonnegative neural residual. The SDF supplies the predicted obstacle geometry directly, while the network learns the additional unsafe region induced by the robot dynamics and obstacle motion. Consequently, no state marked as occupied by the predicted terminal geometry can be classified as safe, regardless of residual-regression error. This failure-set exclusion principle is independent of the dynamic setting and is reused in the subsequent contributions. The method remains compatible with real-time MPC and improves the learned safe-set approximation and collision avoidance over the evaluated direct-regression alternative.

\paragraph{ORN-CBF: observation-conditioned safety filtering for static environments (ICRA 2026).}
Presented in \cref{ch:orn-cbf}, the \emph{observation-conditioned residual neural CBF} (ORN-CBF) transfers the observation-conditioned and residual representation principles to a safety-filter interface in unknown static environments \cite{derajic_orn-cbf_2026}. The observed SDF minus a nonnegative, hypernetwork-generated residual yields a barrier candidate that approximates the reachability-based safe set while excluding the observed failure set by construction. Unlike an MPC terminal constraint, this field and its gradient are queried at the current state at control rate and are replaced when a new observation arrives. The chapter formulates this observation-update issue explicitly, introduces positional cropping to reduce offline training cost, and integrates the resulting barrier function into a real-time CBF-QP safety filter.

\paragraph{CN-CBF: composite safety filtering for moving obstacles (IROS 2026).}
Presented in \cref{ch:cn-cbf}, the \emph{composite neural CBF} (CN-CBF) extends reachability-supervised safety filtering to a varying number of moving obstacles \cite{derajic_cn-cbf_2026}. It replaces a joint scene-conditioned value function with one residual barrier for the relative dynamics of the robot and a single obstacle. In relative coordinates, collision is a fixed region even while the obstacle moves, so one offline reachability computation and one shared learned model cover every obstacle of the modeled class. At runtime, a smooth aggregation combines the pairwise barriers of the currently observed obstacles into one composite barrier candidate. CN-CBF handles a variable obstacle count without retraining and provides a compositional alternative to the prediction-conditioned joint representation of RNTC-MPC.

\paragraph{VertexCBF: grid-free reachability supervision (CoRL 2026).}
Presented in \cref{ch:vertex-cbf}, \emph{VertexCBF} removes the dependence on grid-based reachability solvers that the four preceding methods share, for the control-affine dynamics and convex polytope input sets used in the CBF-QP formulation here \cite{derajic_vertexcbf_2026}. For such systems, the HJ Hamiltonian is maximized at a vertex of the input polytope, so the maximization over the input set can be restricted to its finitely many vertices. A label generator built on this property runs a finite-horizon tree search over vertex inputs in parallel on a graphics processing unit (GPU) and produces supervision labels independently at sparsely sampled states. A physics-informed loss derived from the stationary variational inequality and the residual parameterization of the preceding chapters complete the training objective. The result is a set of neural barrier candidates for systems with up to 13 states, well beyond the range of the grid-based solvers used elsewhere in the dissertation. The chapter also contributes a closed-loop validation protocol that assesses a learned barrier by rollouts, because no tractable numerical reference is available for such systems.

\paragraph{Unified framework and validation.}
Beyond the five methods, the dissertation contributes a common framework and cross-method analysis. It identifies the shared offline--online pipeline, compares conditioned and compositional representations across the two controller interfaces, and separates the choice of representation from the choice of supervision. The experiments span simulation and hardware studies on ground and aerial robots in static and dynamic environments, with comparisons against multiple baseline planners and safety filters, and a benchmark over 15 dynamical systems for the supervision study.

\bigskip

The five contributions map onto the research questions as follows. The first four methods address real-time safety representations (question~\ref{rq:realtime}), and among them RNTC-MPC and CN-CBF also treat moving obstacles (question~\ref{rq:dynamic}). The residual representation and the guarantee-scope analysis, which run through all technical chapters, address the structural and empirical safety properties of question~\ref{rq:structure}, while VertexCBF addresses the scalable supervision of question~\ref{rq:supervision}. \Cref{ch:conclusion} consolidates these answers.

\section{Dissertation Structure}
\label{sec:structure}

\Cref{ch:preliminaries} develops the mathematical foundations used throughout the dissertation, including reachability analysis, MPC, CBFs, and neural implicit representations. It then introduces the shared framework for observation-conditioned safety functions, offline reachability supervision, efficient online inference, residual failure-set exclusion, and the scope of claims made for learned safety functions.

\begin{figure}[tb]
  \centering
  \pgfmathsetlengthmacro{\outlinecolsep}{4mm}
\pgfmathsetlengthmacro{\outlinepartw}{(\textwidth-\outlinecolsep)/2-1pt}
\pgfmathsetlengthmacro{\outlinepartcx}{(\outlinepartw+\outlinecolsep)/2}
\pgfmathsetlengthmacro{\outlineinnerw}{\outlinepartw-8mm}
\begin{tikzpicture}[
    font=\footnotesize,
    arrow/.style={-{Latex[length=2.4mm, width=2.2mm]}, line width=0.8pt},
    joint/.style={line width=0.8pt},
    chapter/.style={
      draw=black!70, fill=black!10, rounded corners=1.5mm, line width=0.6pt,
      align=center, inner xsep=2mm, inner ysep=2mm,
      execute at begin node={\hyphenpenalty=10000\exhyphenpenalty=10000\relax},
    },
    chainchapter/.style={chapter, text width=68mm, minimum height=8mm},
    partchapter/.style={chapter, text width=\outlineinnerw, minimum height=11mm},
    partbox/.style={
      draw=black!60, fill=black!3, dashed, rounded corners=3.5mm,
      line width=1.1pt, inner xsep=0mm, inner ysep=2mm,
      minimum width=\outlinepartw,
    },
    parttitle/.style={
      align=center, text width=\outlinepartw-3mm, minimum height=8mm, inner sep=1mm,
    },
  ]

  \node[chainchapter] (ch1) at (0, 0) {\textbf{Chapter 1:} Introduction};
  \node[chainchapter, below=4mm of ch1] (ch2) {\textbf{Chapter 2:} Preliminaries and Common Framework};

  \coordinate (parttop) at ($(ch2.south)+(0,-9mm)$);

  \node[parttitle, anchor=north] (p1title) at (parttop -| {-\outlinepartcx,0})
    {\textbf{Part I:}\\ Reachability-Informed Model Predictive Control};
  \node[partchapter, below=2mm of p1title] (ch3)
    {\textbf{Chapter 3:} Neural Terminal Constraints in Static Environments};
  \node[partchapter, below=3.5mm of ch3] (ch4)
    {\textbf{Chapter 4:} Neural Terminal Constraints in Dynamic Environments};

  \node[parttitle, anchor=north] (p2title) at (parttop -| {\outlinepartcx,0})
    {\textbf{Part II:}\\ Reachability-Informed Safety Filters};
  \node[partchapter, below=2mm of p2title] (ch5)
    {\textbf{Chapter 5:} Neural Control Barrier Functions in Static Environments};
  \node[partchapter, below=3.5mm of ch5] (ch6)
    {\textbf{Chapter 6:} Neural Control Barrier Functions in Dynamic Environments};

  \begin{scope}[on background layer]
    \node[partbox, fit=(p1title)(ch3)(ch4)] (part1) {};
    \node[partbox, fit=(p2title)(ch5)(ch6)] (part2) {};
  \end{scope}

  \coordinate (part3top) at ($(part2.south -| 0,0)+(0,-9mm)$);
  \node[parttitle, anchor=north] (p3title) at (part3top)
    {\textbf{Part III:}\\ Grid-Free Reachability Supervision};
  \node[partchapter, below=2mm of p3title] (ch7)
    {\textbf{Chapter 7:} Vertex-Restricted Supervision for Neural Control Barrier Functions};

  \begin{scope}[on background layer]
    \node[partbox, fit=(p3title)(ch7)] (part3) {};
  \end{scope}

  \node[chainchapter, anchor=north] (ch8) at ($(part3.south -| 0,0)+(0,-9mm)$)
    {\textbf{Chapter 8:} Conclusions and Future Work};

  \draw[arrow] (ch1) -- (ch2);
  \coordinate (split) at ($(ch2.south)!0.5!(ch2.south |- part1.north)$);
  \draw[joint] (ch2.south) -- (split);
  \draw[arrow] (split) -| (part1.north);
  \draw[arrow] (split) -| (part2.north);
  \coordinate (merge) at ($(part3.north)!0.5!(part3.north |- part1.south)$);
  \draw[joint] (part1.south) |- (merge);
  \draw[joint] (part2.south) |- (merge);
  \draw[arrow] (merge) -- (part3.north);
  \draw[arrow] (part3.south) -- (ch8.north);
\end{tikzpicture}
  \caption[Organization of the dissertation]{Organization of the dissertation. After the introduction and the preliminaries, Parts~I and~II group the technical chapters by controller interface, and within each part the development progresses from static to dynamic environments. Part~III varies a third design choice, the origin of the reachability supervision. The concluding chapter compares the methods of the three parts.}
  \label{fig:thesis-outline}
\end{figure}
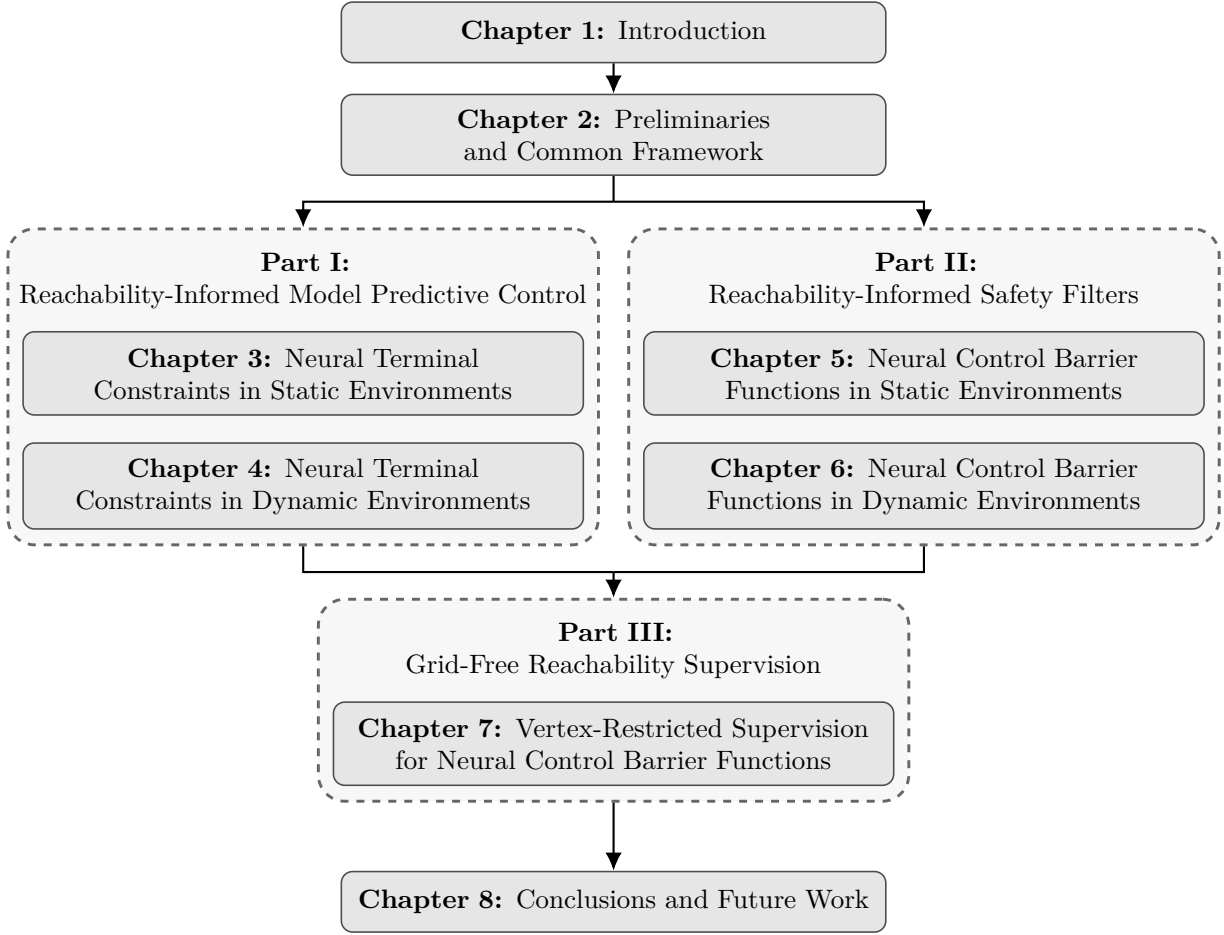

Each technical chapter motivates the problem, reviews related work, states the problem formally, and develops the method, followed by the experiments, a discussion of the results and their scope, and a short summary. The chapters are grouped into three parts, as \cref{fig:thesis-outline} illustrates. Part~I studies neural terminal constraints for MPC, with \cref{ch:ntc-mpc} for unknown static environments and \cref{ch:rntc-mpc} for moving obstacles. Part~II studies CBF safety filters, with \cref{ch:orn-cbf} for unknown static environments and \cref{ch:cn-cbf} for a varying number of moving obstacles. Part~III varies the origin of the reachability supervision. \Cref{ch:vertex-cbf} keeps the safety-filter interface and the residual representation of the preceding chapters and replaces the grid-based solver by labels generated through vertex-restricted search together with a physics-informed loss. The same supervision could also serve the MPC interface of Part~I, but only for robot models with control-affine dynamics and polytopic input constraints, which the MPC methods do not assume in general. This combination is not evaluated here. Finally, \cref{ch:conclusion} answers the research questions, compares the five methods, separates the safety properties that hold by theory or by construction from those demonstrated experimentally, and outlines future work.

\chapter{Preliminaries and Common Framework}
\label{ch:preliminaries}

This chapter introduces the notation and the mathematical tools shared by the technical chapters, which are model predictive control (MPC), control barrier functions (CBFs), Hamilton--Jacobi (HJ) reachability, and learned implicit safety representations. The development starts with system dynamics and distinguishes instantaneous constraint satisfaction from long-term safety. It then presents MPC and CBF safety filters, followed by HJ reachability and its connection to barrier functions. The final sections describe observation-dependent safety functions and their learned representations, including the assumptions needed when observations change. Method-specific models, architectures, and implementation details follow in the corresponding chapters.

\section{System Dynamics}
\label{sec:prelim-system-dynamics}

The robot's \emph{state} \(x\in\mathcal{X}\subseteq\mathbb{R}^{n_x}\) is a vector that collects quantities such as position, heading, and velocity. The \emph{control input} \(u\in\mathcal{U}\subseteq\mathbb{R}^{n_u}\) collects the commanded quantities through which the robot's state is controlled, such as acceleration and steering rate. When present, an exogenous input \(d\in\mathcal{D}\subseteq\mathbb{R}^{n_d}\) models a disturbance or the decision of another agent that is not controlled. The input sets \(\mathcal{U}\) and \(\mathcal{D}\) are assumed compact. Compactness of \(\mathcal{U}\) encodes the actuation limits central to this dissertation and guarantees that the extrema appearing in later sections are attained. Continuous time is denoted by \(t\in\mathbb{R}\) and discrete time by \(k\in\mathbb{Z}\).

\subsection{Continuous-Time Systems}
\label{subsec:prelim-continuous-time-systems}

The continuous-time model used in the reachability and barrier-function formulations is
\begin{equation}
  \dot{x}(t)
  =
  f\bigl(x(t),u(t),d(t)\bigr),
  \qquad
  x(t)\in\mathcal{X},\quad
  u(t)\in\mathcal{U},\quad
  d(t)\in\mathcal{D}.
  \label{eq:prelim-general-continuous-dynamics}
\end{equation}
When no disturbance or adversarial input is modeled, the dynamics are written as
\begin{equation}
  \dot{x}(t)
  =
  f\bigl(x(t),u(t)\bigr).
  \label{eq:prelim-undisturbed-continuous-dynamics}
\end{equation}
Throughout, the vector field \(f\) is assumed continuous in all arguments and locally Lipschitz continuous in \(x\), and trajectories are assumed not to escape to infinity in finite time. For every initial state and every pair of admissible input signals, a unique continuous trajectory then exists on the horizon of interest \cite{khalil_nonlinear_2002,mitchell_time-dependent_2005}. Explicit time dependence is suppressed whenever it is clear from context and not essential to the discussion.

Admissible input signals are measurable functions of time taking values in the respective input sets. For an interval \([t,T]\), the sets of admissible control and disturbance signals are
\begin{equation}
  \mathbb{U}_{[t,T]}
  :=
  \bigl\{u:[t,T]\rightarrow\mathcal{U},\;\text{measurable}\bigr\},
  \qquad
  \mathbb{D}_{[t,T]}
  :=
  \bigl\{d:[t,T]\rightarrow\mathcal{D},\;\text{measurable}\bigr\},
  \label{eq:prelim-admissible-signal-spaces}
\end{equation}
respectively, with the analogous definitions \(\mathbb{U}_{[t,\infty)}\) and \(\mathbb{D}_{[t,\infty)}\) on infinite horizons. Given an initial state \(x\) at time \(t\), the state reached at time \(\tau\geq t\) under admissible input signals \(u(\cdot)\) and \(d(\cdot)\) is denoted by \(\xi_{x,t}^{u,d}(\tau)\). For undisturbed systems, the notation is shortened to \(\xi_{x,t}^{u}(\tau)\). Although the dynamics in \cref{eq:prelim-general-continuous-dynamics} are time invariant, the initial time is retained in the trajectory notation, because \cref{subsec:prelim-time-varying-hj} considers constraints that depend explicitly on time.

\paragraph{Disturbance information pattern.}
When a disturbance is present, the safety question becomes a two-player game in which the controller tries to keep the system safe while the disturbance tries to violate safety. The answer depends on what each player is assumed to know when choosing an input, called the \emph{information pattern}. The pattern adopted in this dissertation allows the disturbance to react to the control but not to anticipate it, which is formalized through strategies. A map \(\gamma:\mathbb{U}_{[t,T]}\rightarrow\mathbb{D}_{[t,T]}\) is a \emph{nonanticipative strategy} if, whenever two control signals coincide almost everywhere on \([t,s]\) for some \(s\leq T\), the corresponding disturbance signals \(\gamma[u]\) also coincide almost everywhere on \([t,s]\). The set of nonanticipative strategies on \([t,T]\) is denoted by \(\Gamma_{[t,T]}\), with \(\Gamma_{[t,\infty)}\) defined analogously. Granting the disturbance this instantaneous informational advantage is the conservative choice for safety analysis, because any safety statement that holds against such strategies also holds against any less informed disturbance \cite{evans_differential_1984,mitchell_time-dependent_2005}.

\subsection{Discrete-Time Systems}
\label{subsec:prelim-discrete-time-systems}

Because the MPC methods are formulated in discrete time, a discrete-time model is also introduced. Let \(\Delta t>0\) denote the sampling period, which serves as the prediction interval of the MPC planners in \cref{ch:ntc-mpc,ch:rntc-mpc}, let \(t_k=t_0+k\Delta t\), and define \(x_k=x(t_k)\). The discrete-time dynamics are written as
\begin{equation}
  x_{k+1}
  =
  F_\Delta(x_k,u_k,d_k),
  \qquad
  x_k\in\mathcal{X},\quad
  u_k\in\mathcal{U},\quad
  d_k\in\mathcal{D}.
  \label{eq:prelim-general-discrete-dynamics}
\end{equation}
When no disturbance or adversarial input is modeled, this becomes
\begin{equation}
  x_{k+1}
  =
  F_\Delta(x_k,u_k).
  \label{eq:prelim-discrete-dynamics}
\end{equation}
The function \(F_\Delta\) may be obtained from a continuous-time model through exact discretization, numerical integration, system identification, or a task-specific approximation.

Continuous-time and discrete-time models are used side by side in this dissertation. Safety analysis via HJ reachability and safety filtering via CBFs operate on the continuous-time models \cref{eq:prelim-general-continuous-dynamics,eq:prelim-undisturbed-continuous-dynamics}, while the MPC planners of \cref{ch:ntc-mpc,ch:rntc-mpc} optimize over the discrete-time model \cref{eq:prelim-discrete-dynamics}. Safety information computed in continuous time enters the discrete-time problem through constraint functions evaluated at the sampling instants.

\section{Safety Constraints and Invariant Sets}
\label{sec:prelim-constraints-safety-conventions}

This section distinguishes instantaneous constraint satisfaction from dynamics-aware safety. It defines constraint and failure sets, specifies signed-distance collision measures, and introduces controlled invariance and the maximal safe set as the relevant notions of long-term safety.

\subsection{Constraint and Failure Sets}
\label{sec:prelim-constraint-set}

Let \(c:\mathcal{X}\rightarrow\mathbb{R}\) denote a \emph{constraint function}. Throughout this dissertation, \(c(x)\geq0\) indicates constraint satisfaction, and \(c(x)<0\) indicates failure. The corresponding \emph{constraint set} and \emph{failure set} are
\begin{equation}
  \mathcal{C}
  :=
  \{x\in\mathcal{X}\mid c(x)\geq0\},
  \qquad
  \mathcal{F}
  :=
  \{x\in\mathcal{X}\mid c(x)<0\}
  =
  \mathcal{X}\setminus\mathcal{C},
  \label{eq:prelim-constraint-and-failure-sets}
\end{equation}
respectively. The zero level \(c(x)=0\) therefore belongs to the constraint set. In a collision-avoidance context, the constraint function is a distance measure that already includes a physical inflation margin, so the boundary \(c(x)=0\) is a state of zero remaining margin rather than a state of physical contact.

Membership in \(\mathcal{C}\) describes instantaneous constraint satisfaction and does not imply dynamic safety. A trajectory that starts in a state \(x\in\mathcal{C}\) may still enter \(\mathcal{F}\) unavoidably because of the system dynamics, actuation limits, disturbances, or environment evolution. The braking example in \cref{subsec:prelim-controlled-invariant-safe-sets} illustrates this distinction.

\subsection{Signed-Distance Collision Constraints}
\label{sec:prelim-geometric-constraint-functions}

Throughout the dissertation, the constraint function primarily encodes collision avoidance, which is commonly described by a signed-distance function (SDF). Let \(\mathcal{A}\subset\mathbb{R}^{n_p}\) be a closed obstacle region in the robot's workspace, and define the Euclidean distance from a point \(y\in\mathbb{R}^{n_p}\) to a set \(\mathcal{Y}\subseteq\mathbb{R}^{n_p}\) by
\begin{equation}
  \operatorname{dist}(y,\mathcal{Y})
  :=
  \inf_{z\in\mathcal{Y}}
  \lVert y-z\rVert_2.
  \label{eq:prelim-distance-to-set}
\end{equation}
Under the sign convention used here, the signed distance to \(\mathcal{A}\) is
\begin{equation}
  \operatorname{sd}_{\mathcal{A}}(y)
  :=
  \operatorname{dist}(y,\mathcal{A})
  -
  \operatorname{dist}\bigl(y,\mathbb{R}^{n_p}\setminus\mathcal{A}\bigr).
  \label{eq:prelim-signed-distance}
\end{equation}
It is positive outside the obstacle, zero on its boundary, and negative in its interior. The signed-distance function is Lipschitz continuous and differentiable almost everywhere \cite{osher_level_2003}. Away from geometric singularities, it satisfies
\begin{equation}
  \left\lVert
  \nabla_y\operatorname{sd}_{\mathcal{A}}(y)
  \right\rVert_2
  =
  1.
  \label{eq:prelim-signed-distance-gradient-norm}
\end{equation}
An SDF therefore provides a distance measure and gradients for optimization-based planning and control. It can be computed efficiently from common perception outputs such as occupancy grids \cite{felzenszwalb_distance_2012,oleynikova_signed_2016-1,oleynikova_voxblox_2017}.

For a robot modeled as a point in the workspace with position \(p(x)\), a geometric collision-avoidance constraint with a required minimum distance \(r\geq 0\) can be defined as
\begin{equation}
  c(x)
  :=
  \operatorname{sd}_{\mathcal{A}}\bigl(p(x)\bigr)-r.
  \label{eq:prelim-point-robot-sdf-constraint}
\end{equation}
For a robot with state-dependent nonzero footprint \(\mathcal{R}(x)\subset\mathbb{R}^{n_p}\), one may instead use
\begin{equation}
  c(x)
  :=
  \inf_{y\in\mathcal{R}(x)}
  \operatorname{sd}_{\mathcal{A}}(y).
  \label{eq:prelim-footprint-sdf-constraint}
\end{equation}
Both constraints are purely geometric, since they measure the obstacle distance at one instant and do not account for how fast the robot moves or how quickly it can brake or turn. The remainder of the chapter addresses this limitation.

\subsection{Controlled-Invariant Safe Sets}
\label{subsec:prelim-controlled-invariant-safe-sets}

The dynamics-aware notion of safety is built from invariance. Consider first an autonomous system \(\dot{x}=f(x)\) with a unique solution for any initial state. A set \(\mathcal{S}\subseteq\mathcal{X}\) is \emph{forward invariant} for this system if every trajectory that starts in \(\mathcal{S}\) remains in \(\mathcal{S}\) for all future times. Forward invariance formalizes system viability, that is, the ability of the system to keep operating inside an admissible region.

For systems with a control input, the corresponding notion requires that \emph{some} admissible control achieve invariance. A set \(\mathcal{S}\subseteq\mathcal{C}\) is called a \emph{safe set} if it is \emph{controlled invariant}. For the undisturbed continuous-time system, this means
\begin{equation}
  \forall x\in\mathcal{S},
  \quad
  \exists u(\cdot)\in\mathbb{U}_{[t,\infty)}
  \quad\text{ s.t. }\quad
  \xi_{x,t}^{u}(\tau)\in\mathcal{S}
  \quad
  \forall \tau\geq t.
  \label{eq:prelim-continuous-safe-set}
\end{equation}
For the discrete-time system in \cref{eq:prelim-discrete-dynamics}, controlled invariance means
\begin{equation}
  \forall x_k\in\mathcal{S},
  \quad
  \exists u_k\in\mathcal{U}
  \quad\text{ s.t. }\quad
  F_\Delta(x_k,u_k)\in\mathcal{S}.
  \label{eq:prelim-discrete-safe-set}
\end{equation}
With disturbances, safety must hold against every admissible disturbance strategy. For nonanticipative strategies \(\Gamma_{[t,\infty)}\) of \cref{subsec:prelim-continuous-time-systems}, a set \(\mathcal{S}\) is \emph{robustly controlled invariant} if
\begin{equation}
  \forall x\in\mathcal{S},\quad
  \forall\gamma\in\Gamma_{[t,\infty)},\quad
  \exists u(\cdot)\in\mathbb{U}_{[t,\infty)}
  \quad\text{ s.t. }\quad
  \xi_{x,t}^{u,\gamma[u]}(\tau)\in\mathcal{S}
  \quad
  \forall\tau\geq t.
  \label{eq:prelim-robust-controlled-invariance}
\end{equation}

The \emph{maximal safe set}, denoted by \(\mathcal{S}^{\star}\), is the largest controlled-invariant, or robustly controlled-invariant, subset of \(\mathcal{C}\). For the undisturbed continuous-time system, it can be written as
\begin{equation}
  \mathcal{S}^{\star}
  :=
  \left\{
  x\in\mathcal{C}
  \;\middle|\;
  \exists u(\cdot)\in\mathbb{U}_{[t,\infty)}
  \text{ s.t. }
  \xi_{x,t}^{u}(\tau)\in\mathcal{C}
  \quad
  \forall \tau\geq t
  \right\}.
  \label{eq:prelim-maximal-safe-set}
\end{equation}
Under the adopted dynamics, constraints, and admissibility assumptions, \(\mathcal{S}^{\star}\) corresponds to the \emph{viability kernel} of \(\mathcal{C}\) studied in viability theory \cite{aubin_viability_1990,aubin_viability_2011,blanchini_survey_1999}. With disturbances, the analogous robust maximal safe set uses the quantifiers in \cref{eq:prelim-robust-controlled-invariance}. Every controlled-invariant or robustly controlled-invariant safe set \(\mathcal{S}\) satisfies
\begin{equation}
  \mathcal{S}
  \subseteq
  \mathcal{S}^{\star}
  \subseteq
  \mathcal{C}.
  \label{eq:prelim-safe-set-inclusions}
\end{equation}
The second inclusion is strict whenever some states satisfy the constraint now but are dynamically unable to avoid future failure, which is the typical situation in robot navigation.

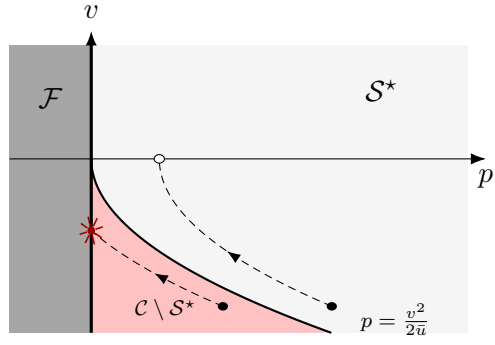
\begin{figure}[tb]
  \centering
  \colorlet{brtline}{red!62!black}
\colorlet{brtfill}{red!24}

\begin{tikzpicture}[font=\small, >={Latex[length=2mm]}, xscale=1.2, yscale=1.0,
    traj/.style={densely dashed, postaction={decorate,
          decoration={markings, mark=at position 0.5 with {\arrow{Latex[length=1.8mm]}}}}},
    crash/.pic={
      \foreach \a in {15,60,105,150,195,240,285,330}{
        \draw[brtline, line width=0.6pt] (\a:0.07) -- (\a:0.19);}
      \fill[brtline] (0,0) circle (1.7pt);
    }]
  \fill[black!4] (0,-2.3) rectangle (4.15,1.5);
  \fill[brtfill] (0,-2.3) -- plot[smooth, domain=-2.3:0, variable=\v] ({\v*\v/2}, \v) -- cycle;
  \fill[black!35] (-0.9,-2.3) rectangle (0,1.5);
  \draw[thick] plot[smooth, domain=-2.3:0, variable=\v] ({\v*\v/2}, \v);
  \draw[->] (-0.9,0) -- (4.35,0) node[below] {$p$};
  \draw[very thick,->] (0,-2.3) -- (0,1.7) node[above] {$v$};
  \fill (1.451,-1.95) circle (1.6pt);
  \draw[traj] plot[samples=60, domain=-1.95:-0.9487, variable=\v] ({-0.45+\v*\v/2}, \v);
  \pic[xscale={1/1.2}] at (0,-0.9487) {crash};
  \fill (2.651,-1.95) circle (1.6pt);
  \draw[traj] plot[samples=60, domain=-1.95:0, variable=\v] ({0.75+\v*\v/2}, \v);
  \draw[fill=white] (0.75,0) circle (1.6pt);
  \node at (-0.45,0.8) {$\mathcal{F}$};
  \node at (3.2,0.9) {$\mathcal{S}^{\star}$};
  \node[font=\scriptsize] at (0.82,-2.0) {$\mathcal{C}\setminus\mathcal{S}^{\star}$};
  \node[font=\scriptsize, anchor=west] at (2.85,-2.12) {$p=\tfrac{v^{2}}{2\bar{u}}$};
\end{tikzpicture}
  \caption[Braking example, constraint set and maximal safe set of the double integrator]{Braking example in the state space of the double integrator \(\dot{p}=v\), \(\dot{v}=u\) with \(\lvert u\rvert\leq\bar{u}\) and a wall at \(p<0\). The failure set \(\mathcal{F}\) (dark) collects the collision states, and the constraint set \(\mathcal{C}\) is its complement. The horizontal axis is the position \(p\) and the vertical axis is the velocity \(v\). Within \(\mathcal{C}\), the states to the left of the braking parabola \(p=v^{2}/(2\bar{u})\) (shaded red) are collision-free at the current instant but cannot avoid future collision, because their distance to the wall is smaller than the distance needed to brake. The maximal safe set \(\mathcal{S}^{\star}\) (light) excludes them. The dashed curves show two trajectories under maximal braking that start with the same approach speed, and both are horizontal translates of the braking parabola. The left one starts inside \(\mathcal{C}\setminus\mathcal{S}^{\star}\) and reaches the wall with nonzero speed (cross), while the right one starts in \(\mathcal{S}^{\star}\) and stops at a positive distance from the wall (open circle).}
  \label{fig:prelim-braking-sets}
\end{figure}

\paragraph{Braking-distance example.}
The following example makes the gap between \(\mathcal{C}\) and \(\mathcal{S}^{\star}\) concrete. Consider a robot moving along a line toward a wall. The robot is modeled as a double integrator with position \(p\), velocity \(v\), state \(x=(p,v)\), dynamics \(\dot{p}=v\), \(\dot{v}=u\), and actuation limit \(\lvert u\rvert\leq\bar{u}\). The wall occupies \(p<0\), so a natural constraint function is the distance to the wall, \(c(x)=p\), with \(\mathcal{F}=\{x\mid p<0\}\). Whether a state is safe in the dynamic sense depends on the velocity as well as on the position. A robot moving away from the wall (\(v\geq0\)) is safe whenever \(p\geq0\). A robot moving toward the wall (\(v<0\)) must be at least its braking distance \(v^{2}/(2\bar{u})\) away from the wall, because no admissible input can stop it sooner. The maximal safe set is therefore
\begin{equation}
  \mathcal{S}^{\star}
  =
  \bigl\{(p,v)\mid v\geq0,\;p\geq0\bigr\}
  \cup
  \Bigl\{(p,v)\;\Big|\;v<0,\;p-\tfrac{v^{2}}{2\bar{u}}\geq0\Bigr\},
  \label{eq:prelim-braking-maximal-safe-set}
\end{equation}
which is a strict subset of \(\mathcal{C}=\{p\geq0\}\), as illustrated in \cref{fig:prelim-braking-sets}. The states in \(\mathcal{C}\setminus\mathcal{S}^{\star}\) are exactly the inevitable collision states discussed in \cref{ch:introduction}, since they are geometrically collision-free but dynamically already unsafe. For this two-dimensional example, the boundary of \(\mathcal{S}^{\star}\) can be derived by hand. For realistic robot models with more state variables, nonlinear dynamics, and irregular obstacle geometry, a computational framework is needed, which \cref{sec:prelim-hj-reachability} presents after the two controller frameworks that use the resulting sets.

\section{Model Predictive Control}
\label{sec:prelim-mpc}

Model predictive control is the first of the two controller frameworks used in this dissertation. At every sampling instant, the controller uses the dynamics model to predict the robot's motion a short time into the future and optimizes this predicted motion with respect to a cost function and constraints. It applies only the first control input of the optimized plan and repeats the procedure at the next instant with updated measurements \cite{mayne_constrained_2000,rawlings_model_2017,schwenzer_review_2021}. Replanning at every step makes MPC a feedback controller despite its planning character, and the explicit constraints make it well suited to safety-critical navigation. This section states the generic problem and then isolates the terminal constraint set as the component responsible for safety beyond the prediction horizon.

Consider the discrete-time model in \cref{eq:prelim-discrete-dynamics}. At time step \(k\), the MPC prediction variables are
\begin{equation}
  x_{0\mid k},x_{1\mid k},\ldots,x_{N\mid k},
  \qquad
  u_{0\mid k},u_{1\mid k},\ldots,u_{N-1\mid k},
\end{equation}
where \(N\) is the length of the prediction horizon, \(x_{i\mid k}\) denotes the state predicted for time step \(k+i\) using the information available at time \(k\), and \(x_{0\mid k}=x_k\). The predicted dynamics satisfy
\begin{equation}
  x_{i+1\mid k}
  =
  F_\Delta(x_{i\mid k},u_{i\mid k}),
  \qquad
  i=0,\ldots,N-1.
  \label{eq:prelim-mpc-predicted-dynamics}
\end{equation}

A generic MPC problem is
\begin{subequations}
  \label{eq:prelim-mpc-problem}
  \begin{align}
    \min_{\substack{x_{0\mid k},\ldots,x_{N\mid k} \\
    u_{0\mid k},\ldots,u_{N-1\mid k}}}
    \quad &
    \ell_{f\mid k}(x_{N\mid k})
    +
    \sum_{i=0}^{N-1}
    \ell_{i\mid k}(x_{i\mid k},u_{i\mid k})
    \label{eq:prelim-mpc-objective}
    \\
    \text{ s.t. }\quad
          &
    x_{0\mid k}=x_k,
    \label{eq:prelim-mpc-initial-condition}
    \\
          &
    x_{i+1\mid k}
    =
    F_\Delta(x_{i\mid k},u_{i\mid k}),
    \quad
    i=0,\ldots,N-1,
    \label{eq:prelim-mpc-dynamics}
    \\
          &
    u_{i\mid k}\in\mathcal{U},
    \quad
    i=0,\ldots,N-1,
    \label{eq:prelim-mpc-input-constraints}
    \\
          &
    x_{i\mid k}\in\mathcal{C}_{i\mid k},
    \quad
    i=0,\ldots,N-1,
    \label{eq:prelim-mpc-stage-constraint-set}
    \\
          &
    x_{N\mid k}\in\mathcal{C}_{f\mid k}.
    \label{eq:prelim-mpc-terminal-constraint-set}
  \end{align}
\end{subequations}
Here, \(\ell_{i\mid k}\) and \(\ell_{f\mid k}\) are the stage and terminal costs, respectively. The set \(\mathcal{C}_{i\mid k}\) is the \emph{stage constraint set} predicted for stage \(i\) using the information available at time \(k\). In navigation, it typically encodes collision avoidance through the signed-distance constraints of \cref{sec:prelim-geometric-constraint-functions}. In a time-invariant, fully observed setting, it reduces to the constant set \(\mathcal{C}\). In general, it may depend on \(i\) because the environment is predicted to evolve, and on \(k\) because the prediction is replaced when new information becomes available. The quadratic tracking objective used throughout this dissertation is
\begin{align}
  \ell_{i\mid k}(x,u)
   & =
  (x-x_{i\mid k}^{\mathrm{ref}})^{\top}
  Q
  (x-x_{i\mid k}^{\mathrm{ref}})
  +
  (u-u_{i\mid k}^{\mathrm{ref}})^{\top}
  R
  (u-u_{i\mid k}^{\mathrm{ref}}),
  \label{eq:prelim-mpc-quadratic-stage-cost}
  \\
  \ell_{f\mid k}(x)
   & =
  (x-x_{N\mid k}^{\mathrm{ref}})^{\top}
  Q_f
  (x-x_{N\mid k}^{\mathrm{ref}}),
  \label{eq:prelim-mpc-quadratic-terminal-cost}
\end{align}
where \(x_{i\mid k}^{\mathrm{ref}}\) and \(u_{i\mid k}^{\mathrm{ref}}\) are the stage reference values, \(x_{N\mid k}^{\mathrm{ref}}\) is the reference state at the end of the prediction horizon, and \(Q,Q_f\succeq 0\) and \(R\succ 0\) are weighting matrices.

\paragraph{Terminal constraint set.}
The set \(\mathcal{C}_{f\mid k}\) in \cref{eq:prelim-mpc-terminal-constraint-set} is the \emph{terminal constraint set}. It restricts the final predicted state and is the only constraint of the optimization problem \cref{eq:prelim-mpc-problem} through which system behavior beyond the prediction horizon can be enforced. The terminal cost \(\ell_{f\mid k}\) also reflects post-horizon behavior, but it only penalizes the final state and cannot exclude an inadmissible one. Increasing \(N\) does not resolve this limitation, since the computational cost grows with the horizon length and a finite horizon alone does not certify safety for all future times. The terminal set is chosen such that reaching it guarantees the desired post-horizon property, such as recursive feasibility, stability, or safety, and common choices include invariant sets and reachability-based approximations. Like the stage constraints, the terminal set may depend on the information available at time \(k\). In a time-invariant setting, the shorter notation \(\mathcal{C}_f\) is used.

The central property associated with the terminal set is \emph{recursive feasibility}, which means that if the MPC problem is feasible at some time step, it remains feasible at all subsequent steps, so that the controller always has an admissible plan. For time-invariant constraints, a standard sufficient condition is the existence of a terminal controller \(\kappa_f\) satisfying
\begin{equation}
  \kappa_f(x)\in\mathcal{U},
  \qquad
  F_\Delta\bigl(x,\kappa_f(x)\bigr)\in\mathcal{C}_f,
  \qquad
  \forall x\in\mathcal{C}_f,
  \label{eq:prelim-mpc-terminal-invariance}
\end{equation}
together with \(\mathcal{C}_f\subseteq\mathcal{C}\). In the terminology of \cref{subsec:prelim-controlled-invariant-safe-sets}, condition \cref{eq:prelim-mpc-terminal-invariance} requires \(\mathcal{C}_f\) to be a controlled-invariant safe set. Recursive feasibility then follows from a shifting argument, in which the optimal input sequence at time \(k\), shifted by one step and extended with \(\kappa_f\), is a feasible candidate at time \(k+1\) \cite{mayne_constrained_2000,kerrigan_invariant_2000,rawlings_model_2017}. Because any controlled-invariant safe set satisfies this condition, larger terminal sets are preferred, as they restrict the planner less. The ideal terminal set is the maximal safe set \(\mathcal{S}^{\star}\), which has also been used directly in viability-based predictive control \cite{liniger_viability_2015}.

When the constraints are time-varying or observation-dependent, this argument no longer applies, because the terminal set \(\mathcal{C}_{f\mid k}\) itself changes from one time step to the next. The shifted candidate sequence must then be admissible for the problem solved at time \(k+1\), whose terminal set \(\mathcal{C}_{f\mid k+1}\) is generally different from \(\mathcal{C}_{f\mid k}\). Invariance of each individual terminal set is not sufficient. Recursive feasibility additionally requires a compatibility condition linking successive terminal sets, and it presumes that the prediction model and the information update are consistent with the true system and its sensing process. What these requirements imply when the terminal set is produced by a learned model is discussed in \cref{subsec:prelim-learned-guarantee-scope}. The remaining challenge, addressed in \cref{ch:ntc-mpc,ch:rntc-mpc}, is to construct a suitable terminal set from obstacle geometry observed only at runtime.

\section{Control Barrier Functions}
\label{sec:prelim-cbf}

The second control framework uses a safety filter to supervise a nominal controller that handles the task objective. Regardless of the controller's design, the filter modifies the proposed control command as little as necessary to satisfy the safety condition. Control barrier functions are a widely used tool for constructing such filters. Their modern control form was introduced in \cite{wieland_constructive_2007,ames_control_2014}, and comprehensive surveys are provided in \cite{ames_control_2019,garg_advances_2024}. This section introduces the scalar function \(h\) and the candidate safe set it represents, states the differential condition that makes \(h\) a control barrier function, and presents the resulting invariance guarantee in relation to the safety conventions of \cref{sec:prelim-constraints-safety-conventions}.

The standard first-order CBF formulation is stated for the control-affine system
\begin{equation}
  \dot{x}
  =
  f_0(x)+g(x)u,
  \qquad
  x\in\mathcal{X},\quad
  u\in\mathcal{U},
  \label{eq:prelim-control-affine-dynamics}
\end{equation}
where \(f_0:\mathcal{X}\rightarrow\mathbb{R}^{n_x}\) and \(g:\mathcal{X}\rightarrow\mathbb{R}^{n_x\times n_u}\) are locally Lipschitz. The control-affine structure covers a broad class of robot models encountered in practice, including those considered in this dissertation, and makes the constraints below linear in the control input.

The starting point is a scalar function \(h:\mathcal{X}\rightarrow\mathbb{R}\), continuously differentiable on an open set containing \(\mathcal{X}\), that implicitly represents the region to be kept safe. As with the constraint function \(c\) of \cref{sec:prelim-constraints-safety-conventions}, its sign classifies states and its zero level set forms the boundary. The set certified by the standard CBF theory is the nonnegative superlevel set
\begin{equation}
  \mathcal{S}_h
  :=
  \{x\in\mathcal{X}\mid h(x)\geq0\},
  \label{eq:prelim-cbf-closed-set}
\end{equation}
and the control objective is to keep the state in \(\mathcal{S}_h\) for all time. Because trajectories of \cref{eq:prelim-control-affine-dynamics} are continuous, the state can leave \(\mathcal{S}_h\) only by crossing its boundary, where \(h=0\). By Nagumo's classical theorem on set invariance, the set is rendered invariant if, wherever \(h(x)=0\), some admissible input prevents \(h\) from decreasing further \cite{nagumo_uber_1942,blanchini_survey_1999}. However, enforcing such a requirement only on the boundary is impractical, because it constrains the input on a set of measure zero and allows the system to approach the boundary at full speed. The CBF condition extends the requirement to the entire domain, allowing \(h\) to decrease away from the boundary but only at a regulated rate. The regulation is expressed by an extended class-\(\mathcal{K}\) function, defined next.

\paragraph{Extended class-\(\mathcal{K}\) functions.}
A function \(\alpha:I\rightarrow\mathbb{R}\), defined on an interval \(I\subseteq\mathbb{R}\) containing the origin, is an \emph{extended class-\(\mathcal{K}\) function} if it is continuous and strictly increasing and satisfies \(\alpha(0)=0\). A simple example is the linear function \(\alpha(s)=\gamma s\) with a constant gain \(\gamma>0\), which is the choice used by the safety filters in this dissertation. In the CBF condition below, \(\alpha\) regulates how fast the barrier value may decrease. Far from the boundary, where \(h\) is large, a fast decrease is permitted, while at the boundary, where \(h=0\), the barrier value must not decrease at all.

\paragraph{Control barrier function.}
The function \(h\) is a \emph{control barrier function} for \(\mathcal{S}_h\) if there exists an extended class-\(\mathcal{K}\) function \(\alpha\), defined on an interval containing the range of \(h\), such that
\begin{equation}
  \sup_{u\in\mathcal{U}}
  \left[
    L_{f_0}h(x)+L_gh(x)u
    \right]
  \geq
  -\alpha\bigl(h(x)\bigr),
  \qquad
  \forall x\in\mathcal{X},
  \label{eq:prelim-cbf-condition}
\end{equation}
where \(L_{f_0}h(x)\) and \(L_{g}h(x)\) are the Lie derivatives of \(h\) along \(f_0\) and \(g\), defined as
\begin{equation}
  L_{f_0}h(x)
  :=
  \nabla_xh(x)^{\top}f_0(x),
  \qquad
  L_gh(x)
  :=
  \nabla_xh(x)^{\top}g(x).
  \label{eq:prelim-lie-derivatives}
\end{equation}
The supremum in \cref{eq:prelim-cbf-condition} is taken over the admissible input set \(\mathcal{U}\), so a valid CBF certifies that safety is achievable \emph{within the actuation limits}. Equivalently, the CBF-admissible input set at state \(x\) is
\begin{equation}
  K_{\mathrm{cbf}}(x)
  :=
  \left\{
  u\in\mathcal{U}
  \;\middle|\;
  L_{f_0}h(x)+L_gh(x)u+\alpha\bigl(h(x)\bigr)\geq0
  \right\}.
  \label{eq:prelim-cbf-admissible-inputs}
\end{equation}

\paragraph{Forward invariance.}
Assume that \(\mathcal{S}_h\) is nonempty and that zero is a regular value of \(h\), meaning that \(\nabla_x h(x)\neq0\) wherever \(h(x)=0\). Under the standard existence and uniqueness assumptions, any locally Lipschitz feedback law \(u=\kappa(x)\) satisfying
\begin{equation}
  \kappa(x)\in K_{\mathrm{cbf}}(x),
  \qquad
  \forall x\in\mathcal{S}_h,
  \label{eq:prelim-cbf-feedback-condition}
\end{equation}
renders the closed set \(\mathcal{S}_h\) forward invariant \cite{ames_control_2014,ames_control_2019}.

\paragraph{Certified set and its size.}
For the invariant set \(\mathcal{S}_h\) to represent physical safety, it must exclude every failure state,
\begin{equation}
  \mathcal{S}_h
  \subseteq
  \mathcal{C}.
  \label{eq:prelim-cbf-set-inclusion-requirement}
\end{equation}
Under \cref{eq:prelim-cbf-set-inclusion-requirement}, the set \(\mathcal{S}_h\) is a safe set in the sense of \cref{subsec:prelim-controlled-invariant-safe-sets}, because from every initial state in \(\mathcal{S}_h\) the feedback law \(\kappa\) generates an admissible input whose trajectory remains in \(\mathcal{S}_h\). The inclusion \cref{eq:prelim-safe-set-inclusions} then gives
\begin{equation}
  \mathcal{S}_h
  \subseteq
  \mathcal{S}^{\star}
  \subseteq
  \mathcal{C}.
  \label{eq:prelim-cbf-maximal-safe-relation}
\end{equation}
A valid barrier can therefore certify at most the maximal safe set. A larger certified set \(\mathcal{S}_h\) admits more initial states and can reduce conservatism, but intervention frequency also depends on the barrier parameterization and the filter design.

\paragraph{Relative degree.}
At states where \(L_gh(x)=0\), the input does not appear in the constraint defining \(K_{\mathrm{cbf}}(x)\), and the CBF inequality must hold through the drift term \(L_{f_0}h(x)\) alone. For barrier candidates whose input dependence appears only in higher time derivatives, the first-order condition is insufficient. Examples include purely position-based distance functions for systems with acceleration inputs. Exponential and higher-order CBF constructions restore an enforceable, input-dependent condition \cite{nguyen_exponential_2016,xiao_high-order_2022}. The reachability-based barriers used in this dissertation approximate value functions that depend on the full state, so \(L_gh(x)\) is generically nonzero and the first-order condition \cref{eq:prelim-cbf-condition} applies directly.

\subsection{Time-Varying and Robust Barrier Functions}
\label{subsec:prelim-time-varying-robust-cbfs}

When obstacles move, the safe region itself changes over time, and the barrier must account for this motion. For a continuously differentiable time-varying barrier \(h:\mathcal{X}\times\mathbb{R}\rightarrow\mathbb{R}\), define the certified set \(\mathcal{S}_h(t):=\{x\in\mathcal{X}\mid h(x,t)\geq0\}\). The CBF condition becomes
\begin{equation}
  \sup_{u\in\mathcal{U}}
  \left[
    \frac{\partial h}{\partial t}(x,t)
    +
    L_{f_0}h(x,t)
    +
    L_gh(x,t)u
    \right]
  \geq
  -\alpha\bigl(h(x,t)\bigr).
  \label{eq:prelim-time-varying-cbf-condition}
\end{equation}
The partial derivative accounts for modeled continuous motion of the represented safe set, for instance when an obstacle moves \cite{igarashi_time-varying_2019}. Under the same regularity assumptions as in the time-invariant case, a feedback law satisfying \cref{eq:prelim-time-varying-cbf-condition} renders the time-varying certified set \(\mathcal{S}_h(t)\) forward invariant.

For dynamics with an adversarial disturbance, a sufficient robust condition requires the inequality to hold for the worst admissible disturbance \cite{jankovic_robust_2018} and reads
\begin{equation}
  \sup_{u\in\mathcal{U}}
  \inf_{d\in\mathcal{D}}
  \left[
    \frac{\partial h}{\partial t}(x,t)
    +
    \nabla_xh(x,t)^\top f(x,u,d)
    \right]
  \geq
  -\alpha\bigl(h(x,t)\bigr).
  \label{eq:prelim-robust-cbf-condition}
\end{equation}
The technical chapters specify how time dependence and obstacle motion enter their particular barrier constructions.

\subsection{CBF-QP Safety Filtering}
\label{subsec:prelim-cbf-qp-safety-filter}

The practical appeal of CBFs is that the safety condition for control-affine systems can be enforced by a simple optimization problem solved at a high rate. Given a nominal control command \(u_{\mathrm{nom}}\), the filtered control input is computed by the quadratic program (QP)
\begin{subequations}
  \label{eq:prelim-cbf-qp}
  \begin{align}
    u^\star
    =
    \operatorname*{arg\,min}_{u\in\mathcal{U}}
    \quad              &
    \frac{1}{2}\lVert u-u_{\mathrm{nom}}\rVert_2^2
    \label{eq:prelim-cbf-qp-objective}
    \\
    \text{ s.t. }\quad &
    L_{f_0}h(x)
    +
    L_gh(x)u
    +
    \alpha\bigl(h(x)\bigr)
    \geq0.
    \label{eq:prelim-cbf-qp-constraint}
  \end{align}
\end{subequations}
When \(\mathcal{U}\) is convex and described by linear inequalities, the CBF-QP is convex because the barrier constraint is affine in \(u\). Under suitable regularity conditions, its solution is locally Lipschitz in the state, so the closed loop is well posed \cite{ames_control_2019}. The filter is \emph{minimally invasive} in the sense that the nominal command is passed through unchanged whenever it already satisfies the barrier constraint. The resulting safety guarantee rests on three premises. The function \(h\) satisfies the CBF condition \cref{eq:prelim-cbf-condition} under the bounded input set, the QP remains feasible along the closed-loop trajectory, and the certified set contains no failure states, as required by \cref{eq:prelim-cbf-set-inclusion-requirement}. The extent to which learned barriers satisfy these premises is discussed in \cref{subsec:prelim-learned-guarantee-scope}.

At states where no admissible input satisfies the barrier constraint, the problem \cref{eq:prelim-cbf-qp} has no solution and the filter returns no input. In practice, the barrier constraint is relaxed by a nonnegative slack variable \(\varsigma\) that is penalized in the objective,
\begin{equation}
  \min_{u\in\mathcal{U},\,\varsigma\geq0}
  \quad
  \frac{1}{2}\lVert u-u_{\mathrm{nom}}\rVert_2^2
  +
  \varrho\,\varsigma^{2},
  \qquad
  \text{ s.t. }\quad
  L_{f_0}h(x)+L_gh(x)u+\alpha\bigl(h(x)\bigr)\geq-\varsigma,
  \label{eq:prelim-cbf-qp-slack}
\end{equation}
with a penalty weight \(\varrho>0\). The relaxed problem always has a solution, and this solution coincides with that of \cref{eq:prelim-cbf-qp} whenever the barrier constraint is inactive at the solution of the hard problem, for example when the nominal command already satisfies it. When the constraint is active, the quadratic penalty generally leaves a small positive slack for every finite weight, and this slack decreases as the weight grows. A positive slack records a violation of the barrier constraint under the chosen penalty weight, but it does not by itself show that no admissible input satisfies the hard constraint. The invariance guarantee applies only while the slack is zero. The safety filters of \cref{ch:orn-cbf,ch:cn-cbf,ch:vertex-cbf} use this relaxation, with the barrier constraint relaxed and the input limits kept as hard constraints.

The filter requires a valid dynamics-aware barrier for the current environment. Constructing one is difficult for general nonlinear systems, especially when geometry is revealed only at runtime \cite{hsu_safety_2024,wabersich_data-driven_2023}. \Cref{ch:orn-cbf,ch:cn-cbf} learn observation-dependent barrier candidates from reachability analysis and examine which guarantees they retain.

\section{Hamilton--Jacobi Reachability Analysis}
\label{sec:prelim-hj-reachability}

The two preceding sections showed that controller safety depends on a suitable set. MPC requires a controlled-invariant terminal set, and safety filtering requires a valid barrier with a large certified set. Hamilton--Jacobi reachability analysis characterizes and computes such sets. Grounded in optimal control theory and dynamic programming, it determines a \emph{value function} for specified dynamics, actuation limits, constraints, and disturbance assumptions. The sign of this function classifies every state, with nonnegative values where failure can be avoided and negative values where it cannot \cite{bardi_optimal_1997,lygeros_reachability_2004,mitchell_time-dependent_2005,fisac_reach-avoid_2015,bansal_hamilton-jacobi_2017}. Although the framework accommodates a variety of problem specifications \cite{herbert_safe_real-world_2020}, this dissertation focuses on safety analysis where the target set is the failure set to be avoided. The finite-horizon formulation is developed first, followed by its infinite-horizon limit, which recovers the maximal safe set, the extension to time-varying constraints, and the numerical cost that motivates the learned representations of \cref{sec:prelim-neural-hypernetworks}.

\subsection{Finite-Horizon Avoidance}
\label{subsec:prelim-finite-horizon-hj}

Consider the dynamics in \cref{eq:prelim-general-continuous-dynamics}, the time-invariant constraint function \(c\), and the associated constraint and failure sets \(\mathcal{C}\) and \(\mathcal{F}\) defined in \cref{eq:prelim-constraint-and-failure-sets}. Finite-horizon avoidance characterizes the initial states from which the control can keep the system out of \(\mathcal{F}\) over the entire horizon \([t,T]\), regardless of the disturbance. To answer this question, define the \emph{reachability payoff} as the smallest constraint value attained along the trajectory,
\begin{equation}
  J\bigl(x,t,u(\cdot),d(\cdot)\bigr)
  :=
  \min_{\tau\in[t,T]}
  c\bigl(\xi_{x,t}^{u,d}(\tau)\bigr),
  \label{eq:prelim-hj-payoff}
\end{equation}
which is nonnegative exactly when the trajectory never enters the failure set. Using the nonanticipative disturbance strategies \(\Gamma_{[t,T]}\) of \cref{subsec:prelim-continuous-time-systems}, the optimal value of the avoidance game is
\begin{equation}
  V(x,t)
  :=
  \inf_{\gamma\in\Gamma_{[t,T]}}
  \sup_{u(\cdot)\in\mathbb{U}_{[t,T]}}
  J\bigl(x,t,u(\cdot),\gamma[u](\cdot)\bigr).
  \label{eq:prelim-finite-horizon-value-function}
\end{equation}
In this differential game, the controller seeks to maximize the minimum constraint margin over the time horizon, while the disturbance seeks to minimize it using a nonanticipative strategy. For an undisturbed system, the infimum over disturbance strategies is omitted. The set of initial states from which failure cannot be avoided, even under optimal control, forms the \emph{backward reachable tube} (BRT), defined as
\begin{equation}
  \mathcal{B}(t)
  :=
  \left\{
  x\in\mathcal{X}
  \;\middle|\;
  \exists \gamma\in\Gamma_{[t,T]},\;
  \forall u(\cdot)\in\mathbb{U}_{[t,T]},\;
  \exists \tau\in[t,T]
  \text{ s.t. }
  \xi_{x,t}^{u,\gamma[u]}(\tau)\in\mathcal{F}
  \right\}.
  \label{eq:prelim-finite-horizon-brt}
\end{equation}
In navigation terms, the BRT is the region around an obstacle from which a collision can no longer be prevented within the horizon, for instance because the braking distance exceeds the remaining distance to the obstacle.

The value function in \cref{eq:prelim-finite-horizon-value-function} can be computed by dynamic programming. Let \(p\in\mathbb{R}^{n_x}\) denote the costate argument. Under the adopted control--disturbance order, the Hamiltonian of the system is defined as
\begin{equation}
  H(x,p)
  :=
  \max_{u\in\mathcal{U}}
  \min_{d\in\mathcal{D}}
  p^{\top}f(x,u,d).
  \label{eq:prelim-hji-hamiltonian}
\end{equation}
For the undisturbed dynamics in \cref{eq:prelim-undisturbed-continuous-dynamics}, it reduces to
\begin{equation}
  H(x,p)
  :=
  \max_{u\in\mathcal{U}}
  p^{\top}f(x,u).
  \label{eq:prelim-hjb-hamiltonian}
\end{equation}
Because the payoff takes a minimum over time, the value function is generally not differentiable everywhere and is understood in the sense of viscosity solutions \cite{crandall_viscosity_1983,evans_differential_1984,bardi_optimal_1997}. Under this interpretation, the value function satisfies the Hamilton--Jacobi--Isaacs variational inequality
\begin{equation}
  \min
  \left\{
  \frac{\partial V}{\partial t}(x,t)
  +
  H\bigl(x,\nabla_xV(x,t)\bigr),
  \;
  c(x)-V(x,t)
  \right\}
  =
  0,
  \qquad
  V(x,T)=c(x),
  \label{eq:prelim-hji-vi-finite-horizon}
\end{equation}
whose obstacle term \(c(x)-V(x,t)\) encodes the running minimum in the payoff \cite{barron_bellman_1989,mitchell_time-dependent_2005,fisac_reach-avoid_2015}. The BRT and the corresponding finite-horizon avoidance set are the negative and the nonnegative level sets of \(V\), so that
\begin{equation}
  \mathcal{B}(t)
  =
  \{x\in\mathcal{X}\mid V(x,t)<0\},
  \qquad
  \mathcal{S}_V(t)
  :=
  \{x\in\mathcal{X}\mid V(x,t)\geq0\}
  =
  \mathcal{X}\setminus\mathcal{B}(t).
  \label{eq:prelim-brt-value-function}
\end{equation}
The set \(\mathcal{S}_V(t)\) contains the states from which the modeled failure set can be avoided over \([t,T]\), despite the worst-case disturbance. Since the payoff includes the initial time, \(\mathcal{S}_V(t)\subseteq\mathcal{C}\), and this inclusion is typically strict for the reason illustrated by the braking example. Unlike the infinite-horizon maximal safe set, \(\mathcal{S}_V(t)\) is not generally controlled invariant beyond the terminal time \(T\).

The obstacle term in \cref{eq:prelim-hji-vi-finite-horizon} implies the pointwise ordering
\begin{equation}
  V(x,t)\leq c(x),
  \qquad
  \forall x\in\mathcal{X},\quad
  \forall t\leq T.
  \label{eq:prelim-hj-value-constraint-ordering}
\end{equation}
The value function is therefore never larger than the instantaneous constraint value, because it additionally accounts for what the dynamics may make unavoidable. This ordering defines the nonnegative residual function
\begin{equation}
  r(x,t)
  :=
  c(x)-V(x,t),
  \qquad
  r(x,t)
  \geq
  0.
  \label{eq:prelim-hj-residual-decomposition}
\end{equation}
The residual measures how much the dynamics reduce the safety margin relative to the instantaneous constraint. When \(c\) is a distance field, it can be interpreted as a dynamics-induced loss of geometric distance to the obstacles. This exact ordering underlies the residual representations of \cref{subsec:prelim-structure-preserving-residual}.

\subsection{The Infinite-Horizon Maximal Safe Set}
\label{subsec:prelim-infinite-horizon-hj}
\label{subsec:prelim-stationary-hj}

Finite-horizon avoidance answers a time-limited question, while the controllers of \cref{sec:prelim-mpc,sec:prelim-cbf} require sets in which the robot can remain indefinitely. Although \(c\), \(\mathcal{C}\), and \(\mathcal{F}\) are time invariant, the finite-horizon value function depends on the time remaining until the terminal time \(T\). When the backward propagation converges as the horizon tends to infinity, the limit defines the stationary infinite-horizon value function
\begin{equation}
  V_\infty(x)
  :=
  \lim_{t\rightarrow-\infty}V(x,t).
  \label{eq:prelim-stationary-hj-value}
\end{equation}
The limiting value function satisfies the stationary variational inequality
\begin{equation}
  \min
  \left\{
  H\bigl(x,\nabla_xV_\infty(x)\bigr),
  \;
  c(x)-V_\infty(x)
  \right\}
  =
  0.
  \label{eq:prelim-stationary-hji-vi}
\end{equation}
Its nonnegative superlevel set is
\begin{equation}
  \mathcal{S}_{V_\infty}
  :=
  \{x\in\mathcal{X}\mid V_\infty(x)\geq0\}.
  \label{eq:prelim-stationary-hj-safe-set}
\end{equation}
Under matching dynamics, constraint sets, admissible inputs, disturbance assumptions, and control--disturbance quantifier order, \(\mathcal{S}_{V_\infty}\) coincides with the maximal safe set introduced in \cref{eq:prelim-maximal-safe-set}, that is,
\begin{equation}
  \mathcal{S}_{V_\infty}
  =
  \mathcal{S}^{\star}.
  \label{eq:prelim-hj-maximal-safe-set}
\end{equation}
The maximal safe set is the ideal target for an MPC terminal constraint or CBF candidate under the modeled dynamics, actuation limits, and disturbances. When it is used in a discrete-time MPC problem, the discretized model must be consistent with the continuous-time model of the HJ solution, as stated in \cref{subsec:prelim-discrete-time-systems}.

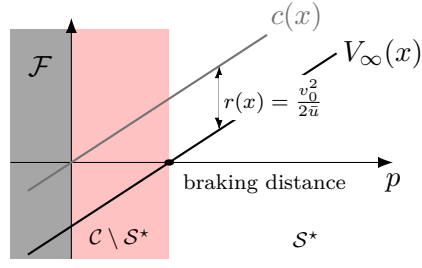
\begin{figure}[tb]
  \centering
  \colorlet{brtfill}{red!24}

\begin{tikzpicture}[font=\small, >={Latex[length=2mm]}, xscale=1.15, yscale=0.75]
  \fill[brtfill] (0,-1.7) rectangle (1.125,2.35);
  \fill[black!35] (-0.7,-1.7) rectangle (0,2.35);
  \draw[->] (-0.7,0) -- (3.7,0) node[below] {$p$};
  \draw[->] (0,-1.7) -- (0,2.55);
  \draw[thick, black!55] (-0.5,-0.5) -- (2.25,2.25) node[anchor=south west, inner sep=1pt] {$c(x)$};
  \draw[thick] (-0.5,-1.625) -- (3.05,1.925) node[anchor=west, inner sep=2pt] {$V_{\infty}(x)$};
  \draw[{Latex[length=1.6mm]}-{Latex[length=1.6mm]}] (1.7,1.7) -- (1.7,0.575)
  node[midway, right, font=\scriptsize, fill=white, inner sep=1.5pt] {$r(x)=\tfrac{v_0^{2}}{2\bar{u}}$};
  \fill (1.125,0) circle (1.6pt);
  \node[below right, font=\scriptsize, inner sep=2pt] at (1.22,-0.1) {braking distance};
  \node at (-0.36,1.7) {$\mathcal{F}$};
  \node[font=\scriptsize] at (0.56,-1.35) {$\mathcal{C}\setminus\mathcal{S}^{\star}$};
  \node[font=\scriptsize] at (2.7,-1.35) {$\mathcal{S}^{\star}$};
\end{tikzpicture}
  \caption[Braking example, constraint function and stationary value function]{Braking example, continued. Constraint function \(c\) and stationary value function \(V_\infty\) along a slice of the state space with constant approach speed \(v_0<0\). The value function runs below the constraint function by exactly the braking distance \(v_0^{2}/(2\bar{u})\), which is the residual \(r=c-V_\infty\) of \cref{eq:prelim-hj-residual-decomposition}. Its zero crossing is shifted by the braking distance in front of the wall, and the shaded region between the two zero crossings contains the states that satisfy the constraint but are dynamically unable to avoid collision.}
  \label{fig:prelim-braking-values}
\end{figure}

\paragraph{Braking-distance example continued.}
For the double integrator of \cref{fig:prelim-braking-sets}, the stationary value function can be written in closed form. With \(c(x)=p\), the optimal control strategy for negative velocity is maximal braking, so the smallest future distance to the wall from state \((p,v)\) is
\begin{equation}
  V_\infty(p,v)
  =
  \begin{cases}
    p,                         & v\geq0, \\[2pt]
    p-\dfrac{v^{2}}{2\bar{u}}, & v<0,
  \end{cases}
  \label{eq:prelim-braking-value-function}
\end{equation}
and its nonnegative superlevel set is exactly the maximal safe set in \cref{eq:prelim-braking-maximal-safe-set}. The residual \(r(x)=c(x)-V_\infty(x)\) equals the braking distance \(v^{2}/(2\bar{u})\) for \(v<0\) and vanishes for \(v\geq0\). It is a purely dynamics-induced quantity, independent of the wall position. \Cref{fig:prelim-braking-values} visualizes \(c\), \(V_\infty\), and \(r\) along a slice with constant approach speed. The learned representations of \cref{subsec:prelim-structure-preserving-residual} use the same decomposition of dynamic safety into geometric obstacle distance minus a dynamics-induced correction.

\subsection{Time-Varying Constraints}
\label{subsec:prelim-time-varying-hj}

The constraint function may depend explicitly on time, \(c:\mathcal{X}\times[t,T]\rightarrow\mathbb{R}\), to represent moving obstacles or other constraints with specified evolution over the horizon \cite{fisac_reach-avoid_2015}. This differs from replacing an observation-conditioned constraint when new information arrives, as discussed in \cref{sec:prelim-environment-conditioning}. The finite-horizon construction extends by replacing \(c(x)\) with \(c(x,\tau)\), so that
\begin{equation}
  J\bigl(x,t,u(\cdot),d(\cdot)\bigr)
  :=
  \min_{\tau\in[t,T]}
  c\!\left(
  \xi_{x,t}^{u,d}(\tau),
  \tau
  \right),
  \label{eq:prelim-time-varying-hj-payoff}
\end{equation}
while the value retains the control--disturbance optimization in \cref{eq:prelim-finite-horizon-value-function}. The variational inequality likewise replaces its obstacle term by \(c(x,t)-V(x,t)\) and uses the terminal condition \(V(x,T)=c(x,T)\). Hence \(V(x,t)\leq c(x,t)\) and \(r(x,t):=c(x,t)-V(x,t)\geq0\) remain valid. If only the constraint is time varying, the Hamiltonian is unchanged. Time-varying dynamics or input sets instead result in a time-dependent Hamiltonian \(H(x,p,t)\).

A stationary infinite-horizon limit need not exist when the constraint geometry continues to evolve. Time remains an explicit argument of the value function. The method-specific construction for predicted obstacle evolution is developed in \cref{ch:rntc-mpc}. \Cref{ch:cn-cbf} avoids explicit time dependence in the offline reachability problem by formulating it in relative coordinates of the robot and an obstacle, in which the pairwise collision set is stationary.

\subsection{Numerical Solution and Scalability}
\label{subsec:prelim-hj-numerical}

In practice, the variational inequalities \cref{eq:prelim-hji-vi-finite-horizon,eq:prelim-stationary-hji-vi} are solved numerically. The standard approach discretizes the state space on a grid and propagates the value function backward in time with level-set methods until the horizon is covered or, for the infinite-horizon problem, until the value function converges to a stationary solution \cite{osher_level_2003,mitchell_time-dependent_2005,bansal_hamilton-jacobi_2017}. The stored values are initialized at every grid node with the terminal condition \(V(x,T)=c(x)\), and one backward step of size \(\Delta t\) proceeds in two stages. The first stage propagates the Hamilton--Jacobi part of \cref{eq:prelim-hji-vi-finite-horizon}, which, using \(\partial V/\partial t=-H\), gives the tentative update
\begin{equation}
  \tilde{V}(x)
  =
  V(x,t)
  +
  \Delta t\,
  H\bigl(x,\nabla_xV(x,t)\bigr),
  \label{eq:prelim-hj-numerical-pde-step}
\end{equation}
where the spatial gradient is approximated by upwind finite differences on the stored values and the inner optimization over \(u\) and \(d\) inside \(H\) is carried out node by node. The second stage imposes the obstacle term by the pointwise minimum \(V(x,t-\Delta t)=\min\{\tilde{V}(x),c(x)\}\), which caps each updated value at the constraint value and enforces the ordering \cref{eq:prelim-hj-value-constraint-ordering}. Repeating the two stages marches the solution backward from \(T\) to the initial time of interest, and for the stationary problem \cref{eq:prelim-stationary-hji-vi} the iteration is continued until successive value arrays differ by less than a tolerance. The result is a table of value samples over the grid and time, from which values and gradients at arbitrary states are obtained by interpolation. The learned safety functions of \cref{ch:ntc-mpc,ch:rntc-mpc,ch:orn-cbf,ch:cn-cbf} are supervised by such numerically computed solutions, while \cref{ch:vertex-cbf} generates its supervision without a grid, for the reason given next.

The principal limitation of grid-based solvers is the \emph{curse of dimensionality}, since memory and computation grow exponentially with the state dimension. This typically confines grid-based solutions to roughly five or six state variables, with computation times of seconds to minutes \cite{bansal_hamilton-jacobi_2017}. A substantial literature extends the tractable range through system decomposition, efficient warm starting, learned classification of safe states, and neural approximation of the HJ solution \cite{chen_decomposition_2018,herbert_reachability-based_2019,rubies-royo_classification-based_2019,bansal_deepreach_2021}. \Cref{ch:vertex-cbf} develops a further alternative for control-affine dynamics with polytopic input sets, in which supervision is generated by a search over the vertices of the input polytope instead of by a numerical solver. A second limitation is specific to the setting of this dissertation. Even when a single HJ problem is tractable, its solution is valid only for the constraint function used in that computation, so every new observation would require a new backward propagation. For these two reasons, HJ reachability is treated throughout this dissertation as an \emph{offline} analysis tool, and the online availability of its results is achieved by learning, as developed in \cref{sec:prelim-neural-hypernetworks}.

\section{HJ Reachability and Control Barrier Functions}
\label{sec:prelim-hj-cbf-relation}

HJ reachability and CBFs both represent safety through the sign of a scalar function, but they address complementary parts of the safety-control problem. HJ reachability constructs a value function by dynamic programming, whose nonnegative superlevel set characterizes the states from which failure can be avoided for a specified dynamics model, constraint function, horizon, and control--disturbance quantifier structure. A CBF instead supplies a local differential condition whose pointwise enforcement renders a chosen set forward invariant. HJ reachability constructs a large dynamics-aware avoidance set, which in the stationary infinite-horizon setting is the maximal safe set, and the CBF framework provides an efficient interface for filtering a nominal control input online \cite{choi_robust_2021,wabersich_data-driven_2023,hsu_safety_2024}.

For the infinite-horizon, time-invariant problem, the relation is particularly direct. Under matching model and disturbance assumptions, the nonnegative superlevel set \(\{x\mid V_\infty(x)\geq0\}\) is the maximal safe set \(\mathcal{S}^{\star}\), and it is exactly the type of set that the CBF theorem certifies. Wherever \(V_\infty\) is differentiable, the ordering \cref{eq:prelim-hj-value-constraint-ordering} makes the second term of the stationary variational inequality \cref{eq:prelim-stationary-hji-vi} nonnegative, so that the first term satisfies \(H\bigl(x,\nabla_xV_\infty(x)\bigr)\geq0\). An admissible control can therefore render the directional derivative of \(V_\infty\) nonnegative at such states. Since \(\alpha\bigl(V_\infty(x)\bigr)\geq0\) wherever \(V_\infty(x)\geq0\), the CBF inequality \cref{eq:prelim-cbf-condition} holds at every differentiable point of \(\mathcal{S}^{\star}\) and for every extended class-\(\mathcal{K}\) function \(\alpha\). The value function \(V_\infty\) is thus a natural CBF candidate whose boundary is that of the maximal safe set rather than that of a conservative hand-designed subset.

Two restrictions apply to this conclusion. The argument verifies the CBF inequality only on the certified set \(\mathcal{S}^{\star}\), while the definition in \cref{sec:prelim-cbf} requires it on all of \(\mathcal{X}\). The invariance argument is unaffected, because it applies the condition only at states in \(\mathcal{S}^{\star}\), as in \cref{eq:prelim-cbf-feedback-condition}, but the result is weaker than the definition. The argument further relies on the gradient \(\nabla_xV_\infty\), which need not exist at every state, as HJ value functions are in general only viscosity solutions and may be nonsmooth on parts of their domain \cite{crandall_viscosity_1983,choi_robust_2021}. Consequently, \(V_\infty\) is a CBF candidate rather than a barrier function in the classical sense.

Control-barrier value functions (CBVFs) formalize this connection by modifying the reachability objective so that the resulting value function also encodes an admissible CBF decay rate \cite{choi_robust_2021}. For this dissertation, the practical conclusion is that the infinite-horizon HJ value function is a principled source of barrier functions, which, subject to the regularity caveats above, certifies a set as large as the dynamics permit. The scope of the guarantees retained by a learned approximation of such a value function is summarized in \cref{subsec:prelim-learned-guarantee-scope}.

\section{Observation-Dependent Safety Functions}
\label{sec:prelim-environment-conditioning}

The formalism developed so far treats the constraint function, and hence the failure set, as given. In the navigation problems of this dissertation, the robot instead discovers the environment through onboard sensing. The available information covers only a local region, it is affected by occlusion, and it is refreshed at a finite rate, so every safety-relevant quantity derived from it depends on the current observation. This section introduces observations and the safety functions built from them, specializes these functions to the discrete sequence of updates produced by a perception pipeline, and analyzes their time derivative, which is the form required by the barrier conditions of \cref{sec:prelim-cbf}.

\subsection{Observations and Safety Functions}
\label{subsec:prelim-observation-model}
\label{subsec:prelim-observation-dependent-fields}

Let \(e(t)\in\mathcal{E}\) denote the \emph{environment state}, which collects the quantities that determine the failure set, such as the geometry of the occupied region and the positions and velocities of moving obstacles. The robot has no direct access to \(e\) and receives instead the \emph{observation}
\begin{equation}
  o(t)
  =
  \zeta\bigl(e(t),x(t)\bigr)
  \in\mathcal{O},
  \label{eq:prelim-observation-model}
\end{equation}
where \(\zeta:\mathcal{E}\times\mathcal{X}\rightarrow\mathcal{O}\) is the \emph{observation map} and \(\mathcal{O}\) is the observation space. The dependence on the robot state reflects the finite range and the limited field of view of onboard sensing, together with the occlusions caused by the observed geometry itself. Throughout the dissertation, \(o\) denotes the output of the perception pipeline, an occupancy grid over a local sensing window or a list of estimated obstacle states in the technical chapters. The quantity that a learned model receives is computed from \(o\) and is introduced separately as the conditioning input in \cref{sec:prelim-neural-hypernetworks}.

Observations can change because the environment evolves or because the robot moves, and two special cases recur in the technical chapters. The first case is a static environment observed egocentrically. Occupancy maps, local signed-distance fields, and point clouds are expressed relative to the current robot pose and cover a window that moves with the robot. If the environment state is constant, \(e(t)\equiv e\), the observation varies only through the robot state and is written \(o=o(x)\). The methods of \cref{ch:ntc-mpc,ch:orn-cbf} operate in this case.

The second case is a dynamic environment observed in a fixed frame. When the observation is a list of positions and velocity vectors of moving obstacles expressed in world coordinates, its value does not depend on where the robot is, as long as the obstacles concerned remain within detection range. The observation then varies only with time and is written \(o=o(t)\). A change in the set of detected obstacles is not described by this dependence and is treated as a discrete update below. The methods of \cref{ch:rntc-mpc,ch:cn-cbf} operate in this case.

\paragraph{Safety fields.}
A safety function built from local information is a scalar field on the joint space of states and observations,
\begin{equation}
  \phi:
  \mathcal{X}\times\mathcal{O}\rightarrow\mathbb{R},
  \qquad
  (x,o)\mapsto\phi(x,o).
  \label{eq:prelim-observation-dependent-function}
\end{equation}
The sign convention of \cref{sec:prelim-constraint-set} is retained, so nonnegative values indicate membership in the set represented by the field. For a fixed observation, this set is
\begin{equation}
  \mathcal{S}(o)
  :=
  \{x\in\mathcal{X}\mid\phi(x,o)\geq0\}.
  \label{eq:prelim-conditioned-safe-set}
\end{equation}
Whether membership in \(\mathcal{S}(o)\) establishes instantaneous constraint satisfaction or dynamic safety depends on the field and its validity.

The form \cref{eq:prelim-observation-dependent-function} covers geometric constraints, reachability values, and barrier candidates. For example, \(c(x,o)=\operatorname{sd}_{\mathcal{A}(o)}(p(x))\) measures the distance to the observed geometry \(\mathcal{A}(o)\). For a fixed static scene, \(V_\infty(x,o)\) is the stationary HJ value for that constraint, whereas predicted moving geometry requires a time-dependent value. A barrier \(h(x,o)\) supplies the field used by the safety filter. Along a trajectory, these fields are evaluated as \(\phi\bigl(x(t),o(t)\bigr)\), with the observation dependence determined by the two cases above. A fixed field, as in \cref{ch:vertex-cbf}, omits \(o\).

\paragraph{Observation updates.}
A perception pipeline delivers observations at discrete instants rather than continuously. Let \(j\in\mathbb{N}_0\) index the updates, let \(t_j\) denote the instant of update \(j\), and let \(\Delta t_o=t_{j+1}-t_j\) denote the update period. The observation available at update \(j\) is
\begin{equation}
  o_j
  =
  \zeta\bigl(e(t_j),x(t_j)\bigr),
  \label{eq:prelim-observation-update}
\end{equation}
and it remains the robot's only description of the environment on the interval \([t_j,t_{j+1})\). The associated field is \(\phi(\,\cdot\,,o_j)\) with the safe set \(\mathcal{S}(o_j)\). The update period is typically longer than the control period, and \cref{ch:orn-cbf,ch:cn-cbf} index observations separately from control steps. When a new observation is available at every control step, as in \cref{ch:ntc-mpc,ch:rntc-mpc}, the same index is used for both.

\paragraph{Fixed-observation notation.}
Whenever the observation is held fixed, \(\phi(\,\cdot\,,o)\) is an ordinary scalar field on the state space. In that situation the field is written
\begin{equation}
  \phi(\,\cdot\mid o):
  \mathcal{X}\rightarrow\mathbb{R},
  \qquad
  x\mapsto\phi(x\mid o),
  \label{eq:prelim-observation-conditioned-function}
\end{equation}
to indicate that the observation acts as a parameter rather than as a second time-dependent argument. A field written in this form is said to be \emph{conditioned} on the argument after the bar. The notation is only a way of writing the field and does not by itself introduce a modeling assumption. It is appropriate wherever the analysis concerns a single fixed observation, as within one planning problem in \cref{ch:ntc-mpc}, and it becomes a deliberate design choice in \cref{ch:orn-cbf}, where the barrier is held fixed between two updates. The conditions under which that choice is justified follow from the time derivative studied next. The same convention applies to the learned fields of \cref{sec:prelim-neural-hypernetworks}, which are conditioned on an input derived from the observation.

\subsection{Observation Dynamics and Updates}
\label{subsec:prelim-observation-time-derivative}

The barrier conditions of \cref{sec:prelim-cbf} constrain the derivative of the safety function along the closed-loop trajectory, so the observation dependence must be differentiated as well. This subsection matters primarily for the safety filters of \cref{ch:orn-cbf,ch:cn-cbf}. Assume that \(\mathcal{O}\subseteq\mathbb{R}^{n_o}\) and that \(\phi\) is continuously differentiable in both arguments.\footnote{Observation spaces are finite dimensional in practice, since an occupancy grid is a finite array of cells and a list of obstacle states is a finite vector. When the observation is modeled as an element of a function space, the second term below becomes a directional derivative and the case distinction is unchanged.} Along a trajectory \(x(t)\) with an observation \(o(t)\), the chain rule gives
\begin{equation}
  \frac{d}{dt}\phi\bigl(x(t),o(t)\bigr)
  =
  \nabla_x\phi(x,o)^{\top}\dot{x}
  +
  \nabla_o\phi(x,o)^{\top}\dot{o}.
  \label{eq:prelim-observation-total-derivative}
\end{equation}
With the control-affine dynamics \cref{eq:prelim-control-affine-dynamics} and with \(\phi\) in the role of the barrier candidate \(h\), the CBF condition \cref{eq:prelim-cbf-condition} becomes
\begin{equation}
  \sup_{u\in\mathcal{U}}
  \left[
    L_{f_0}\phi(x,o)
    +
    L_g\phi(x,o)u
    +
    \nabla_o\phi(x,o)^{\top}\dot{o}
    \right]
  \geq
  -\alpha\bigl(\phi(x,o)\bigr).
  \label{eq:prelim-observation-cbf-condition}
\end{equation}
The first two terms are the Lie derivatives of \cref{eq:prelim-lie-derivatives}. The third term requires the \emph{observation dynamics} \(\dot{o}\), that is, a model of how the perceived description of the environment evolves. Its structure follows the cases of \cref{subsec:prelim-observation-model}.

\paragraph{Static environment with egocentric observations.}
Here \(o=o(x)\) and \(\dot{o}=(\partial o/\partial x)\,\dot{x}\), so that \cref{eq:prelim-observation-total-derivative} reduces to
\begin{equation}
  \frac{d}{dt}\phi\bigl(x(t),o(x(t))\bigr)
  =
  \left[
    \nabla_x\phi(x,o)
    +
    \left(\frac{\partial o}{\partial x}\right)^{\!\top}
    \nabla_o\phi(x,o)
    \right]^{\top}
  \dot{x}.
  \label{eq:prelim-observation-derivative-static}
\end{equation}
The derivative depends only on the state and the control input, and no explicit time dependence appears. The observation enters through the Jacobian \(\partial o/\partial x\), which describes two effects of robot motion. The first effect is \emph{displacement}, since the sensing window moves with the robot and the geometry that the observation already contains is expressed relative to the new window. The second effect is \emph{renewal}, since geometry that was outside the window enters at its boundary and geometry that leaves the window is removed. If the field depends only on the geometry that the observation represents, expressed in coordinates fixed to the environment, then displacement leaves the value of the field at a given state unchanged, while renewal can change it, and \cref{ch:orn-cbf} treats this case in detail. Holding the field fixed between two observation updates is a different operation. It differentiates the field \(\phi(\,\cdot\mid o_j)\) for a fixed observation instead of the composed function \(\phi\bigl(x,o(x)\bigr)\), so no term of \cref{eq:prelim-observation-derivative-static} is neglected.

\paragraph{Moving obstacles observed in a fixed frame.}
Here \(o=o(t)\), so the last term of \cref{eq:prelim-observation-cbf-condition} does not depend on the control input and plays the role of the explicit partial time derivative in the time-varying condition \cref{eq:prelim-time-varying-cbf-condition}. If the obstacles follow an assumed motion model with a bounded input \(d\in\mathcal{D}\), then \(\dot{o}\) is supplied by that model, and taking the infimum over \(d\) yields the robust condition \cref{eq:prelim-robust-cbf-condition}. \Cref{ch:cn-cbf} retains this term explicitly.

\paragraph{General case.}
When an egocentric observation covers a scene that contains moving obstacles, both mechanisms act simultaneously and both terms are required. The model must then describe the sensing geometry and the obstacle motion together.

\paragraph{Modeled and unmodeled observation dynamics.}
Observation dynamics can describe changes to known geometry, including displacement within the sensing window \cite{dawson_learning_2022} and tracked motion under an assumed model. They cannot determine unseen geometry beyond the sensing range or behind an occlusion. Newly visible geometry may enter abruptly, making \(\dot{o}\) and the total derivative \cref{eq:prelim-observation-total-derivative} undefined at that instant. The differential condition is enforced between such updates, and safety across them requires additional sensing and update assumptions.

\paragraph{Sensing adequacy.}
Let \(\mathcal{F}(e)\subseteq\mathcal{X}\) denote the failure set induced by the true environment state and let \(\mathcal{F}(o)\subseteq\mathcal{X}\) denote the failure set represented by the observation, so that \(\mathcal{F}(e)\setminus\mathcal{F}(o)\) collects the failure states that the robot has not yet perceived. Let \(\mathcal{B}_\infty(\mathcal{F})\) denote the infinite-horizon backward reachable tube of a failure set \(\mathcal{F}\), defined as in \cref{eq:prelim-finite-horizon-brt} with \(T\rightarrow\infty\), which collects the states from which entry into \(\mathcal{F}\) can no longer be avoided under the dynamics and actuation limits. The setting of this dissertation assumes that
\begin{equation}
  x(t)
  \notin
  \mathcal{B}_\infty
  \bigl(
  \mathcal{F}(e(t))\setminus\mathcal{F}(o(t))
  \bigr),
  \qquad
  \forall t,
  \label{eq:prelim-sensing-adequacy}
\end{equation}
which means that the failure states not yet revealed by the sensing remain far enough away, in the dynamic rather than the purely geometric sense, for the robot to still avoid them once they are revealed. Condition \cref{eq:prelim-sensing-adequacy} is a requirement on the sensing configuration and the operating speed rather than on the safety function, since a controller cannot act on information that it does not have. The sensing range, the field of view, the occlusion structure, the update period, and the robot speed all enter, because the extent of \(\mathcal{B}_\infty\) grows with the robot speed and with the distance needed to stop or to turn away.

\paragraph{Cross-update safety.}
At a discrete update, replacing \(o_j\) by \(o_{j+1}\) may shrink the represented safe set. Continued safety requires at minimum
\begin{equation}
  x(t_{j+1})
  \in
  \mathcal{S}(o_{j+1}).
  \label{eq:prelim-cross-update-safety}
\end{equation}
A condition enforced under the previous field cannot guarantee membership in the updated set. Sensing adequacy \cref{eq:prelim-sensing-adequacy} does not imply \cref{eq:prelim-cross-update-safety} either, because avoiding newly revealed geometry separately need not be compatible with avoiding the previously observed geometry. Membership in the updated set must be checked at the update or protected by an additional margin. The technical chapters state the conditions relevant to each sensing and obstacle model.

\paragraph{Consequences for the methods.}
In a static environment, a field held fixed between two updates remains consistent with the environment that it represents. Its derivative is \(\nabla_x\phi(x\mid o_j)^{\top}\dot{x}\) alone, and the offline computation becomes a family of stationary reachability problems indexed by the observation, which is the construction of \cref{ch:ntc-mpc,ch:orn-cbf}. When obstacles move, the represented safe set drifts while a fixed field remains unchanged, so obstacle motion is represented explicitly, through a time-varying constraint function in \cref{ch:rntc-mpc} or through relative coordinates in \cref{ch:cn-cbf}.

\section{Learned Implicit Safety Functions}
\label{sec:prelim-neural-hypernetworks}

Learning connects offline HJ analysis to the observation-dependent functions needed online. This section defines the conditioning input, neural implicit fields, the residual parameterization, and hypernetworks, then summarizes their shared pipeline and guarantee scope.

\paragraph{Conditioning input.}
A learned safety function is not evaluated on the raw observation \cref{eq:prelim-observation-model}, but on a quantity derived from it. This quantity is called the \emph{conditioning input} and is written
\begin{equation}
  w
  =
  \eta(o)
  \in\mathcal{W},
  \label{eq:prelim-conditioning-input}
\end{equation}
where \(\eta\) is a fixed map applied by the perception and prediction stack before the model is queried, and \(\mathcal{W}\) is the conditioning space. Conditioning a value function on such a variable follows the parameter-conditioned reachability formulation of \cite{borquez_parameter-conditioned_2023}. The methods of this dissertation define \(\eta\) differently. In \cref{ch:ntc-mpc,ch:orn-cbf}, the observation is an occupancy grid and \(w\) is the signed-distance field computed from it over the sensing window. In \cref{ch:rntc-mpc}, the observed obstacle states are propagated by a motion model, and \(w\) collects the signed-distance fields of the predicted geometry at selected future instants. In \cref{ch:cn-cbf}, \(w\) consists of the observed obstacle states themselves, so that the conditioning input coincides with the observation. \Cref{ch:vertex-cbf} learns a single fixed field, for which \(w\) is omitted, because its subject is the supervision rather than the environment representation. The first three methods hold \(w\) fixed until the next update, so their learned fields use the fixed-argument notation \cref{eq:prelim-observation-conditioned-function} and are written as \(\hat{\phi}_\Theta(x\mid w)\), with the safe set \cref{eq:prelim-conditioned-safe-set} evaluated for the corresponding conditioning input. \Cref{ch:cn-cbf} instead differentiates the obstacle states with respect to time, so its barrier is written as a joint function of state and observation \cref{eq:prelim-observation-dependent-function}.

\subsection{Neural Implicit Safety Representations}
\label{subsec:prelim-neural-implicit-sets}

A neural scalar field
\begin{equation}
  \hat{\phi}_{\Theta}:
  \mathcal{X}\rightarrow\mathbb{R},
  \label{eq:prelim-neural-scalar-field}
\end{equation}
parameterized by a real-valued vector \(\Theta\), represents a set implicitly through its sign or level sets, in the same way that geometry is represented by learned signed-distance or occupancy fields in computer vision and graphics \cite{park_deepsdf_2019,mescheder_occupancy_2019,mildenhall_nerf_2020}. Under the convention used for HJ safety functions, the predicted safe set is defined by
\begin{equation}
  \hat{\mathcal{S}}_{\Theta}
  :=
  \{x\in\mathcal{X}\mid \hat{\phi}_{\Theta}(x)\geq0\}.
  \label{eq:prelim-neural-implicit-safe-set}
\end{equation}
For an environment-dependent family, the field is conditioned as \(\hat{\phi}_{\Theta}(x\mid w)\), yielding a different implicit set for each conditioning input \(w\in\mathcal{W}\).

Several approximation objectives must be distinguished. A small value error need not imply a small safe-set error, because even a small bias can move the zero level set. Conversely, accurate classification or boundary recovery does not determine values far from the boundary. Neither value nor set accuracy guarantees accurate state gradients \(\nabla_x\hat\phi_\Theta(x)\), which are required by gradient-based MPC and by the Lie derivatives in CBF constraints. The training objectives of the technical chapters address these distinctions explicitly.

The networks of this dissertation are differentiable almost everywhere, and the state gradient \(\nabla_x\hat{\phi}_{\Theta}\) is available efficiently through automatic differentiation, so the same learned field can serve as a scalar inequality constraint in an MPC problem and as a barrier function in a CBF-QP. The quality of these derivatives depends on the architecture. Networks with smooth activation functions, such as the sinusoidal representations used in implicit geometry and in learned reachability \cite{sitzmann_implicit_2020,bansal_deepreach_2021}, are preferable whenever the accuracy of the gradient matters, while rectified linear unit (ReLU) networks produce gradients that are discontinuous across activation boundaries. Automatic differentiation returns exact derivatives of the implemented network, but it does not by itself establish the regularity assumed by the CBF theorem. In an MPC setting, smoothness mainly influences the conditioning of the numerical optimization, while a classical CBF guarantee treats sufficient differentiability of \(h\) as a standing assumption.

\subsection{Residual Parameterization with Failure-Set Exclusion}
\label{subsec:prelim-structure-preserving-residual}

An unconstrained neural value-function approximation can make errors anywhere, including inside the observed failure set, where errors are unacceptable. The HJ ordering in \cref{eq:prelim-hj-value-constraint-ordering,eq:prelim-hj-residual-decomposition} suggests a parameterization that rules out such errors by construction. For a conditioning input \(w\), let \(c(x\mid w)\) denote the corresponding constraint function and parameterize
\begin{equation}
  \hat{V}_{\Theta}(x\mid w)
  :=
  c(x\mid w)-\hat{r}_{\Theta}(x\mid w),
  \qquad
  \hat{r}_{\Theta}(x\mid w)\geq0,
  \label{eq:prelim-structure-preserving-residual}
\end{equation}
where the nonnegativity of the learned residual \(\hat{r}_{\Theta}\) is enforced architecturally, for example through a ReLU output activation. A ReLU is the most direct choice, but its zero derivative on the inactive region can stall gradient-based training, so the technical chapters use smooth relaxations of the ReLU instead. The nonnegative residual enforces the pointwise ordering
\begin{equation}
  \hat{V}_{\Theta}(x\mid w)
  \leq
  c(x\mid w),
  \qquad
  \left\{x\mid\hat{V}_{\Theta}(x\mid w)\geq0\right\}
  \subseteq
  \left\{x\mid c(x\mid w)\geq0\right\}.
  \label{eq:prelim-residual-safe-set-inclusion}
\end{equation}
Regardless of the residual approximation error, the learned safe set cannot include states that the represented constraint function classifies as failures, whether it serves as an MPC terminal set or as a CBF candidate set. In the braking example, no nonnegative residual, regardless of its accuracy, could declare a state inside the wall as safe. The property is one-sided, because \(\hat{V}_{\Theta}\leq c\) does not imply \(\hat{V}_{\Theta}\leq V\), so a state outside the failure set but inside the true backward reachable tube can still be predicted safe if the residual is underestimated there. The decomposition also simplifies the learning problem, because the network no longer needs to reproduce the constraint geometry, which is supplied exactly by \(c\), but only the dynamics-induced correction, such as the braking distance in \cref{fig:prelim-braking-values}. Later chapters instantiate this principle for time-varying terminal constraints, observation-conditioned barriers, pairwise relative-state barriers, and fixed barriers trained without a grid-based solver. In the last case, developed in \cref{ch:vertex-cbf}, the parameterization serves a second purpose beyond exclusion. It makes the uninformative constant solutions of the stationary variational inequality harder for the network to represent, although it does not remove them from the solution set of the equation.

\subsection{Hypernetwork Conditioning}
\label{subsec:prelim-hypernetworks}

A conditioning input may be supplied directly as an additional input to a neural field or used to generate some or all of its parameters. A \emph{hypernetwork} follows the second option and generates those parameters \cite{ha_hypernetworks_2017,chauhan_brief_2023}. A hypernetwork \(H_{\Psi}\) with trainable parameters \(\Psi\) maps \(w\) to the parameters \(\Theta(w)\) of a \emph{main network} \(M_{\Theta(w)}\), so that
\begin{equation}
  \Theta(w)
  =
  H_{\Psi}(w),
  \qquad
  \hat{\phi}_{\Psi}(x\mid w)
  =
  M_{\Theta(w)}(x).
  \label{eq:prelim-hypernetwork}
\end{equation}
This architecture separates the potentially expensive processing of the conditioning input from the repeated state queries of the neural scalar field \(\hat{\phi}_{\Psi}(x\mid w)\). The hypernetwork is evaluated once when the conditioning input changes, whether at every planning cycle or only at a slower observation update, and the compact main network is evaluated and differentiated many times within the online optimization. Compared with concatenating the conditioning input to the state input of a single large network, the hypernetwork approach can represent the same function family with far fewer online computations per query \cite{galanti_modularity_2020}. The later chapters specify which environment representations are used as \(w\) and how the scalar field is structured.

\subsection{Offline Supervision and Online Inference}
\label{subsec:prelim-offline-online-safety-learning}

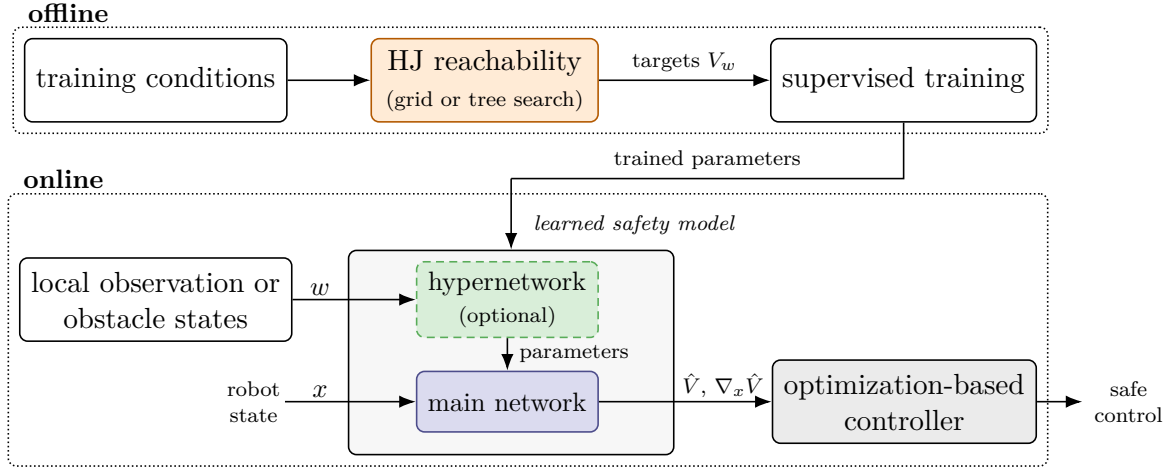
\begin{figure}[tb]
  \centering
  \colorlet{hjfill}{orange!16}%
\colorlet{hjline}{orange!70!black}%
\colorlet{hypfill}{green!55!black!15}%
\colorlet{hypline}{green!50!black!65}%
\colorlet{mlpfill}{blue!55!black!14}%
\colorlet{mlpline}{blue!35!black!70}%
\colorlet{ctrlfill}{black!8}%

\begin{tikzpicture}[
    font=\small,
    blk/.style={draw, semithick, rounded corners=3pt, align=center,
                inner xsep=1.5mm, inner ysep=1.4mm, minimum height=11mm},
    sub/.style={draw, semithick, rounded corners=3pt, align=center,
                font=\footnotesize, inner xsep=1.5mm, inner ysep=1.2mm,
                minimum width=2.4cm, minimum height=8mm},
    arrow/.style={-{Latex[length=2.2mm]}, semithick},
    wire/.style={semithick},
    sig/.style={font=\footnotesize, inner sep=1.5pt},
    note/.style={font=\scriptsize, align=center, inner sep=1.5pt},
    gl/.style={font=\scriptsize\itshape},
    stg/.style={font=\footnotesize\bfseries, inner sep=1.5pt},
  ]

  \node[blk, minimum width=3.3cm] (cond) at (1.80,5.90) {training conditions};
  \node[blk, fill=hjfill, draw=hjline, minimum width=3.0cm] (hj) at (6.15,5.90)
  {HJ reachability\\{\scriptsize (grid or tree search)}};
  \node[blk, minimum width=3.4cm] (train) at (11.69,5.90) {supervised training};
  \draw[arrow] (cond) -- (hj);
  \draw[arrow] (hj) -- node[note, above, yshift=1.5pt] {targets $V_{w}$} (train);

  \node[draw, semithick, rounded corners=3pt, fill=black!3,
    minimum width=4.3cm, minimum height=2.7cm] (model) at (6.50,2.30) {};
  \node[gl, anchor=south west] (modellbl) at ([shift={(0.15,0.06)}]model.north)
  {learned safety model};
  \node[sub, fill=hypfill, draw=hypline, densely dashed, minimum height=10mm]
  (hyper) at (6.45,3.00) {hypernetwork\\{\scriptsize (optional)}};
  \node[sub, fill=mlpfill, draw=mlpline] (net) at (6.45,1.65) {main network};
  \draw[arrow] (hyper.south) -- (net.north);
  \node[note, anchor=west] at (6.55,2.28) {parameters};

  \draw[wire] (train.south) -- (train.south |- 0,4.60) -- (model.north |- 0,4.60);
  \draw[arrow] (model.north |- 0,4.60) -- (model.north);
  \node[note, anchor=south] at ($(train.south |- 0,4.67)!0.5!(model.north |- 0,4.67)$)
  {trained parameters};

  \node[blk, minimum width=3.3cm] (obs) at (1.80,3.00)
  {local observation or\\obstacle states};
  \node[note, anchor=east] (state) at ([xshift=-0.10cm]obs.east |- net.west) {robot\\state};
  \node[blk, fill=ctrlfill] (ctrl) at (11.70,1.65)
  {optimization-based\\controller};
  \draw[arrow] (obs.east) -- (hyper.west);
  \node[sig, anchor=south] at ($(obs.east)!0.5!(model.west |- obs.east)$) {$w$};
  \draw[arrow] (state.east) -- (net.west);
  \node[sig, anchor=south] at ($(obs.east |- net.west)!0.5!(model.west |- net.west)$) {$x$};
  \draw[arrow] (net.east) -- (ctrl.west);
  \node[note, anchor=south] at ($(model.east |- net.east)!0.5!(ctrl.west)$)
  {$\hat{V}$, $\nabla_{x}\hat{V}$};
  \draw[arrow] (ctrl.east) -- ++(0.60,0);
  \node[note, anchor=west] at ([xshift=0.70cm]ctrl.east) {safe\\control};

  \node[draw, densely dotted, semithick, rounded corners=3pt, inner sep=4pt,
    fit=(cond)(hj)(train)] (offbox) {};
  \node[draw, densely dotted, semithick, rounded corners=3pt, inner sep=4pt,
    fit=(obs)(state)(model)(modellbl)(ctrl)] (onbox) {};
  \node[stg, anchor=south west] at ([xshift=1.5mm]offbox.north west) {offline};
  \node[stg, anchor=south west] at ([xshift=1.5mm]onbox.north west) {online};
\end{tikzpicture}
  \caption[Offline--online pipeline shared by the methods of this dissertation]{Offline--online pipeline shared by the methods of this dissertation. Offline, a solver computes HJ value functions \(V_w\) for a set of training conditioning inputs, and these solutions supervise the learned safety model. \Cref{ch:ntc-mpc,ch:rntc-mpc,ch:orn-cbf,ch:cn-cbf} obtain them from a grid-based solver, while \cref{ch:vertex-cbf} replaces this block by vertex-restricted search combined with a physics-informed loss. Online, the current conditioning input \(w\) selects the scalar field \(\hat{V}(\,\cdot\mid w)\) whose value and gradient the optimization-based controller evaluates. The compact main network takes the robot state \(x\) as input and is conditioned on \(w\), either through parameters generated by the optional hypernetwork (dashed) once per update, as in \cref{ch:ntc-mpc,ch:rntc-mpc,ch:orn-cbf}, or directly when the hypernetwork is omitted. The main network is evaluated and differentiated at every query.}
  \label{fig:prelim-pipeline}
\end{figure}

The pipeline in \cref{fig:prelim-pipeline} separates the supervision source from the learned representation. Each conditioning input \(w\) specifies a target field \(V_w(x)\). For a fixed field, \(w\) is omitted. \Cref{ch:ntc-mpc,ch:rntc-mpc,ch:orn-cbf,ch:cn-cbf} use grid-based HJ targets, while \cref{ch:vertex-cbf} uses vertex-restricted search and a physics-informed loss. The learned model approximates
\begin{equation}
  \hat{V}_{\Psi}(x\mid w)
  \approx
  V_w(x)
  \label{eq:prelim-conditioned-value-approximation}
\end{equation}
over a distribution of training conditioning inputs, following the general strategy of approximating HJ solutions with neural networks \cite{rubies-royo_classification-based_2019,bansal_deepreach_2021}. At runtime, the conditioning input \(w_j\) formed from the latest observation selects a scalar field that can be queried repeatedly at candidate states, while a fixed field requires no online update. The expensive reachability analysis is thus performed offline, once per training environment, and the learned model generalizes the result to unseen conditioning inputs.

\subsection{Guarantee Scope for Learned Safety Functions}
\label{subsec:prelim-learned-guarantee-scope}

A neural network fitted to reachability data does not automatically inherit the properties of the exact solution, so the guarantees of a learned safety function must be stated explicitly \cite{dawson_safe_2023,so_how_2024}. An architectural constraint such as \cref{eq:prelim-structure-preserving-residual} guarantees exclusion of the represented failure set. Value-regression and classification metrics instead assess approximation of an HJ value function or safe set. Neither establishes satisfaction of the HJ variational inequality, validity of the CBF condition \cref{eq:prelim-cbf-condition}, or recursive feasibility of an MPC scheme, which require separate assumptions and verification. Empirical collision avoidance is a closed-loop performance observation that supports, but does not replace, a formal certificate. Moreover, the accuracy of a learned field is limited by the accuracy of its training targets. The learned fields of \cref{ch:ntc-mpc,ch:rntc-mpc,ch:orn-cbf,ch:cn-cbf} use numerically computed HJ solutions as targets, whose accuracy is limited by discretization, finite computational domains, and solver error. \Cref{ch:vertex-cbf} replaces those targets by a variational-inequality residual and by labels from a truncated search, which removes the solver error but introduces errors from time discretization, horizon truncation, and incomplete search. In either case, a neural approximation cannot inherit a stronger guarantee than the target and the model used to train it. Each technical chapter reports, for its method, which properties hold by construction, which are measured, which are inherited from the numerical target under its model assumptions, and which are demonstrated experimentally, and \cref{ch:conclusion} collects these statements.

\section{Summary}
\label{sec:prelim-summary}

This chapter distinguished instantaneous constraint satisfaction from membership in a controlled-invariant safe set. MPC terminal constraints and CBFs provide two interfaces for preserving safety, with the maximal safe set as the ideal stationary target. HJ reachability characterizes this set, and the ordering \(V\leq c\) separates the geometric obstacle distance from the dynamics-induced reduction in safety margin. Its computational cost motivates offline analysis and learned online representations.

Observation-dependent fields require both a state derivative and, where modeled, an observation derivative. New information also requires assumptions on the sensing and a check of membership in the updated set. Neural implicit fields and hypernetworks support efficient queries, while residual parameterization excludes the represented failure set by construction. Value, set, and gradient accuracy, CBF validity, and MPC recursive feasibility remain distinct properties.

\part{Reachability-Informed Model Predictive Control}
\chapter{Neural Terminal Constraints in Static Environments}
\label{ch:ntc-mpc}

This chapter presents \emph{neural terminal constraint MPC} (NTC-MPC), which estimates the maximal safe set of an unknown static environment from local observations and uses it as an MPC terminal constraint. It addresses research question~\ref{rq:realtime} through the MPC interface and is based on the author's article \cite{derajic_learning_2025}, published in IEEE Robotics and Automation Letters (RA-L) in 2025.

\section{Motivation}
\label{sec:ntc-motivation}

An MPC local planner converts the latest observation into a short motion that makes progress toward a goal supplied by a global route, subject to explicit constraints on the dynamics, the actuation limits, and the distance to the obstacles. The optimization reasons only over a finite prediction horizon, so a terminal constraint set is needed that requires the final predicted state to admit a safe continuation (\cref{sec:prelim-mpc}). The question addressed in this chapter is how that set should be chosen when the obstacles are unknown before deployment.

\begin{figure}[tbp]
  \centering
  \input{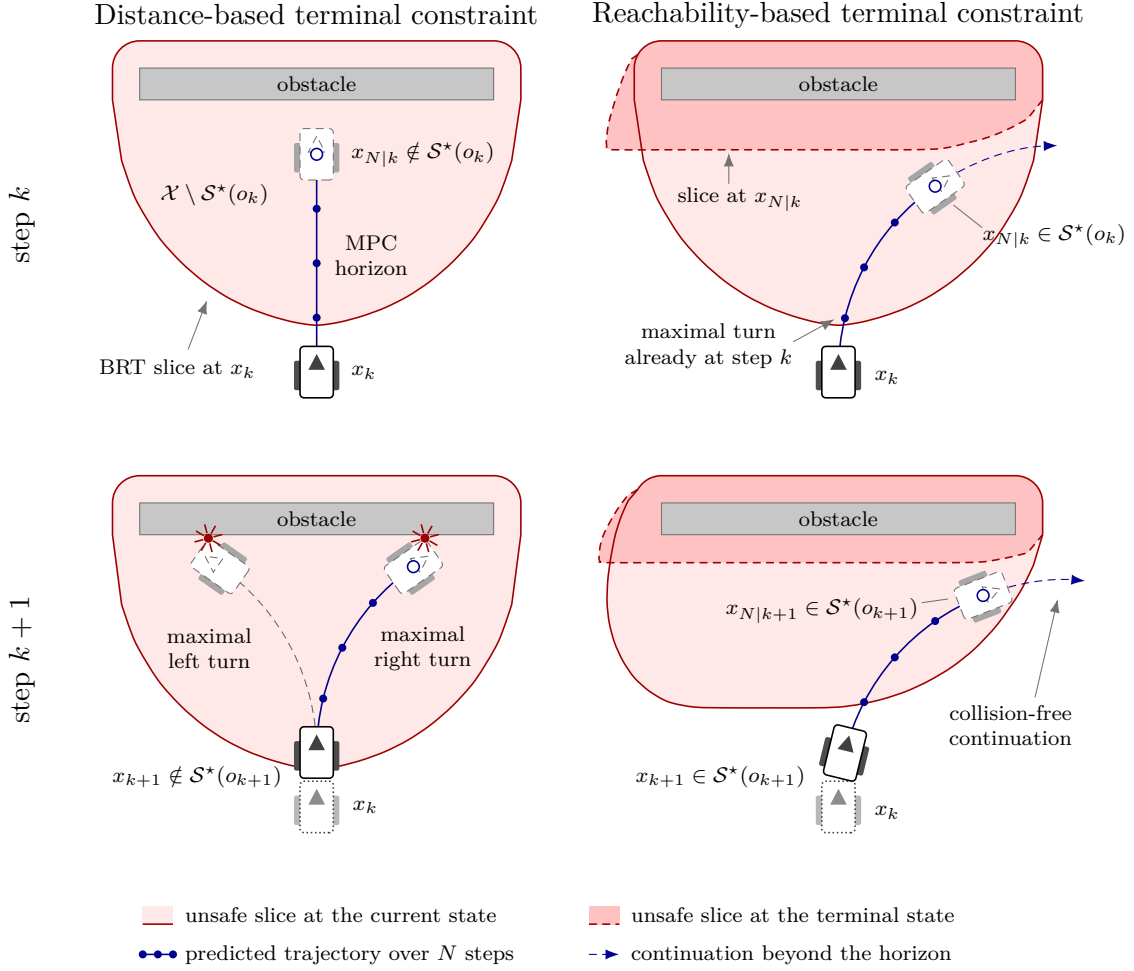}
  \caption[Effect of the terminal constraint on an MPC local planner near a static obstacle]{Effect of the terminal constraint on an MPC local planner that approaches a static obstacle, shown at two consecutive planning steps. Each red-shaded region is a slice of the backward reachable tube (BRT) of the observed obstacle, that is, of the set \(\mathcal{X}\setminus\mathcal{S}^{\star}(o_{k})\) of states from which collision can no longer be avoided under the actuation limits. The slice depends on the heading, so a solid outline belongs to the current state and a dashed outline to the terminal state. In the lower row the dotted body marks the position at step \(k\). Left: the terminal constraint is distance based, so every state predicted at step \(k\) keeps a nonnegative distance to the obstacle and the planner continues straight ahead. At step \(k+1\) the robot has entered the BRT, and every admissible input sequence collides at some later instant. Right: the reachability-based terminal constraint \(x_{N\mid k}\in\mathcal{S}^{\star}(o_{k})\) makes the planner steer away already at step \(k\), so the robot remains outside the BRT one step later.}
  \label{fig:ntc-intro-illustration}
\end{figure}

\Cref{fig:ntc-intro-illustration} shows how a short-horizon plan can satisfy every distance constraint yet enter an obstacle's backward reachable tube (BRT). The unavoidable collision lies beyond the prediction horizon, so infeasibility appears only after recovery is impossible. A longer horizon reduces this risk at additional computational cost. A controlled-invariant terminal set instead requires a safe continuation beyond the horizon, with the maximal safe set \(\mathcal{S}^{\star}\) as the least restrictive target.

The difficulty specific to this setting is that such a set cannot be precomputed. Every observation induces a different maximal safe set, new observations arrive at the perception rate, and solving the corresponding Hamilton--Jacobi (HJ) problem numerically takes seconds even for low-dimensional robot models (\cref{subsec:prelim-hj-numerical}). Classical terminal-set design assumes a fixed and known constraint geometry, while the planner considered here requires an environment-dependent terminal set within milliseconds.

NTC-MPC resolves this difficulty by learning the mapping from observations to terminal sets, following the offline--online pipeline of \cref{sec:prelim-neural-hypernetworks}. The offline stage solves HJ reachability problems for many representative local observations and retains the resulting value functions as supervision targets. At runtime, a hypernetwork is evaluated once per new observation and generates the parameters of a compact main network, which the solver then evaluates and differentiates at every optimization iteration. As a result, the high-dimensional observation is processed once per planning cycle rather than once per solver iteration. In addition, the proposed radially weighted regression loss concentrates approximation accuracy near the safe-set boundary, where sign errors determine whether a state is treated as safe or unsafe. The experiments in \cref{sec:ntc-experiments} show that the resulting planner avoids collisions more reliably than the evaluated baselines, with the largest margins at short prediction horizons.

\section{Related Work}
\label{sec:ntc-related-work}

MPC local planners commonly derive their collision costs and constraints directly from the latest local observation. SDFs can be computed efficiently from the detected occupied space and supply distance values and gradients for trajectory optimization \cite{oleynikova_signed_2016-1,oleynikova_voxblox_2017}. Depending on the formulation, this information enters the optimizer as soft cost terms, convex inner approximations, or exact distance-based hard constraints \cite{gaertner_collision-free_2021,schoels_nmpc_2020,zhang_optimization-based_2021}. Other formulations impose sampled-data or discrete-time CBF conditions along a reference corridor or along the prediction horizon, or evaluate occupancy costs across candidate rollouts \cite{roque_corridor_2022,zeng_safety-critical_2021,mohamed_autonomous_2022}. Learning-based variants retain this stage-wise structure but learn the constraint itself, for example by fitting a barrier function to LiDAR and odometry data at every control step \cite{lafmejani_nmpc-lbf_2022} or by embedding learned collision scores and dynamics models directly into the optimizer \cite{jacquet_n-mpc_2024,salzmann_real-time_2023}. These methods support effective obstacle avoidance within the planned trajectory, but a positive obstacle distance over the horizon alone cannot exclude terminal states from which a collision is already inevitable under the dynamics and the actuation limits \cite{fraichard_short_2007,hwang_safe_2024}.

In classical MPC, this post-horizon requirement is the role of a controlled-invariant terminal set \cite{mayne_constrained_2000,kerrigan_invariant_2000}. When the set satisfies the state constraints and admits a terminal control law that keeps the state inside it, the shifted solution from the preceding control step establishes recursive feasibility. Candidate constructions include maximal or multi-step control-invariant regions, viability kernels, and discrete-time CBF certificates \cite{johansson_stable_2024,zhao_nonlinear_2024,liniger_viability_2015,katriniok_discrete-time_2023}. Viability information has also been used to select viable controls, to analyze horizon requirements without a terminal set, and to project measured states back into the viable region \cite{liniger_real-time_2019,ruswurm_interplay_2021,yeganegi_robust_2022}. All these constructions start from a prescribed constraint geometry and cannot generate a new kernel from a runtime observation. For unknown environments, multitrajectory MPC instead pairs the task trajectory with a safe trajectory that ends in a stopping state within observed free space \cite{saccani_multitrajectory_2022}. It achieves persistent obstacle avoidance under its assumptions, but the resulting terminal condition remains more restrictive than the maximal safe set.

Here, numerical approximations of the maximal safe set are obtained through HJ reachability, which is too expensive to repeat within a planning cycle. The scalable reachability formulations reviewed in \cref{subsec:intro-hj-reachability} reduce this cost by exploiting system structure or by reusing earlier solutions \cite{herbert_reachability-based_2019,chen_decomposition_2018,rubies-royo_classification-based_2019,bansal_deepreach_2021}, and parameter-conditioned reachability covers structured low-dimensional variation of the environment \cite{borquez_parameter-conditioned_2023,jeong_parameterized_2024,nakamura_online_2023}. Online differential dynamic programming offers a further alternative, solving a finite-horizon reach--avoid problem by iterative local quadratic approximation around a nominal trajectory \cite{ramesh_kumar_fast_2023}. Closest to the sensor-driven setting, grid-based HJ updates treat unseen space as occupied and revise the affected parts of the safe set as free space is observed \cite{bajcsy_efficient_2019}. The reported updates average one second and yield a standalone safety controller rather than an MPC terminal constraint. None of these methods maps a high-dimensional local observation to the corresponding maximal terminal safe set at planning rates, which is what NTC-MPC provides through observation-conditioned HJ value prediction and a single terminal constraint.

\section{Problem Statement}
\label{sec:ntc-problem-statement}

This section formalizes local navigation in an unknown static environment within the MPC formulation of \cref{sec:prelim-mpc}. The goal is to design a terminal constraint set that represents the maximal safe set of the currently observed environment.

\paragraph{Setting.}
The robot follows \(\dot{x}=f(x,u)\), with \(x\in\mathcal{X}\subseteq\mathbb{R}^{n_x}\) and \(u\in\mathcal{U}\subseteq\mathbb{R}^{n_u}\), where compact \(\mathcal{U}\) encodes the actuation limits. HJ analysis uses this continuous-time model, while MPC uses \(x_{k+1}=F_\Delta(x_k,u_k)\), obtained by numerical integration with prediction interval \(\Delta t>0\), as in \cref{sec:prelim-system-dynamics}.

The environment contains an unknown, static occupied region. At every control step \(k\), the robot receives a local observation \(o_k\in\mathcal{O}\) in the sense of \cref{eq:prelim-observation-model}. Because a new observation is available at every planning step in this chapter, observations are indexed directly by \(k\) rather than by a separate update index. Concretely, the observation is an occupancy grid over a square sensing window centered at the robot's position, the representation that the mapping stack produces from the onboard sensor data \cite{elfes_using_1989,hornung_octomap_2013}, in which every cell is marked as either free or occupied. Let \(\mathcal{A}(o_k)\subset\mathbb{R}^{2}\) denote the obstacle geometry represented by this map, including any footprint inflation or physical safety margin applied by the mapping stack, and let \(p(x)\) denote the workspace position of the robot. The SDF used throughout this chapter is obtained from the occupancy grid by a distance transform, which runs in time linear in the number of cells and yields the signed distance \(\operatorname{sd}_{\mathcal{A}(o_k)}\) sampled on the same grid \cite{felzenszwalb_distance_2012,oleynikova_voxblox_2017}. Any other perception output from which such a distance field can be computed, for instance a truncated signed-distance volume or a point cloud, could take the place of the occupancy grid without changing the formulation. Following \cref{eq:prelim-point-robot-sdf-constraint}, the observation-dependent constraint function is
\begin{equation}
  c(x\mid o_k)
  :=
  \operatorname{sd}_{\mathcal{A}(o_k)}\bigl(p(x)\bigr).
  \label{eq:ntc-observed-constraint}
\end{equation}
Nonnegative values therefore place the robot outside the represented obstacle geometry, and the corresponding constraint set is \(\mathcal{C}(o_k):=\{x\in\mathcal{X}\mid c(x\mid o_k)\geq0\}\). The robot's task is to track a reference provided by a global planner, typically the segment between two consecutive waypoints, using only the local observations.

Following \cref{eq:prelim-conditioning-input}, the learned model of \cref{sec:ntc-method} does not receive the occupancy grid itself. Its conditioning input is the single-channel image
\begin{equation}
  w_k
  :=
  \eta(o_k)
  \in\mathcal{W},
  \label{eq:ntc-conditioning-input}
\end{equation}
which collects the signed distances \(\operatorname{sd}_{\mathcal{A}(o_k)}\) sampled on the cells of the sensing window. The same image supplies the values and gradients of the stage constraints \cref{eq:ntc-observed-constraint}, so one distance transform per planning cycle serves both purposes.

\paragraph{The observed maximal safe set.}
Because the environment is revealed only through observations, safety must be defined relative to the observed geometry. For an observation \(o\), let \(V_\infty(\,\cdot\mid o)\) denote the stationary infinite-horizon HJ value function obtained by solving the variational inequality \cref{eq:prelim-stationary-hji-vi} for the continuous-time model \cref{eq:prelim-undisturbed-continuous-dynamics} with the constraint function \(c(\,\cdot\mid o)\). Its nonnegative superlevel set,
\begin{equation}
  \mathcal{S}^{\star}(o)
  :=
  \left\{x\in\mathcal{X}\mid V_\infty(x\mid o)\geq0\right\},
  \label{eq:ntc-observed-maximal-safe-set}
\end{equation}
is the maximal safe set for the observed environment, that is, the largest set of states from which the observed failure set can be avoided indefinitely under the dynamics and actuation limits (\cref{eq:prelim-hj-maximal-safe-set}). Since the environment is static, the observation, and hence the field \(V_\infty(\,\cdot\mid o_k)\), is held fixed within each planning problem, in the fixed-observation form \cref{eq:prelim-observation-conditioned-function}. New information enters only when the next problem is posed with \(o_{k+1}\).

\paragraph{MPC problem.}
At every step \(k\), the local planner solves the MPC problem \cref{eq:prelim-mpc-problem} over the discretized model \cref{eq:prelim-discrete-dynamics}, with the quadratic tracking costs \cref{eq:prelim-mpc-quadratic-stage-cost,eq:prelim-mpc-quadratic-terminal-cost}, the stage constraint sets
\begin{equation}
  \mathcal{C}_{i\mid k}
  =
  \left\{x\in\mathcal{X}\mid c(x\mid o_k)\geq0\right\},
  \qquad
  i=0,\ldots,N-1,
  \label{eq:ntc-stage-constraint-sets}
\end{equation}
and a terminal constraint set \(\mathcal{C}_{f\mid k}\) that remains to be designed. Because \(x_{0\mid k}=x_k\), the constraint at \(i=0\) checks the measured current state, and the remaining stage constraints keep the predicted states collision-free over the horizon. Following \cref{sec:prelim-mpc}, the ideal terminal set is the maximal safe set for the current observation,
\begin{equation}
  \mathcal{C}_{f\mid k}
  =
  \mathcal{S}^{\star}(o_k),
  \label{eq:ntc-ideal-terminal-set}
\end{equation}
because it is the least restrictive controlled-invariant candidate set. The set \(\mathcal{S}^{\star}(o_k)\) is invariant for the continuous-time model, while the planner imposes it on the terminal state of a discrete-time prediction. Provided that the discretized model \(F_\Delta\) reproduces the continuous-time trajectories at the sampling instants, this invariance supports the usual recursive-feasibility argument while imposing no smaller terminal region than necessary. \Cref{subsec:ntc-ocp} returns to the consequences of the remaining mismatch between the two models.

\paragraph{Design problem.}
The goal is to construct, for each incoming observation \(o_k\), a tractable representation of \(\mathcal{S}^{\star}(o_k)\) with three properties. It must be available within a small fraction of the planning period, generalize to observations not seen during offline computation, and provide values and gradients efficiently inside a numerical optimizer. Exact online computation of \(V_\infty(\,\cdot\mid o_k)\) is excluded by the computational complexity of HJ reachability (\cref{subsec:prelim-hj-numerical}), so the method must approximate the mapping from the conditioning input \cref{eq:ntc-conditioning-input} to \(V_\infty(\,\cdot\mid o)\) at runtime.

\paragraph{Assumptions.}
The setting uses three assumptions. First, the environment is static, so the occupied region does not change over time, although different parts of it enter the sensing window as the robot moves. Navigation among moving obstacles is treated in \cref{ch:rntc-mpc}. Second, the sensing window is large enough, relative to the robot speed and the degree of occlusion, that every obstacle is observed before the robot can enter its BRT. This is the chapter-specific form of the sensing-adequacy condition \cref{eq:prelim-sensing-adequacy}, and it depends on the sensing configuration and the environment rather than on the terminal-set estimator alone. Third, the continuous-time model \cref{eq:prelim-undisturbed-continuous-dynamics} used for the reachability analysis is accurate and its discretization \(F_\Delta\) used for planning is consistent with it, so disturbances, model mismatch, and discretization error are not modeled explicitly in this chapter.

\section{Observation-Conditioned Neural Terminal Constraints}
\label{sec:ntc-method}

This section first explains why reachability information is placed in the terminal constraint rather than in all stage constraints. It then introduces the hypernetwork representation and its HJ-supervised training, including a loss that emphasizes the safe-set boundary, and finally assembles the optimal control problem.

\subsection{Reachability-Based Terminal Set Design}
\label{subsec:ntc-terminal-set-design}

The first design decision is where the learned safety information should enter the MPC problem. One option is to replace the geometric stage constraints \cref{eq:ntc-stage-constraint-sets} by reachability-based constraints at every stage. The option pursued here is to keep the geometric stage constraints and to add a single reachability-based constraint at the terminal stage. Two arguments favor this second choice.

The first argument is structural. Within the horizon, the stage constraints keep the sampled predicted states outside the represented obstacle geometry, and only the terminal constraint can guarantee that a safe continuation exists \emph{after} the horizon. With the maximal safe set \(\mathcal{S}^{\star}(o_k)\) as the terminal set, as in \cref{eq:ntc-ideal-terminal-set}, the finite-horizon plan cannot end in a state from which collision with the observed obstacles is inevitable. This argument assumes the exact set and a fixed observation, and the learned and observation-updated implementation is discussed in \cref{subsec:ntc-ocp}.

The second argument is computational. Applying the learned function at all \(N+1\) stages makes the cost of value and gradient queries grow with the horizon, while one terminal evaluation adds a constant cost per optimizer iteration. Under the ideal assumptions above, geometric stage constraints and an exact terminal set already suffice, so learned constraints at every stage would add no safety benefit.

\subsection{Hypernetwork Architecture}
\label{subsec:ntc-architecture}

The mapping to be learned, \(w\mapsto V_\infty(\,\cdot\mid o)\), takes the high-dimensional conditioning input \cref{eq:ntc-conditioning-input} and returns a scalar field over the state space. A model realizing this mapping must satisfy two competing requirements. Generalization across a wide range of obstacle configurations requires an expressive model, while the representation queried by the optimizer must remain small because it is evaluated and differentiated many times per planning cycle.

\begin{figure}[tb]
  \centering
  \colorlet{hypfill}{green!55!black!15}%
\colorlet{hypline}{green!50!black!65}%
\colorlet{sdffill}{orange!16}%
\colorlet{sdfline}{orange!70!black}%
\colorlet{mpcfill}{teal!14}%
\colorlet{mpcline}{teal!55!black}%
\colorlet{mlpfill}{blue!55!black!14}%
\colorlet{mlpline}{blue!55!black!60}%
\colorlet{robotfill}{yellow!65}%
\colorlet{wheelfill}{black!68}%
\colorlet{headfill}{blue!55!black!14}%

\begin{tikzpicture}[
    font=\small,
    blk/.style={draw, semithick, rounded corners=3pt, align=center,
                inner sep=1mm},
    sig/.style={font=\footnotesize, align=center, inner sep=1pt},
    arrow/.style={-{Latex[length=2.2mm]}, semithick},
    wire/.style={semithick},
    unit/.style={circle, draw=black, fill=mlpfill, semithick,
                 inner sep=0pt, minimum size=2.7mm},
  ]

  \fill[mpcfill, rounded corners=8pt] (0.30,1.15) rectangle (7.00,5.32);
  \draw[mpcline, semithick, rounded corners=8pt] (0.30,1.15) rectangle (7.00,5.32);
  \node[anchor=south east] at (6.85,1.28) {MPC};

  \node[blk, rounded corners=6pt, minimum width=3.30cm, minimum height=2.50cm]
    (hyper) at (6.30,7.25) {};
  \node[above=0.6mm of hyper] {Hypernetwork};

  \filldraw[fill=hypfill, draw=hypline, semithick]
    (7.60,8.30) -- (6.05,7.65) -- (6.05,6.85) -- (7.60,6.20) -- cycle;
  \node at (6.95,7.25) {CNN};
  \filldraw[fill=hypfill, draw=hypline, semithick]
    (5.11,6.25) rectangle (5.66,8.25);
  \node[font=\footnotesize] at (5.385,7.25) {FC};
  \draw[arrow] (6.05,7.25) -- (5.70,7.25);

  \node[blk, fill=sdffill, draw=sdfline, minimum width=3.40cm,
        minimum height=1.45cm] (sdf) at (11.65,7.25) {SDF\\computation};

  \node[blk, fill=white, minimum width=2.70cm, minimum height=1.50cm]
    (main) at (3.68,4.00) {};
  \node[above=0.6mm of main] {Main network};

  \foreach \i/\y in {a/4.28,b/3.72} {\node[unit] (in\i) at (3.05,\y) {};}
  \foreach \i/\y in {a/4.46,b/4.00,c/3.54} {\node[unit] (hid\i) at (3.68,\y) {};}
  \node[unit] (out) at (4.31,4.00) {};
  \foreach \i in {a,b}{\foreach \j in {a,b,c}{
    \draw[black, line width=0.35pt] (in\i) -- (hid\j);}}
  \foreach \j in {a,b,c}{
    \draw[black, line width=0.35pt] (hid\j) -- (out);}

  \node[blk, fill=white, minimum width=3.60cm, minimum height=0.85cm]
    (nlp) at (3.63,1.98) {NLP Solver};

  \draw[wire] (main.east) -- (5.88,4.00) -- (5.88,1.98);
  \draw[arrow] (5.88,1.98) -- (nlp.east);
  \node[sig, anchor=south] at (5.88,4.10) {$\hat{V}_{\Psi}(x_{N\mid k})$};
  \draw[wire] (nlp.west) -- (1.32,1.98) -- (1.32,4.00);
  \draw[arrow] (1.32,4.00) -- (main.west);
  \node[sig, anchor=south west] at (1.38,4.10) {$x_{N\mid k}$};

  \draw[wire] (0.30,6.00) -- (2.08,6.00);
  \draw[arrow] (2.08,6.00) -- (2.08,5.32);
  \node[sig, anchor=south west] at (0.32,6.08) {Reference};

  \draw[wire] (5.11,7.25) -- (3.61,7.25) -- (3.61,5.62);
  \draw[arrow] (3.61,5.62) -- (3.61,5.32);
  \node[sig, anchor=south] at (3.55,7.33) {$\Theta(w_k)$};

  \draw[wire] (sdf.west) -- (7.95,7.25);
  \draw[arrow] (7.95,7.25) -- (7.62,7.25);
  \draw[wire] (8.95,7.25) -- (8.95,4.20);
  \draw[arrow] (8.95,4.20) -- (7.00,4.20);
  \fill (8.95,7.25) circle (1.7pt);
  \node[sig, anchor=south] at (9.40,7.33) {$w_k$};
  \node[sig, anchor=south] at (7.97,4.28) {$c(\,\cdot\mid w_k)$};

  \node[blk, fill=white, rounded corners=6pt, minimum width=2.70cm,
        minimum height=2.90cm] (robot) at (11.65,3.28) {};
  \foreach \yl/\yu in {3.35/4.15, 2.17/2.97} {
    \fill[wheelfill, rounded corners=1.2pt] (10.57,\yl) rectangle (10.83,\yu);
    \fill[wheelfill, rounded corners=1.2pt] (12.47,\yl) rectangle (12.73,\yu);}
  \filldraw[fill=robotfill, draw=black, semithick]
    (10.80,2.01) rectangle (12.50,4.31);
  \filldraw[fill=headfill, draw=black, semithick]
    (11.24,3.54) -- (12.06,3.54) -- (11.65,4.18) -- cycle;

  \draw[arrow] (7.00,3.30) -- (robot.west);
  \node[sig, anchor=south] at (8.66,3.38) {Control $u^{\star}_{0\mid k}$};

  \draw[arrow] (11.65,4.73) -- (sdf.south);
  \node[sig, anchor=east] at (11.53,5.60) {Sensor data};

  \draw[wire] (11.65,1.83) -- (11.65,0.38) -- (3.32,0.38);
  \draw[arrow] (3.32,0.38) -- (3.32,1.15);
  \node[sig, anchor=south] at (8.66,0.46) {Feedback $x_k$};
\end{tikzpicture}
  \caption[Architecture of the NTC-MPC local planner]{Architecture of the NTC-MPC local planner. From the current sensor data, an SDF is computed and used both as the geometric stage constraint of the MPC problem and as the conditioning input \(w_k\) of the hypernetwork. The hypernetwork, a convolutional backbone with a fully connected head, outputs the parameters of the compact main network, which represents the estimated value function \(\hat{V}_{\Psi}(\,\cdot\mid w_k)\). The nonlinear program (NLP) solver queries the main network for the value and gradient of the terminal constraint at every optimizer iteration, while the hypernetwork is evaluated only once per planning cycle.}
  \label{fig:ntc-architecture}
\end{figure}
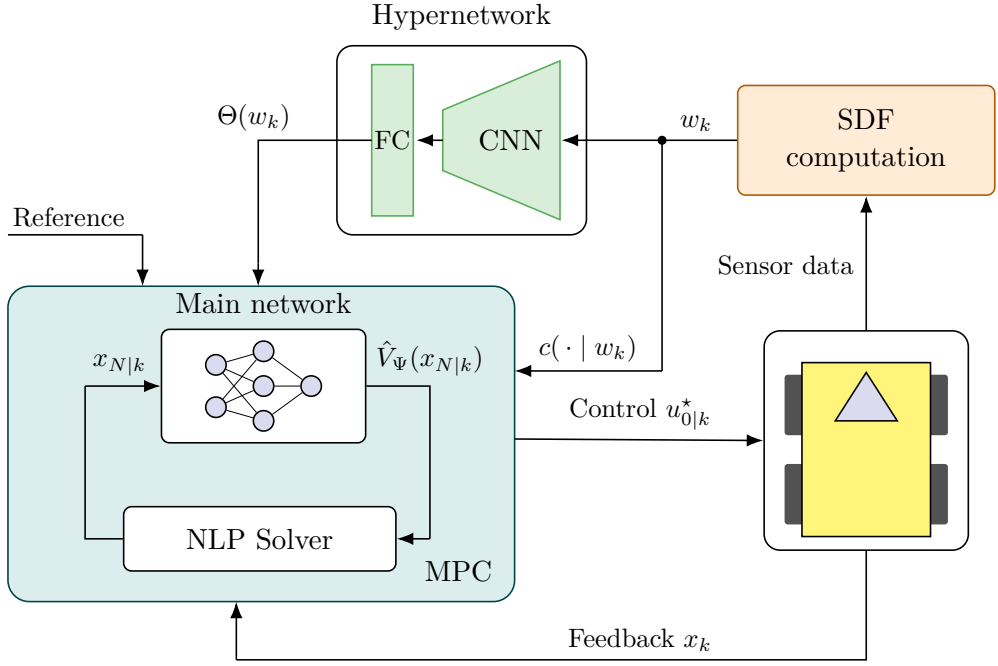

NTC-MPC resolves this conflict with the hypernetwork construction of \cref{subsec:prelim-hypernetworks}, illustrated in \cref{fig:ntc-architecture}. A hypernetwork \(H_{\Psi}\) with trainable parameters \(\Psi\) processes the conditioning input and outputs the parameters of a main network \(M_{\Theta(w)}\),
\begin{equation}
  \Theta(w)
  =
  H_{\Psi}(w),
  \qquad
  \hat{V}_{\Psi}(x\mid w)
  =
  M_{\Theta(w)}(x),
  \label{eq:ntc-hypernetwork}
\end{equation}
so that the learned scalar field \(\hat{V}_{\Psi}(\,\cdot\mid w)\) approximates the value function \(V_\infty(\,\cdot\mid o)\) and, through its superlevel set, the maximal safe set \(\mathcal{S}^{\star}(o)\). Since the conditioning input is a single-channel SDF image over the sensing window, the hypernetwork consists of a deep convolutional neural network (CNN) backbone followed by a fully connected head whose output dimension equals the number of parameters of the main network. The main network is a small multilayer perceptron (MLP) that takes a state \(x\) and returns the scalar \(\hat{V}_{\Psi}(x\mid w)\). Its early layers use sinusoidal activation functions, which suit implicit field regression and yield smooth gradients \cite{sitzmann_implicit_2020}, while the remaining layers use scaled exponential linear unit (SELU) activations.

This split matches how the planner uses the model. The conditioning input changes once per planning cycle, so the expressive hypernetwork is evaluated once per cycle, outside the optimization loop, and the optimizer queries only the main network, whose gradients are provided by automatic differentiation. Compared with the single-network architecture evaluated in \cref{subsec:ntc-training-eval}, the factorization retains an expressive scene encoder while reducing the size of the network queried inside the optimization loop by orders of magnitude \cite{ha_hypernetworks_2017,galanti_modularity_2020}.

\subsection{Reachability-Supervised Training}
\label{subsec:ntc-training}

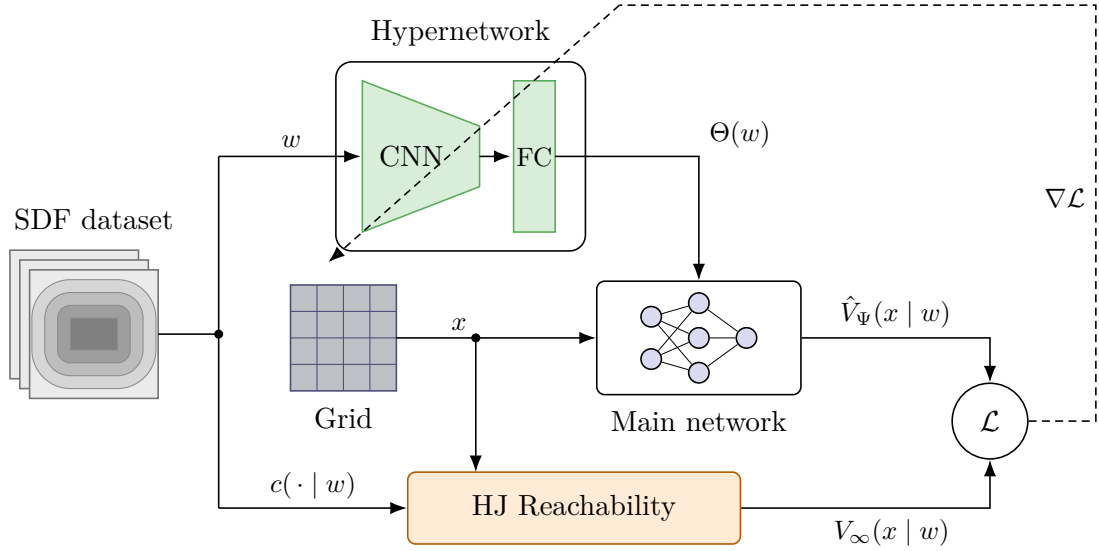
\begin{figure}[tb]
  \centering
  \colorlet{hypfill}{green!55!black!15}%
\colorlet{hypline}{green!50!black!65}%
\colorlet{hjfill}{orange!16}%
\colorlet{hjline}{orange!70!black}%
\colorlet{mlpfill}{blue!55!black!14}%
\colorlet{gridfill}{blue!10!black!25}%
\colorlet{gridline}{blue!25!black!70}%
\colorlet{sdffill}{black!8}%
\colorlet{sdfline}{black!55}%
\colorlet{obsfill}{black!50}%

\begin{tikzpicture}[
    font=\small,
    blk/.style={draw, semithick, rounded corners=3pt, align=center,
                inner sep=1mm},
    sig/.style={font=\footnotesize, align=center, inner sep=1pt},
    arrow/.style={-{Latex[length=2.2mm]}, semithick},
    wire/.style={semithick},
    grad/.style={-{Latex[length=2.2mm]}, semithick, densely dashed},
    unit/.style={circle, draw=black, fill=mlpfill, semithick,
                 inner sep=0pt, minimum size=2.7mm},
  ]

  \foreach \dx/\dy in {-0.26/0.26, -0.13/0.13} {
    \filldraw[fill=sdffill, draw=sdfline, semithick]
      (0.40+\dx,2.35+\dy) rectangle (2.10+\dx,4.05+\dy);}
  \filldraw[fill=sdffill, draw=sdfline, semithick]
    (0.40,2.35) rectangle (2.10,4.05);
  \begin{scope}
    \clip (0.40,2.35) rectangle (2.10,4.05);
    \foreach \d/\lvl in {0.51/16, 0.34/26, 0.17/38} {
      \filldraw[fill=black!\lvl, draw=black!45, thin, rounded corners=\d cm]
        (0.94-\d,2.99-\d) rectangle (1.56+\d,3.41+\d);}
  \end{scope}
  \filldraw[fill=obsfill, draw=black!45, thin] (0.94,2.99) rectangle (1.56,3.41);
  \node[anchor=south] at (1.25,4.45) {SDF dataset};

  \node[blk, rounded corners=6pt, minimum width=3.30cm, minimum height=2.50cm]
    (hyper) at (6.10,5.55) {};
  \node[above=0.6mm of hyper] {Hypernetwork};

  \filldraw[fill=hypfill, draw=hypline, semithick]
    (4.80,6.55) -- (6.35,5.95) -- (6.35,5.15) -- (4.80,4.55) -- cycle;
  \node at (5.45,5.55) {CNN};
  \filldraw[fill=hypfill, draw=hypline, semithick]
    (6.80,4.55) rectangle (7.35,6.55);
  \node[font=\footnotesize] at (7.075,5.55) {FC};
  \draw[arrow] (6.35,5.55) -- (6.76,5.55);

  \filldraw[fill=gridfill, draw=gridline, semithick]
    (3.85,2.45) rectangle (5.25,3.85);
  \foreach \i in {1,2,3} {
    \draw[gridline, thin] (3.85+0.35*\i,2.45) -- (3.85+0.35*\i,3.85);
    \draw[gridline, thin] (3.85,2.45+0.35*\i) -- (5.25,2.45+0.35*\i);}
  \node[anchor=north] at (4.55,2.37) {Grid};

  \node[blk, fill=white, minimum width=2.70cm, minimum height=1.50cm]
    (main) at (9.25,3.15) {};
  \node[below=0.6mm of main] {Main network};

  \foreach \i/\y in {a/3.43,b/2.87} {\node[unit] (in\i) at (8.62,\y) {};}
  \foreach \i/\y in {a/3.61,b/3.15,c/2.69} {\node[unit] (hid\i) at (9.25,\y) {};}
  \node[unit] (out) at (9.88,3.15) {};
  \foreach \i in {a,b}{\foreach \j in {a,b,c}{
    \draw[black, line width=0.35pt] (in\i) -- (hid\j);}}
  \foreach \j in {a,b,c}{
    \draw[black, line width=0.35pt] (hid\j) -- (out);}

  \node[blk, fill=hjfill, draw=hjline, minimum width=4.40cm,
        minimum height=0.95cm] (hj) at (7.60,0.90) {HJ Reachability};

  \node[circle, draw, semithick, fill=white, inner sep=0pt,
        minimum size=1.00cm] (loss) at (13.10,2.05) {$\mathcal{L}$};

  \draw[wire] (2.10,3.20) -- (2.90,3.20);
  \draw[wire] (2.90,5.55) -- (2.90,0.90);
  \draw[arrow] (2.90,5.55) -- (4.78,5.55);
  \draw[arrow] (2.90,0.90) -- (hj.west);
  \fill (2.90,3.20) circle (1.7pt);
  \node[sig, anchor=south] at (3.86,5.63) {$w$};
  \node[sig, anchor=south] at (4.15,0.98) {$c(\,\cdot\mid w)$};

  \draw[wire] (7.35,5.55) -- (9.25,5.55) -- (9.25,4.20);
  \draw[arrow] (9.25,4.20) -- (main.north);
  \node[sig, anchor=south west] at (9.35,5.62) {$\Theta(w)$};

  \draw[wire] (5.25,3.15) -- (6.30,3.15);
  \draw[wire] (6.30,3.15) -- (6.30,1.55);
  \draw[arrow] (6.30,3.15) -- (main.west);
  \draw[arrow] (6.30,1.55) -- (6.30,1.375);
  \fill (6.30,3.15) circle (1.7pt);
  \node[sig, anchor=south east] at (6.22,3.21) {$x$};

  \draw[wire] (main.east) -- (13.10,3.15) -- (13.10,2.75);
  \draw[arrow] (13.10,2.75) -- (loss.north);
  \node[sig, anchor=south] at (11.85,3.25) {$\hat{V}_{\Psi}(x\mid w)$};
  \draw[wire] (hj.east) -- (13.10,0.90) -- (13.10,1.35);
  \draw[arrow] (13.10,1.35) -- (loss.south);
  \node[sig, anchor=north] at (11.80,0.80) {$V_\infty(x\mid w)$};

  \draw[densely dashed, semithick]
    (loss.east) -- (14.50,2.05) -- (14.50,7.55) -- (8.20,7.55);
  \draw[grad] (8.20,7.55) -- (4.35,4.15);
  \node[sig, anchor=east] at (14.40,5.00) {$\nabla\mathcal{L}$};
\end{tikzpicture}
  \caption[Training procedure for the NTC model]{Training procedure for the NTC model. Each SDF conditioning input \(w\) from the dataset is processed by the hypernetwork into the main-network parameters \(\Theta(w)\), and all states of a state-space grid are propagated through the main network to predict the value function over the grid. HJ reachability analysis, using the same SDF as the constraint function \(c(\,\cdot\mid w)\), provides the numerical target values \(V_\infty(\,\cdot\mid w)\). The loss \(\mathcal{L}\) compares predicted and target value functions, and its gradient updates the hypernetwork parameters \(\Psi\). The main-network parameters are outputs of the hypernetwork rather than free variables. The figure shows a single training pair, so the batch and grid indices of \cref{eq:ntc-rwmse-loss} are omitted.}
  \label{fig:ntc-training}
\end{figure}

The hypernetwork is trained offline in a supervised fashion, following the pipeline of \cref{subsec:prelim-offline-online-safety-learning}, which \cref{fig:ntc-training} shows for NTC-MPC. The training data consist of a set of local observations, the conditioning input derived from each of them, and the numerically computed HJ value function over a state-space grid.

The observations are obtained by collecting occupancy grid maps at random poses in a simulation environment. Each map is converted into the corresponding conditioning input \(w^{(m)}=\eta(o^{(m)})\) by the distance transform of \cref{eq:ntc-conditioning-input}, and rotation and mirroring augment the set at no additional sensing cost. For each observation \(o^{(m)}\) in the dataset \(\{o^{(m)}\}_{m=1}^{M}\), the numerical HJ target is computed for the continuous-time model with a grid-based solver as described in \cref{subsec:prelim-hj-numerical}. The constraint function is the observed \(c(\,\cdot\mid o^{(m)})\) of \cref{eq:ntc-observed-constraint}, and the value function is propagated backward in time until it converges. The result approximates the stationary infinite-horizon value function \(V_\infty(\,\cdot\mid o^{(m)})\) of \cref{eq:prelim-stationary-hj-value}, whose nonnegative region is the maximal safe set. The states at which the targets are available form a uniform grid \(\{x_j\}_{j=1}^{K}\subset\mathcal{X}\) determined by the state limits and the desired resolution.

Training then uses minibatch gradient-based optimization of the hypernetwork parameters \(\Psi\). For each conditioning input in a batch, the hypernetwork generates the main-network parameters and the full state grid is propagated through the main network. The loss then compares the predicted values \(\hat{V}_{\Psi}(x_j\mid w^{(m)})\) with the targets \(V_\infty(x_j\mid w^{(m)})\), which implements the approximation \cref{eq:prelim-conditioned-value-approximation}. Only \(\Psi\) is trainable, since the main network has no free parameters of its own. Every reachability problem is solved offline, and the trained hypernetwork must generalize the mapping to unseen observations. The choice of the loss function determines where in the state space the finite model capacity is emphasized.

\paragraph{Radially weighted regression loss.}
\label{subsec:ntc-rwmse}
The natural first choice of training objective is the mean squared error (MSE) between predicted and numerical target values over the grid. MSE assigns the same weight to squared errors everywhere, while the downstream use of the learned field is not uniform. The terminal constraint classifies states as safe or unsafe by the \emph{sign} of \(\hat{V}_{\Psi}\), so the classification depends only on the location of the zero level set. A small value error far inside either region may not change this classification, while the same error near the boundary can turn an unsafe state into a predicted-safe one. This mismatch between a uniformly weighted regression objective and a boundary-sensitive use case motivates a weighted objective.

To concentrate accuracy where classification is decided, the training uses the \emph{radially weighted mean squared error} (RWMSE). Let a training dataset contain \(M\) observations, evaluated at the \(K\) grid states, and abbreviate the target values as \(V^{(m)}_j:=V_\infty(x_j\mid w^{(m)})\) and the predictions as \(\hat{V}^{(m)}_j:=\hat{V}_{\Psi}(x_j\mid w^{(m)})\). The loss is
\begin{align}
  \mathcal{L}_{\mathrm{RWMSE}}(\Psi)
   & =
  \frac{1}{M}\sum_{m=1}^{M}
  \frac{1}{K}\sum_{j=1}^{K}
  \omega_{m,j}
  \left(
  V^{(m)}_j-\hat{V}^{(m)}_j
  \right)^{2},
  \label{eq:ntc-rwmse-loss}
  \\
  \omega_{m,j}
   & =
  1+\alpha\exp\!\left(
  -\beta\left(V^{(m)}_j\right)^{2}
  \right),
  \qquad
  \alpha,\beta>0.
  \label{eq:ntc-rwmse-weight}
\end{align}
The weight \(\omega_{m,j}\) attains its maximum \(1+\alpha\) exactly on the numerical zero level set, where \(V^{(m)}_j=0\), and decays toward one as \(\lvert V^{(m)}_j\rvert\) grows. The hyperparameter \(\alpha\) sets the additional weight assigned near the boundary, and \(\beta\) determines how quickly that emphasis decays with the magnitude of the target value. The term \enquote{radially weighted} refers to the symmetric decay with \(\lvert V^{(m)}_j\rvert\) and does not require the value to equal a Euclidean distance in state space. In the limit \(\alpha\rightarrow0\), RWMSE recovers standard MSE. Emphasizing the boundary is also consistent with invariance analysis, in which Nagumo-type conditions constrain the dynamics precisely on the boundary \cite{ames_control_2019}.

Two properties of RWMSE should be stated precisely. First, it remains a global regression objective, because all states contribute and only the \emph{relative} emphasis shifts toward the boundary. Second, it is neither one-sided nor conservative, since it does not constrain the sign of the prediction error and cannot guarantee that the predicted safe set is contained in the true one. Its contribution is an improved recovery of the zero level set, which \cref{subsec:ntc-training-eval} quantifies and \cref{subsec:ntc-simulations} connects to closed-loop success rates.

\subsection{NTC-MPC Formulation}
\label{subsec:ntc-ocp}

With these components, the complete planner is obtained as follows. Given the current state \(x_k\), the current conditioning input \(w_k\), and the main-network parameters \(\Theta(w_k)=H_{\Psi}(w_k)\) computed once for this cycle, NTC-MPC solves
\begin{subequations}
  \label{eq:ntc-mpc-problem}
  \begin{align}
    \min_{\substack{x_{0\mid k},\ldots,x_{N\mid k}                              \\ u_{0\mid k},\ldots,u_{N-1\mid k}}}
    \quad &
    \ell_{f\mid k}(x_{N\mid k})
    +
    \sum_{i=0}^{N-1}
    \ell_{i\mid k}(x_{i\mid k},u_{i\mid k})
    \label{eq:ntc-mpc-objective}
    \\
    \text{ s.t. }\quad
          &
    x_{0\mid k}=x_k,
    \label{eq:ntc-mpc-initial-condition}
    \\
          &
    x_{i+1\mid k}
    =
    F_\Delta(x_{i\mid k},u_{i\mid k}),
    \quad
    i=0,\ldots,N-1,
    \label{eq:ntc-mpc-dynamics}
    \\
          &
    u_{i\mid k}\in\mathcal{U},
    \quad
    i=0,\ldots,N-1,
    \label{eq:ntc-mpc-input-constraints}
    \\
          &
    x_{i\mid k}\in\mathcal{X},
    \quad
    i=0,\ldots,N,
    \label{eq:ntc-mpc-state-constraints}
    \\
          &
    c(x_{i\mid k}\mid o_k)\geq0,
    \quad
    i=0,\ldots,N-1,
    \label{eq:ntc-mpc-stage-safety}
    \\
          &
    \hat{V}_{\Psi}(x_{N\mid k}\mid w_k)\geq0,
    \label{eq:ntc-terminal}
  \end{align}
\end{subequations}
with the quadratic tracking costs of \cref{sec:prelim-mpc}. The geometric stage constraints \cref{eq:ntc-mpc-stage-safety} keep the sampled predicted states collision-free within the horizon, and the neural terminal constraint \cref{eq:ntc-terminal} keeps the final predicted state inside the predicted safe set. The corresponding terminal constraint set is
\begin{equation}
  \mathcal{C}_{f\mid k}
  =
  \left\{
  x\in\mathcal{X}
  \mid
  \hat{V}_{\Psi}(x\mid w_k)\geq0
  \right\}
  \approx
  \mathcal{S}^{\star}(o_k).
  \label{eq:ntc-learned-terminal-set}
\end{equation}
The terminal constraint restricts the admissible terminal states but does not define a navigation objective or preferred direction, so it can be combined with different stage costs, references, and local goals.

From an implementation standpoint, the optimization problem defined by \cref{eq:ntc-mpc-problem} is a nonlinear program (NLP) in which the main-network parameters \(\Theta(w_k)\), the sampled SDF, the estimated state, and the reference enter as time-varying parameters rather than as decision variables. The problem structure is identical at every cycle, which allows the solver code to be generated and compiled once, and only the parameter values change when a new observation arrives. Because the main network is composed of elementary differentiable operations, it integrates directly into the NLP computational graph, and automatic differentiation supplies the derivatives required by the solver without finite-difference approximations. Each cycle requires one hypernetwork inference and one NLP solution.

The classical recursive-feasibility argument would apply if the terminal set were the exact maximal safe set for a fixed observation, a terminal feedback law kept that set invariant under the same discrete-time model and constraints used by MPC, and the initial problem were feasible. The implemented scheme does not satisfy all of these premises, so recursive feasibility and collision avoidance are assessed empirically in the next section, and \cref{sec:ntc-discussion} states the scope of the results.

\section{Experimental Evaluation}
\label{sec:ntc-experiments}

The evaluation measures safe-set accuracy, collision avoidance, solve time, transfer to hardware without fine-tuning, and the computational benefit of the hypernetwork. Generalization is tested in environments and obstacle configurations absent from training.\footnote{The implementation is available at \url{https://github.com/bojan-derajic/NTC-MPC}.}

\subsection{Robot Models}
\label{subsec:ntc-robot-models}

Two nonholonomic robot models of increasing complexity are used. They are chosen so that their maximal safe sets differ qualitatively from the geometric free space, which exposes the difference between distance-based and reachability-based constraints. Both are stated in continuous time, which is the form used by the HJ computations of \cref{subsec:ntc-training-eval}, while the planners optimize over their forward-Euler discretization, as specified in \cref{subsec:ntc-baselines}.

\paragraph{Dubins car.}
The first model is a first-order unicycle with constant forward speed and bounded turning rate, commonly called the Dubins car. The state \(x=(p_x,p_y,\theta)\) collects the planar position \(p(x)=(p_x,p_y)\) and the heading angle \(\theta\), the control input is the angular velocity \(u=\omega\), and the continuous-time dynamics are
\begin{equation}
  \dot{x}
  =
  \begin{bmatrix}
    v\cos\theta \\
    v\sin\theta \\
    \omega
  \end{bmatrix},
  \label{eq:ntc-dubins-dynamics}
\end{equation}
with fixed speed \(v=\SI{0.5}{\meter\per\second}\) and actuation limit \(\omega\in[-0.25,0.25]\)~\si{\radian\per\second}. This model describes systems whose forward speed cannot be modulated quickly or at all, such as fixed-wing aircraft or ground vehicles for which braking is unavailable or undesirable. Because the robot cannot stop, a state heading directly toward an obstacle at close range may already admit no collision-free continuation, which makes terminal-set reasoning particularly consequential.

\paragraph{Dynamic unicycle.}
The second model is a second-order unicycle that captures the velocity dynamics of differential-drive and skid-steer platforms. The state \(x=(p_x,p_y,\theta,v,\omega)\) additionally contains the linear and angular velocities, the control input \(u=(a_v,a_\omega)\) consists of the linear and angular accelerations, and the dynamics are
\begin{equation}
  \dot{x}
  =
  \begin{bmatrix}
    v\cos\theta \\
    v\sin\theta \\
    \omega      \\
    a_v         \\
    a_\omega
  \end{bmatrix},
  \label{eq:ntc-unicycle-dynamics}
\end{equation}
with state and actuation limits \(v\in[-1.0,1.0]\)~\si{\meter\per\second}, \(\omega\in[-0.5,0.5]\)~\si{\radian\per\second}, \(a_v\in[-0.25,0.25]\)~\si{\meter\per\square\second}, and \(a_\omega\in[-1.0,1.0]\)~\si{\radian\per\square\second}. Here the safe set depends on the velocities through braking and turning distances, in direct analogy to the braking example of \cref{subsec:prelim-controlled-invariant-safe-sets}, and its five-dimensional domain approaches the practical limit of grid-based HJ solvers (\cref{subsec:prelim-hj-numerical}).

\subsection{Learning and Computational Evaluation}
\label{subsec:ntc-training-eval}
\label{subsec:ntc-single-network}

\paragraph{Datasets and targets.}
The training observations were collected in a simulated warehouse environment (\cref{fig:ntc-sim-environments}, left). The dataset was built from \num{2500} local occupancy grid maps of size \(6\times6\)~\si{\meter} (\(100\times100\) cells at \SI{0.06}{\meter} resolution), recorded at random poses. Each original map and its vertically mirrored version were retained in four orientations, separated by \(90^{\circ}\), which yielded eight variants per map and \num{20000} SDF conditioning inputs in total. For the Dubins car, the numerical HJ targets were computed on a \(100\times100\times20\) grid over \((p_x,p_y,\theta)\), and for the dynamic unicycle on a \(50\times50\times7\times7\times7\) grid over the five-dimensional state space. The value functions were propagated until temporal convergence with the \texttt{hj\_reachability} toolbox\footnote{\url{https://github.com/StanfordASL/hj_reachability}}. The \num{20000} conditioning inputs were divided into 80\% for training and 20\% for validation after the augmentation, so rotated or mirrored variants of the same original map can appear in both sets. The validation metrics thus assess recovery of the safe-set boundary on held-out variants rather than transfer to unseen scenes, and generalization to new geometry is assessed separately through the unseen house environment of the closed-loop study in \cref{subsec:ntc-simulations}.

\begin{table}[tb]
  \caption{NTC-MPC configuration shared by both robot models.}
  \label{tab:ntc-training-config-shared}
  \centering
  \begin{tabular}{@{}ll@{}}
    \toprule
    Observation (SDF) size     & \(100\times100\) cells, \(6\times6\)~\si{\meter}      \\
    Dataset size (train/val)   & \num{20000} SDFs (80\%/20\%)                          \\
    Hypernetwork               & CNN backbone with fully connected head                \\
    Hypernetwork parameters    & \(\sim4.8\times10^{6}\)                               \\
    Main network hidden layers & \([32,32,32,16,16,16,8,8,8]\), sine + SELU            \\
    Main network output        & 1 (value function)                                    \\
    Loss                       & RWMSE, \(\alpha=1000\), \(\beta=10\)                  \\
    Optimizer                  & Adam                                                  \\
    \bottomrule
  \end{tabular}
\end{table}

\begin{table}[tb]
  \caption{NTC-MPC configuration specific to each robot model.}
  \label{tab:ntc-training-config-model}
  \centering
  \begin{tabular}{@{}lll@{}}
    \toprule
                             & Dubins car               & Dynamic unicycle                    \\
    \midrule
    State-space grid         & \(100\times100\times20\) & \(50\times50\times7\times7\times7\) \\
    Main network inputs      & 3                        & 5                                   \\
    Main network parameters  & \num{3601}               & \num{3665}                          \\
    Epochs / batch size      & 150 / 32                 & 100 / 16                            \\
    Training time (RTX~3090) & \(\sim\)\SI{7.5}{\hour}  & \(\sim\)\SI{18}{\hour}              \\
    \bottomrule
  \end{tabular}
\end{table}

\paragraph{Model and training configuration.}
\Cref{tab:ntc-training-config-shared} lists the hypernetwork and main-network configuration shared by both robot models, and \cref{tab:ntc-training-config-model} lists the entries that differ between them. In the main network, the first three hidden layers use sinusoidal activations and the remaining ones use SELU. The two models differ only in the dimension of the state space, which determines the number of main-network inputs, the resolution of the HJ target grid, and the number of main-network parameters, and in the training schedule.

\begin{figure}[tb]
  \centering
  \includegraphics[width=\textwidth]{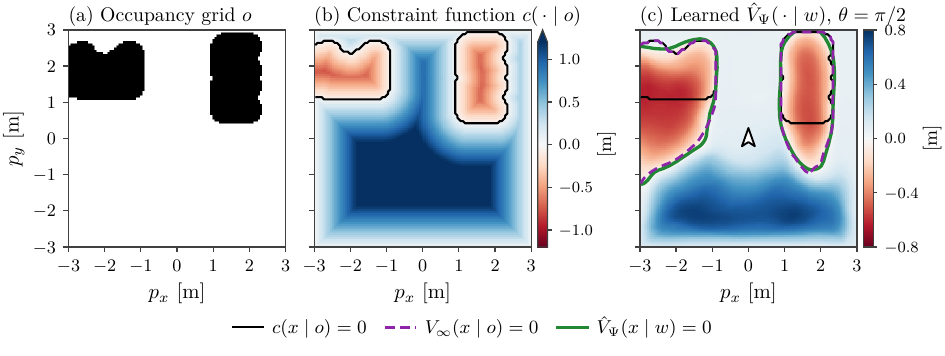}
  \caption[Learned safe set for a validation observation of the Dubins car]{Learned safe set for a sample observation from the validation dataset (Dubins car). (a)~Local occupancy grid \(o\). (b)~The corresponding SDF \(c(\,\cdot\mid o)\), whose zero level set (black) outlines the obstacles. (c)~Slice of the learned value function \(\hat{V}_{\Psi}(\,\cdot\mid w)\) at heading \(\theta=\pi/2\), that is, for a robot at the marked position (dart) moving upward. Red shading marks the states predicted to be unsafe. The predicted safe-set boundary \(\hat{V}_{\Psi}=0\) (green) closely follows the numerical HJ boundary \(V_\infty=0\) (dashed). The unsafe region extends beyond the obstacle outlines, most visibly below the obstacles, which the robot approaches head-on at this heading.}
  \label{fig:ntc-ogm-sdf-vf}
\end{figure}

\paragraph{Qualitative results.}
\Cref{fig:ntc-ogm-sdf-vf} shows a validation observation for the Dubins car. The unsafe region extends beyond the obstacles, especially below them, where the approaching constant-speed robot cannot turn away in time. The learned boundary closely follows the numerical HJ boundary at the displayed heading, capturing information absent from the SDF.

\begin{figure}[tb]
  \centering
  \includegraphics[width=\textwidth]{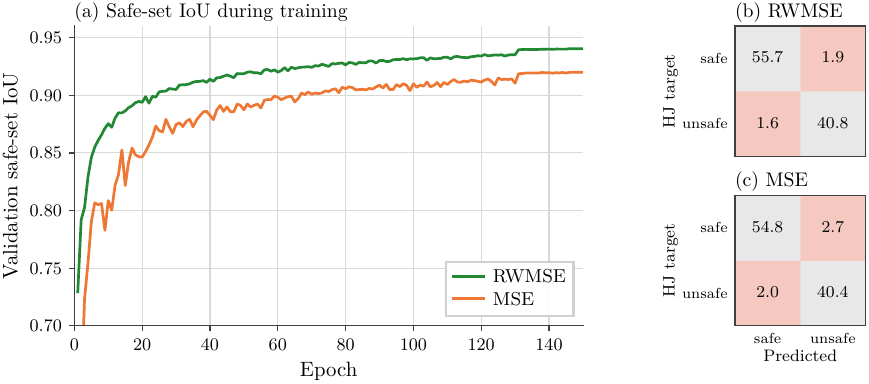}
  \caption[Effect of the RWMSE and MSE training objectives for the Dubins car]{Effect of the training objective for the Dubins car model, evaluated on the validation set. (a)~Safe-set IoU over training epochs for models trained with RWMSE and MSE under identical settings and initialization. (b),~(c)~Confusion matrices of the safe/unsafe classification for RWMSE and MSE, with entries in percent of all grid states. RWMSE yields a consistently higher IoU and fewer misclassifications on both sides of the boundary.}
  \label{fig:ntc-mse-vs-rwmse}
\end{figure}

\paragraph{Effect of the RWMSE loss.}
To isolate the effect of the loss function, training was repeated with the standard MSE loss under identical settings and weight initialization. Since loss values of differently weighted objectives are not directly comparable, the comparison used two task-relevant metrics on the validation set. The first is the intersection over union (IoU) between the predicted safe set and the numerical HJ safe set, that is, the ratio between the volume of their intersection and the volume of their union over the evaluation grid. The second is the confusion matrix of the induced classification of states as safe or unsafe. \Cref{fig:ntc-mse-vs-rwmse} presents the results for the Dubins car. RWMSE attained a higher IoU throughout training and converged to approximately \(0.94\), compared with \(0.92\) for MSE. The confusion matrices show reductions in both error types, and in particular the fraction of all states that were numerically unsafe but predicted safe decreased from \(2.0\%\) to \(1.6\%\). This improvement required no additional data and did not change the architecture or online inference cost.

\paragraph{Online cost of the learned mapping.}
The trained hypernetwork requires about \SI{2}{\milli\second} per inference on the GPU for both robot models. Solving the corresponding HJ problem numerically for a single observation takes about \SI{1}{\second} for the Dubins car and \SI{2}{\second} for the dynamic unicycle on the same hardware.
The learned model accelerates the mapping from observation to safe set by roughly three orders of magnitude, which turns an offline analysis tool into a component that can be used online in planning.

\paragraph{Architecture comparison.}
The two-network design is compared with a single network conditioned on the SDF input. For this purpose, a single MLP with three hidden layers of 512~neurons is considered, following the parameter-conditioned reachability approach \cite{borquez_parameter-conditioned_2023}, in which the SDF is treated as a vector of \enquote{virtual states} appended to the robot state at the network input. With a \(100\times100\) SDF, the joint input has more than \(10^{4}\) dimensions, roughly three orders of magnitude more than the parameter dimensions reported in \cite{borquez_parameter-conditioned_2023}, and the model has about \(5.6\times10^{6}\) parameters, the same order as the hypernetwork.

The comparison identifies two prohibitive costs. Online, the entire conditioned network lies inside the optimization loop, so every solver iteration evaluates the full network and propagates state derivatives through its approximately \(5.6\times10^{6}\) parameters. Using this model as the terminal constraint for the Dubins car with \(N=10\) requires about \SI{1.6}{\second} per MPC solution, compared with about \SI{5.5}{\milli\second} for NTC-MPC, a reduction in solve time by more than two orders of magnitude. Offline, training requires propagating the full state-space grid through the model for every observation. The intermediate activations of the large joint network then occupy about \SI{9}{\giga\byte} of memory per observation, compared with about \SI{0.7}{\giga\byte} for the hypernetwork and main network combined, because only the compact main network is applied to the grid.

\subsection{Planner Baselines and Implementation}
\label{subsec:ntc-baselines}

The evaluation compares three local planners with the same costs, dynamics, discretization, and solver implementation. They differ only in the safety constraints under study, which isolates the effect of the constraint design. The first baseline, SDF-MPC, uses the observed geometric constraint at every stage including the terminal one,
\begin{equation}
  c(x_{i\mid k}\mid o_k)\geq0,
  \qquad
  i=0,\ldots,N,
  \label{eq:ntc-sdf-mpc}
\end{equation}
and thus represents the standard distance-constrained MPC planner. The second baseline, DCBF-MPC, additionally shapes the rate at which the obstacle distance may decrease, using the discrete-time CBF (DCBF) constraint of \cite{zeng_safety-critical_2021},
\begin{equation}
  c(x_{i\mid k}\mid o_k)-c(x_{i-1\mid k}\mid o_k)
  \geq
  -\gamma\,c(x_{i-1\mid k}\mid o_k),
  \qquad
  i=1,\ldots,N,
  \label{eq:ntc-dcbf-mpc}
\end{equation}
with decay parameter \(\gamma\in(0,1]\). Each predicted step may reduce the obstacle distance by at most the fraction \(\gamma\) of its current value, which slows the approach toward obstacles. The measured current state \(x_{0\mid k}=x_k\) is checked against the geometric constraint in SDF-MPC and NTC-MPC. For DCBF-MPC, its nonnegative distance value serves as the initial condition of a recursion, because \cref{eq:ntc-dcbf-mpc} implies \(c(x_{i\mid k}\mid o_k)\geq(1-\gamma)c(x_{i-1\mid k}\mid o_k)\) and thereby propagates a nonnegative predicted distance to all later stages. The proposed method, NTC-MPC, solves \cref{eq:ntc-mpc-problem} with the geometric constraint for stages \(i=0,\ldots,N-1\) and the neural terminal constraint \cref{eq:ntc-terminal} at stage \(N\). To quantify the contribution of the loss function in closed loop, NTC-MPC is evaluated with models trained with both RWMSE and MSE.

All planners are implemented in the CasADi framework \cite{andersson_casadi_2019} with IPOPT as the NLP solver and deployed as a ROS~2 package \cite{macenski_robot_2022}, and the solver code is generated and compiled ahead of deployment. The dynamics are discretized by the forward Euler method with prediction interval \(\Delta t=\SI{0.1}{\second}\), and the planner runs at \SI{10}{\hertz}.
The hypernetwork runs as a PyTorch model on the GPU, while the main network is expressed with CasADi functions so that it becomes part of the NLP computational graph. Its parameters, together with the SDF samples, the estimated state, and the reference, enter the solver as time-varying parameters.

\subsection{Simulation Experiments}
\label{subsec:ntc-simulations}

\begin{figure}[tb]
  \centering
  \includegraphics[width=\textwidth]{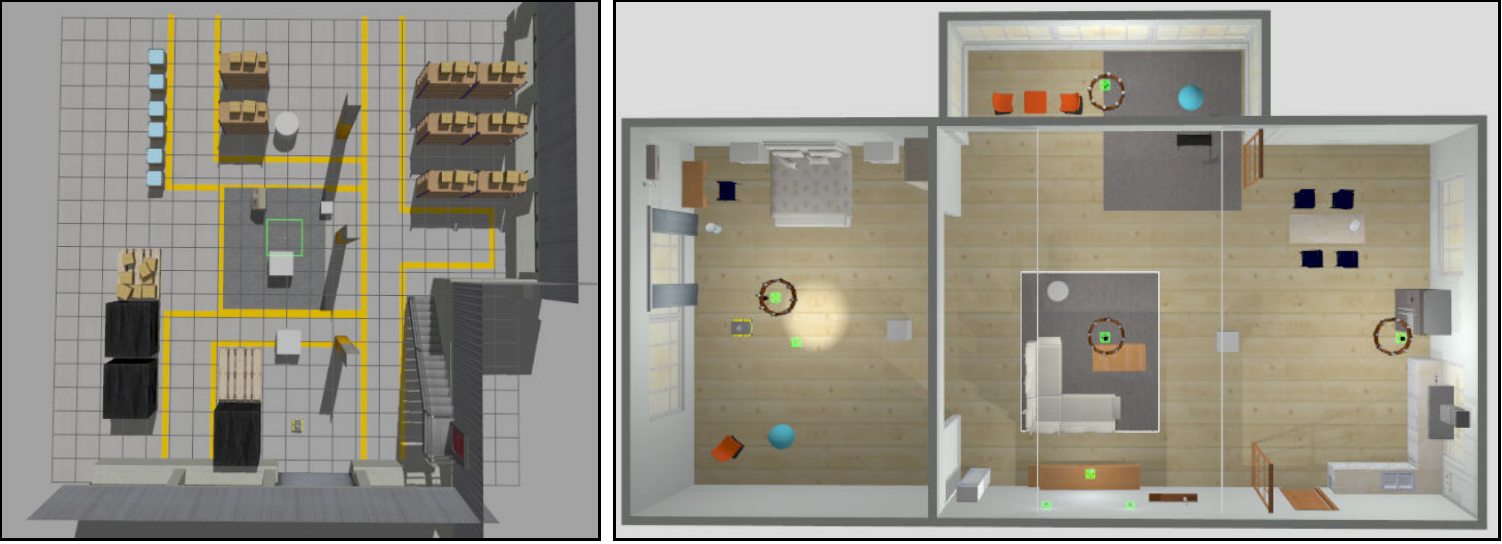}
  \caption[Gazebo simulation environments of the NTC-MPC study]{Gazebo simulation environments. Left: warehouse environment, from which all training observations are collected. Right: house environment, used exclusively for evaluation to assess generalization under a distribution shift in obstacle geometry.}
  \label{fig:ntc-sim-environments}
\end{figure}

\paragraph{Setup.}
The closed-loop evaluation used 3D Gazebo simulations with both robot models and the two environments shown in \cref{fig:ntc-sim-environments}. The models were trained exclusively on warehouse data, and the house environment provided different obstacle geometry and was used only for evaluation. In each experiment, the robot navigated from an initial pose to a goal pose using only local observations, which corresponds to the case in which a global planner supplies two consecutive waypoints. For each robot model and environment, 100 simulations with randomly placed obstacles of different shapes and sizes were run for all planners and horizon lengths \(N\in\{5,10,20,30\}\). A run was successful if the robot reached the goal within a fixed time limit without colliding with an obstacle. The two reported metrics were success rate and NLP computation time.

\begin{figure}[tb]
  \centering
  \includegraphics[width=0.9\textwidth]{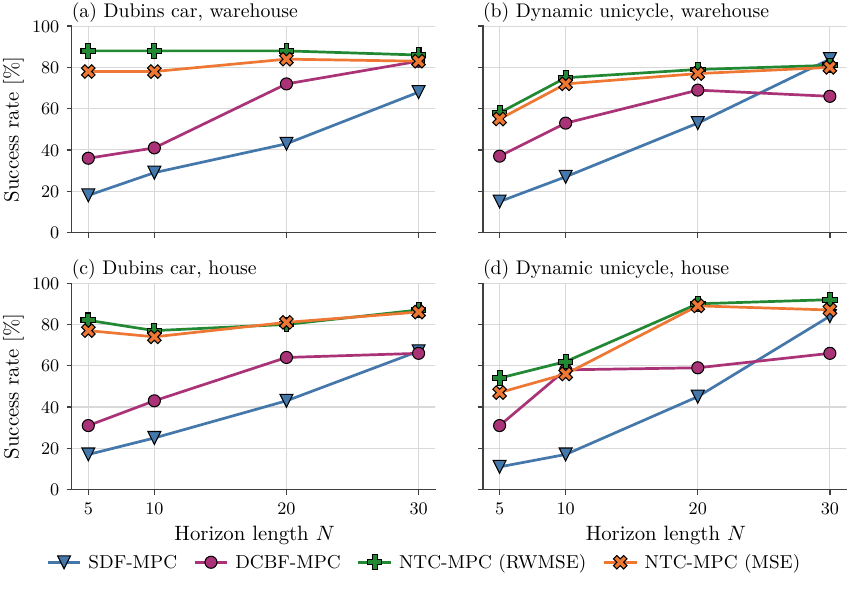}
  \caption[Success rates of the NTC-MPC simulation study]{Success rates in the Monte Carlo simulation study for the two robot models (columns) and the two environments (rows), with 100~runs per configuration. The models are trained on warehouse data only and evaluated in both environments. The performance of the distance-based baselines degrades considerably as the prediction horizon becomes shorter, while the NTC-MPC method maintains higher success rates across all horizon lengths. The RWMSE-trained model also consistently matches or outperforms its MSE-trained counterpart.}
  \label{fig:ntc-success-rates}
\end{figure}

\paragraph{Success rates.}
\Cref{fig:ntc-success-rates} shows the success rates, from which three findings follow. First, the performance of the baselines deteriorated considerably as the horizon became shorter, and SDF-MPC fell below \(20\%\) success at \(N=5\). NTC-MPC, in contrast, maintained higher success rates across the evaluated horizons and exceeded the best baseline by up to 52~percentage points at the shortest horizon. This trend is consistent with the structural argument of \cref{subsec:ntc-terminal-set-design}, since the terminal constraint carries an approximation of infinite-horizon reachability information that a short geometric horizon lacks. Second, the advantage persisted in the unseen house environment with a modest reduction in success rates, which provides evidence of generalization beyond the warehouse training environment. Third, the RWMSE-trained model consistently matched or outperformed its MSE-trained counterpart, which links the improved boundary recovery in \cref{subsec:ntc-training-eval} to better closed-loop outcomes. The experiments did not isolate the causes of the remaining NTC-MPC failures, although approximation errors and local-observability limits are two plausible mechanisms discussed in \cref{sec:ntc-discussion}.

\begin{figure}[tb]
  \centering
  \includegraphics[width=\textwidth]{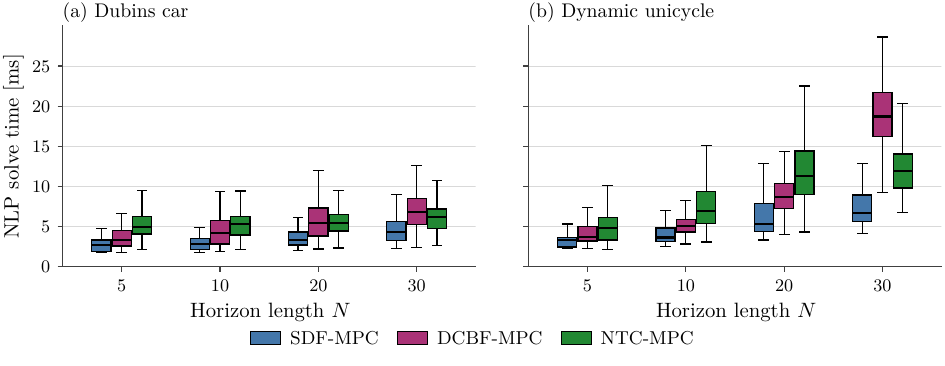}
  \caption[NLP solve times of the three MPC planners]{NLP solve times of the three planners for (a)~the Dubins car and (b)~the dynamic unicycle as functions of the horizon length, pooled over both environments. Each box spans the interquartile range of the per-run mean solve times of the successful runs, the line marks the median, and the whiskers extend to 1.5 times the interquartile range. The neural terminal constraint adds an approximately horizon-independent offset over SDF-MPC, while the stage-wise DCBF-MPC constraints scale faster with the horizon. All values remain considerably below the \SI{100}{\milli\second} planning period.}
  \label{fig:ntc-computation-times}
\end{figure}

\paragraph{Computation times.}
\Cref{fig:ntc-computation-times} reports the NLP computation times. Depending on the robot model and the horizon, the mean solve time of NTC-MPC ranged from \SI{5}{\milli\second} to \SI{13}{\milli\second} and remained considerably below the \SI{100}{\milli\second} planning period. Relative to SDF-MPC, the neural terminal constraint added an approximately constant offset as the horizon grew, which is consistent with the addition of a single learned constraint. DCBF-MPC, in contrast, imposes a distance-rate constraint at every predicted step, so its computation time grew faster with the horizon and exceeded that of NTC-MPC at the longest tested horizon, \(N=30\).

\subsection{Hardware Experiments}
\label{subsec:ntc-hardware}

\begin{figure}[tb]
  \centering
  \includegraphics[width=\textwidth]{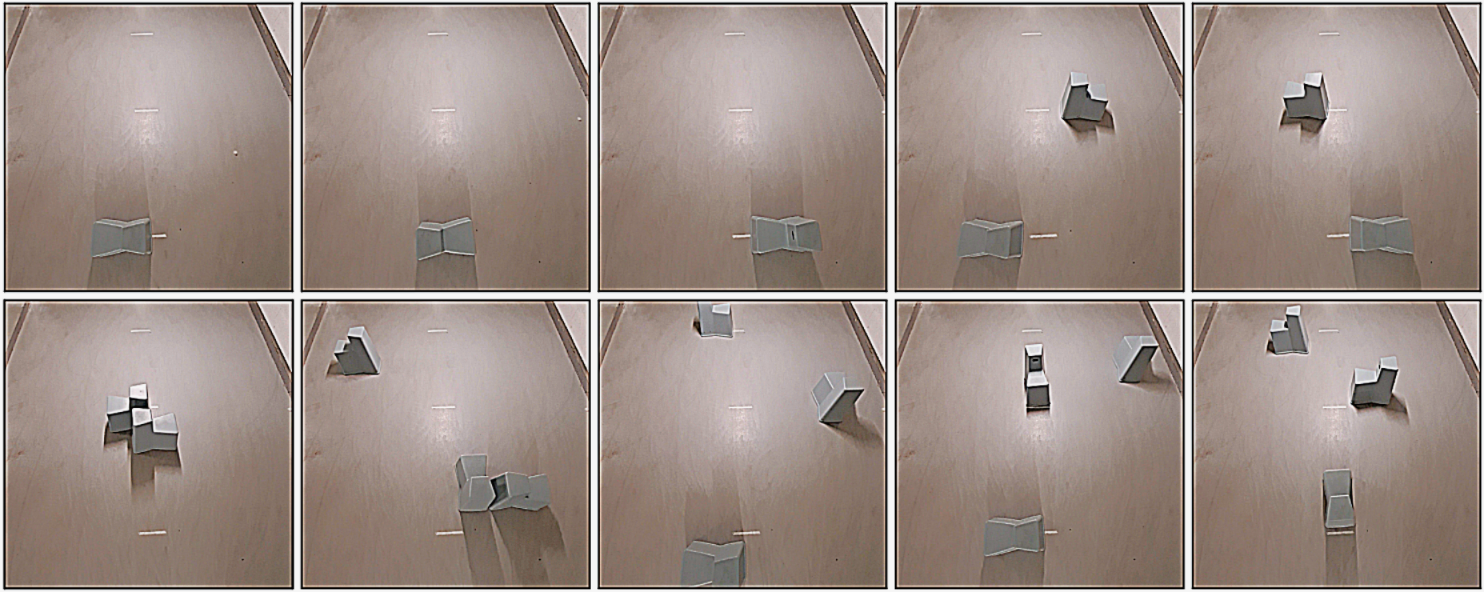}
  \caption[Obstacle configurations of the NTC-MPC hardware experiments]{The ten static obstacle configurations used in the hardware experiments, viewed from the external camera. The robot starts below the lower edge and must reach a waypoint near the upper edge of the scenario image.}
  \label{fig:ntc-hardware-scenarios}
\end{figure}

\paragraph{Setup.}
To assess transfer from simulation to a physical platform, the planners were deployed on a \emph{Continental} delivery robot, a differential-drive platform operated at constant forward speed and modeled as the Dubins car of \cref{eq:ntc-dubins-dynamics}. The robot used a 3D~LiDAR for localization and local-costmap construction, and all computation ran onboard on an Intel NUC11PHi7 mini~PC with an Intel~i7-1165G7 CPU, \SI{32}{\giga\byte} of RAM, and an RTX~2060 GPU. This platform is also used in the hardware experiments of the later chapters. The task mirrored the simulation setting, in which the robot must reach a goal waypoint behind an unknown arrangement of static obstacles using only local observations. Ten obstacle configurations were tested (\cref{fig:ntc-hardware-scenarios}), and each planner attempted every configuration for horizon lengths \(N\in\{5,10,20,30\}\). The hypernetwork trained on simulated warehouse data was transferred to the robot without fine-tuning.

\begin{table}[tb]
  \caption[NTC-MPC hardware experiment results]{Hardware experiment results with the Dubins car model over the ten obstacle configurations of \cref{fig:ntc-hardware-scenarios}.}
  \label{tab:ntc-hardware-results}
  \centering
  \begin{tabular}{@{}l*{8}{c}@{}}
    \toprule
                     & \multicolumn{4}{c}{Success rate [\%]} & \multicolumn{4}{c}{Mean computation time [\si{\milli\second}]}                                                                   \\
    \cmidrule(lr){2-5}\cmidrule(lr){6-9}
    \(N\) [steps]    & 5                                     & 10                                                              & 20          & 30          & 5    & 10   & 20   & 30   \\
    \midrule
    SDF-MPC          & 0                                     & 0                                                               & 30          & 70          & 2.65 & 4.55 & 6.26 & 6.54 \\
    DCBF-MPC         & 0                                     & 10                                                              & 70          & 80          & 4.06 & 5.61 & 5.91 & 7.76 \\
    NTC-MPC (ours)   & \textbf{80}                           & \textbf{90}                                                     & \textbf{90} & \textbf{90} & 4.93 & 5.26 & 6.54 & 6.31 \\
    \bottomrule
  \end{tabular}
\end{table}

\begin{figure}[tb]
  \centering
  \includegraphics[width=\textwidth]{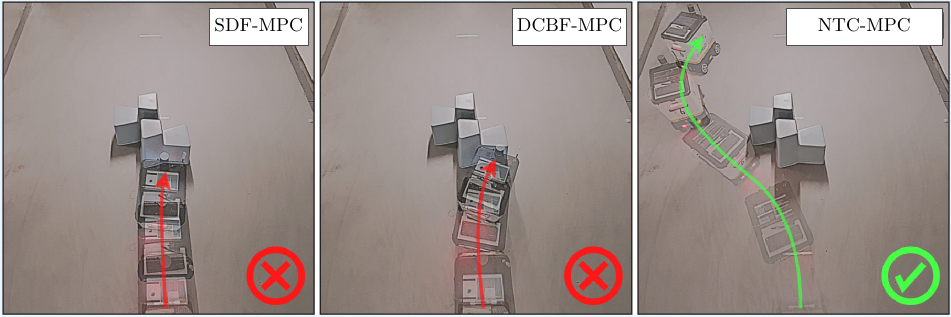}\\[1.5mm]
  \includegraphics[width=\textwidth]{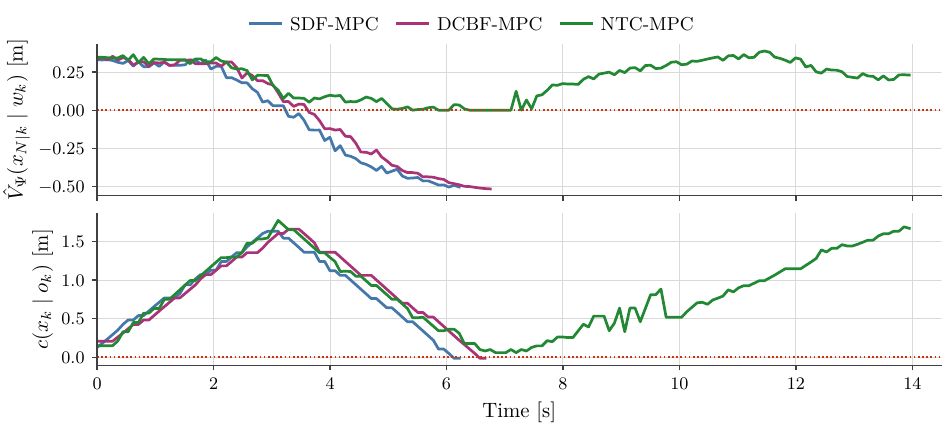}
  \caption[Representative NTC-MPC hardware experiment]{Representative hardware experiment with the Dubins car model and horizon \(N=10\) (\SI{1}{\second}). The upper row shows the overlaid robot motion for the three planners, where both baselines collide with the obstacle (red) while NTC-MPC avoids it (green). The plots show the time evolution of the predicted terminal value \(\hat{V}_{\Psi}(x_{N\mid k}\mid w_k)\) and of the observed obstacle distance \(c(x_k\mid o_k)\), with the zero level marked by the dotted line. In this configuration, the obstacle distance of the baselines remains positive until shortly before the collision, while NTC-MPC begins to avoid the obstacle when its terminal value approaches zero, several seconds earlier.}
  \label{fig:ntc-hardware-exp}
\end{figure}

\paragraph{Results.}
\Cref{tab:ntc-hardware-results} summarizes the outcomes, which reproduce the pattern observed in simulation. With short horizons, the baselines failed in almost all configurations, while NTC-MPC succeeded in \(80\%\) of the configurations at \(N=5\) and in \(90\%\) from \(N=10\) onward, at computation times comparable to those of the baselines. \Cref{fig:ntc-hardware-exp} examines one representative configuration for \(N=10\), corresponding to a \SI{1}{\second} look-ahead. In this configuration, both baselines drove toward the obstacle, and their observed obstacle distance \(c(x_k\mid o_k)\) remained positive until collision was imminent. For NTC-MPC, the predicted terminal value \(\hat{V}_{\Psi}(x_{N\mid k}\mid w_k)\) approached zero earlier, and the planner steered around the obstacle before the obstacle distance at the current state alone indicated imminent contact, which is the intended mechanism of the terminal constraint.

\section{Discussion}
\label{sec:ntc-discussion}

The experiments show that observation-conditioned neural terminal sets improve collision avoidance, especially at short horizons. The following limitations define the scope of this empirical result and motivate the next chapters.

\paragraph{Scope of guarantees.}
The classical recursive-feasibility argument does not apply to the implemented scheme, because the terminal set is a learned approximation whose error is not one-sided, the HJ targets and the MPC problem use different time discretizations, and the observation is updated between problems. A state inside the numerical BRT may thus be predicted safe, so recursive feasibility and collision avoidance are demonstrated empirically under the assumptions of \cref{sec:ntc-problem-statement} rather than guaranteed \cite{lin_generating_2023-1,dawson_safe_2023}, and RWMSE reduces boundary misclassification without being a certificate. This is the central trade-off of the method, because numerical reachability targets are available only offline, while the representation evaluated in real time is approximate. The next chapter excludes one class of this approximation error by construction, and \cref{ch:conclusion} discusses how the remaining error can be bounded.

\paragraph{Structural limitation of direct regression.}
Direct regression imposes no relation between \(\hat{V}_{\Psi}\) and the observed SDF, so a sufficiently large error can classify even a state inside an observed obstacle as safe. \Cref{ch:rntc-mpc} addresses this limitation by subtracting a nonnegative learned residual from the SDF, enforcing failure-set exclusion by construction, and extends the terminal constraint to predicted moving geometry.

\paragraph{Local observability.}
Some failure modes are inherent to local planning rather than to the safety representation. An occluded obstacle enters the observation only when it is revealed, possibly at close range. If the sensing-adequacy assumption is violated at that moment, infeasibility can result regardless of how the terminal set was obtained. Similarly, the local decision to pass an obstacle on the left or on the right may appear symmetric even though one of the two options leads into a dead end beyond the sensing window. Such situations require cooperation with a global planner or explicit reasoning about unobserved space, both of which lie outside the scope of a local terminal constraint.

\paragraph{Scalability of supervision.}
The offline stage inherits the dimensionality limit of grid-based HJ solvers (\cref{subsec:prelim-hj-numerical}), and the five-dimensional unicycle already requires substantial computation per observation. Since supervision is generated offline, other sources of targets could take the place of the grid solver, including system decomposition \cite{chen_decomposition_2018} and neural HJ solvers \cite{bansal_deepreach_2021}. The grid-free supervision of \cref{ch:vertex-cbf} is an option only for robot models with control-affine dynamics and polytopic input constraints, which NTC-MPC does not assume in general. The accuracy of these targets limits the accuracy that can be expected from the learned model. Beyond mobile robots, a hypernetwork that converts a high-dimensional context into a compact constraint for an optimizer is not specific to navigation, and it may serve other applications that require complex terminal sets in real time.

\section{Summary}
\label{sec:ntc-summary}

NTC-MPC uses a hypernetwork to produce a compact terminal value function from each local SDF, with RWMSE training focused near the safety boundary. In simulation, success rates exceeded the best baseline by up to 52~percentage points at mean solve times of \SIrange{5}{13}{\milli\second}. The benefit persisted in an unseen environment and on hardware without fine-tuning, and the hypernetwork reduced the MPC solve time by more than two orders of magnitude compared with the evaluated single-network architecture. These results address question~\ref{rq:realtime} for static-environment MPC. \Cref{ch:rntc-mpc} adds structural failure-set exclusion and extends the approach to moving obstacles.

\chapter{Neural Terminal Constraints in Dynamic Environments}
\label{ch:rntc-mpc}

This chapter presents \emph{residual neural terminal constraint MPC} (RNTC-MPC), based on the author's paper \cite{derajic_residual_2025} at the Conference on Robot Learning (CoRL) 2025. The method extends NTC-MPC to predicted obstacle motion and represents the terminal value as the predicted SDF minus a nonnegative learned residual, which excludes the predicted terminal failure set regardless of approximation error. The method addresses moving obstacles (question~\ref{rq:dynamic}), structural safety properties (question~\ref{rq:structure}), and real-time MPC (question~\ref{rq:realtime}).

\section{Motivation}
\label{sec:rntc-motivation}

\begin{figure}[tbp]
  \centering
  \resizebox{\textwidth}{!}{\input{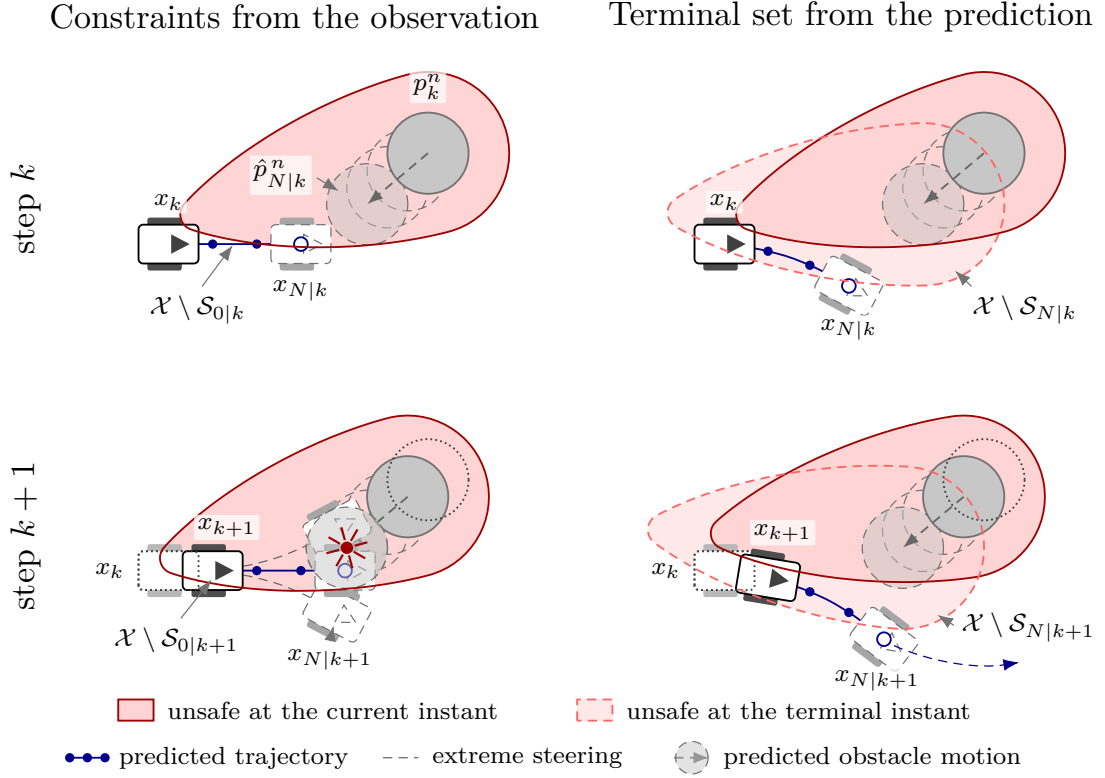}}
  \caption[Effect of the terminal constraint on an MPC local planner among moving obstacles]{Effect of the terminal constraint on an MPC local planner among moving obstacles, shown at two consecutive planning steps. The robot tracks the reference path to the right at constant speed and with a limited turning rate, while an obstacle crosses that path from the front left. Each red region collects the states from which a collision can no longer be avoided under the actuation limits. It extends along the obstacle velocity \emph{relative to the robot}, so it is not aligned with the drawn velocity arrow, and the region at the terminal instant is the region at the current instant displaced and rotated according to the obstacle motion over the horizon. In the lower row the dotted outlines mark the positions at step~\(k\). Left: constraints formed from the current observation refer only to the unsafe region at the current instant, \(\mathcal{X}\setminus\mathcal{S}_{0\mid k}\), so the planner continues straight ahead at step~\(k\). At step~\(k+1\) its state lies inside that region, where continued straight motion and both extreme steering maneuvers end in the predicted obstacle. Right: the terminal constraint \(x_{N\mid k}\in\mathcal{S}_{N\mid k}\) of RNTC-MPC refers to the predicted geometry and is already active at step~\(k\), so the planner turns away and its plan remains feasible one step later.}
  \label{fig:rntc-intro-illustration}
\end{figure}

A robot cannot always rely on other agents to help avoid collisions. Fast riders in public spaces or human-operated forklifts in warehouses may not yield within the available time, leaving a nonholonomic robot with limited actuation to avoid them \cite{fraichard_short_2007,hwang_safe_2024}. This chapter treats such agents as \emph{non-interactive moving obstacles}, whose motion is predicted but not influenced.

This setting changes the terminal-constraint problem of \cref{ch:ntc-mpc} in one essential respect. The safe set is no longer a fixed region attached to the observed geometry, but a region that travels and deforms with the obstacles. The backward reachable tube of each obstacle moves with that obstacle and extends along the obstacle velocity \emph{relative to the robot}, so it contains states that are collision-free at present but from which the robot can no longer escape before the obstacle arrives. The set imposed at the end of the horizon consequently refers to geometry that has not yet been observed, and it must contain the states from which the \emph{predicted} failure set can still be avoided once the horizon ends. Constructing this set requires a motion prediction for the obstacles, a reachability analysis over the predicted evolution, and a repetition of both steps whenever new observations arrive. \Cref{fig:rntc-intro-illustration} illustrates the effect at two consecutive planning steps. A planner whose constraints are formed from the current observation continues straight ahead while the crossing obstacle is still clear of the reference path, and one step later its state lies inside the unsafe region. A terminal constraint that refers to the predicted geometry becomes active one step earlier, so the planner turns away in time and retains a feasible continuation.

A second difficulty, already identified in \cref{sec:ntc-discussion}, concerns the learned representation itself. The terminal function of \cref{ch:ntc-mpc} is unconstrained, and nothing relates its predicted safe set to the observed geometry, so a sufficiently large approximation error can lead the model to classify a state inside a represented obstacle as safe. Moving obstacles aggravate this concern, because the model must now generalize over sequences of predicted geometries rather than over single snapshots, and its errors combine with those of the motion prediction. The residual parameterization of \cref{subsec:prelim-structure-preserving-residual} removes this particular error mode. RNTC-MPC computes the SDF of the predicted geometry numerically, which the stage constraints already require, and learns only the nonnegative residual by which the value function lies below it. The learned terminal constraint then admits no state that the predicted terminal SDF rejects, regardless of the residual approximation error.

RNTC-MPC retains the hypernetwork and MPC interface of \cref{ch:ntc-mpc}. The following sections develop its predicted SDF input, residual terminal constraint, and sign-sensitive training loss, then evaluate approximation quality and closed-loop performance.

\section{Related Work}
\label{sec:rntc-related-work}

This section reviews the work most closely related to MPC-based collision avoidance among moving obstacles. The static-environment counterpart was reviewed in \cref{sec:ntc-related-work}, and the broader context is surveyed in \cref{ch:introduction}.

Interactive planners assume that surrounding agents contribute to avoidance. Optimal reciprocal collision avoidance divides responsibility between each pair \cite{van_den_berg_reciprocal_2011}. MPC extensions derive decentralized velocity constraints \cite{cheng_decentralized_2017}, model human responses in bilevel optimization \cite{samavi_sicnav_2025}, or refine diffusion-based predictions \cite{samavi_sicnav-diffusion_2025}. Learned environment models also supply interaction-aware predictions \cite{zhu_learning_2021}. These methods can avoid the freezing that arises when every agent is treated as unyielding in a dense crowd \cite{trautman_unfreezing_2010}. Their interaction assumptions, however, do not cover the non-interactive encounters considered here.

MPC planners for non-interactive moving obstacles enforce a minimum distance along the predicted obstacle motion through distance functions \cite{oleynikova_signed_2016-1,zhang_optimization-based_2021,brito_model_2019}, velocity obstacles \cite{piccinelli_mpc_2023}, or discrete-time CBF conditions \cite{zeng_safety-critical_2021}. None of these formulations introduces a terminal set. They rely on the finite prediction horizon instead and lengthen it when the closed-loop behavior is unsatisfactory, which increases the computational cost and still leaves the time beyond the final stage unconstrained. This chapter follows the complementary approach and keeps the horizon short, while the reasoning about the time beyond the horizon is concentrated in a terminal set that is constructed in real time. The three constraint families above are the natural baselines, and \cref{sec:rntc-experiments} compares RNTC-MPC with a representative of each.

Obtaining such a terminal set by reachability analysis is more demanding in the dynamic setting than in the static one. With static geometry, one backward propagation converges to a stationary safe set that remains valid for as long as the observation does. A moving failure set instead yields a time-dependent value function, which must be propagated backward from the terminal instant over the entire predicted evolution and recomputed whenever the prediction itself changes. The scalable formulations discussed in \cref{sec:ntc-related-work} reduce parts of this cost but still require the environment evolution to be specified before the computation. Parameter-conditioned reachability solves the problem once for a family of parameter values and can treat obstacle velocities as such parameters \cite{borquez_parameter-conditioned_2023,jeong_parameterized_2024,nakamura_online_2023}. However, the augmented problem grows with each additional parameter, while a scene with several moving obstacles has many degrees of freedom and a varying number of obstacles. The observation-conditioned approach of \cref{ch:ntc-mpc} \cite{derajic_learning_2025} handles high-dimensional observations, but only for static geometry and without any architectural constraint on the approximation error. RNTC-MPC extends that approach to time-varying geometry and additionally enforces predicted-failure-set exclusion by construction, so that one property of the learned constraint follows from the architecture and not from the training outcome, which is the separation that the learned-certificate literature emphasizes \cite{dawson_safe_2023,wabersich_data-driven_2023}. A complementary approach, taken in \cref{ch:cn-cbf}, formulates the reachability problem in relative coordinates, so that each pairwise collision set becomes stationary.

\section{Problem Statement}
\label{sec:rntc-problem-statement}

This section formalizes local navigation among moving obstacles within the MPC formulation of \cref{sec:prelim-mpc}. The goal is to design a terminal constraint that represents the time-varying safe set at the end of the prediction horizon.

\paragraph{Setting.}
The robot and MPC models are those of \cref{sec:ntc-problem-statement}. HJ analysis uses the continuous-time model \(\dot{x}=f(x,u)\) with compact \(\mathcal{U}\), and MPC predicts \(x_{i+1\mid k}=F_\Delta(x_{i\mid k},u_{i\mid k})\) with prediction interval \(\Delta t>0\). Let \(t_k\) denote the time of planning update \(k\), where the time between two consecutive planning updates may differ from \(\Delta t\). At update \(k\), observation \(o_k\) supplies estimated positions \(p^n_k\in\mathbb{R}^2\) and velocities \(v^n_k\in\mathbb{R}^2\) for obstacles \(n\) within the sensing window. Under the constant-velocity model, their predicted positions are
\begin{equation}
  \hat{p}^{\,n}_{i\mid k}
  =
  p^n_k+i\,\Delta t\,v^n_k,
  \qquad
  i=0,\ldots,N.
  \label{eq:rntc-constant-velocity-prediction}
\end{equation}
Let \(\mathcal{A}_{i\mid k}\subset\mathbb{R}^2\) denote the region predicted to be occupied at stage \(i\), that is, the union of the obstacle footprints placed at the predicted positions \cref{eq:rntc-constant-velocity-prediction} and inflated by the robot footprint and a safety margin, exactly as in \cref{eq:ntc-observed-constraint}. With \(p(x)\) denoting the workspace position of the robot, the \emph{predicted constraint function} at stage \(i\) is the SDF-based function
\begin{equation}
  c_{i\mid k}(x)
  :=
  \operatorname{sd}_{\mathcal{A}_{i\mid k}}\bigl(p(x)\bigr),
  \qquad
  i=0,\ldots,N,
  \label{eq:rntc-predicted-constraint}
\end{equation}
so that \(c_{i\mid k}(x)\geq0\) indicates a nonnegative distance to the predicted obstacles at that stage and \(c_{i\mid k}(x)<0\) defines its failure set. Unlike the single static constraint in \cref{ch:ntc-mpc}, this sequence changes with the predicted geometry. The robot still tracks a reference supplied by a global planner using local information.

\paragraph{The time-varying terminal safe set.}
Whether the predicted obstacles can still be avoided after the horizon is represented through the time-varying HJ formulation of \cref{subsec:prelim-time-varying-hj}. Consider the avoidance problem for the undisturbed continuous-time dynamics \cref{eq:prelim-undisturbed-continuous-dynamics}, with the constraint evolution over the MPC horizon given by \cref{eq:rntc-predicted-constraint} and continued over the HJ reachability horizon by the same motion model. Because the geometry continues to evolve, the value function retains an explicit time argument. The predicted geometry available at update \(k\) is collected in a conditioning input \(w_k\), which is defined precisely in \cref{subsec:rntc-time-varying}. Let \(V_{N\mid k}(x):=V\bigl(x,t_k+N\Delta t\bigr)\) denote the value function at the terminal instant under this modeled evolution. Its nonnegative superlevel set,
\begin{equation}
  \mathcal{S}_{N\mid k}
  :=
  \left\{x\in\mathcal{X}\mid V_{N\mid k}(x)\geq0\right\},
  \label{eq:rntc-terminal-safe-set}
\end{equation}
contains the states from which the predicted failure set can be avoided throughout the HJ horizon under the modeled dynamics and actuation limits. In the numerical construction used for supervision, that horizon is extended until the value at the reference instant has converged to the chosen tolerance (\cref{subsec:rntc-training}). The set \cref{eq:rntc-terminal-safe-set} plays the role that \(\mathcal{S}^{\star}(o_k)\) played in \cref{ch:ntc-mpc}, but it refers to predicted future geometry and not the current observation.

\paragraph{Design problem.}
The goal is to construct, at every update \(k\), a representation \(\hat{V}_{\Psi}(\,\cdot\mid w_k)\) of the terminal value function \(V_{N\mid k}\) with the following properties. As in \cref{ch:ntc-mpc}, it must be available within a small fraction of the control period, support efficient value and gradient evaluations inside a numerical optimizer, and generalize to obstacle configurations that were not seen offline. In addition, it must satisfy the \emph{predicted-failure-set exclusion} property
\begin{equation}
  \left\{x\mid \hat{V}_{\Psi}(x\mid w_k)\geq0\right\}
  \subseteq
  \left\{x\mid c_{N\mid k}(x)\geq0\right\}
  \label{eq:rntc-failure-set-exclusion-requirement}
\end{equation}
by construction, which addresses the structural weakness of direct regression.

\paragraph{Assumptions.}
The formulation rests on three assumptions. First, the perception module provides sufficiently accurate position and velocity estimates for every represented obstacle within the sensing window, while perception itself lies outside the scope of this dissertation. Second, the obstacles are non-interactive, and the constant-velocity model \cref{eq:rntc-constant-velocity-prediction} adequately predicts their motion both within the MPC horizon and over the HJ horizon used to define the target. This modeling assumption is revisited in \cref{sec:rntc-discussion}. Third, the sensing window is sufficiently large relative to the robot and obstacle speeds and to the update period, so that the cross-update condition \cref{eq:prelim-cross-update-safety} can be satisfied. Newly detected obstacles then enter the observation near its boundary rather than immediately next to the robot.

\section{Residual Neural Terminal Constraints}
\label{sec:rntc-method}

The method combines predicted-geometry conditioning, a residual hypernetwork, and time-varying HJ supervision with a sign-sensitive loss. These components are then assembled into the MPC problem.

\subsection{Predicted Terminal Safe Sets}
\label{subsec:rntc-time-varying}

The first design question is what information should condition the terminal-set representation. In \cref{ch:ntc-mpc}, the SDF of the current observation sufficed because the failure set was static. Here, \(V_{N\mid k}\) depends on the failure-set evolution from the terminal instant through the specified HJ horizon, which cannot be observed and must be supplied by a motion model. Under the constant-velocity model \cref{eq:rntc-constant-velocity-prediction}, the obstacle positions and velocities at the terminal instant determine the modeled future. For well-separated obstacles, two distinct geometry snapshots near that instant encode these quantities. The later snapshot gives the positions, and the displacement between the two snapshots, divided by their known temporal separation, gives the velocities. The conditioning input is formed from selected snapshots at the end of the predicted SDF sequence,
\begin{equation}
  w_k
  :=
  \left(
  c_{N-q_1\mid k},\;\ldots,\;c_{N-q_L\mid k}
  \right),
  \qquad
  q_1>\cdots>q_L=0,
  \label{eq:rntc-condition}
\end{equation}
where the SDFs are sampled on a regular grid over the sensing window, \(L\) is the number of snapshots, and the fixed nonnegative integers \(q_1,\ldots,q_L\) specify their lags in prediction steps. Two snapshots with a nonzero temporal separation suffice when obstacle correspondence is unambiguous, and \cref{subsec:rntc-training-eval} states the snapshot count and the lags used in the experiments. Motion models that include acceleration or higher-order effects may require more snapshots or additional predicted quantities. Definition \cref{eq:rntc-condition} is the concrete form of the conditioning input \cref{eq:prelim-conditioning-input} for this chapter, with the map \(\eta\) consisting of the motion prediction \cref{eq:rntc-constant-velocity-prediction} followed by the distance transform of the predicted geometry. The model receives predicted geometry rather than the current observation.

A second property of the mapping \(w_k\mapsto V_{N\mid k}\) is essential for training. Under the time-invariant robot dynamics and the constant-velocity obstacle model used here, this mapping is invariant to a common shift in absolute time. The value function at an instant \(\tau\) depends on the geometry at fixed times relative to \(\tau\), and not on the numerical value of \(\tau\) itself. The same mapping can consequently be trained on snapshot sequences ending at an arbitrary reference instant and evaluated on predicted sequences ending at \(t_k+N\Delta t\). One trained model thus applies to any MPC horizon with \(N\geq q_1\), provided that the same temporal separations and motion model are used. The absolute terminal time is not an input to the model, and the obstacle prediction enters only through the SDF sequence.

\subsection{Structure-Preserving Residual Parameterization}
\label{subsec:rntc-residual}

The second design decision is the structure of the learned representation. Direct regression, as in \cref{ch:ntc-mpc}, would approximate \(V_{N\mid k}\) by an unconstrained neural field, so its errors would also be unconstrained. RNTC instead uses the residual parameterization \cref{eq:prelim-structure-preserving-residual}, subtracting a nonnegative learned residual from the predicted terminal constraint function,
\begin{equation}
  \hat{V}_{\Psi}(x\mid w_k)
  :=
  c_{N\mid k}(x)
  -
  \hat{r}_{\Psi}(x\mid w_k),
  \qquad
  \hat{r}_{\Psi}(x\mid w_k)\geq0,
  \label{eq:rntc-residual-parameterization}
\end{equation}
where \(c_{N\mid k}\) is computed numerically from the predicted geometry, exactly as the stage constraints already require, and only the residual \(\hat{r}_{\Psi}\) is learned. This choice does not restrict the target, because by the HJ ordering \cref{eq:prelim-hj-value-constraint-ordering}, extended to time-varying constraints in \cref{subsec:prelim-time-varying-hj}, the true value function equals the constraint function minus a nonnegative residual. The nonnegativity of the residual immediately yields \(\hat{V}_{\Psi}(x\mid w_k)\leq c_{N\mid k}(x)\) for all \(x\) and hence establishes the failure-set exclusion property required in \cref{eq:rntc-failure-set-exclusion-requirement},
\begin{equation}
  \left\{x\mid \hat{V}_{\Psi}(x\mid w_k)\geq0\right\}
  \subseteq
  \left\{x\mid c_{N\mid k}(x)\geq0\right\}.
  \label{eq:rntc-failure-set-exclusion}
\end{equation}
Every state admitted by the learned terminal constraint therefore has a nonnegative distance to the predicted terminal geometry, regardless of the residual approximation error, so the learned terminal constraint is never weaker than the predicted terminal SDF constraint. The property has two limitations. It is one-sided, as noted in \cref{subsec:prelim-structure-preserving-residual}, so an underestimated residual may still classify a state inside the predicted BRT as safe. It is also stated relative to the predicted geometry, so it inherits the perception and motion-prediction errors. The architecture eliminates one important error class, while the remaining errors are reduced only empirically through training.

The parameterization also simplifies the learning problem, because the terminal constraint geometry is supplied directly by \(c_{N\mid k}\) and the network represents only the dynamics-induced correction, analogous to the braking-distance residual in \cref{fig:prelim-braking-values}. The ablation in \cref{subsec:rntc-training-eval} evaluates the resulting learning benefit through convergence speed and safe-set accuracy at equal model size.

Nonnegativity of the residual is enforced by the output activation. For the reason given in \cref{subsec:prelim-structure-preserving-residual}, RNTC enforces it with a smooth relaxation of the ReLU, the shifted exponential linear unit \(\operatorname{ELU}(z)+1\), where
\begin{equation}
  \operatorname{ELU}(z)
  =
  \begin{cases}
    z, & z>0,\\
    \exp(z)-1, & z\leq0.
  \end{cases}
  \label{eq:rntc-elu}
\end{equation}
This output is nonnegative and has a nonzero derivative for every finite input. The residual is strictly positive for finite inputs but can approach zero arbitrarily closely, so where the true residual vanishes, the parameterization approximates it with an arbitrarily small conservative offset.

\subsection{Hypernetwork Architecture}
\label{subsec:rntc-architecture}

RNTC-MPC adopts the hypernetwork factorization of \cref{subsec:ntc-architecture}. The hypernetwork \(H_{\Psi}\) maps \(w_k\) to the parameters of a compact main network \(M_{\Theta(w_k)}\), whose output passes through the shifted ELU:
\begin{equation}
  \Theta(w_k)
  =
  H_{\Psi}(w_k),
  \qquad
  \hat{r}_{\Psi}(x\mid w_k)
  =
  \operatorname{ELU}\bigl(M_{\Theta(w_k)}(x)\bigr)+1.
  \label{eq:rntc-hypernetwork}
\end{equation}
Because the conditioning input \cref{eq:rntc-condition} is a stack of \(L\) SDF images over the sensing window, the hypernetwork is a convolutional neural network with \(L\) input channels followed by a fully connected head. The main network is a small multilayer perceptron with sinusoidal activations in its early layers and SELU activations thereafter, as in \cref{ch:ntc-mpc}.

\begin{figure}[tb]
  \centering
  \colorlet{hypfill}{green!55!black!15}%
\colorlet{hypline}{green!50!black!65}%
\colorlet{sdffill}{orange!16}%
\colorlet{sdfline}{orange!70!black}%
\colorlet{mpcfill}{teal!14}%
\colorlet{mpcline}{teal!55!black}%
\colorlet{mlpfill}{blue!55!black!14}%
\colorlet{robotfill}{yellow!65}%
\colorlet{wheelfill}{black!68}%
\colorlet{headfill}{blue!55!black!14}%

\begin{tikzpicture}[
    font=\small,
    blk/.style={draw, semithick, rounded corners=3pt, align=center,
                inner sep=1mm},
    sig/.style={font=\footnotesize, align=center, inner sep=1pt},
    arrow/.style={-{Latex[length=2.2mm]}, semithick},
    wire/.style={semithick},
    unit/.style={circle, draw=black, fill=mlpfill, semithick,
                 inner sep=0pt, minimum size=2.7mm},
  ]

  \fill[mpcfill, rounded corners=8pt] (0.30,0.75) rectangle (8.60,6.10);
  \draw[mpcline, semithick, rounded corners=8pt] (0.30,0.75) rectangle (8.60,6.10);
  \node[anchor=north east] at (8.42,5.98) {MPC};

  \node[blk, fill=white, rounded corners=6pt, minimum width=3.30cm,
        minimum height=2.50cm] (hyper) at (7.30,8.20) {};
  \node[above=0.6mm of hyper] {Hypernetwork};

  \filldraw[fill=hypfill, draw=hypline, semithick]
    (8.60,9.25) -- (7.05,8.60) -- (7.05,7.80) -- (8.60,7.15) -- cycle;
  \node at (7.95,8.20) {CNN};
  \filldraw[fill=hypfill, draw=hypline, semithick]
    (6.11,7.20) rectangle (6.66,9.20);
  \node[font=\footnotesize] at (6.385,8.20) {FC};
  \draw[arrow] (7.05,8.20) -- (6.70,8.20);

  \node[blk, fill=sdffill, draw=sdfline, minimum width=2.60cm,
        minimum height=1.20cm] (sdf) at (12.20,8.20) {SDF\\computation};
  \node[blk, fill=white, minimum height=1.30cm]
    (track) at (12.20,6.10) {Object Tracking and\\Motion Prediction};

  \draw[arrow] (12.20,6.75) -- (sdf.south);
  \node[sig, anchor=east] at (13.15,7.12) {$\hat{p}^{\,n}_{i\mid k}$};

  \node[blk, fill=white, minimum width=2.70cm, minimum height=1.50cm]
    (main) at (3.35,4.60) {};
  \node[below=0.6mm of main] {Main network};

  \foreach \i/\y in {a/4.88,b/4.32} {\node[unit] (in\i) at (2.72,\y) {};}
  \foreach \i/\y in {a/5.06,b/4.60,c/4.14} {\node[unit] (hid\i) at (3.35,\y) {};}
  \node[unit] (out) at (3.98,4.60) {};
  \foreach \i in {a,b}{\foreach \j in {a,b,c}{
    \draw[black, line width=0.35pt] (in\i) -- (hid\j);}}
  \foreach \j in {a,b,c}{
    \draw[black, line width=0.35pt] (hid\j) -- (out);}

  \node[blk, fill=white, minimum width=1.80cm, minimum height=0.80cm]
    (elu) at (6.05,4.60) {$\operatorname{ELU}+1$};
  \draw[arrow] (main.east) -- (elu.west);

  \node[blk, fill=white, minimum width=2.95cm, minimum height=0.85cm]
    (cterm) at (3.35,2.75) {$c_{N\mid k}(x_{N\mid k})$};

  \node[blk, fill=white, minimum width=4.40cm, minimum height=0.85cm]
    (nlp) at (3.60,1.55) {NLP Solver};

  \node[circle, draw=black, semithick, fill=white, inner sep=0pt,
        minimum size=0.40cm] (sum) at (7.55,3.70) {};
  \draw[wire] (elu.east) -- (7.55,4.60);
  \draw[arrow] (7.55,4.60) -- (sum.north);
  \node[sig, anchor=south] at (7.75,4.68) {$\hat{r}_{\Psi}(x_{N\mid k})$};
  \node[sig, anchor=west] at (7.67,3.97) {$-$};
  \draw[wire] (cterm.east) -- (7.55,2.75);
  \draw[arrow] (7.55,2.75) -- (sum.south);
  \node[sig, anchor=west] at (7.67,3.43) {$+$};

  \draw[wire] (sum.east) -- (8.15,3.70) -- (8.15,1.55);
  \draw[arrow] (8.15,1.55) -- (nlp.east);
  \node[sig, anchor=south] at (7.05,1.65) {$\hat{V}_{\Psi}(x_{N\mid k})$};

  \draw[wire] (nlp.west) -- (0.70,1.55) -- (0.70,4.60);
  \fill (0.70,2.75) circle (1.7pt);
  \draw[arrow] (0.70,4.60) -- (main.west);
  \draw[arrow] (0.70,2.75) -- (cterm.west);
  \node[sig, anchor=south] at (1.42,4.68) {$x_{N\mid k}$};

  \draw[wire] (0.30,6.80) -- (2.10,6.80);
  \draw[arrow] (2.10,6.80) -- (2.10,6.10);
  \node[sig, anchor=south west] at (0.32,6.88) {Reference};

  \draw[wire] (6.11,8.20) -- (3.35,8.20) -- (3.35,5.75);
  \draw[arrow] (3.35,5.75) -- (main.north);
  \node[sig, anchor=south] at (3.8,8.28) {$\Theta(w_k)$};

  \draw[wire] (sdf.west) -- (8.90,8.20);
  \draw[arrow] (8.90,8.20) -- (8.62,8.20);
  \fill (9.90,8.20) circle (1.7pt);
  \node[sig, anchor=south] at (9.30,8.28) {$w_k$};
  \draw[wire] (9.90,8.20) -- (9.90,4.60);
  \draw[arrow] (9.90,4.60) -- (8.60,4.60);
  \node[sig, anchor=east] at (9.80,6.40) {$c_{0\mid k},\ldots,c_{N-1\mid k}$};

  \node[blk, fill=white, rounded corners=6pt, minimum width=2.70cm,
        minimum height=2.90cm] (robot) at (12.20,3.00) {};
  \foreach \yl/\yu in {3.07/3.87, 1.89/2.69} {
    \fill[wheelfill, rounded corners=1.2pt] (11.12,\yl) rectangle (11.38,\yu);
    \fill[wheelfill, rounded corners=1.2pt] (13.02,\yl) rectangle (13.28,\yu);}
  \filldraw[fill=robotfill, draw=black, semithick]
    (11.35,1.73) rectangle (13.05,4.03);
  \filldraw[fill=headfill, draw=black, semithick]
    (11.79,3.26) -- (12.61,3.26) -- (12.20,3.90) -- cycle;

  \draw[arrow] (12.20,4.45) -- (track.south);
  \node[sig, anchor=east] at (14.3,4.95) {Sensor data};

  \draw[arrow] (8.60,3.00) -- (robot.west);
  \node[sig, anchor=south] at (9.72,3.08) {Control $u^{\star}_{0\mid k}$};

  \draw[wire] (12.20,1.55) -- (12.20,0.15) -- (3.00,0.15);
  \draw[arrow] (3.00,0.15) -- (3.00,0.75);
  \node[sig, anchor=south] at (7.60,0.20) {Feedback $x_k$};
\end{tikzpicture}
  \caption[RNTC-MPC architecture]{RNTC-MPC architecture. Tracked obstacle positions and velocities are extrapolated over the MPC horizon and converted into a predicted SDF sequence. The stage-matched SDFs define the collision-avoidance constraints, while selected snapshots near the terminal stage form the conditioning input of the hypernetwork. The hypernetwork runs once per planning cycle and generates the main-network parameters. Inside the nonlinear program, the terminal constraint is the predicted terminal SDF minus the nonnegative main-network output, as in \cref{eq:rntc-residual-parameterization}. The solver evaluates this assembled function and its derivatives repeatedly.}
  \label{fig:rntc-architecture}
\end{figure}
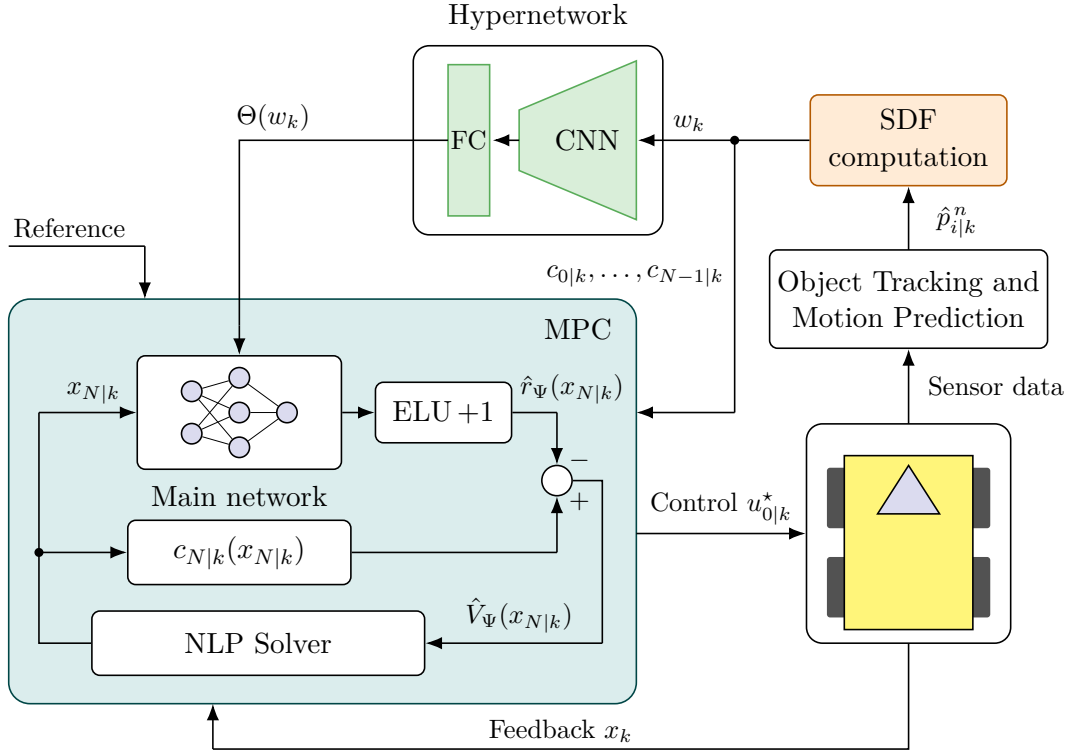

As shown in \cref{fig:rntc-architecture}, one hypernetwork evaluation precedes each MPC solve, and the solver differentiates the terminal SDF minus the residual through automatic differentiation. Fixed-size SDF images keep the main-network query cost independent of the number of obstacles.

\subsection{Reachability-Supervised Training}
\label{subsec:rntc-training}
\label{subsec:rntc-cme}

The hypernetwork is trained offline in a supervised fashion, following \cref{subsec:ntc-training} with two modifications. The training data are obstacle \emph{trajectories} rather than static maps, and the targets are solutions of the \emph{time-varying} HJ problem of \cref{subsec:prelim-time-varying-hj}.

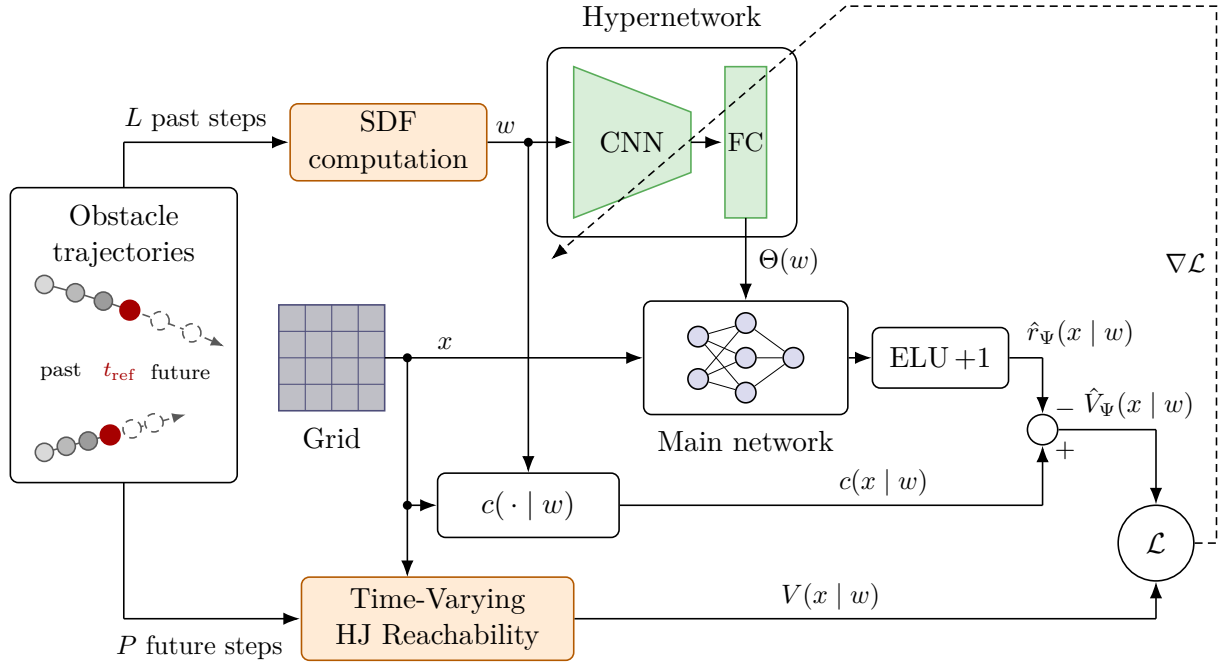
\begin{figure}[tb]
  \centering
  \colorlet{hypfill}{green!55!black!15}%
\colorlet{hypline}{green!50!black!65}%
\colorlet{sdffill}{orange!16}%
\colorlet{sdfline}{orange!70!black}%
\colorlet{mlpfill}{blue!55!black!14}%
\colorlet{gridfill}{blue!10!black!25}%
\colorlet{gridline}{blue!25!black!70}%
\colorlet{obsline}{black!65}%
\colorlet{obspast}{black!25}%
\colorlet{obsnow}{red!65!black}%

\begin{tikzpicture}[
    font=\small,
    blk/.style={draw, semithick, rounded corners=3pt, align=center,
                fill=white, inner sep=1mm},
    sig/.style={font=\footnotesize, align=center, inner sep=1pt},
    arrow/.style={-{Latex[length=2.2mm]}, semithick},
    wire/.style={semithick},
    grad/.style={-{Latex[length=2.2mm]}, semithick, densely dashed},
    unit/.style={circle, draw=black, fill=mlpfill, semithick,
                 inner sep=0pt, minimum size=2.7mm},
  ]

  \node[blk, minimum width=3.00cm, minimum height=3.90cm] (traj) at (1.55,4.65) {};
  \node[anchor=north, align=center] at (1.55,6.52) {Obstacle\\trajectories};

  \draw[wire, obsline]
    (0.50,5.32) .. controls (0.93,5.22) and (1.28,5.10) .. (1.63,4.98);
  \draw[wire, obsline, densely dashed, -{Latex[length=1.8mm]}]
    (1.63,4.98) .. controls (2.03,4.85) and (2.43,4.70) .. (2.88,4.50);
  \foreach \p/\s in {(0.50,5.32)/15,(0.91,5.21)/27,(1.28,5.10)/39}
    {\filldraw[fill=black!\s, draw=obsline, semithick] \p circle (0.12);}
  \foreach \p in {(2.03,4.84),(2.45,4.68)}
    {\filldraw[fill=white, draw=obsline, semithick, densely dashed]
       \p circle (0.12);}
  \filldraw[fill=obsnow, draw=obsnow, semithick] (1.63,4.98) circle (0.13);

  \draw[wire, obsline] (0.50,3.10) -- (1.37,3.33);
  \draw[wire, obsline, densely dashed, -{Latex[length=1.8mm]}]
    (1.37,3.33) -- (2.37,3.60);
  \foreach \p/\s in {(0.50,3.10)/15,(0.79,3.18)/27,(1.08,3.25)/39}
    {\filldraw[fill=black!\s, draw=obsline, semithick] \p circle (0.12);}
  \foreach \p in {(1.66,3.41),(1.95,3.49)}
    {\filldraw[fill=white, draw=obsline, semithick, densely dashed]
       \p circle (0.12);}
  \filldraw[fill=obsnow, draw=obsnow, semithick] (1.37,3.33) circle (0.13);

  \node[sig, font=\scriptsize] at (0.72,4.17) {past};
  \node[sig, font=\scriptsize, obsnow] at (1.50,4.17) {$t_{\mathrm{ref}}$};
  \node[sig, font=\scriptsize] at (2.30,4.17) {future};

  \draw[wire] (1.55,6.60) -- (1.55,7.20);
  \draw[arrow] (1.55,7.20) -- (3.73,7.20);
  \node[sig, anchor=south] at (2.50,7.30) {$L$ past steps};

  \draw[wire] (1.55,2.70) -- (1.55,0.90);
  \draw[arrow] (1.55,0.90) -- (3.88,0.90);
  \node[sig, anchor=south] at (2.55,0.32) {$P$ future steps};

  \node[blk, fill=sdffill, draw=sdfline, minimum width=2.60cm,
        minimum height=1.00cm] (sdf) at (5.05,7.20) {SDF\\computation};

  \node[blk, rounded corners=6pt, minimum width=3.30cm, minimum height=2.50cm]
    (hyper) at (8.80,7.20) {};
  \node[above=0.6mm of hyper] {Hypernetwork};

  \filldraw[fill=hypfill, draw=hypline, semithick]
    (7.50,8.20) -- (9.05,7.60) -- (9.05,6.80) -- (7.50,6.20) -- cycle;
  \node at (8.27,7.20) {CNN};
  \filldraw[fill=hypfill, draw=hypline, semithick]
    (9.50,6.20) rectangle (10.05,8.20);
  \node[font=\footnotesize] at (9.775,7.20) {FC};
  \draw[arrow] (9.05,7.20) -- (9.46,7.20);

  \draw[arrow] (sdf.east) -- (7.48,7.20);
  \fill (6.90,7.20) circle (1.7pt);
  \node[sig, anchor=south] at (6.60,7.28) {$w$};

  \filldraw[fill=gridfill, draw=gridline, semithick]
    (3.60,3.65) rectangle (5.00,5.05);
  \foreach \i in {1,2,3} {
    \draw[gridline, thin] (3.60+0.35*\i,3.65) -- (3.60+0.35*\i,5.05);
    \draw[gridline, thin] (3.60,3.65+0.35*\i) -- (5.00,3.65+0.35*\i);}
  \node[anchor=north] at (4.30,3.57) {Grid};

  \node[blk, minimum width=2.70cm, minimum height=1.50cm]
    (main) at (9.775,4.35) {};
  \node[below=0.6mm of main] {Main network};

  \foreach \i/\y in {a/4.63,b/4.07} {\node[unit] (in\i) at (9.145,\y) {};}
  \foreach \i/\y in {a/4.81,b/4.35,c/3.89} {\node[unit] (hid\i) at (9.775,\y) {};}
  \node[unit] (out) at (10.405,4.35) {};
  \foreach \i in {a,b}{\foreach \j in {a,b,c}{
    \draw[black, line width=0.35pt] (in\i) -- (hid\j);}}
  \foreach \j in {a,b,c}{
    \draw[black, line width=0.35pt] (hid\j) -- (out);}

  \draw[arrow] (9.775,6.20) -- (main.north);
  \node[sig, anchor=west] at (9.90,5.62) {$\Theta(w)$};

  \node[blk, minimum width=1.80cm, minimum height=0.80cm]
    (elu) at (12.35,4.35) {$\operatorname{ELU}+1$};
  \draw[arrow] (main.east) -- (elu.west);

  \node[blk, minimum width=2.40cm, minimum height=0.85cm]
    (cref) at (6.90,2.40) {$c(\,\cdot\mid w)$};
  \draw[arrow] (6.90,7.20) -- (6.90,2.83);

  \draw[wire] (5.00,4.35) -- (5.30,4.35) -- (5.30,1.65);
  \fill (5.30,4.35) circle (1.7pt);
  \fill (5.30,2.40) circle (1.7pt);
  \draw[arrow] (5.30,4.35) -- (main.west);
  \draw[arrow] (5.30,2.40) -- (cref.west);
  \draw[arrow] (5.30,1.65) -- (5.30,1.45);
  \node[sig, anchor=south] at (5.80,4.43) {$x$};

  \node[blk, fill=sdffill, draw=sdfline, minimum width=3.60cm,
        minimum height=1.10cm] (hj) at (5.70,0.90) {Time-Varying\\HJ Reachability};

  \node[circle, draw=black, semithick, fill=white, inner sep=0pt,
        minimum size=0.40cm] (sum) at (13.70,3.40) {};
  \draw[wire] (elu.east) -- (13.70,4.35);
  \draw[arrow] (13.70,4.35) -- (sum.north);
  \node[sig, anchor=south west] at (13.45,4.45) {$\hat{r}_{\Psi}(x\mid w)$};
  \node[sig, anchor=west] at (13.82,3.67) {$-$};
  \draw[wire] (cref.east) -- (13.70,2.40);
  \draw[arrow] (13.70,2.40) -- (sum.south);
  \node[sig, anchor=south] at (11.60,2.48) {$c(x\mid w)$};
  \node[sig, anchor=west] at (13.82,3.13) {$+$};

  \node[circle, draw, semithick, fill=white, inner sep=0pt,
        minimum size=1.00cm] (loss) at (15.20,1.90) {$\mathcal{L}$};

  \draw[wire] (sum.east) -- (15.20,3.40);
  \draw[arrow] (15.20,3.40) -- (loss.north);
  \node[sig, anchor=east] at (15.75,3.75) {$\hat{V}_{\Psi}(x\mid w)$};

  \draw[wire] (hj.east) -- (15.20,0.90);
  \draw[arrow] (15.20,0.90) -- (loss.south);
  \node[sig, anchor=south] at (10.90,0.98) {$V(x\mid w)$};

  \draw[densely dashed, semithick]
    (loss.east) -- (16.00,1.90) -- (16.00,9.00) -- (11.10,9.00);
  \draw[grad] (11.10,9.00) -- (7.20,5.65);
  \node[sig, anchor=east] at (15.90,5.60) {$\nabla\mathcal{L}$};
\end{tikzpicture}
  \caption[RNTC training procedure]{RNTC training procedure. Each obstacle trajectory is split at a reference instant. Selected snapshots before that instant form the conditioning input \(w\) of \cref{eq:rntc-condition}, which the hypernetwork converts into the main-network parameters \(\Theta(w)\). The generated main network produces a nonnegative residual on the state-space grid. The subsequent trajectory segment defines the time-varying failure set used to compute the numerical HJ target \(V(\,\cdot\mid w)\) at the reference instant. Subtracting the predicted residual from the reference-instant SDF yields the value-function estimate used in the training loss. The figure shows a single training pair, so the batch and grid indices of \cref{eq:rntc-cme-loss} are omitted. Signal names match \cref{fig:ntc-training}.}
  \label{fig:rntc-training}
\end{figure}

\Cref{fig:rntc-training} illustrates the construction of one training pair. A trajectory of the surrounding obstacles, synthesized in simulation or recorded in the real world, is split at a reference instant \(t_{\mathrm{ref}}\). The \(L\) selected steps preceding that instant are converted into SDF images over the sensing-window grid and represent the online conditioning input \cref{eq:rntc-condition}. The subsequent segment of \(P\) future steps defines the time-varying failure set evolution after \(t_{\mathrm{ref}}\). A grid-based HJ solver, extended to time-varying failure sets, then propagates the variational inequality backward from the end of this segment to the reference instant and yields the numerical target over the state-space grid. Writing \(w\) for the conditioning input of that pair, this target is denoted \(V(\,\cdot\mid w)\) in the fixed-conditioning notation \cref{eq:prelim-observation-conditioned-function}, and it is the training-time counterpart of the terminal value function \(V_{N\mid k}\). The reachability horizon \(P\) is chosen long enough for the reference-instant value to converge numerically to the chosen tolerance. By the time-shift invariance established in \cref{subsec:rntc-time-varying}, pairs generated at arbitrary reference instants supervise the same relative-time mapping evaluated online at the MPC terminal stage.

Training uses minibatch gradient-based optimization of the hypernetwork parameters \(\Psi\). For each pair, the hypernetwork generates the main-network parameters, and the main network with its shifted ELU output is evaluated on the state grid \(\{x_j\}_{j=1}^{K}\) to obtain the predicted residual. The SDF at the reference instant is evaluated on the same grid, and the subtraction in \cref{eq:rntc-residual-parameterization} assembles the value-function estimate, which the loss compares directly with the numerical HJ target. The training signal acts on the quantity that the controller uses, and a separate residual target is unnecessary.

\paragraph{Sign-sensitive regression loss.}
As discussed in \cref{subsec:ntc-rwmse}, the terminal constraint depends primarily on the \emph{sign} of the learned field. The radially weighted loss of \cref{ch:ntc-mpc} addresses this indirectly by assigning larger squared-error weights near the target boundary. This chapter instead supplements the squared error with a term that responds directly to sign disagreement, which yields the \emph{combined MSE--exponential} (CME) loss. With the abbreviations of \cref{subsec:ntc-rwmse} for a batch of \(M\) training pairs with targets \(V^{(m)}_j:=V(x_j\mid w^{(m)})\) and predictions \(\hat{V}^{(m)}_j:=\hat{V}_{\Psi}(x_j\mid w^{(m)})\) over the grid \(\{x_j\}_{j=1}^{K}\),
\begin{equation}
  \mathcal{L}_{\mathrm{CME}}(\Psi)
  =
  \frac{1}{M}\sum_{m=1}^{M}
  \frac{1}{K}\sum_{j=1}^{K}
  \left[
  \lambda
  \left(
  V^{(m)}_j-\hat{V}^{(m)}_j
  \right)^{2}
  +
  (1-\lambda)
  \exp\!\left(
  -V^{(m)}_j\,\hat{V}^{(m)}_j
  \right)
  \right],
  \qquad
  \lambda\in(0,1].
  \label{eq:rntc-cme-loss}
\end{equation}
The first term is the standard squared regression error and promotes value accuracy. The second term depends on the target and the prediction only through their product, so it decreases when their signs agree and grows exponentially when they disagree. The weight \(\lambda\) trades value matching against sign sensitivity, and the choice \(\lambda=1\) recovers plain MSE. The case \(\lambda=0\) is excluded because, for a nonzero target, the exponential term alone has an unattained infimum rather than a finite minimizer, so the training would drive the prediction to ever larger magnitude.

\begin{figure}[tb]
  \centering
  \includegraphics[width=0.72\textwidth]{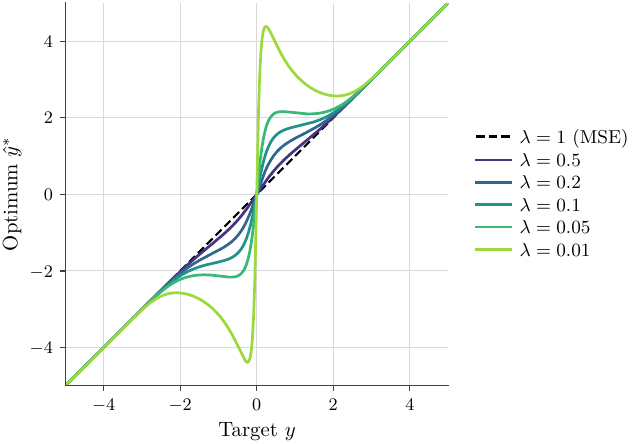}
  \caption[Pointwise optimum of the CME loss]{Pointwise optimum \(\hat{y}^{*}\) of the CME loss \cref{eq:rntc-cme-optimum} as a function of the target \(y\) for different weights \(\lambda\). For \(\lambda=1\) the loss reduces to MSE and the optimum is the identity (dashed). As \(\lambda\) decreases, predictions near the zero level set are displaced away from zero, while the sign, the zero crossing, and the behavior for large \(\lvert y\rvert\) are preserved.}
  \label{fig:rntc-cme-optimum}
\end{figure}

The exponential term deliberately biases the value estimates, so the effect of that bias on the zero level set must be examined. Consider a single target \(y\) and a corresponding prediction \(\hat{y}\). For \(\lambda>0\), the loss \(\lambda(y-\hat{y})^{2}+(1-\lambda)\exp(-y\hat{y})\) is strictly convex in \(\hat{y}\), and its unique minimizer is the stationary point of
\begin{equation}
  -2\lambda(y-\hat{y})
  -
  y(1-\lambda)\exp(-y\hat{y})
  =
  0,
  \label{eq:rntc-cme-stationarity}
\end{equation}
which for \(y\neq0\) admits a closed-form solution in terms of the Lambert \(W\) function \cite{corless_lambertw_1996},
\begin{equation}
  \hat{y}^{*}
  =
  y+\frac{1}{y}\,
  W\!\left(
  \frac{(1-\lambda)\,y^{2}}{2\lambda\exp\!\left(y^{2}\right)}
  \right).
  \label{eq:rntc-cme-optimum}
\end{equation}
The argument of \(W\) is nonnegative, so \(W\) is nonnegative and the correction \(W(\cdot)/y\) has the sign of \(y\). Hence \(\operatorname{sign}(\hat{y}^{*})=\operatorname{sign}(y)\) and \(\lvert\hat{y}^{*}\rvert\geq\lvert y\rvert\). As \(y\rightarrow0\), the correction vanishes and zero maps to zero by continuity. For large \(\lvert y\rvert\), the factor \(\exp(-y^{2})\) suppresses the correction and the optimum approaches the target. \Cref{fig:rntc-cme-optimum} visualizes the resulting mapping. The pointwise CME optimum thus preserves the sign of the target and its zero crossing, at the cost of some value accuracy near the boundary. This property does not guarantee that a finite trained network attains the pointwise optimum, but it explains why the loss is aligned with the classification objective.

Two practical remarks conclude the description of the loss. First, random initialization can produce large values and gradients in the exponential term, which can be avoided by training with plain MSE for an initial phase before switching to CME. Second, the weight \(\lambda\) is a hyperparameter of the objective. \Cref{subsec:rntc-training-eval} reports the value used in the experiments and compares it with plain MSE under otherwise identical conditions.

\subsection{RNTC-MPC Formulation}
\label{subsec:rntc-ocp}

The preceding components are now assembled into the complete planner. Given the current state \(x_k\), the predicted constraint functions \(c_{0\mid k},\ldots,c_{N\mid k}\) of \cref{eq:rntc-predicted-constraint}, and the main-network parameters \(\Theta(w_k)=H_{\Psi}(w_k)\) computed once for this cycle, RNTC-MPC solves
\begin{subequations}
  \label{eq:rntc-mpc-problem}
  \begin{align}
    \min_{\substack{x_{0\mid k},\ldots,x_{N\mid k}                              \\ u_{0\mid k},\ldots,u_{N-1\mid k}}}
    \quad &
    \ell_{f\mid k}(x_{N\mid k})
    +
    \sum_{i=0}^{N-1}
    \ell_{i\mid k}(x_{i\mid k},u_{i\mid k})
    \label{eq:rntc-mpc-objective}
    \\
    \text{ s.t. }\quad
          &
    x_{0\mid k}=x_k,
    \label{eq:rntc-mpc-initial-condition}
    \\
          &
    x_{i+1\mid k}
    =
    F_\Delta(x_{i\mid k},u_{i\mid k}),
    \quad
    i=0,\ldots,N-1,
    \label{eq:rntc-mpc-dynamics}
    \\
          &
    u_{i\mid k}\in\mathcal{U},
    \quad
    i=0,\ldots,N-1,
    \label{eq:rntc-mpc-input-constraints}
    \\
          &
    x_{i\mid k}\in\mathcal{X},
    \quad
    i=0,\ldots,N,
    \label{eq:rntc-mpc-state-constraints}
    \\
          &
    c_{i\mid k}(x_{i\mid k})\geq0,
    \quad
    i=0,\ldots,N-1,
    \label{eq:rntc-mpc-stage-safety}
    \\
          &
    c_{N\mid k}(x_{N\mid k})
    -
    \hat{r}_{\Psi}(x_{N\mid k}\mid w_k)
    \geq0,
    \label{eq:rntc-terminal}
  \end{align}
\end{subequations}
with the quadratic tracking costs of \cref{sec:prelim-mpc}. Relative to NTC-MPC \cref{eq:ntc-mpc-problem}, two substitutions extend the formulation to moving obstacles. First, each stage constraint \cref{eq:rntc-mpc-stage-safety} refers to the predicted geometry at the corresponding stage. Second, the terminal constraint \cref{eq:rntc-terminal} imposes the residual value estimate \(\hat{V}_{\Psi}(x_{N\mid k}\mid w_k)\geq0\), that is, it requires the terminal state to lie in the predicted safe set, which by \cref{eq:rntc-failure-set-exclusion} implies \(c_{N\mid k}(x_{N\mid k})\geq0\).

Implementation follows \cref{subsec:ntc-ocp}. A compiled NLP takes the network weights, predicted SDFs, state estimate, and reference as updated parameters, and automatic differentiation supplies the derivatives. Each cycle adds one hypernetwork inference and repeated evaluation of a single terminal field whose network size is independent of the obstacle count. Predicted-failure-set exclusion \cref{eq:rntc-failure-set-exclusion} holds by construction, while real-time execution and closed-loop collision avoidance are evaluated experimentally in the next section, and \cref{sec:rntc-discussion} states the scope of these claims.

\section{Experimental Evaluation}
\label{sec:rntc-experiments}

The evaluation compares residual and direct regression at equal model capacity, CME and MSE training, closed-loop performance against four baselines, and deployment on a physical robot.\footnote{The implementation is available at \url{https://github.com/bojan-derajic/RNTC-MPC}.}

\subsection{Robot Model}
\label{subsec:rntc-robot-model}

All experiments use the kinematic unicycle, the standard velocity-controlled model of a differential-drive platform. The state \(x=(p_x,p_y,\theta)\) collects the planar position \(p(x)=(p_x,p_y)\) and the heading angle \(\theta\), the control input \(u=(v,\omega)\) consists of the linear and angular velocities, and the continuous-time dynamics are
\begin{equation}
  \dot{x}
  =
  \begin{bmatrix}
    v\cos\theta \\
    v\sin\theta \\
    \omega
  \end{bmatrix},
  \label{eq:rntc-unicycle-dynamics}
\end{equation}
with actuation limits \(v\in[-0.5,0.5]\)~\si{\meter\per\second} and \(\omega\in[-0.5,0.5]\)~\si{\radian\per\second}.

This idealized velocity-controlled model isolates the effect of obstacle motion. Because velocity is the control input, the model can stop instantaneously, so among static obstacles the robot can remain at rest at every collision-free state, and the reachability-based safe set coincides with the free space. Moving obstacles remove this option, because the experimental obstacles travel at up to \SI{1}{\meter\per\second}, twice the maximum speed of the robot, and can reach a robot that stops or reacts too late. The predicted safe set can then be strictly smaller than the predicted free space by a velocity-dependent margin.

\subsection{Training and Safe-Set Approximation}
\label{subsec:rntc-training-eval}

\paragraph{Dataset and targets.}
The training data are generated synthetically. The sensing window is a square of \(8\times8\)~\si{\meter} centered at the robot and discretized into \(100\times100\) cells. Each training scene contains up to four circular obstacles whose initial positions and velocity vectors are sampled randomly, with speeds up to \SI{1}{\meter\per\second}, and whose trajectories follow the constant-velocity model. Each trajectory is split at a reference instant as described in \cref{subsec:rntc-training}. The past segment yields the input SDF pair, consisting of \(L=2\) snapshots taken \SI{0.4}{\second} apart, and the future segment defines the time-varying failure set for the numerical HJ computation. With the prediction interval \(\Delta t=\SI{0.1}{\second}\) of \cref{subsec:rntc-baselines}, this separation corresponds to the lags \((q_1,q_2)=(4,0)\) in \cref{eq:rntc-condition}, so that every evaluated horizon satisfies \(N\geq q_1\). The targets are computed with a grid-based HJ solver\footnote{\url{https://github.com/StanfordASL/hj_reachability}, modified to handle time-varying failure sets.} on a \(100\times100\times30\) grid over \((p_x,p_y,\theta)\). Generating the full set of \num{40000} training pairs took approximately \SI{20}{\hour} on two RTX~3090 GPUs. As in \cref{ch:ntc-mpc}, part of the data is held out for validation.

\begin{table}[tb]
  \caption{RNTC-MPC model and training configuration.}
  \label{tab:rntc-training-config}
  \centering
  \begin{tabular}{@{}ll@{}}
    \toprule
    Robot model                & kinematic unicycle \cref{eq:rntc-unicycle-dynamics}                       \\
    Sensing window             & \(8\times8\)~\si{\meter}, \(100\times100\) cells                          \\
    Obstacles per scene        & up to 4, speeds \(\leq\SI{1}{\meter\per\second}\), constant velocity      \\
    Condition (SDF snapshots)  & \(L=2\), lags \((q_1,q_2)=(4,0)\), \SI{0.4}{\second} apart                 \\
    Dataset size               & \num{40000} SDF-sequence/value-function pairs                             \\
    State-space grid           & \(100\times100\times30\) over \((p_x,p_y,\theta)\)                        \\
    Hypernetwork               & CNN backbone with fully connected head                                    \\
    Hypernetwork parameters    & \(\sim9.4\times10^{6}\)                                                  \\
    Main network hidden layers & \([36,36,36,18,18,18,9,9,9]\), sine + SELU                                \\
    Main network output        & \(\operatorname{ELU}+1\) (nonnegative residual)                           \\
    Main network parameters    & \num{4519}                                                                \\
    Loss                       & CME, \(\lambda=0.1\)                                                      \\
    Optimizer / epochs / batch & Adam / 100 / 40                                                           \\
    Learning rate              & \(10^{-4}\), reduced \(10\times\) at epochs 85 and 95                     \\
    Training time (RTX~3090)   & \(\sim\)\SI{15}{\hour}                                                    \\
    \bottomrule
  \end{tabular}
\end{table}

\paragraph{Model and training configuration.}
\Cref{tab:rntc-training-config} summarizes the setup. The hypernetwork is a CNN with two input channels, one per SDF snapshot, followed by a fully connected head, and it has approximately \(9.4\times10^{6}\) trainable parameters. The main network is an MLP with three inputs and nine hidden layers of widths \([36,36,36,18,18,18,9,9,9]\). Its first three layers use sinusoidal activations, the remaining hidden layers use SELU, and the output uses the shifted ELU of \cref{eq:rntc-hypernetwork}. The main network has \num{4519} parameters, which is also the output dimension of the hypernetwork. Training uses Adam for 100~epochs with a batch size of 40 and the CME weight \(\lambda=0.1\) selected by grid search, without the optional MSE warm start mentioned in \cref{subsec:rntc-cme}.

\begin{figure}[tb]
  \centering
  \includegraphics[width=\textwidth]{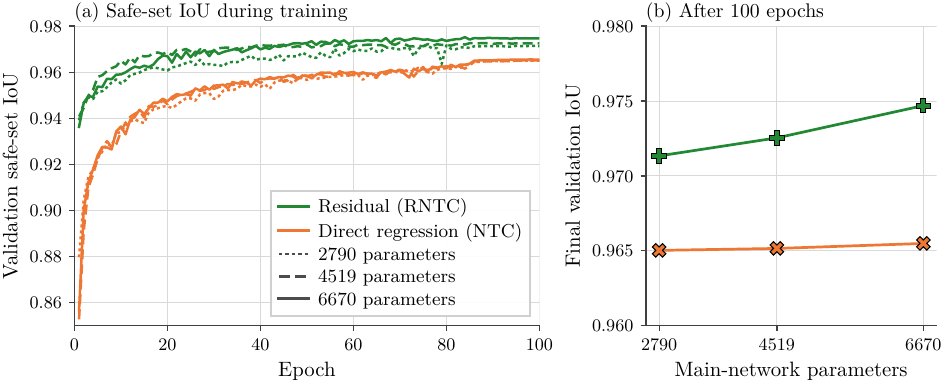}
  \caption[Residual parameterization versus direct value regression at equal model capacity]{Residual parameterization versus direct value regression at equal model capacity, evaluated on the validation set for three main-network sizes (\num{2790}, \num{4519}, and \num{6670} parameters). (a)~Safe-set intersection over union (IoU) over training epochs, with the residual models in green and the direct-regression models in orange, and the line pattern indicating the network size. The residual models start higher, converge roughly five times faster, and end higher. (b)~Final IoU after 100~epochs. The residual models reach a higher IoU at every size, and the smallest residual model outperforms the largest direct-regression model.}
  \label{fig:rntc-iou-ablation}
\end{figure}

\paragraph{Residual parameterization versus direct regression.}
The first ablation isolates the residual parameterization. Its direct-regression counterpart uses the same hypernetwork, the same SDF input, and the same training data, but it predicts the value function directly, without the SDF constraint function and the nonnegative residual. Both variants are trained at three main-network sizes and evaluated by the safe-set intersection over union (IoU) on the validation set. \Cref{fig:rntc-iou-ablation} shows three consistent results. At the earliest reported epochs, the residual models already have a higher IoU, about \(0.94\) compared with \(0.86\), because the SDF constraint function supplies the scene geometry from the outset. After approximately 10~epochs they reach the IoU that direct regression attains only after about 50~epochs. After 100~epochs, they also end at a higher IoU for every tested size, approximately \(0.971\)--\(0.975\) compared with about \(0.965\). In addition to enforcing predicted-failure-set exclusion, the residual parameterization makes the supervised approximation problem easier in this ablation.

\begin{figure}[tb]
  \centering
  \includegraphics[width=0.8\textwidth]{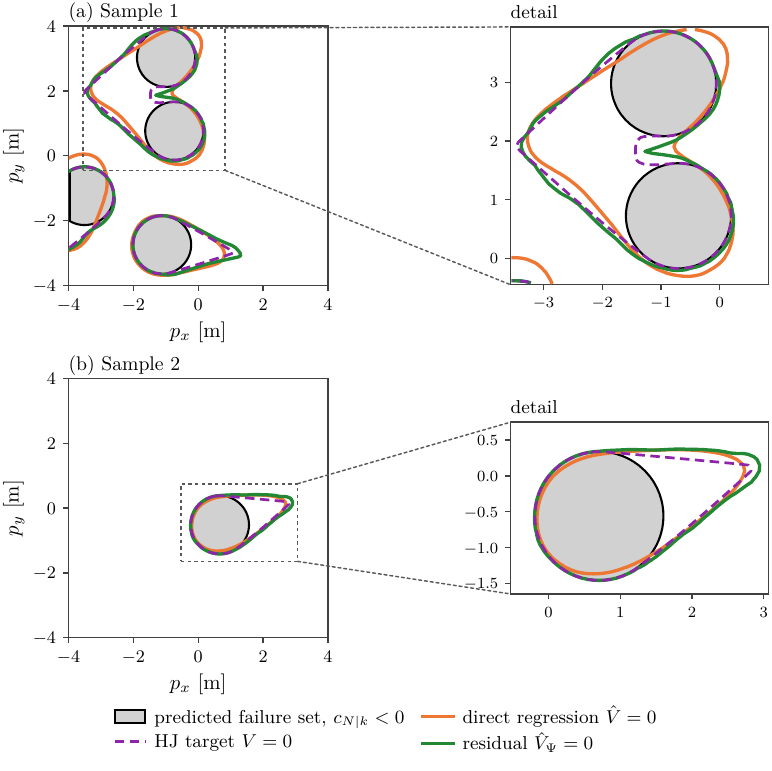}
  \caption[Qualitative comparison of residual and direct-regression estimates]{Qualitative comparison for two validation samples (a)~and~(b), each shown as a heading slice over the full sensing window and, connected by the dotted lines, as a magnified detail of the marked rectangle. Gray disks mark the predicted failure set, in which the terminal constraint function \(c_{N\mid k}\) is negative. The curves are the zero level sets of the numerical HJ target (dashed), of the direct-regression estimate (orange), and of the residual estimate (green). The direct-regression boundary enters the predicted failure set in several locations, while the residual estimate excludes that region by construction.}
  \label{fig:rntc-ntc-vs-rntc-slices}
\end{figure}

\Cref{fig:rntc-ntc-vs-rntc-slices} visualizes the structural difference. In both validation scenes, the direct-regression boundary enters the predicted SDF failure set and assigns positive values to some represented occupied states, a failure mode that the residual estimate excludes by \cref{eq:rntc-failure-set-exclusion}.

\begin{figure}[tb]
  \centering
  \includegraphics[width=0.5\textwidth]{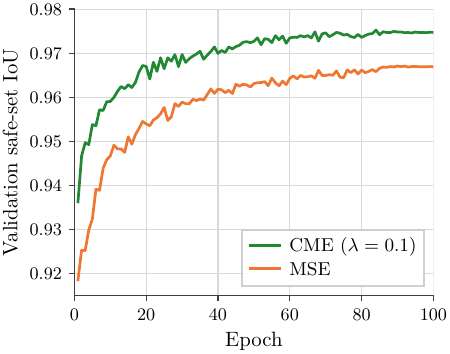}
  \caption[Effect of the CME and MSE training objectives]{Effect of the training objective on the safe-set IoU, evaluated on the validation set over training epochs for the same residual model trained with the CME loss (\(\lambda=0.1\)) and with plain MSE under identical settings. CME converges faster, reaching the final MSE level after about 20~epochs, and ends at a higher IoU (\(\approx0.975\) versus \(\approx0.967\)).}
  \label{fig:rntc-cme-vs-mse}
\end{figure}

\paragraph{Effect of the CME loss.}
The second ablation changes only the training objective. The same residual model is trained once with the CME loss and once with plain MSE under otherwise identical conditions. As shown in \cref{fig:rntc-cme-vs-mse}, CME reaches the final MSE validation IoU after roughly 20 of the 100~epochs and finishes at approximately \(0.975\), compared with \(0.967\) for MSE. This experiment supports the use of sign-sensitive training whenever safe-set classification, rather than value regression alone, is the principal objective.

\subsection{Planner Baselines and Implementation}
\label{subsec:rntc-baselines}

The closed-loop evaluation compares five planners with the same MPC costs, robot model, prediction discretization, and constant-velocity obstacle predictions. They differ only in their collision-avoidance constraints, so the comparison isolates the effect of the constraint construction. The first two baselines are the dynamic-environment counterparts of those in \cref{subsec:ntc-baselines}. SDF-MPC uses the predicted geometric constraint at every stage including the terminal one, \(c_{i\mid k}(x_{i\mid k})\geq0\) for \(i=0,\ldots,N\). DCBF-MPC \cite{zeng_safety-critical_2021} additionally imposes the discrete-time CBF condition \(c_{i\mid k}(x_{i\mid k})-c_{i-1\mid k}(x_{i-1\mid k})\geq-\gamma\,c_{i-1\mid k}(x_{i-1\mid k})\) for \(i=1,\ldots,N\), with decay parameter \(\gamma\in(0,1]\) as in \cref{eq:ntc-dcbf-mpc}, evaluated along the time-varying constraint sequence. The third baseline, VO-MPC \cite{piccinelli_mpc_2023}, formulates velocity-obstacle constraints within the MPC problem. For each obstacle, the commanded velocity is kept outside the cone of velocities that lead to a collision under the current relative motion. The fourth planner, NTC-MPC, uses the direct-regression ablation of \cref{subsec:rntc-training-eval} as its terminal constraint, which isolates the contribution of the residual structure in closed loop. The fifth planner is RNTC-MPC as developed in this chapter.

All planners are implemented in the CasADi framework \cite{andersson_casadi_2019} with IPOPT as the NLP solver and MA57 for the linear systems \cite{duff_ma57---code_2004}, and they are deployed as ROS~2 packages \cite{macenski_robot_2022} with solver code compiled before deployment. The internal MPC prediction grid uses \(\Delta t=\SI{0.1}{\second}\), while the receding-horizon problem is solved at the control frequency of \SI{20}{\hertz}. The planner replans twice per prediction interval, and a horizon of \(N=10\) still denotes a prediction horizon of \SI{1}{\second}.
The hypernetwork runs in PyTorch on the GPU and requires about \SI{2}{\milli\second} per inference on an RTX~3090. The main network is expressed directly in CasADi, and its weights, the sampled predicted SDFs, the state estimate, and the reference are updated as solver parameters at each planning cycle.

\subsection{Simulation Experiments}
\label{subsec:rntc-simulations}

\begin{figure}[tb]
  \centering
  \includegraphics[width=0.8\textwidth]{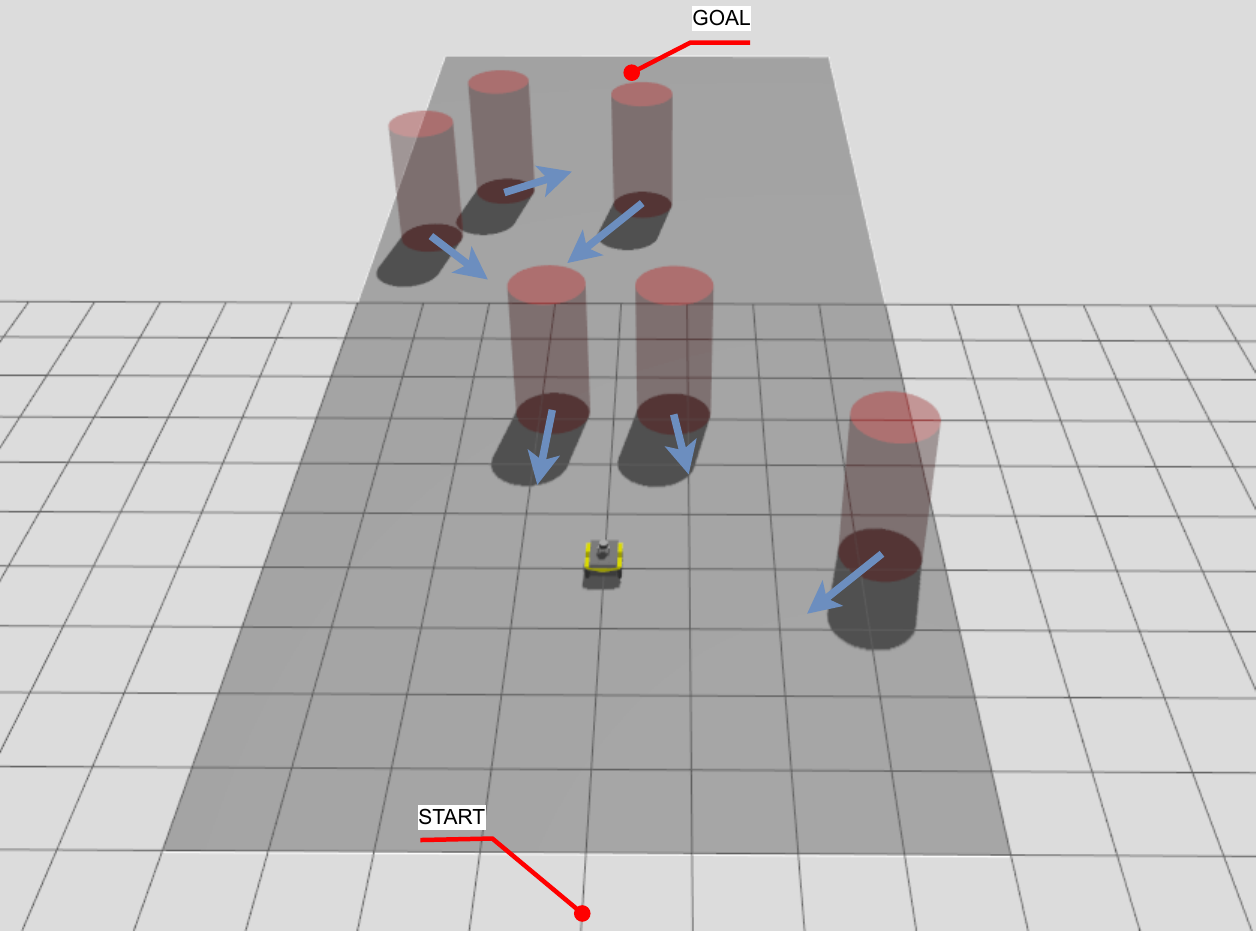}
  \caption[Gazebo simulation environment of the RNTC-MPC study]{Gazebo simulation environment. Six cylindrical obstacles with randomly initialized positions and velocity vectors (arrows) move at constant speed within a corridor and reverse direction at its edges. The robot must navigate from the start position to the goal position ahead of it, using only local observations.}
  \label{fig:rntc-gazebo-env}
\end{figure}

\paragraph{Setup.}
The closed-loop evaluation uses physics-based 3D simulations in Gazebo (\cref{fig:rntc-gazebo-env}). The robot must travel \SI{15}{\meter} along a straight reference path from a start position to a goal position, while avoiding six moving cylindrical obstacles with a radius of \SI{0.5}{\meter}. The obstacles move within a corridor of \SI{16}{\meter} length and \SI{10}{\meter} width that contains the reference path. Their initial positions are drawn uniformly within the corridor, their velocity vectors have random directions and a speed of \SI{1}{\meter\per\second}, and each obstacle reverses its velocity when it leaves the corridor, so that the obstacles remain in the path of the robot. A total of 100 random scenarios is generated once and reused across all planners and horizon lengths \(N\in\{5,10,20,30\}\). The simulator provides accurate obstacle positions and velocities within the sensing range, which isolates the planning problem from perception errors. A run fails if the robot collides or does not reach the goal within a fixed time limit, and besides the success rate and the NLP computation time, the evaluation reports the travel time from start to goal and the per-run mean \(d_{\mathrm{mean}}\) and maximum \(d_{\mathrm{max}}\) of the lateral deviation \(d\) from the straight reference path. Small deviations are important in applications such as cleaning or inspection, where following the path is itself the objective.

\begin{figure}[tb]
  \centering
  \includegraphics[width=\textwidth]{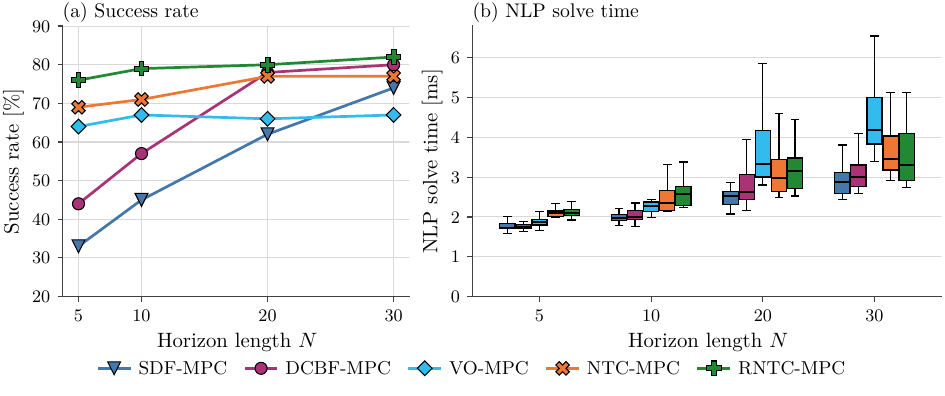}
  \caption[Success rates and NLP solve times of the RNTC-MPC simulation study]{Monte Carlo simulation results over 100~randomized scenarios. (a)~Success rate against the prediction-horizon length. RNTC-MPC achieves the highest rate at every tested horizon and consistently exceeds direct-regression NTC-MPC. (b)~Per-run mean NLP solve times over the scenarios completed by all planners, which remain considerably below the \SI{50}{\milli\second} control-update period. Boxes span the interquartile range, lines mark the median, and whiskers extend to 1.5 times the interquartile range.}
  \label{fig:rntc-sim-results}
\end{figure}

\paragraph{Success rates.}
\Cref{fig:rntc-sim-results}(a) reports the success rates. RNTC-MPC has the highest rate at every tested horizon, from \(76\%\) at \(N=5\) to \(82\%\) at \(N=30\). At \(N=5\), it exceeds SDF-MPC by 43 and DCBF-MPC by 32~percentage points. SDF-MPC increases from \(33\%\) to \(74\%\) and DCBF-MPC from \(44\%\) to \(80\%\) as the horizon grows, which illustrates their stronger horizon dependence, while VO-MPC remains close to \(66\%\) at every horizon. Direct-regression NTC-MPC is the strongest baseline at the two shorter horizons and follows the same broad trend as RNTC-MPC, but it remains 3--8~percentage points lower at every horizon, which suggests that the approximation gains of \cref{subsec:rntc-training-eval} translate into fewer collisions. No method reaches \(100\%\). The obstacles move faster than the robot and reverse direction at the corridor boundary, which violates the constant-velocity prediction used by every planner. Both effects are possible causes of the remaining failures, but the experiment does not isolate them.

\paragraph{Computation times.}
\Cref{fig:rntc-sim-results}(b) reports the NLP solve times, which lie in the low-millisecond range for all planners. The solve time of RNTC-MPC grows from roughly \SI{2}{\milli\second} at \(N=5\) to \SI{3.3}{\milli\second} at \(N=30\), which is close to NTC-MPC, slightly above the geometric baselines, and below VO-MPC at long horizons. Together with the approximately \SI{2}{\milli\second} hypernetwork inference, these values remain considerably below the \SI{50}{\milli\second} control-update budget.

\begin{figure}[tb]
  \centering
  \includegraphics[width=\textwidth]{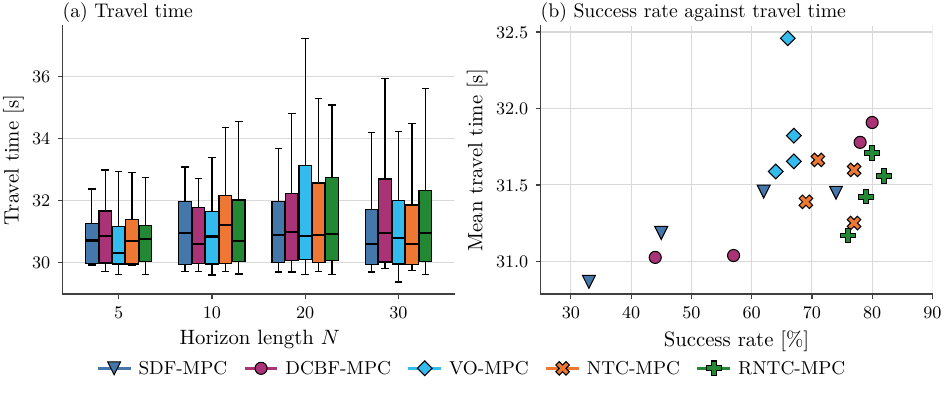}
  \caption[Motion efficiency in the RNTC-MPC simulation study]{Motion efficiency in simulation over the scenarios completed by all planners. (a)~Travel times from start to goal, which are similar across methods. (b)~Mean travel time against success rate, with one marker per method and horizon. Higher success and shorter travel time are preferable, so the lower right region is favorable. RNTC-MPC achieves the highest success rate at every tested horizon with a mean travel time close to the lowest.}
  \label{fig:rntc-travel-time}
\end{figure}

\paragraph{Motion quality.}
To assess the performance cost of improved collision avoidance, travel time and lateral deviation are compared across planners as well. \Cref{fig:rntc-travel-time}(a) shows similar travel times across the planners, and the trade-off in \cref{fig:rntc-travel-time}(b) combines the mean travel time with the success rate. RNTC-MPC has the highest success rate at every tested horizon, and its mean travel time remains within \SI{0.4}{\second} of that of the fastest planner.

\begin{figure}[tb]
  \centering
  \includegraphics[width=\textwidth]{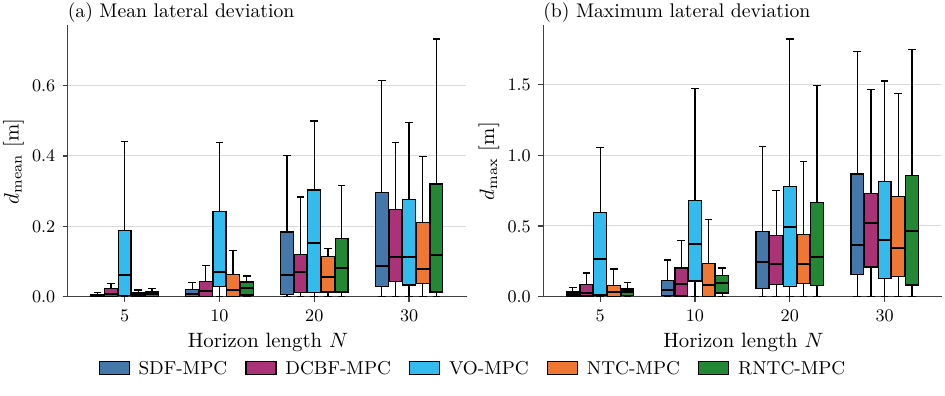}
  \caption[Lateral deviations in the RNTC-MPC simulation study]{(a)~Per-run mean and (b)~per-run maximum lateral deviation from the reference path in the simulation study, shown over the horizon length for the scenarios completed by all planners. VO-MPC deviates most at the three shorter horizons, which reflects its evasive velocity constraints, while RNTC-MPC remains comparable to the geometric baselines. The deviations of all methods are larger at long horizons than at short ones, since a longer look-ahead triggers earlier and wider avoidance.}
  \label{fig:rntc-lat-dev}
\end{figure}

\begin{figure}[tb]
  \centering
  \includegraphics[width=\textwidth]{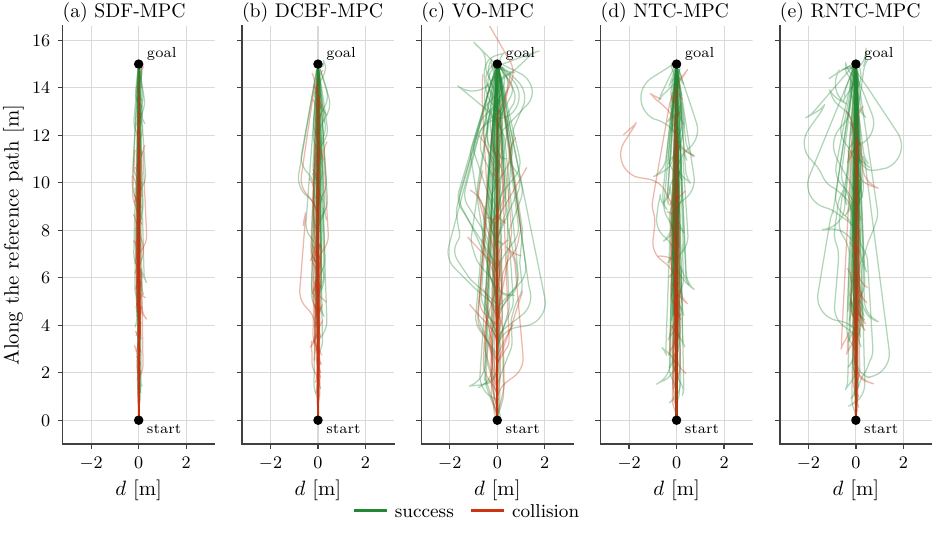}
  \caption[Executed trajectories in the RNTC-MPC simulation study]{Executed trajectories over all 100~simulation runs for each planner at \(N=10\), corresponding to a \SI{1}{\second} prediction horizon, plotted as lateral deviation \(d\) against the distance along the reference path. Green denotes successful runs and red denotes collisions. SDF-MPC remains close to the reference but collides frequently, VO-MPC deviates widely from the reference, and the planners with learned terminal constraints achieve moderate deviations with fewer failures.}
  \label{fig:rntc-sim-trajectories}
\end{figure}

\Cref{fig:rntc-lat-dev} reports the lateral deviations. VO-MPC has the largest deviations, particularly at short horizons, and RNTC-MPC remains comparable to the geometric baselines while completing more trials without collision. The deviations increase with the horizon for all methods, since a longer look-ahead enables earlier and wider avoidance. The trajectories in \cref{fig:rntc-sim-trajectories} show the same pattern, in which SDF-MPC remains close to the reference but collides frequently, VO-MPC deviates widely from the reference, and RNTC-MPC combines moderate deviations with fewer failures.

\subsection{Hardware Experiments}
\label{subsec:rntc-hardware}

\paragraph{Setup.}
The hardware evaluation uses the delivery robot of \cref{subsec:ntc-hardware}, which additionally carries stereo cameras. The complete pipeline of object detection and tracking, localization, constant-velocity prediction, hypernetwork inference, and MPC optimization runs onboard. Hypernetwork inference takes about \SI{4.5}{\milli\second}, and the planner executes at \SI{20}{\hertz}. Each of SDF-MPC, DCBF-MPC, VO-MPC, and RNTC-MPC is evaluated in ten crossing-pedestrian trials at \(N=10\), which corresponds to a prediction horizon of \SI{1}{\second}. Five of these trials contain one pedestrian, and five contain two. The initial pedestrian positions and velocities vary between trials, and the pedestrians move faster than the maximum robot speed of \SI{0.5}{\meter\per\second}.

\begin{table}[tb]
  \caption[RNTC-MPC hardware experiment results]{Hardware experiment results in the pedestrian-avoidance task, with 10~trials per method and horizon \(N=10\), corresponding to \SI{1}{\second}. Values are reported as mean \(\pm\) standard deviation where applicable.}
  \label{tab:rntc-hardware-results}
  \centering
  \small
  \setlength{\tabcolsep}{4pt}
  \begin{tabular}{@{}lccccc@{}}
    \toprule
    Method & Success rate [\%] & \(d_{\mathrm{mean}}\) [\si{\meter}] & \(d_{\mathrm{max}}\) [\si{\meter}] & Opt.\ time [\si{\milli\second}] & Travel time [\si{\second}] \\
    \midrule
    SDF-MPC         & 40.0           & 0.03 {\scriptsize\(\pm\)~0.01} & 0.08 {\scriptsize\(\pm\)~0.01} & 2.93 {\scriptsize\(\pm\)~0.25} & 13.03 {\scriptsize\(\pm\)~1.73} \\
    DCBF-MPC        & 70.0           & 0.11 {\scriptsize\(\pm\)~0.06} & 0.36 {\scriptsize\(\pm\)~0.22} & 3.09 {\scriptsize\(\pm\)~0.38} & 13.98 {\scriptsize\(\pm\)~0.93} \\
    VO-MPC          & 70.0           & 0.19 {\scriptsize\(\pm\)~0.13} & 0.53 {\scriptsize\(\pm\)~0.33} & 6.94 {\scriptsize\(\pm\)~2.61} & 14.67 {\scriptsize\(\pm\)~1.55} \\
    RNTC-MPC (ours) & \textbf{100.0} & 0.04 {\scriptsize\(\pm\)~0.02} & 0.10 {\scriptsize\(\pm\)~0.08} & 4.20 {\scriptsize\(\pm\)~0.56} & 12.76 {\scriptsize\(\pm\)~0.89} \\
    \bottomrule
  \end{tabular}
\end{table}

\paragraph{Results.}
\Cref{tab:rntc-hardware-results} summarizes the outcomes. RNTC-MPC completes all ten trials without collision, 30~percentage points above DCBF-MPC and VO-MPC, the best baselines at \(70\%\). It also attains the lowest mean travel time, with lateral deviations close to those of SDF-MPC, whose small deviations do not indicate better avoidance, since that planner collides in six of the ten trials. DCBF-MPC and VO-MPC complete more trials than SDF-MPC but make larger detours, and VO-MPC also has the highest mean optimization time. Because each method is evaluated only ten times, the hardware percentages have substantial sampling uncertainty and should be interpreted as evidence from the tested trials rather than as general success rates.

\begin{figure}[tb]
  \centering
  \includegraphics[width=\textwidth]{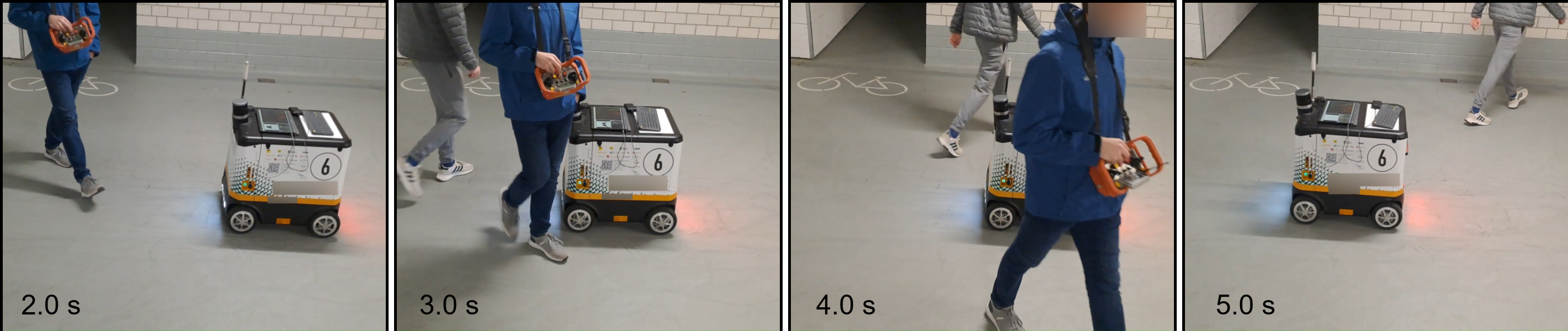}
  \caption[RNTC-MPC hardware experiment among crossing pedestrians]{Successful collision avoidance by RNTC-MPC during the hardware experiments with a prediction horizon of \SI{1}{\second}. The robot slows down before the crossing pedestrian and resumes its path once the predicted pedestrian trajectory no longer constrains the terminal state.}
  \label{fig:rntc-hardware-exp}
\end{figure}

\begin{figure}[tb]
  \centering
  \includegraphics[width=\textwidth]{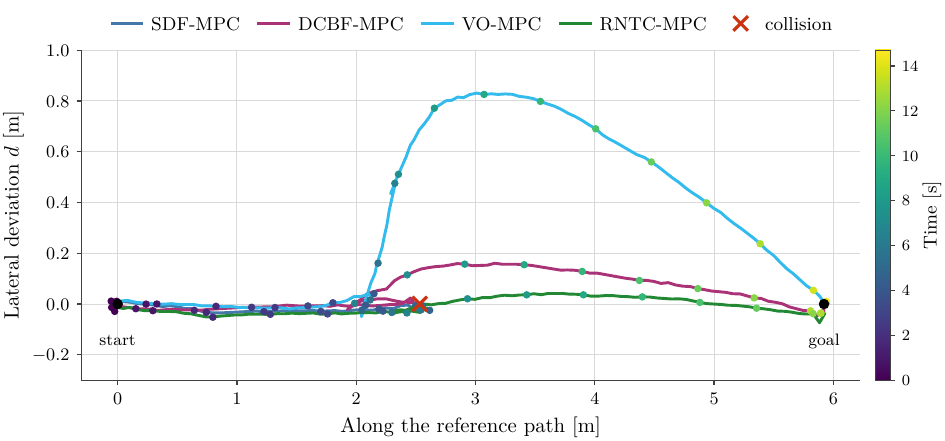}
  \caption[Executed trajectories in a representative RNTC-MPC hardware trial]{Executed trajectories of the four planners in one representative hardware scenario with a single crossing pedestrian, with one run per planner under the same experimental setting, plotted as lateral deviation \(d\) against the distance along the reference path, with the dots along each trajectory colored by time. SDF-MPC fails to react and collides after about \SI{2.5}{\meter} (red cross). VO-MPC evades with a detour of up to \SI{0.8}{\meter} and DCBF-MPC with up to \SI{0.16}{\meter}, while RNTC-MPC resolves the encounter with a deviation of only a few centimeters by adjusting its timing along the path.}
  \label{fig:rntc-hardware-traj}
\end{figure}

\begin{figure}[tb]
  \centering
  \includegraphics[width=\textwidth]{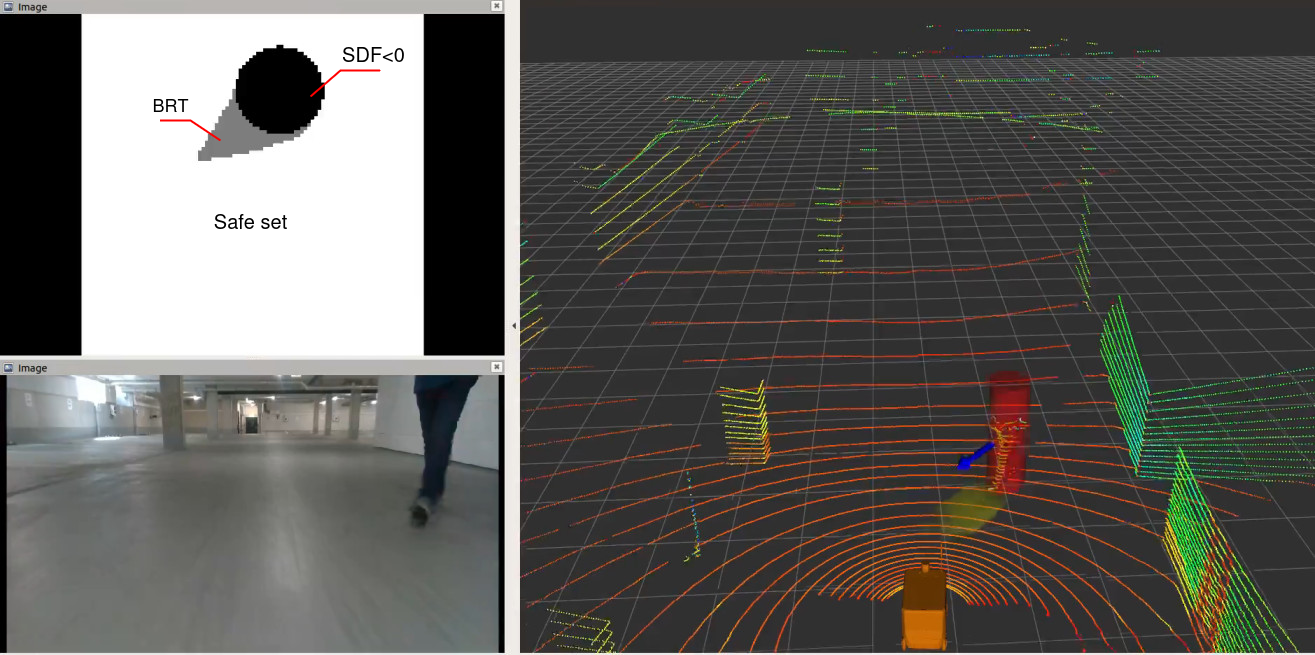}
  \caption[Onboard view during an RNTC-MPC hardware trial]{Onboard view during a hardware trial. Top left: boundary of the predicted terminal safe set relative to the predicted pedestrian motion. Bottom left: front-camera view. Right: detected pedestrian and predicted trajectory used to construct the SDF sequence.}
  \label{fig:rntc-hardware-screen}
\end{figure}

The qualitative results illustrate the characteristics of the individual planners. In \cref{fig:rntc-hardware-traj}, RNTC-MPC reduces its speed along the reference path and allows the pedestrian to pass with only a small lateral deviation. The other planners either deviate farther or, in the case of SDF-MPC in this scenario, react too late to avoid the collision. \Cref{fig:rntc-hardware-exp} shows the encounter from an external viewpoint, and \cref{fig:rntc-hardware-screen} shows the predicted pedestrian trajectory, the resulting SDF sequence, and the estimated terminal-set boundary as computed onboard. Together with the timing measurements, these trials demonstrate real-time execution of the complete perception--prediction--planning pipeline on the physical robot.

\section{Discussion}
\label{sec:rntc-discussion}

The experiments support prediction-conditioned terminal constraints in the tested dynamic environments and show improved approximation with residual learning. The exclusion property is structural, and the following discussion separates it from the empirical results.

\paragraph{Scope of guarantees.}
Predicted-failure-set exclusion \cref{eq:rntc-failure-set-exclusion} holds by construction and is the property that distinguishes the method structurally from \cref{ch:ntc-mpc}. It is stated relative to the supplied terminal geometry, so it does not cover unobserved or occluded obstacles, motion-prediction errors, or states outside the predicted failure set, where an underestimated residual can remain optimistic. Collision avoidance and recursive feasibility are demonstrated in the tested scenarios under the assumptions of \cref{sec:rntc-problem-statement} rather than certified formally. The actual obstacle motion may differ from its prediction, and the terminal set changes as the predictions are refreshed, so consecutive sets need not satisfy the compatibility condition of \cref{sec:prelim-mpc}. A recursive-feasibility argument that tolerates this per-cycle exchange of learned terminal sets remains open, and \cref{ch:conclusion} places it among the steps toward a formal closed-loop result.

\paragraph{Dependence on the obstacle-motion model.}
The constant-velocity model enters the method twice, online through the predicted SDF sequence and offline through the trajectories used to compute the training targets. Replanning at \SI{20}{\hertz} refreshes the nominal prediction frequently, but it does not bound the prediction error between updates. The simulation obstacles reverse direction at the corridor boundaries, which directly violates the prediction model and thereby demonstrates this limitation. The architecture itself can accept snapshots from another predictor, including the rule-based and learned models surveyed in \cite{korbmacher_review_2022}, provided that the training targets are generated under a compatible evolution model. Higher-order models may require more snapshots or different conditioning inputs. Representing prediction \emph{uncertainty} is a more fundamental extension, since a distribution of future trajectories requires stochastic or robust terminal sets, for example following \cite{zhang_trajectory_2021}, rather than the single nominal evolution used here.

\paragraph{Scalability of supervision.}
As in \cref{ch:ntc-mpc}, the offline stage inherits the dimensionality limit of grid-based HJ solvers. Generating \num{40000} time-varying targets required approximately \SI{20}{\hour} on two GPUs, and the cost grows exponentially with the dimension of the robot state. Grid-free supervision, such as self-supervised HJ learning \cite{bansal_deepreach_2021}, could remove this limit, while the vertex-restricted search of \cref{ch:vertex-cbf} applies only to robot models with control-affine dynamics and polytopic input constraints, which RNTC-MPC does not assume in general. Approximation methods with statistical error guarantees \cite{taheri_statistical_2021} could further strengthen the connection between training metrics and deployment claims.

\paragraph{Scope of the residual principle.}
The residual parameterization does not require moving obstacles. In a static environment, it becomes a residual variant of the terminal constraint of \cref{ch:ntc-mpc} built on the observed SDF and retains failure-set exclusion, although whether it yields the same learning benefit in that setting would have to be assessed in a separate ablation. The construction is also not specific to MPC, since it produces a state-space field whose nonnegative superlevel set approximates a reachability-based safe set. \Cref{ch:orn-cbf,ch:cn-cbf} reuse this structure as a CBF candidate for real-time safety filtering.

\section{Summary}
\label{sec:rntc-summary}

RNTC-MPC combines predicted SDF snapshots with a nonnegative residual to exclude the predicted terminal failure set by construction. Ablations show faster learning and higher validation IoU than direct regression and plain MSE. Across 100 Gazebo scenarios, it achieved the highest success rate at every tested horizon with NLP solution times in the millisecond range, and it completed all ten hardware trials, exceeding the best baseline by 30~percentage points at the lowest mean travel time. The results address dynamic-environment MPC (questions~\ref{rq:dynamic} and~\ref{rq:realtime}) and structural exclusion (question~\ref{rq:structure}). The next two chapters reuse residual fields in CBF safety filters for static and dynamic environments.

\part{Reachability-Informed Safety Filters}
\chapter{Neural Control Barrier Functions in Static Environments}
\label{ch:orn-cbf}

This chapter presents the \emph{observation-conditioned residual neural CBF} (ORN-CBF), based on the author's paper \cite{derajic_orn-cbf_2026} at the IEEE International Conference on Robotics and Automation (ICRA) 2026. The method transfers the hypernetwork of \cref{ch:ntc-mpc} and residual structure of \cref{ch:rntc-mpc} to a CBF-QP safety filter for unknown static environments. It addresses real-time safety filtering (question~\ref{rq:realtime}) and structural failure-set exclusion (question~\ref{rq:structure}).

\section{Motivation}
\label{sec:orn-cbf-motivation}

A safety filter supervises the nominal control command at control rate and modifies it as little as necessary to satisfy a safety condition \cite{hsu_safety_2024,wabersich_data-driven_2023}. For control-affine dynamics, this yields the small CBF-QP of \cref{eq:prelim-cbf-qp}, which can run at hundreds of hertz for the systems considered here. The effectiveness of the filter depends on the barrier function \(h\), for two reasons. First, the CBF condition \cref{eq:prelim-cbf-condition} must be satisfiable within the actuation limits, which a barrier derived from the geometric obstacle distance alone cannot generally ensure (\cref{subsec:prelim-controlled-invariant-safe-sets}). Second, an unnecessarily small certified set makes the filter intervene earlier and more often than the dynamics require, which degrades task performance. The stationary Hamilton--Jacobi (HJ) value function addresses both issues, because it accounts for the dynamics and the actuation limits and its safe set is the maximal safe set (\cref{sec:prelim-hj-cbf-relation}). Computing this value function is an offline procedure for a single fixed environment (\cref{subsec:prelim-hj-numerical}), while the robots considered here perceive the obstacle geometry only at runtime. The problem addressed in this chapter is to supply such a barrier at control rate for the locally observed environment.

\begin{figure}[tbp]
  \centering
  \input{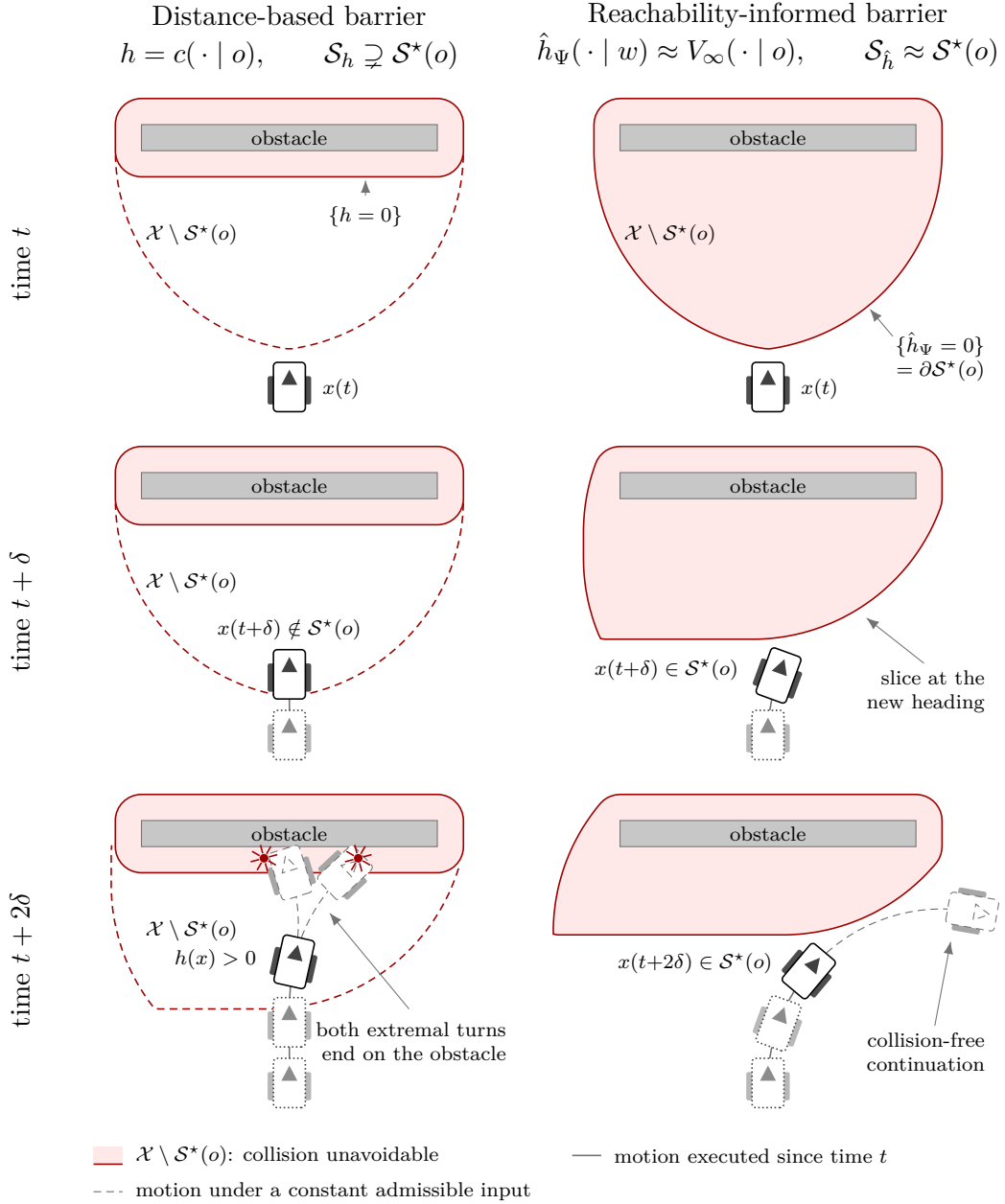}
  \caption[Effect of the barrier candidate on a CBF-QP safety filter in a static environment]{Effect of the barrier candidate on a CBF-QP safety filter at three instants separated by a short interval \(\delta\). The constant-speed Dubins car of \cref{subsec:orn-cbf-robot-models} cannot brake and is driven toward a goal behind an obstacle. The red region is the slice at the current heading of the set \(\mathcal{X}\setminus\mathcal{S}^{\star}(o)\) from which a collision can no longer be avoided, and it changes only because the robot turns. Left: the barrier is the observed obstacle distance \(c(\,\cdot\mid o)\), enforced through the second-order exponential CBF of \cref{subsec:orn-cbf-sim-qq}. The robot turns away only once the distance to the obstacle has become small. By then the state has crossed \(\partial\mathcal{S}^{\star}(o)\), and at \(t+2\delta\) both extremal turns end in a collision. Right: the zero level set of \(\hat{h}_{\Psi}(\,\cdot\mid w)\approx V_\infty(\,\cdot\mid o)\) coincides with \(\partial\mathcal{S}^{\star}(o)\), so the filter acts already at \(t\) and keeps the state inside \(\mathcal{S}^{\star}(o)\), where an admissible input always remains.}
  \label{fig:orn-cbf-intro-illustration}
\end{figure}

\Cref{fig:orn-cbf-intro-illustration} illustrates why the obstacle distance alone is insufficient. Its nonnegative region can contain states from which bounded steering or braking cannot avoid collision. Higher-order CBFs address relative degree, but their certified sets depend on the chosen functions and gains, and feasibility under actuation limits is not automatic. A barrier following the maximal safe-set boundary instead identifies when intervention is needed while recovery remains possible.

Two requirements distinguish this problem from its MPC-based counterpart in \cref{ch:ntc-mpc,ch:rntc-mpc}. First, the filter evaluates the barrier and its state gradient several hundred times per second, so the learned field must be smooth and inexpensive to differentiate, and an error in this gradient propagates directly into the applied input. Second, every new observation replaces the barrier and may shrink the predicted safe set discontinuously. The MPC methods of the preceding chapters incorporate each new observation when they replan, while a filter that runs continuously must address this discontinuity explicitly through the cross-update condition of \cref{sec:prelim-environment-conditioning}.

ORN-CBF learns a nonnegative residual from offline stationary HJ solutions and subtracts it from the observed SDF. A hypernetwork updates the compact residual network once per observation. Since the filter queries only near the current position, training uses a small central positional patch, reducing its cost by roughly an order of magnitude relative to full-window NTC-MPC training. The evaluation compares the filter with MPC planners and an exponential CBF and tests transfer to ground and aerial robots.

\section{Related Work}
\label{sec:orn-cbf-related-work}

The broader literature on safe navigation is surveyed in \cref{ch:introduction}, and the MPC-oriented literature on reachability constraints in \cref{sec:ntc-related-work,sec:rntc-related-work}. This section addresses the central design question of ORN-CBF, which is how to obtain a barrier candidate that is minimally conservative and valid for a static environment that is revealed only at runtime by onboard sensors.

Methods for enforcing a barrier function that is already available are well established, as reviewed in \cref{subsec:intro-safety-mechanisms,sec:prelim-cbf} \cite{ames_control_2019,garg_advances_2024,nguyen_exponential_2016,xiao_high-order_2022,jankovic_robust_2018,igarashi_time-varying_2019}. Constructing a valid barrier for a general nonlinear system with state and input constraints remains the central difficulty of the field \cite{hsu_safety_2024}. Learning-based methods make this construction tractable where manual design is not feasible. Examples are barriers learned from LiDAR measurements \cite{srinivasan_synthesis_2020} or expert demonstrations \cite{robey_learning_2020}, neural CBFs synthesized for systems with limited control authority \cite{liu_safe_2022}, and Gaussian-process formulations that account for uncertainty \cite{li_robust_2023}. These constructions share a structural limitation, because they do not relate the learned safe set to the largest safe set that the dynamics admit, and the learned set can consequently deviate from it in either direction. Where it is smaller than the maximal safe set, the filter discards recoverable states and intervenes more often than the dynamics require. Where it is larger, the filter admits states from which a collision is already unavoidable within the actuation limits \cite{so_how_2024,bansal_hamilton-jacobi_2017}.

Reachability-grounded constructions relate the certified set to the maximal safe set directly. Control-barrier value functions (CBVFs) modify the reachability objective so that the resulting value function encodes an admissible barrier decay rate \cite{choi_robust_2021}. Subsequent work iteratively refines hand-crafted or learned candidates to enlarge their certified sets \cite{tonkens_refining_2022}, and local reachability analysis patches potentially unsafe regions of a neural CBF to improve scalability \cite{tonkens_patching_2024}. The certified sets obtained in this way are close to maximal. All of these methods have been demonstrated only for environments that are fully known at computation time, and this assumption does not hold in the present setting.

A complementary line of work constructs barriers directly from onboard sensing. Training a neural SDF approximation of the current observation online avoids any offline distribution assumption, but the per-observation training restricts the approach to slow dynamics \cite{long_learning_2021}. Discrete-time CBFs defined over sensor readings require a one-step-ahead prediction of the observation. The demonstrated formulation uses planar laser scans, and extending its observation-update model to modalities with far more measurements, such as dense point clouds or camera images, remains a separate modeling and computational challenge \cite{dawson_learning_2022}. The work closest to this chapter uses a neural approximation of the value function of a state-dependent Riccati formulation as the barrier, computed from planar laser scans \cite{harms_neural_2024-2}. These observation-based barriers are available at runtime, but their safe sets do not approximate the maximal one, and no structural mechanism prevents the learned functions from classifying observed-occupied states as safe.

ORN-CBF combines observation-conditioned HJ approximation with the residual and hypernetwork designs of the preceding chapters \cite{derajic_learning_2025,derajic_residual_2025}, which provide failure-set exclusion and efficient control-rate queries. Positional cropping further reduces training cost by exploiting the local query region of a safety filter.

\section{Problem Statement}
\label{sec:orn-cbf-problem-statement}

This section formalizes mobile robot navigation in an unknown static environment within the safety-filtering framework of \cref{sec:prelim-cbf}. The goal is to design an observation-dependent CBF candidate whose value and gradient define the constraint of the CBF-QP.

\paragraph{Setting.}
The robot is described by the control-affine continuous-time model \cref{eq:prelim-control-affine-dynamics},
\begin{equation}
  \dot{x}
  =
  f_0(x)+g(x)u,
  \qquad
  x\in\mathcal{X},\quad
  u\in\mathcal{U},
  \label{eq:orn-cbf-dynamics}
\end{equation}
where \(f_0\) and \(g\) are locally Lipschitz and the compact set \(\mathcal{U}\) encodes the actuation limits. The full state is measured at every control instant. A nominal controller of arbitrary design supplies a command \(u_{\mathrm{nom}}(t)\) at the same rate, and its commands are not assumed to be safe. The environment contains an unknown static occupied region that the robot perceives through local observations. Unlike in \cref{ch:ntc-mpc}, observations arrive more slowly than control commands and are indexed separately, following \cref{sec:prelim-environment-conditioning}. Observation \(o_j\in\mathcal{O}\) arrives at time \(t_j\), with update period \(\Delta t_o=t_{j+1}-t_j\). It is an occupancy grid over a square sensing window of half-width \(\rho\), centered at the robot position \(p\bigl(x(t_j)\bigr)\). Let \(\mathcal{A}(o_j)\subset\mathbb{R}^{2}\) denote the occupied region identified by the observation, inflated by the robot footprint and a physical safety margin. The region \(\mathcal{A}(o_j)\) and the state \(x\) are expressed in the world frame, while the grid quantities derived from the observation are indexed in the coordinates of the sensing window. A world position is mapped to a window position by a translation that is fixed while the observation \(o_j\) is in use, and the window is not rotated, so the heading and the velocity components of the state are the same in both frames. Following \cref{sec:prelim-geometric-constraint-functions}, the observation-conditioned constraint function is
\begin{equation}
  c(x\mid o_j)
  :=
  \operatorname{sd}_{\mathcal{A}(o_j)}\bigl(p(x)\bigr),
  \label{eq:orn-cbf-observed-constraint}
\end{equation}
so that \(c(x\mid o_j)\geq0\) means a nonnegative distance to the observed physical obstacles. The corresponding observation-dependent constraint set is \(\mathcal{C}(o_j):=\{x\in\mathcal{X}\mid c(x\mid o_j)\geq0\}\). Following \cref{eq:prelim-conditioning-input}, the learned model does not receive the occupancy grid itself. Its conditioning input
\begin{equation}
  w_j
  :=
  \eta(o_j)
  \in\mathcal{W}
  \label{eq:orn-cbf-conditioning-input}
\end{equation}
is the discretized SDF, that is, the signed distances to \(\mathcal{A}(o_j)\) sampled on the cells of the sensing window. It is computed once per observation update with a linear-time distance transform \cite{maurer_linear_2003} and serves two purposes, as described in \cref{subsec:orn-cbf-architecture}. It generates the learned residual, and it represents the constraint function \(c(\,\cdot\mid o_j)\).

\paragraph{Reachability-based safety representation.}
As in \cref{ch:ntc-mpc}, safety is defined relative to the observed geometry. For an observation \(o\), let \(V_\infty(\,\cdot\mid o)\) denote the stationary infinite-horizon HJ value function obtained by solving the variational inequality \cref{eq:prelim-stationary-hji-vi} for the dynamics \cref{eq:orn-cbf-dynamics} with the constraint function \(c(\,\cdot\mid o)\). Its nonnegative superlevel set \(\mathcal{S}^{\star}(o)=\{x\mid V_\infty(x\mid o)\geq0\}\) is the maximal safe set for the observed environment (\cref{eq:prelim-hj-maximal-safe-set}). By \cref{sec:prelim-hj-cbf-relation}, the value function \(V_\infty(\,\cdot\mid o)\) is the natural barrier candidate, because its safe set is as large as the dynamics and the actuation limits permit.

\paragraph{Design problem.}
The goal is to construct, from each incoming observation \(o_j\), a barrier candidate \(\hat{h}(\,\cdot\mid w_j):\mathcal{X}\rightarrow\mathbb{R}\) whose value and state gradient can be evaluated at any queried state and that
\begin{enumerate}
  \item is available within a small fraction of the observation period \(\Delta t_o\), without solving any reachability problem online,
  \item approximates the reachability-based value function \(V_\infty(\,\cdot\mid o_j)\), and hence the maximal safe set, rather than the geometric obstacle distance alone,
  \item satisfies the \emph{failure-set exclusion} property by construction,
        \begin{equation}
          \left\{x\mid \hat{h}(x\mid w_j)\geq0\right\}
          \subseteq
          \mathcal{C}(o_j),
          \label{eq:orn-cbf-exclusion-requirement}
        \end{equation}
        so that no observed-occupied state is ever certified as safe, regardless of the approximation error, and
  \item can be evaluated and differentiated at the current state fast enough that the CBF-QP \cref{eq:prelim-cbf-qp} runs at the control rate.
\end{enumerate}
The third requirement transfers the exclusion property of \cref{ch:rntc-mpc} to the set that the filter seeks to preserve, since invariance alone would not ensure physical safety if that set contained occupied states.

\paragraph{Standing assumptions.}
The setting rests on four assumptions. First, the environment is static, and navigation among moving obstacles is treated in \cref{ch:cn-cbf} with the same safety-filter interface. Second, the dynamics model \cref{eq:orn-cbf-dynamics} is accurate and the full state is available, so disturbances and estimation errors are not modeled explicitly. Third, new information enters an updated observation only near the boundary of the sensing window, for example when previously distant geometry enters the sensing range. This assumption excludes obstacles that appear close to the robot when an occluding object moves away. Fourth, the sensing window, the update period, and the robot speed jointly satisfy the cross-update safety condition of \cref{subsec:orn-cbf-observation-conditioning}, which requires the robot to remain outside the backward reachable tube (BRT) of any newly revealed obstacle at the instant at which the barrier is replaced.

\section{Observation-Conditioned Residual Neural Barriers}
\label{sec:orn-cbf-method}

The method first separates fixed-observation filtering from safety across updates. It then introduces the residual parameterization, the hypernetwork and its training on cropped data, and the resulting CBF-QP.

\subsection{Observation-Conditioned Barrier Formulation}
\label{subsec:orn-cbf-observation-conditioning}

The first design question is how an observation should enter the barrier function. Following \cref{subsec:prelim-observation-dependent-fields}, the barrier is in general a joint function \(h(x,o)\) of state and observation. Differentiating it along the closed loop converts the CBF condition \cref{eq:prelim-observation-cbf-condition} into the following requirement on the applied input,
\begin{equation}
  \nabla_xh(x,o)^{\top}\dot{x}
  +
  \nabla_oh(x,o)^{\top}\dot{o}
  \geq
  -\alpha\bigl(h(x,o)\bigr),
  \label{eq:orn-cbf-joint-condition}
\end{equation}
which requires the \emph{observation dynamics} \(\dot{o}\). The environment is assumed to be static, so the observation depends on time only through the robot state and \(\dot{o}=(\partial o/\partial x)\,\dot{x}\), as in \cref{eq:prelim-observation-derivative-static}. As discussed in \cref{subsec:prelim-observation-time-derivative}, the Jacobian \(\partial o/\partial x\) can be modeled only in part. A sensor model predicts how obstacles that the robot has already seen move within the sensing window as the robot advances \cite{dawson_learning_2022}, but it cannot predict the appearance of unseen obstacles. A model of \(\dot{o}\) is thus incomplete for the newly revealed geometry, which is the geometry that determines safety.

ORN-CBF instead holds the world-frame field \(h(\,\cdot\mid o_j)\) fixed between observation updates, as in \cref{eq:prelim-observation-conditioned-function}. In a static environment, this field continues to describe the same observed geometry during that interval, and its derivative contains only the state term. This follows from holding the field fixed rather than from neglecting part of the derivative of \(h(x,o(x))\). New observations replace the field at discrete instants, whose safety conditions are considered below. Between updates, the applied input must satisfy
\begin{equation}
  \nabla_xh(x\mid o_j)^{\top}\dot{x}
  \geq
  -\alpha\bigl(h(x\mid o_j)\bigr),
  \qquad
  t\in[t_j,t_{j+1}),
  \label{eq:orn-cbf-conditioned-condition}
\end{equation}
with an extended class-\(\mathcal{K}\) function \(\alpha\) as in \cref{sec:prelim-cbf}. Both \cref{eq:orn-cbf-joint-condition,eq:orn-cbf-conditioned-condition} are conditions on the input that the filter applies, and the corresponding existence condition is obtained by taking the supremum over \(u\in\mathcal{U}\) as in \cref{eq:prelim-cbf-condition}. This choice has a second practical implication, since for each fixed observation the barrier is time-invariant in the state, so that the training targets can be generated with the standard stationary HJ methods of \cref{subsec:prelim-hj-numerical}, without any model of the observation dynamics.

\begin{figure}[tb]
  \centering
  \colorlet{brtline}{red!62!black}
\colorlet{brtnow}{red!14}
\colorlet{brtpastfill}{red!5}
\colorlet{obsfill}{black!22}
\colorlet{obsline}{black!55}
\colorlet{winline}{black!70}
\colorlet{gridline}{black!14}

\def\brtoutline{%
  (-0.053,1.433) (-0.129,1.443) (-0.217,1.460) (-0.314,1.483) (-0.416,1.512) (-0.522,1.548)
  (-0.630,1.589) (-0.739,1.636) (-0.848,1.688) (-0.955,1.744) (-1.059,1.805) (-1.160,1.869)
  (-1.256,1.937) (-1.347,2.008) (-1.432,2.080) (-1.510,2.154) (-1.580,2.229) (-1.642,2.304)
  (-1.695,2.377) (-1.740,2.449) (-1.775,2.518) (-1.800,2.582) (-1.815,2.639) (-1.820,2.680)
  (-1.820,3.580) (-1.815,3.643) (-1.798,3.703) (-1.772,3.760) (-1.736,3.811) (-1.691,3.856)
  (-1.640,3.892) (-1.583,3.918) (-1.523,3.935) (-1.460,3.940) (1.460,3.940) (1.523,3.935)
  (1.583,3.918) (1.640,3.892) (1.691,3.856) (1.736,3.811) (1.772,3.760) (1.798,3.703)
  (1.815,3.643) (1.820,3.580) (1.820,2.680) (1.815,2.639) (1.800,2.582) (1.775,2.518)
  (1.740,2.449) (1.695,2.377) (1.642,2.304) (1.580,2.229) (1.510,2.154) (1.432,2.080)
  (1.347,2.008) (1.256,1.937) (1.160,1.869) (1.059,1.805) (0.955,1.744) (0.848,1.688)
  (0.739,1.636) (0.630,1.589) (0.522,1.548) (0.416,1.512) (0.314,1.483) (0.217,1.460)
  (0.129,1.443) (0.053,1.433) (0.000,1.430)}
\def\brtregion#1{\draw[#1] plot coordinates {\brtoutline} -- cycle;}
\def\obstacle#1{\draw[#1] (-1.46,2.68) rectangle (1.46,3.58);}

\begin{tikzpicture}[
  font=\small, >={Latex[length=1.8mm]},
  lab/.style={font=\scriptsize, align=center, inner sep=1.5pt},
  ttl/.style={font=\small, align=center},
  dimlab/.style={font=\scriptsize, fill=white, inner sep=1.5pt},
  dim/.style={<->, thin, >={Latex[length=1.7mm]}},
  ext/.style={black!45, thin, densely dotted},
  win/.style={draw=winline, semithick},
  gridlines/.style={draw=gridline, line width=0.3pt, step=0.23},
  brtfilled/.style={fill=brtnow, draw=none},
  brtedge/.style={draw=brtline, semithick},
  brtfilledpast/.style={fill=brtpastfill, draw=none},
  brtedgepast/.style={draw=brtline!45, thin, densely dashed},
  obs/.style={fill=obsfill, draw=obsline, thin},
  obspast/.style={fill=black!8, draw=black!35, thin, densely dashed},
  robot/.pic={
    \fill[black!70, rounded corners=0.8pt] (-0.24,0.19) rectangle (0.15,0.30);
    \fill[black!70, rounded corners=0.8pt] (-0.24,-0.30) rectangle (0.15,-0.19);
    \draw[semithick, fill=white, rounded corners=1.5pt] (-0.34,-0.22) rectangle (0.34,0.22);
    \fill[black!75] (0.24,0) -- (0.03,0.11) -- (0.03,-0.11) -- cycle;
  },
  pastrobot/.pic={
    \fill[black!28, rounded corners=0.8pt] (-0.24,0.19) rectangle (0.15,0.30);
    \fill[black!28, rounded corners=0.8pt] (-0.24,-0.30) rectangle (0.15,-0.19);
    \draw[densely dotted, black!75, semithick, fill=white, rounded corners=1.5pt]
      (-0.34,-0.22) rectangle (0.34,0.22);
    \fill[black!45] (0.24,0) -- (0.03,0.11) -- (0.03,-0.11) -- cycle;
  },
  ]

  \begin{scope}
    \node[ttl] at (0,4.45) {(a)\; $t=t_j$};

    \draw[gridlines] (-2.53,-2.53) grid (2.53,2.53);

    \brtregion{brtfilledpast}
    \obstacle{obspast}
    \brtregion{brtedgepast}
    \draw[win] (-2.53,-2.53) rectangle (2.53,2.53);
    \node[lab, anchor=south west, text=winline] at (-2.41,-2.41) {$o_j$};
    \node[lab, text=black!55] at (0,3.00) {$\mathcal{F}$};
    \node[lab, text=brtline!55] at (0,2.05) {$\mathcal{B}_{\infty}(\mathcal{F})$};

    \pic[rotate=90] at (0,0) {robot};
    \node[lab, anchor=east] at (-0.42,0) {$x(t_j)$};
    \draw[ext] (0,-0.42) -- (0,-1.87);
    \draw[dim] (0,-1.60) -- (2.53,-1.60) node[midway, dimlab] {$\rho$};
  \end{scope}

  \begin{scope}[shift={(7.90,0)}]
    \node[ttl] at (0,4.45) {(b)\; $t=t_{j+1}=t_j+\Delta t_o$};

    \draw[gridlines] (-2.53,-1.75) grid (2.53,3.31);
    \draw[black!45, thin, densely dashed] (-2.53,2.53) -- (2.53,2.53);

    \begin{scope}
      \clip (-2.60,3.31) rectangle (2.60,4.20);
      \brtregion{brtfilledpast}
      \obstacle{obspast}
      \brtregion{brtedgepast}
    \end{scope}
    \begin{scope}
      \clip (-2.53,-1.75) rectangle (2.53,3.31);
      \brtregion{brtfilled}
      \obstacle{obs}
      \brtregion{brtedge}
    \end{scope}
    \draw[win] (-2.53,-1.75) rectangle (2.53,3.31);
    \node[lab, anchor=south west, text=winline] at (-2.41,-1.63) {$o_{j+1}$};
    \node[lab] at (0,3.00) {$\mathcal{F}$};
    \node[lab, text=brtline] at (0,2.05) {$\mathcal{B}_{\infty}(\mathcal{F})$};

    \draw[ext] (-2.53,3.31) -- (-2.88,3.31);
    \draw[ext] (-2.53,2.53) -- (-2.88,2.53);
    \draw[dim] (-2.74,2.53) -- (-2.74,3.31);
    \node[lab, anchor=east] at (-2.92,2.92) {$v_{\max}\Delta t_o$};
    \draw[dim] (-2.74,0.78) -- (-2.74,2.53);
    \node[lab, anchor=east] at (-2.92,1.655) {$\rho-v_{\max}\Delta t_o$};
    \draw[ext] (1.90,2.68) -- (2.30,2.68);
    \draw[ext] (0.16,1.43) -- (2.30,1.43);
    \draw[dim] (2.15,1.43) -- (2.15,2.68) node[midway, dimlab] {$\beta_{\mathcal{B}}$};

    \pic[rotate=90] at (0,0) {pastrobot};
    \node[lab, anchor=west, text=black!55] at (0.42,0) {$x(t_j)$};
    \pic[rotate=90] at (0,0.78) {robot};
    \node[lab, anchor=west] at (0.42,0.78) {$x(t_{j+1})$};
  \end{scope}

  \draw[ext] (0.42,0) -- (7.48,0);
  \draw[ext] (7.48,0.78) -- (4.62,0.78);
  \draw[dim] (4.75,0) -- (4.75,0.78);
  \node[lab, anchor=east] at (4.61,0.39) {$v_{\max}\Delta t_o$};
\end{tikzpicture}
  \caption[The critical moment of an observation update]{The critical moment of an observation update. Geometry that the current observation does not contain is drawn faded and with dashed boundaries. (a)~At \(t_j\), a previously unobserved obstacle lies beyond the boundary of the sensing window of half-width \(\rho\). Neither the failure set \(\mathcal{F}\) (grey) nor its BRT \(\mathcal{B}_{\infty}(\mathcal{F})\) (red) is part of the observation \(o_j\). (b)~Between two updates the robot travels at most \(v_{\max}\Delta t_o\), and the window is recentered at the new robot position. The new observation thus gains a band of that depth beyond its previous boundary (dashed) and now contains most of the obstacle. Under the boundary-appearance assumption, previously unseen obstacles can appear only inside this band, and hence at least \(\rho-v_{\max}\Delta t_o\) away from the robot. Continued safety requires the state to lie outside \(\mathcal{B}_{\infty}(\mathcal{F})\) at the instant at which the barrier is replaced.}
  \label{fig:orn-cbf-new-obs}
\end{figure}
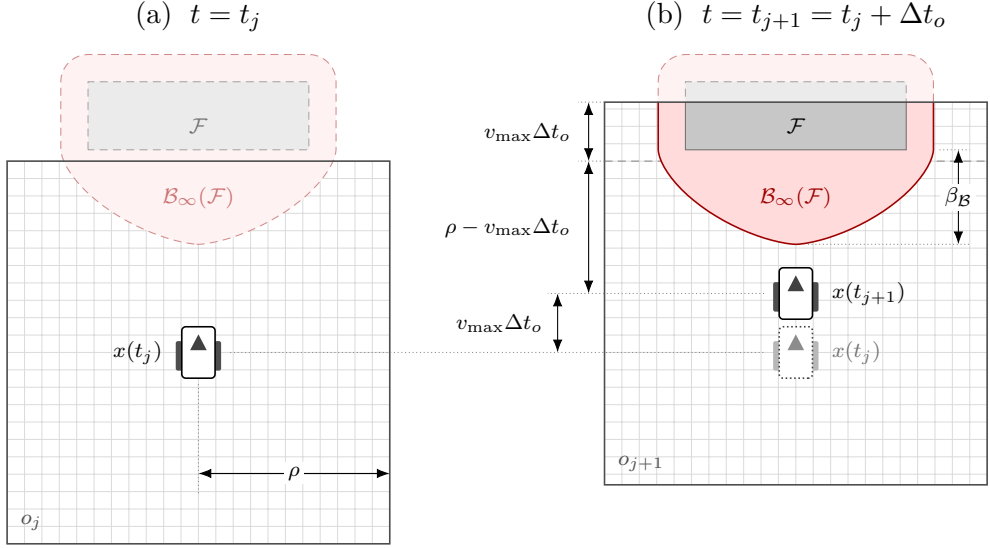

\paragraph{Cross-update safety.}
Conditioning does not remove the discontinuity at an update instant. Replacing \(o_j\) by \(o_{j+1}\) changes both the barrier and its predicted safe set in one step. As formalized in \cref{eq:prelim-cross-update-safety}, continued safety requires the state at the update instant to lie in the safe set of the \emph{new} barrier,
\begin{equation}
  x(t_{j+1})
  \in
  \left\{x\mid h(x\mid o_{j+1})\geq0\right\},
  \label{eq:orn-cbf-cross-update}
\end{equation}
because a CBF inequality enforced under the previous observation carries no information about obstacles that this observation did not contain. \Cref{fig:orn-cbf-new-obs} illustrates the worst case. Under the standing assumptions, the environment is static, previously unseen obstacles appear only at the window boundary, and the robot speed is bounded by \(v_{\max}\). The robot travels at most \(v_{\max}\Delta t_o\) between two updates. Since the sensing window is recentered at the robot's position, previously unseen obstacles can enter the next observation only within a band of the same depth along the window boundary, so a newly revealed failure set is separated from the robot by at least \(\rho-v_{\max}\Delta t_o\). Let \(\beta_{\mathcal{B}}\) denote the largest distance, measured in the workspace, by which the BRT of an admissible obstacle extends beyond the failure set of that obstacle. Consider first the case in which the barrier is the exact conditioned value function \(V_\infty(\,\cdot\mid o_{j+1})\), whose zero level set is the boundary of the BRT of the entire geometry represented by \(o_{j+1}\). A sufficient condition for the robot to remain outside the BRT of every newly revealed obstacle is then
\begin{equation}
  \rho-v_{\max}\Delta t_o
  >
  \beta_{\mathcal{B}}.
  \label{eq:orn-cbf-sensing-adequacy}
\end{equation}
The margin \(\beta_{\mathcal{B}}\) depends on the robot dynamics, the actuation limits, and the admissible obstacle shapes. For the double integrator of the braking example in \cref{subsec:prelim-controlled-invariant-safe-sets}, it is of the order of the braking distance \(v_{\max}^{2}/(2\bar{u})\). For a vehicle that cannot stop, it is determined by the minimum turning geometry. Condition \cref{eq:orn-cbf-sensing-adequacy} is the concrete form of the sensing-adequacy requirement \cref{eq:prelim-sensing-adequacy} for the present configuration. It removes the geometric mechanism by which an update can reveal an obstacle too close to be evaded. Remaining outside each of these tubes individually does not by itself place the state in \(\mathcal{S}^{\star}(o_{j+1})\), because avoidability of each newly revealed obstacle separately does not imply avoidability of their union, as \cref{ch:cn-cbf} discusses for the analogous pairwise composition. The condition also does not guarantee \cref{eq:orn-cbf-cross-update} for the \emph{learned} barrier, whose approximation error may displace the predicted boundary. That condition must still be verified at the update instants or protected by an additional margin on the learned barrier.

\subsection{Residual Barrier Parameterization}
\label{subsec:orn-cbf-residual}

The target is the stationary HJ value \(V_\infty(\,\cdot\mid o)\), whose nonnegative region is the maximal safe set. ORN-CBF approximates this field with the residual structure of \cref{ch:rntc-mpc}. The ordering \(V_\infty(x\mid o)\leq c(x\mid o)\) permits the observed SDF to supply geometry directly while a network learns the nonnegative difference,
\begin{equation}
  \hat{h}_{\Psi}(x\mid w_j)
  :=
  c(x\mid o_j)
  -
  \hat{r}_{\Psi}(x\mid w_j),
  \qquad
  \hat{r}_{\Psi}(x\mid w_j)\geq0,
  \label{eq:orn-cbf-residual-parameterization}
\end{equation}
where \(c(\,\cdot\mid o_j)\) is computed from the observation and only \(\hat{r}_{\Psi}\) is learned.

\paragraph{Failure-set exclusion guarantee.}
The requirement \cref{eq:orn-cbf-exclusion-requirement} follows from the nonnegativity of the residual, exactly as in \cref{ch:rntc-mpc}. The learned barrier satisfies \(\hat{h}_{\Psi}(x\mid w_j)\leq c(x\mid o_j)\), so the predicted safe set \(\mathcal{S}_{\hat{h}}(w_j):=\{x\mid\hat{h}_{\Psi}(x\mid w_j)\geq0\}\) satisfies
\begin{equation}
  \mathcal{S}_{\hat{h}}(w_j)
  \subseteq
  \mathcal{C}(o_j).
  \label{eq:orn-cbf-exclusion}
\end{equation}
Every state in the predicted safe set has a nonnegative distance to the observed obstacles. This inclusion does not establish invariance or cover unobserved geometry, and it concerns the set used for current-state filtering, while \cref{ch:rntc-mpc} imposed it at the predicted terminal state.

\paragraph{Smooth residual output.}
The output activation enforces the nonnegativity of the residual. ORN-CBF uses the softplus function, a smooth relaxation of the ReLU,
\begin{equation}
  \operatorname{softplus}(z)
  =
  \log\bigl(1+\exp(z)\bigr).
  \label{eq:orn-cbf-softplus}
\end{equation}
Its derivative is positive for every finite input, so no part of the output can become permanently inactive. The shifted exponential linear unit of \cref{eq:rntc-elu} shares this property, and softplus is used here because it has a single closed-form expression. The hidden layers use sinusoidal activations, which suit implicit field regression when the accuracy of the gradient matters \cite{sitzmann_implicit_2020,bansal_deepreach_2021}. The constraint function \(c(\,\cdot\mid o_j)\) is available as the discretized SDF, and a numerical interpolant converts these samples into a function that can be evaluated and differentiated at any queried state. This interpolation step limits how smooth the assembled barrier can be, because a bilinear interpolant is only piecewise differentiable and its gradient is discontinuous across cell boundaries. A smooth residual network alone does not make the assembled barrier globally continuously differentiable, and \cref{sec:orn-cbf-discussion} discusses this limitation.

\subsection{Network Architecture and Training}
\label{subsec:orn-cbf-architecture}
\label{subsec:orn-cbf-training}

The filter queries the barrier at least an order of magnitude more often than observations arrive in the evaluated configurations, so one hypernetwork evaluation per observation serves many control updates, and ORN-CBF adopts the hypernetwork factorization of \cref{subsec:ntc-architecture} for this reason. A convolutional hypernetwork \(H_{\Psi}\) with a fully connected head processes the conditioning input \(w_j\) of \cref{eq:orn-cbf-conditioning-input} once per observation update and generates the parameters of a compact main network, a multilayer perceptron (MLP) with sinusoidal hidden activations and the softplus output \cref{eq:orn-cbf-softplus}. The construction yields the pair of mappings
\begin{equation}
  \Theta(w_j)
  =
  H_{\Psi}(w_j),
  \qquad
  \hat{r}_{\Psi}(x\mid w_j)
  =
  \operatorname{softplus}\bigl(M_{\Theta(w_j)}(x)\bigr).
  \label{eq:orn-cbf-hypernetwork}
\end{equation}
The main network receives the state in the window coordinates introduced in \cref{sec:orn-cbf-problem-statement}. Its positional inputs are the robot position relative to the window center \(p\bigl(x(t_j)\bigr)\), and its remaining inputs, such as the heading or the velocity, are unchanged. This matches the training data described below, which are generated on a grid attached to the window. For a fixed observation, the map from the world state to the network input is a constant translation, so it leaves the state gradient \(\nabla_x\hat{h}_{\Psi}\) and the Lie derivatives unchanged, and the barrier can be written as a function of the world state without any further transformation.

\begin{figure}[tb]
  \centering
  \ifdefined\ornarchunit\else\newlength{\ornarchunit}\fi
\setlength{\ornarchunit}{0.68pt}%
\newcommand{\ornarchzoom}{1.08}%

\newcommand{\ornlwa}{0.68pt}
\newcommand{\ornlwb}{0.54pt}
\newcommand{\ornlwc}{0.34pt}

\newcommand{\ornTviii}{\fontsize{5.44}{6.53}\selectfont}
\newcommand{\ornTix}{\fontsize{6.12}{7.34}\selectfont}
\newcommand{\ornTx}{\fontsize{6.80}{8.16}\selectfont}
\newcommand{\ornTxi}{\fontsize{7.48}{8.98}\selectfont}
\newcommand{\ornTxii}{\fontsize{8.16}{9.79}\selectfont}
\newcommand{\ornMix}{\fontsize{6.49}{7.79}\selectfont}
\newcommand{\ornMx}{\fontsize{7.21}{8.65}\selectfont}
\newcommand{\ornMxi}{\fontsize{7.93}{9.52}\selectfont}
\newcommand{\ornMxii}{\fontsize{8.65}{10.38}\selectfont}
\newcommand{\ornMixs}{\fontsize{6.49}{6.90}\selectfont}

\definecolor{ornfiltfill}{HTML}{D5E8D4}%
\definecolor{ornfiltline}{HTML}{82B366}%
\definecolor{ornnetfill}{HTML}{DAE8FC}%
\definecolor{ornnetline}{HTML}{6C8EBF}%
\definecolor{ornhypfill}{HTML}{E1D5E7}%
\definecolor{ornhypline}{HTML}{9673A6}%
\definecolor{ornnode}{HTML}{1A5FB4}%
\definecolor{ornquery}{HTML}{FF0000}%
\definecolor{ornlvlA}{HTML}{E5A50A}%
\definecolor{ornlvlB}{HTML}{FF7800}%
\definecolor{ornlvlC}{HTML}{ED333B}%
\definecolor{ornlvlD}{HTML}{A51D2D}%

\begin{tikzpicture}[
    x=\ornarchunit, y=-\ornarchunit,
    line width=\ornlwa, line cap=butt, line join=miter,
    arw/.style={-{Stealth[length=4.30pt, width=4.76pt, inset=0.90pt]}},
    arws/.style={-{Stealth[length=3.44pt, width=3.81pt, inset=0.72pt]}},
    box/.style={draw, rounded corners=#1\ornarchunit, inner sep=0pt, outer sep=0pt},
    dsh/.style={dash pattern=on 3\ornarchunit off 3\ornarchunit},
    dashbox/.style={box=27, dsh},
    txt/.style={inner sep=0pt, outer sep=0pt, align=center},
    dot/.style={circle, fill, inner sep=0pt, outer sep=0pt,
                minimum size=5\ornarchunit},
    zoom/.style={shift={({(1-\ornarchzoom)*160},{(1-\ornarchzoom)*90})},
                 scale=\ornarchzoom, transform shape},
  ]

  \useasboundingbox (-27,-8) rectangle (646.1,188.5);

  \draw[box=6, fill=ornfiltfill, draw=ornfiltline] (13,72) rectangle (93,112);
  \node[txt, font=\ornTxi] at (53,92) {ORN-CBF\\safety filter};

  \draw[arw] (110,45) -- (53,45) -- (53,72);
  \draw[arw] (110,135) -- (53,135) -- (53,112);
  \draw[arw] (-22,92) -- (13,92);
  \draw[arw] (93,92) -- (128,92);

  \node[txt, font=\ornMxii] at (81,37)     {$o_j$};
  \node[txt, font=\ornTix]  at (85,52.5)   {Observation};
  \node[txt, font=\ornMxii] at (84,127.5)  {$x$};
  \node[txt, font=\ornTix]  at (82.5,142.5){State feedback};
  \node[txt, font=\ornMxii] at (-10,83.0)  {$u_{\mathrm{nom}}$};
  \node[txt, font=\ornTix]  at (-10.5,107)  {Nominal\\control};
  \node[txt, font=\ornMxii] at (111.4,83.0){$u^{\star}$};
  \node[txt, font=\ornTix]  at (120,107.0) {Safe\\control};

  \draw[dsh] (102,70) -- (160,{90-45*\ornarchzoom});

  \begin{scope}[zoom]
  \draw[dashbox] (160,0) rectangle (608.25,180);
  \node[txt, font=\ornTxii\bfseries] at (563.25,25) {ORN-CBF};

  \draw[box=5.25] (184,17.5) rectangle (243,52.5);
  \node[txt, font=\ornTix] at (213.5,35) {SDF\\Comput.};
  \draw[arw] (167,35) -- (184,35);
  \node[txt, font=\ornMxii] at (170.7,27) {$o_j$};

  \draw (243,35) -- (251.25,35);
  \node[dot] (sdftap) at (253.75,35) {};
  \node[txt, font=\ornMx] at (251.5,26) {$w_j$};
  \draw[arw] (256.25,35) -- (270.02,35);
  \draw[arw] (253.75,37.5) -- (253.75,150) -- (343.75,150);

  \draw[box=7.5] (261,10) rectangle (336.01,60);
  \draw[fill=ornhypfill, draw=ornhypline]
    (270.02,15.11) -- (309.34,25.0) -- (309.34,45.0) -- (270.02,54.89) -- cycle;
  \node[txt, font=\ornTx] at (289.68,35) {CNN};
  \draw[arws] (309.34,35) -- (319.75,35);
  \draw[box=0.7, fill=ornhypfill, draw=ornhypline] (319.75,17.49) rectangle (329.75,52.5);
  \foreach \yy in {21, 28, 34.99, 42, 49}
    \draw[fill=ornnetfill, draw=ornnetline] (324.75,\yy) circle[radius=2];
  \node[txt, font=\ornTviii] at (298.5,66.0) {Hypernetwork};

  \draw[arws] (329.75,35) -- (372.72,35) -- (372.72,60);
  \node[txt, font=\ornMix] at (352.1,29) {$\Theta$};

  \node[dot] (xtap) at (320.75,130) {};
  \draw (303.25,130) -- (320.75,130);
  \node[txt, font=\ornMxii] at (306.7,121) {$x$};
  \draw[arw] (320.75,127.5) -- (320.75,80.1) -- (343.25,80.1);
  \draw[arw] (323.25,130) -- (343.75,130);

  \draw[box=6.055] (343.25,60) rectangle (402.197,100.368);
  \draw[line width=\ornlwc]
    (369.19,66.962) -- (358.783,72.734)  (369.19,79.949) -- (358.783,72.734)
    (369.19,92.936) -- (358.783,72.734)  (369.19,66.962) -- (358.783,87.164)
    (369.19,79.949) -- (358.783,87.164)  (369.19,92.936) -- (358.783,87.164)
    (386.298,79.949) -- (377.316,66.962) (386.298,79.949) -- (377.316,79.949)
    (386.298,79.949) -- (377.316,92.936);
  \foreach \nx/\ny in {354.72/72.734, 354.72/87.164,
                       373.25/66.962, 373.25/79.949, 373.25/92.936,
                       390.36/79.949}
    \draw[fill=ornnetfill, draw=ornnetline] (\nx,\ny) circle[radius=4.063];
  \node[txt, font=\ornTviii] at (372.5,106.5) {Main network};

  \draw[box=6] (343.75,120) rectangle (401.25,160);
  \draw[line width=\ornlwc, dash pattern=on 1.5\ornarchunit off 1.5\ornarchunit]
    (359.38,122.5) -- (359.38,157.5) (384.38,122.5) -- (384.38,157.5)
    (355,126.93) -- (390,126.93)     (355,151.93) -- (390,151.93);
  \draw (362,129.5) -- (374,135.5)  (362,149.5) -- (374,140.5)
        (379,135.5) -- (382,129.5)  (379,140.5) -- (382,149.5);
  \foreach \nx/\ny in {359.5/127, 384.5/127, 359.5/152, 384.5/152}
    \fill[ornnode] (\nx,\ny) circle[radius=2.5];
  \fill[ornquery] (376.5,138) circle[radius=2.5];
  \node[txt, font=\ornTviii] at (372.5,166.5) {SDF interpolation};

  \draw[arw] (402.197,80.184) -- (458.25,80);
  \node[txt, font=\ornMixs] at (430,70.5) {$\hat{r}_{\Psi}$};
  \draw[arw] (401.25,140) -- (464.25,140) -- (464.25,86);
  \node[txt, font=\ornMix] at (428.4,133) {$c$};
  \draw[fill=white] (464.25,80) circle[radius=6];
  \node[txt, font=\ornTx] at (447.75,85) {$-$};
  \node[txt, font=\ornTx] at (456.25,93) {$+$};

  \draw[box=5.25] (423.25,20) rectangle (483.25,55);
  \node[txt, font=\ornTx] at (453.25,37.5) {System\\dynamics};
  \draw[arw] (403.25,37.556) -- (423.25,37.5);
  \node[txt, font=\ornMxii] at (407.1,28.5) {$x$};
  \draw[arw] (483.25,37.5) -- (512.25,37.5) -- (512.25,74);
  \node[txt, font=\ornMix] at (502.5,28.3) {$f_{0}, g$};

  \draw[arw] (470.25,80) -- (506.25,80);
  \node[txt, font=\ornMixs] at (488.25,70.5) {$\hat{h}_{\Psi}$};
  \draw[fill=white] (512.25,80) circle[radius=6];
  \fill (512.25,80) circle[radius=1.5];

  \draw[arw] (518.25,80) -- (545.75,80) -- (545.75,100);
  \node[txt, font=\ornMx] at (565,68)
    {$\hat{h}_{\Psi}$\\$L_{f_{0}}\hat{h}_{\Psi}$\\$L_{g}\hat{h}_{\Psi}$};

  \draw[box=6] (513.25,100) rectangle (578.25,140);
  \foreach \cx/\cy/\rx/\ry/\col in {%
      545.479/121.854/25.744/13.957/ornlvlA,
      545.747/121.679/18.235/9.595/ornlvlB,
      545.532/121.854/11.799/6.979/ornlvlC,
      545.318/121.679/5.363/2.617/ornlvlD}
    \draw[line width=\ornlwb, draw=\col, rotate around={10:(\cx,\cy)},
          dash pattern=on 2.4\ornarchunit off 2.4\ornarchunit]
      (\cx,\cy) ellipse[x radius=\rx, y radius=\ry];
  \draw[line width=\ornlwb]
    (541.274,104.756) -- (528.313,104.756) -- (519.732,116.969) --
    (532.604,127.437) -- (551.912,116.969) -- (541.041,104.582);
  \node[txt, font=\ornTviii] at (545.75,145.5) {CBF-QP};

  \draw[arw] (489.855,119.9) -- (513.25,120);
  \node[txt, font=\ornMxi] at (494.8,111.8) {$u_{\mathrm{nom}}$};
  \draw[arw] (578.25,120) -- (603.25,120);
  \node[txt, font=\ornMxii] at (589.85,110) {$u^{\star}$};
  \end{scope}
\end{tikzpicture}
  \caption[Architecture of the ORN-CBF safety filter]{Architecture of the ORN-CBF safety filter. The discretized SDF \(w_j\) computed from the observation \(o_j\) is used twice, as the conditioning input of the hypernetwork, which generates the parameters \(\Theta\) of the main network representing the nonnegative residual \(\hat{r}_{\Psi}\), and as the input of the interpolant representing the observed constraint function \(c(\,\cdot\mid o_j)\). Barrier value and gradient are assembled by subtraction as in \cref{eq:orn-cbf-residual-parameterization}, and the Lie derivatives \(L_{f_0}\hat{h}_{\Psi}\) and \(L_g\hat{h}_{\Psi}\) parameterize the CBF-QP \cref{eq:orn-cbf-qp} together with \(\hat{h}_{\Psi}\). The softplus output of the main network is not drawn as a separate stage. The hypernetwork runs once per observation update, and the main network and the interpolant are queried at control rate.}
  \label{fig:orn-cbf-architecture}
\end{figure}
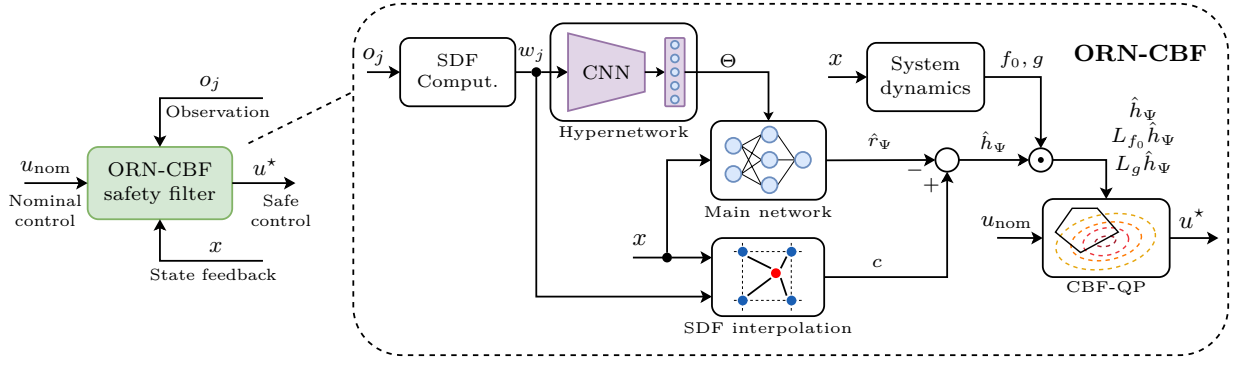

The same conditioning input also enters the interpolant that represents the constraint function \(c(\,\cdot\mid o_j)\), and the barrier is assembled by subtraction as in \cref{eq:orn-cbf-residual-parameterization}. At points where the interpolant is differentiable, automatic differentiation returns the gradient \(\nabla_x\hat{h}_{\Psi}(x\mid w_j)\) of the represented function, from which the Lie derivatives \(L_{f_0}\hat{h}_{\Psi}\) and \(L_g\hat{h}_{\Psi}\) follow by \cref{eq:prelim-lie-derivatives}. \Cref{fig:orn-cbf-architecture} shows the complete pipeline from observation to safety filter.

The computational benefit of this factorization, measured for the MPC interface in \cref{subsec:ntc-single-network}, applies equally here, because the filter also queries the field many times per conditioning input.

\paragraph{Reachability supervision and positional cropping.}
The hypernetwork is trained offline by supervised learning as in \cref{subsec:ntc-training}. Local observations are collected in simulation and converted into the conditioning inputs \(w^{(m)}\) of \cref{eq:orn-cbf-conditioning-input}, and for each observation a grid-based HJ solver propagates the stationary variational inequality to convergence, using the observed SDF as the constraint function, which yields the target \(V_\infty(\,\cdot\mid o^{(m)})\) over a state-space grid. The training objective is the radially weighted mean squared error (RWMSE) of \cref{subsec:ntc-rwmse}, whose weighting concentrates model capacity near the safe-set boundary, where the sign of the barrier determines the filter behavior. As in \cref{subsec:rntc-training}, the loss is applied to the assembled barrier estimate \cref{eq:orn-cbf-residual-parameterization} rather than to the residual in isolation.

The new element of the training procedure is positional cropping. It exploits a difference between the two controller interfaces, which query their learned fields over regions that differ widely in size. An MPC terminal constraint is evaluated at \emph{predicted} states up to \(N\Delta t\,v_{\max}\) ahead of the robot, so its field must be accurate across most of the sensing window. A safety filter evaluates the barrier only at the \emph{current} state. Since the observation is recentered at the robot position at every update, the current state remains within a ball of radius \(v_{\max}\Delta t_o\) around the window center while that observation is in use, which is the same bound as in the cross-update analysis of \cref{subsec:orn-cbf-observation-conditioning}. The learned field then has to represent the value function only on a small central patch of the positional coordinates, while the remaining state dimensions, such as heading and velocity, must be covered completely. The HJ targets are still computed on the full grid, because the boundary conditions require the complete window, but only the central positional patch is stored and used for training.

Positional cropping reduces the size of the stored dataset and the per-epoch evaluation cost by the ratio of the patch area to the window area. \Cref{subsec:orn-cbf-training-eval} quantifies the resulting reduction relative to the full-window procedure of \cref{ch:ntc-mpc}. Compared with the MPC interface, the safety-filter interface thus demands a higher query rate and greater smoothness, but less spatial coverage, and positional cropping exploits exactly this difference.

\subsection{ORN-CBF Safety Filter}
\label{subsec:orn-cbf-filter}

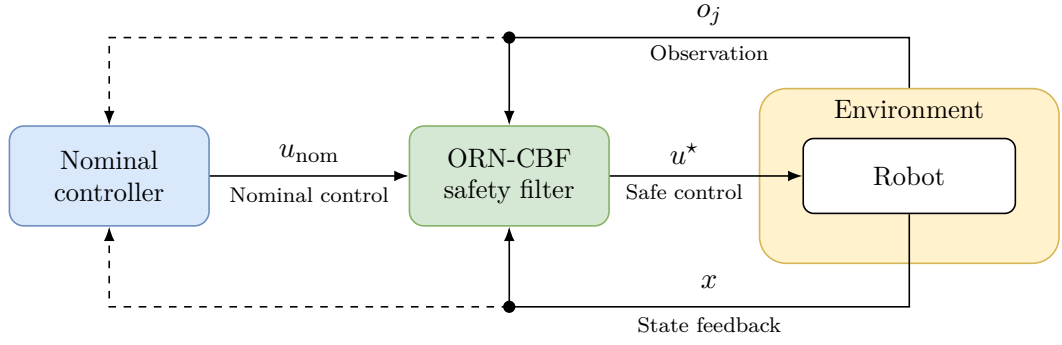
\begin{figure}[tb]
  \centering
  \definecolor{sfactrlfill}{HTML}{DAE8FC}%
\definecolor{sfactrlline}{HTML}{6C8EBF}%
\definecolor{sfafiltfill}{HTML}{D5E8D4}%
\definecolor{sfafiltline}{HTML}{82B366}%
\definecolor{sfaenvfill}{HTML}{FFF2CC}%
\definecolor{sfaenvline}{HTML}{D6B656}%

\begin{tikzpicture}[
    font=\footnotesize,
    blk/.style={draw, semithick, rounded corners=5.5pt, align=center,
                inner sep=0pt, minimum width=2.64cm, minimum height=1.32cm},
    ctrl/.style={blk, fill=sfactrlfill, draw=sfactrlline},
    filt/.style={blk, fill=sfafiltfill, draw=sfafiltline},
    sys/.style={draw, semithick, rounded corners=4pt, align=center, fill=white,
                inner sep=0pt, minimum width=2.805cm, minimum height=0.99cm},
    env/.style={draw=sfaenvline, semithick, rounded corners=10pt,
                fill=sfaenvfill, inner sep=0pt,
                minimum width=3.96cm, minimum height=2.31cm},
    wire/.style={semithick},
    arrow/.style={wire, -{Latex[length=1.9mm]}},
    optarrow/.style={arrow, dash pattern=on 1mm off 1mm},
    sig/.style={font=\small, inner sep=0pt},
    dsc/.style={font=\scriptsize, inner sep=0pt},
  ]

  \node[env] (env) at (10.58,0) {};
  \node[inner sep=0pt] at (10.58,0.891) {Environment};
  \node[sys] (sys) at (10.58,0) {Robot};
  \node[ctrl] (nom) at (0,0) {Nominal\\controller};
  \node[filt] (filt) at (5.29,0) {ORN-CBF\\safety filter};

  \coordinate (jt) at (5.29,1.83);
  \coordinate (jb) at (5.29,-1.73);

  \draw[arrow] (nom.east) -- (filt.west);
  \draw[arrow] (filt.east) -- (sys.west);
  \node[sig] at (2.645,0.30) {$u_{\mathrm{nom}}$};
  \node[dsc] at (2.645,-0.25) {Nominal control};
  \node[sig] at (7.605,0.30) {$u^{\star}$};
  \node[dsc] at (7.605,-0.21) {Safe control};

  \draw[wire] (env.north) -- (10.58,1.83) -- (jt);
  \draw[arrow] (jt) -- (filt.north);
  \draw[optarrow] (jt) -- (0,1.83) -- (nom.north);
  \fill (jt) circle (2.3pt);
  \node[sig] at (7.935,2.13) {$o_j$};
  \node[dsc] at (7.935,1.63) {Observation};

  \draw[wire] (sys.south) -- (10.58,-1.73) -- (jb);
  \draw[arrow] (jb) -- (filt.south);
  \draw[optarrow] (jb) -- (0,-1.73) -- (nom.south);
  \fill (jb) circle (2.3pt);
  \node[sig] at (7.935,-1.40) {$x$};
  \node[dsc] at (7.935,-1.96) {State feedback};
\end{tikzpicture}
  \caption[Closed-loop arrangement of the ORN-CBF safety filter]{Closed-loop arrangement of the ORN-CBF safety filter for a robot that operates on local observations. A nominal controller of arbitrary design issues the command \(u_{\mathrm{nom}}\). From the current observation \(o_j\), ORN-CBF generates the barrier candidate \(\hat{h}_{\Psi}(\,\cdot\mid w_j)\) once per observation update, as detailed in \cref{fig:orn-cbf-architecture}. The CBF-QP \cref{eq:orn-cbf-qp} then evaluates this barrier at the measured state \(x\) at control rate, modifies the command minimally if needed, and forwards the filtered control \(u^{\star}\) to the robot. The dashed arrows indicate that the nominal controller may use the same two signals, on which the filter places no requirement.}
  \label{fig:orn-cbf-safety-filtering}
\end{figure}

\Cref{fig:orn-cbf-safety-filtering} shows the closed-loop arrangement of the complete safety filter. At each control instant \(t\in[t_j,t_{j+1})\), ORN-CBF receives the current state \(x\) and the nominal command \(u_{\mathrm{nom}}\), reuses the main-network parameters \(\Theta(w_j)=H_{\Psi}(w_j)\) computed at the latest observation update, and solves the CBF-QP \cref{eq:prelim-cbf-qp},
\begin{subequations}
  \label{eq:orn-cbf-qp}
  \begin{align}
    u^{\star}
    =
    \operatorname*{arg\,min}_{u\in\mathcal{U}}
    \quad               &
    \frac{1}{2}\lVert u-u_{\mathrm{nom}}\rVert_2^{2}
    \label{eq:orn-cbf-qp-objective}
    \\
    \text{ s.t. }\quad  &
    L_{f_0}\hat{h}_{\Psi}(x\mid w_j)
    +
    L_g\hat{h}_{\Psi}(x\mid w_j)\,u
    +
    \gamma\,\hat{h}_{\Psi}(x\mid w_j)
    \geq0,
    \label{eq:orn-cbf-qp-constraint}
  \end{align}
\end{subequations}
where the extended class-\(\mathcal{K}\) function is chosen linear, \(\alpha(s)=\gamma s\) with gain \(\gamma>0\), as is standard in practice. The robot applies the resulting input \(u^{\star}\), and a nominal command that already satisfies the barrier constraint passes through unchanged. For the robots considered here, \(\mathcal{U}\) is described by linear inequalities and \cref{eq:orn-cbf-qp-constraint} is affine in \(u\), so the optimization is a small convex QP whose size does not depend on the environment complexity, because all geometry enters through the scalar \(\hat{h}_{\Psi}\) and its Lie derivatives.

The main network and the SDF interpolant are implemented in a framework that supports automatic differentiation, which returns exact derivatives where the represented barrier is differentiable. At interpolation cell boundaries and other nonsmooth points, it returns the derivative selected by its computational branch, which is not a classical derivative of a globally smooth field. The main-network weights and the SDF samples enter the QP as runtime parameters, while its structure is fixed and can be precompiled before deployment. Each observation update requires one hypernetwork inference, and each control instant requires one evaluation of the main network and the interpolant with gradients, followed by one QP solve.

Failure-set exclusion \cref{eq:orn-cbf-exclusion} holds by construction. Whether \(\hat{h}_{\Psi}(\,\cdot\mid w_j)\) satisfies the CBF condition is assessed through the closed-loop behavior rather than verified analytically, and because feasibility of the QP \cref{eq:orn-cbf-qp} cannot be ensured in advance, the implementation uses the relaxed form \cref{eq:prelim-cbf-qp-slack}, which always returns an input. Accurate value regression and closed-loop collision avoidance are tested in the next section, and \cref{sec:orn-cbf-discussion} states the scope of these claims.

\section{Experimental Evaluation}
\label{sec:orn-cbf-experiments}

The evaluation examines training cost, barrier approximation, closed-loop safety against MPC and exponential-CBF baselines, generalization to unseen geometry, and transfer to hardware.\footnote{The implementation is available at \url{https://github.com/bojan-derajic/ORN-CBF}.} The direct-regression ablation, \emph{observation-conditioned neural CBF} (ON-CBF), predicts the value without an SDF residual structure and therefore lacks the exclusion guarantee.

\subsection{Robot Models}
\label{subsec:orn-cbf-robot-models}

The evaluation uses two robot models with complementary characteristics. The first model cannot change its linear speed and allows only limited steering, while the second model has bounded acceleration and cannot change its velocity instantaneously.

\paragraph{Dubins car.}
The ground robot is modeled as the Dubins car \cref{eq:ntc-dubins-dynamics}, with a constant forward speed of \(v=\SI{1.0}{\meter\per\second}\) and an angular-velocity input limited to \(\omega\in[-0.5,0.5]\)~\si{\radian\per\second}, which differ from the limits used in \cref{ch:ntc-mpc}. Because the robot cannot stop, its maximal safe set differs substantially from the geometric free space, and a purely distance-based barrier is inadequate.

\paragraph{Planar double integrator.}
The aerial robot is a quadrotor operating at constant altitude. It is modeled as a planar double integrator with state \(x=(p_x,p_y,v_x,v_y)\), control input \(u=(a_x,a_y)\), and dynamics
\begin{equation}
  \dot{x}
  =
  \begin{bmatrix}
    v_x \\ v_y \\ a_x \\ a_y
  \end{bmatrix},
  \label{eq:orn-cbf-double-integrator}
\end{equation}
with limits \(v_x,v_y\in[-2,2]\)~\si{\meter\per\second} and \(a_x,a_y\in[-1,1]\)~\si{\meter\per\square\second}. This model describes quadrotors with limited acceleration authority, for example when they transport loads. It is the multidimensional counterpart of the braking example in \cref{subsec:prelim-controlled-invariant-safe-sets}, where the unsafe region extends beyond the obstacle by the braking distance required at the current approach speed. The model also illustrates the relative-degree issue of \cref{sec:prelim-cbf}, since a position-based distance constraint has relative degree two for acceleration inputs, which motivates the exponential-CBF baseline of \cref{subsec:orn-cbf-sim-qq}.

\subsection{Training Setup and Learned Barriers}
\label{subsec:orn-cbf-training-eval}

\begin{table}[tb]
  \caption{ORN-CBF data generation and Hamilton--Jacobi targets for the two robot models.}
  \label{tab:orn-cbf-training-config-data}
  \centering
  \begin{tabular}{@{}lll@{}}
    \toprule
                                     & Dubins car                                        & Planar double integrator                           \\
    \midrule
    Training environment             & warehouse                                         & forest-like                                        \\
    Observations collected           & \num{2500}                                        & \num{2000}                                         \\
    After augmentation (\(\times8\)) & \num{20000}                                       & \num{16000}                                        \\
    Sensing window                   & \(6\times6\)~\si{\meter}, \SI{0.06}{\meter}\,res. & \(6\times6\)~\si{\meter}, \SI{0.075}{\meter}\,res. \\
    HJ state-space grid              & \(100\times100\times30\)                          & \(80\times80\times20\times20\)                     \\
    HJ computation (RTX~3090)        & \(\sim\)\SI{1.17}{\hour}                          & \(\sim\)\SI{16.3}{\hour}                           \\
    Stored positional patch          & \(16\times16\)                                    & \(16\times16\)                                     \\
    Stored grid per observation      & \(16\times16\times30\)                            & \(16\times16\times20\times20\)                     \\
    \bottomrule
  \end{tabular}
\end{table}

\paragraph{Training data and targets.}
\Cref{tab:orn-cbf-training-config-data} summarizes the data generation and the HJ targets. For the Dubins car, \num{2500} occupancy grids are collected in a simulated warehouse, each covering \(6\times6\)~\si{\meter} at a resolution of \SI{0.06}{\meter}. For the double integrator, \num{2000} maps of the same size are collected at a resolution of \SI{0.075}{\meter} in a forest-like environment with cylindrical obstacles. Each map and its horizontally flipped version are retained in four orientations, separated by \(90^{\circ}\), which yields \num{20000} and \num{16000} SDF observations, respectively, split into training and validation data after augmentation as in \cref{ch:ntc-mpc}. The stationary HJ targets are computed with a grid-based solver.\footnote{\url{https://github.com/StanfordASL/hj_reachability}} The Dubins-car grid has size \(100\times100\times30\) over \((p_x,p_y,\theta)\), and solving the corresponding problems for the whole dataset requires about \SI{1.17}{\hour} of GPU time. The double-integrator grid has size \(80\times80\times20\times20\) over \((p_x,p_y,v_x,v_y)\), for which the dataset requires about \SI{16.3}{\hour}. Following \cref{subsec:orn-cbf-training}, only the central \(16\times16\) positional patch of each solution is stored and used for training, while the non-positional dimensions are kept in full. For the Dubins car, this reduces the stored data and the per-epoch computation by a factor of about \num{39} relative to the complete window.

\begin{table}[tb]
  \caption{ORN-CBF configuration shared by both robot models.}
  \label{tab:orn-cbf-training-config-shared}
  \centering
  \begin{tabular}{@{}ll@{}}
    \toprule
    Hypernetwork               & CNN backbone with fully connected head                                   \\
    Main network hidden layers & \([32,32,32,16,16,16,8,8,8]\), sinusoidal                \\
    Main network output        & softplus (nonnegative residual)                          \\
    Loss                       & RWMSE (\cref{subsec:ntc-rwmse})                          \\
    Optimizer / epochs / batch & Adam / 100 / 16                                          \\
    Learning rate              & \(10^{-4}\), reduced \(10\times\) at epochs 85 and 95     \\
    \bottomrule
  \end{tabular}
\end{table}

\begin{table}[tb]
  \caption{ORN-CBF configuration specific to each robot model.}
  \label{tab:orn-cbf-training-config-model}
  \centering
  \begin{tabular}{@{}lll@{}}
    \toprule
                             & Dubins car              & Planar double integrator \\
    \midrule
    Main network inputs      & 3                       & 4                        \\
    Main network parameters  & \num{3601}              & \num{3633}              \\
    Training time (RTX~3090) & \(\sim\)\SI{0.5}{\hour} & \(\sim\)\SI{2.9}{\hour}  \\
    \bottomrule
  \end{tabular}
\end{table}

\paragraph{Network architecture and optimization.}
\Cref{tab:orn-cbf-training-config-shared} lists the model and optimization settings shared by both robot models, and \cref{tab:orn-cbf-training-config-model} lists the entries that differ between them. Full-window training of the Dubins-car model in \cref{ch:ntc-mpc} required \SI{7.5}{\hour} for 150~epochs (\cref{tab:ntc-training-config-model}). At an equal number of epochs, positional cropping thus reduces the training time by roughly an order of magnitude.

\begin{figure}[tb]
  \centering
  \includegraphics[width=\textwidth]{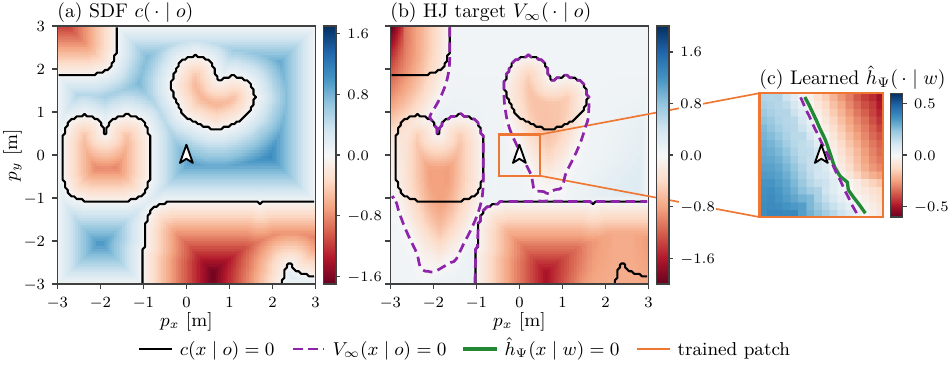}
  \caption[Learned ORN-CBF for the Dubins car on a validation observation]{A learned ORN-CBF for the Dubins car on a validation observation. (a)~The observed SDF \(c(\,\cdot\mid o)\) with the obstacle outlines \(c=0\) (black) and the robot position at heading \(\theta=\pi/2\) (dart, moving upward). (b)~The numerical HJ target \(V_\infty(\,\cdot\mid o)\) at the same heading with its zero level set (dashed). Red shading marks the unsafe states, which extend beyond the obstacle outlines in the directions in which the robot approaches the obstacles head-on, and the orange square marks the central \(16\times16\) positional patch on which the residual network is trained. (c)~The learned barrier \(\hat{h}_{\Psi}(\,\cdot\mid w)\) on this patch whose zero level set (green) closely follows the HJ boundary. The asymmetry of the unsafe region reflects the turning limitation of the constant-speed vehicle.}
  \label{fig:orn-cbf-sample-cbf}
\end{figure}

\paragraph{Barrier approximation.}
\Cref{fig:orn-cbf-sample-cbf} visualizes a learned barrier for the Dubins car on a validation observation. As in \cref{ch:ntc-mpc}, the learned unsafe region is visibly larger than the occupied region and depends on the heading. States from which the constant-speed vehicle can no longer turn away in time are classified as unsafe although their distance to the obstacles is positive. A geometric barrier does not account for this effect, while the learned barrier provides it at the observation rate.

\subsection{Ground-Robot Simulation}
\label{subsec:orn-cbf-sim-gr}

\begin{figure}[tb]
  \centering
  \includegraphics[width=0.9\textwidth]{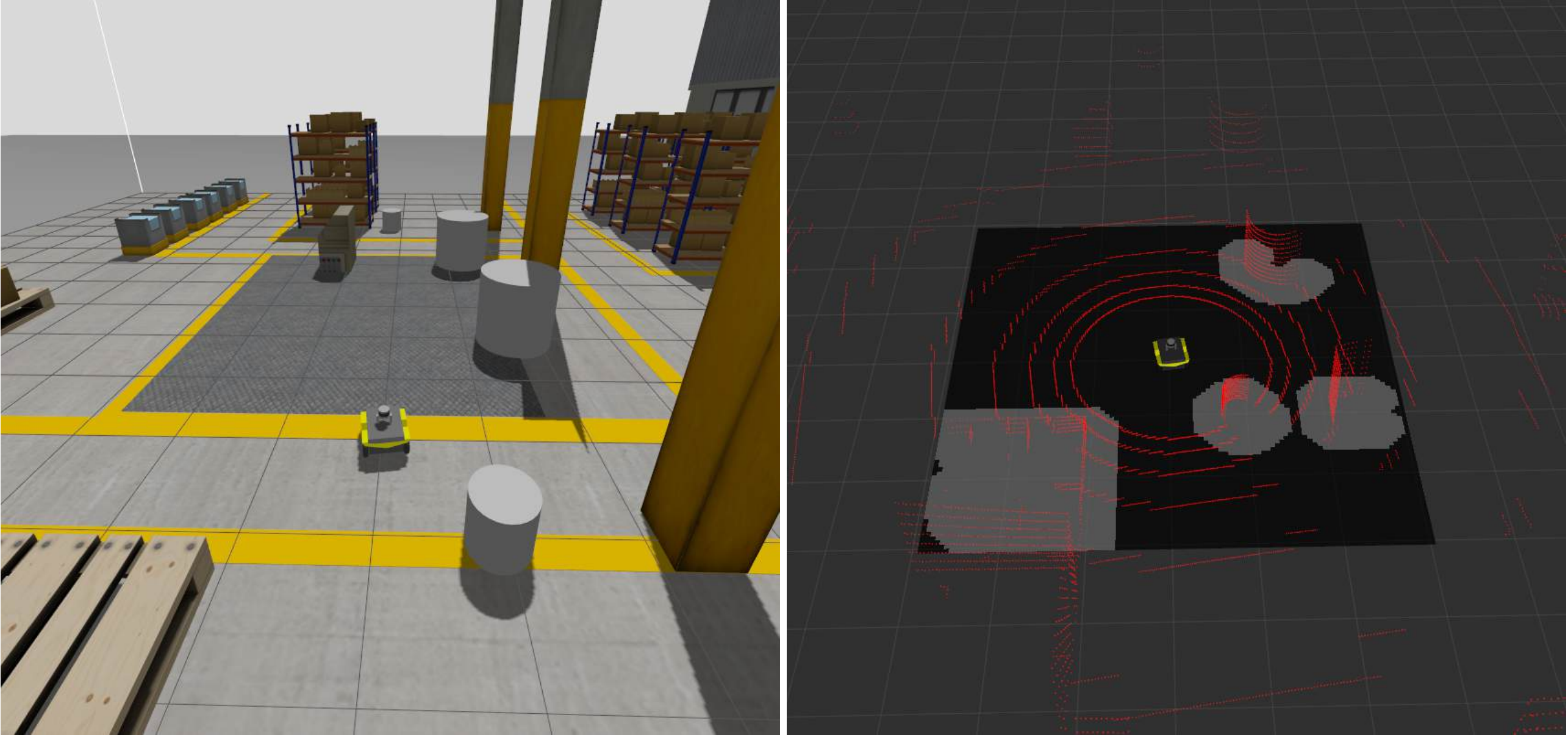}
  \caption[Ground-robot simulation environment of the ORN-CBF study]{Ground-robot simulation study. Left: scene from the Gazebo warehouse environment. Right: the corresponding local observation from which the ORN-CBF barrier is generated.}
  \label{fig:orn-cbf-gr-sim}
\end{figure}

\paragraph{Experimental setup and baselines.}
The first closed-loop study evaluates the safety filter for the ground robot in physics-based Gazebo 3D simulations of a warehouse environment (\cref{fig:orn-cbf-gr-sim}). The task is the local navigation problem of the previous chapters, in which the robot must reach a goal position behind randomly distributed obstacles using only local observations. The nominal controller is the SDF-MPC local planner of \cref{subsec:ntc-baselines}, and either ORN-CBF or its direct-regression ablation ON-CBF is placed between this planner and the robot. Three MPC planners of \cref{subsec:ntc-baselines} operate without a safety filter, SDF-MPC, DCBF-MPC, and NTC-MPC, the last retrained for the Dubins-car speed and turning-rate limits used in this chapter. NTC-MPC and ORN-CBF follow the same reachability-supervision principle for static obstacles, but NTC-MPC supplies a terminal constraint to the MPC problem, while ORN-CBF supplies a barrier to a QP filter that operates downstream of a separate planner. Related observation-based neural CBFs \cite{dawson_learning_2022,harms_neural_2024-2} assume different observation modalities and are not directly applicable in this setting.

All MPC planners run at \SI{20}{\hertz} with prediction interval \(\Delta t=\SI{0.1}{\second}\), implemented in CasADi \cite{andersson_casadi_2019} with the IPOPT NLP solver and deployed as ROS~2 nodes \cite{macenski_robot_2022}. The CBF-QP filters are solved with qpOASES and run at \SI{200}{\hertz} with gain \(\gamma=2.6\), while the costmap updates at \SI{20}{\hertz}. The deployed filters use the relaxed problem \cref{eq:prelim-cbf-qp-slack} rather than the hard-constrained form \cref{eq:orn-cbf-qp}, so that the solver always returns an admissible input, and the returned input solves \cref{eq:orn-cbf-qp} whenever the slack is zero. For each method and each horizon \(N\in\{5,10,20,30\}\), 100~runs with randomized obstacle configurations are performed, and any collision counts as a failure.

With this observation update rate, the robot travels \(v_{\max}\Delta t_o=\SI{0.05}{\meter}\) between two updates, which leaves \(\rho-v_{\max}\Delta t_o=\SI{2.95}{\meter}\) on the left-hand side of the sensing condition \cref{eq:orn-cbf-sensing-adequacy}. The margin \(\beta_{\mathcal{B}}\) on the right-hand side is determined by the turning geometry of the constant-speed robot. Its minimum turning radius is \(v/\omega_{\max}=\SI{2}{\meter}\), and a robot heading straight at a flat wall advances by that radius while turning to travel parallel to it, so \(\beta_{\mathcal{B}}\) is of the order of \SI{2}{\meter} for such geometry. Condition \cref{eq:orn-cbf-sensing-adequacy} therefore holds with a margin of roughly \SI{1}{\meter} for this geometry. This estimate is a plausibility check rather than a bound over the evaluated scenes, because concave geometry, such as the inner corner of two shelves, enlarges \(\beta_{\mathcal{B}}\) and can exceed that margin.

\begin{figure}[tb]
  \centering
  \includegraphics[width=0.6\textwidth]{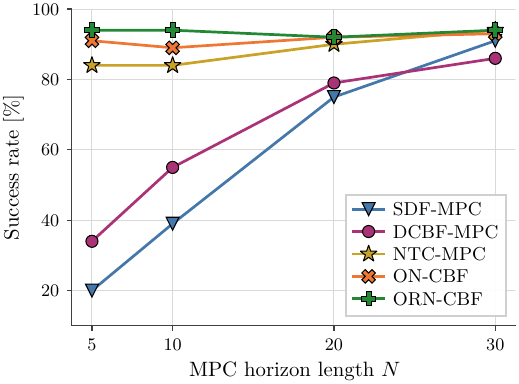}
  \caption[Success rates of the ORN-CBF ground-robot simulation study]{Success rates of the ground-robot simulation study over the MPC prediction-horizon length, with 100~randomized runs per configuration. The safety filters ORN-CBF and ON-CBF are combined with the plain SDF-MPC planner. Both outperform every MPC-based planner for \(N\leq20\), including the reachability-informed NTC-MPC of \cref{ch:ntc-mpc}, which reaches a similar rate at \(N=30\). ORN-CBF achieves the highest rates overall.}
  \label{fig:orn-cbf-gr-success}
\end{figure}

\paragraph{Simulation results.}
\Cref{fig:orn-cbf-gr-success} shows the obtained success rates, and three observations summarize the study. First, the plain SDF-MPC planner combined with ORN-CBF outperforms every unfiltered planner for \(N\leq20\), including SDF-MPC itself, DCBF-MPC, and NTC-MPC, and reaches the same success rate as NTC-MPC at \(N=30\). One plausible explanation is the difference between the controller interfaces. The filter evaluates the reachability boundary at the current state at \SI{200}{\hertz}, while the planners constrain predicted states once per planning cycle. This difference can make the filter less sensitive to prediction and approximation errors within the horizon, although the experiment does not isolate the effect. Second, the advantage is largest at short horizons, where the geometric baselines degrade considerably for the reasons discussed in \cref{subsec:ntc-simulations}. Both filtered variants remain nearly horizon-independent, and the direct-regression variant ON-CBF also outperforms every baseline at the shortest horizon. Third, ORN-CBF consistently matches or exceeds ON-CBF, which agrees with the approximation advantage of residual learning observed in \cref{subsec:rntc-training-eval}.

\subsection{Quadrotor Simulation}
\label{subsec:orn-cbf-sim-qq}

\begin{figure}[tb]
  \centering
  \includegraphics[width=0.9\textwidth]{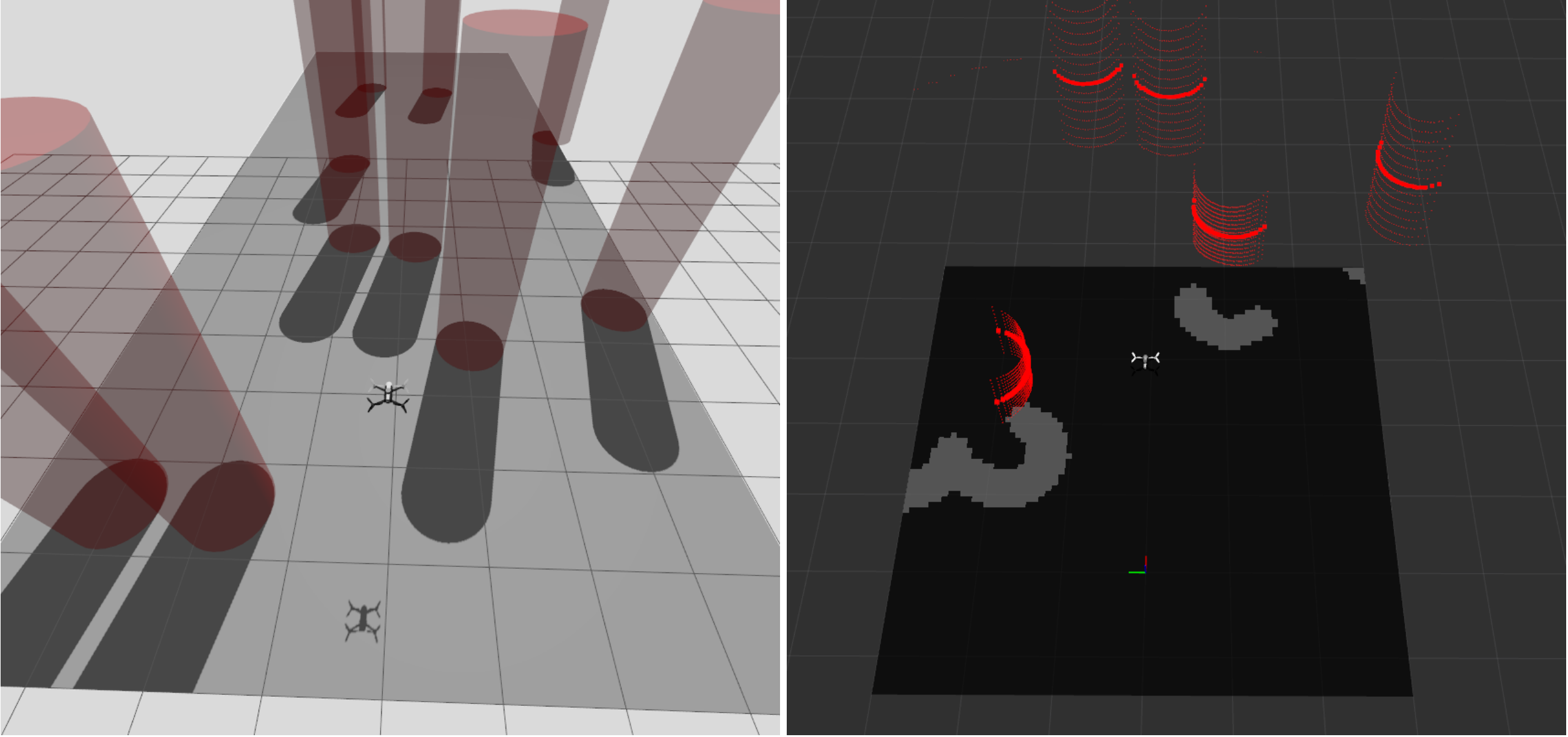}
  \caption[Quadrotor simulation environment of the ORN-CBF study]{Quadrotor simulation study. Left: scene from the Gazebo forest-like environment. Right: the corresponding local observation.}
  \label{fig:orn-cbf-qq-sim}
\end{figure}

\paragraph{Experimental setup and baselines.}
The second study examines the quadrotor model in a forest-like environment (\cref{fig:orn-cbf-qq-sim}) and shifts the focus to the \emph{generalization} of the barrier itself. The nominal controller is a linear-quadratic regulator (LQR) with \(Q=\operatorname{diag}(10,10,1,1)\) and \(R=\operatorname{diag}(1,1)\). It has no obstacle awareness, so every avoidance maneuver originates from the filter. The classical baseline is a second-order exponential CBF (ECBF) \cite{nguyen_exponential_2016} constructed from the observed SDF. Its order follows from the relative degree of the SDF under double-integrator dynamics, and its gain vector \(K_{\mathrm{e}}=[1.0,\,2.0]\) is tuned in the training environment. ORN-CBF and ON-CBF are trained in the forest-like environment with cylindrical obstacles of \SI{0.5}{\meter} radius and evaluated both in this domain and in an out-of-domain environment with radii from \SIrange{0.2}{1.0}{\meter}. The ECBF uses the same fixed, hand-tuned gains in both environments. Each configuration contains 200~scenarios with 10~randomly placed obstacles. The CBF-QPs run at \SI{200}{\hertz} with \(\gamma=0.7\), and the costmap updates at \SI{5}{\hertz}.

\begin{figure}[tb]
  \centering
  \includegraphics[width=0.85\textwidth]{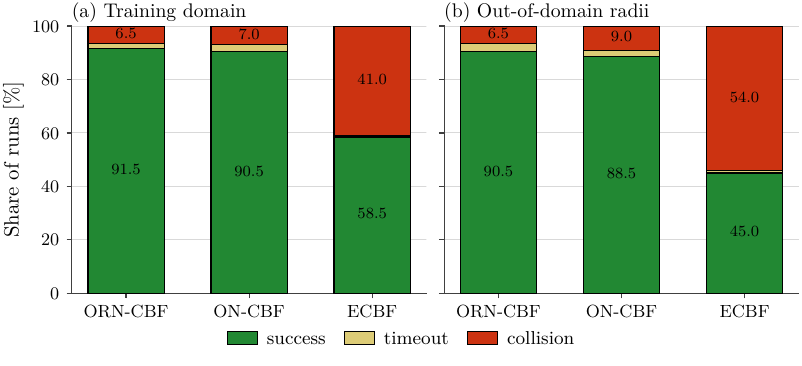}
  \caption[Outcomes of the ORN-CBF quadrotor simulation study]{Outcomes of the quadrotor simulation study over 200~randomized scenarios per configuration, split into successes, timeouts, and collisions. (a)~In the training domain, the learned filters succeed in more than \(90\%\) of the runs, compared with \(58.5\%\) for the hand-tuned ECBF. (b)~In the out-of-domain environment with unseen obstacle radii, the success rates of the learned filters decrease by at most two percentage points, while that of the ECBF decreases to \(45.0\%\).}
  \label{fig:orn-cbf-qq-results}
\end{figure}

\begin{figure}[tb]
  \centering
  \includegraphics[width=\textwidth]{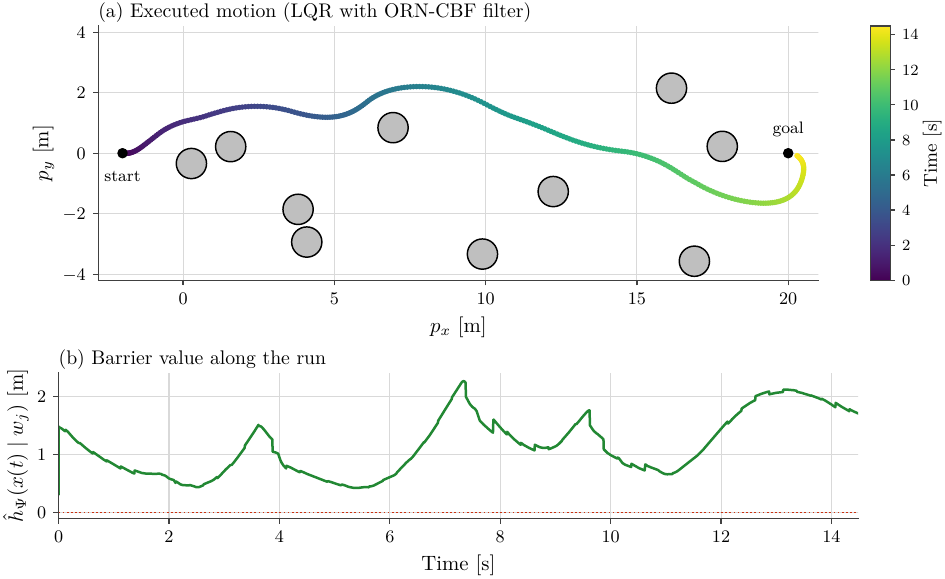}
  \caption[Representative ORN-CBF quadrotor scenario]{Representative quadrotor scenario. (a)~Executed motion under the LQR nominal controller with the ORN-CBF safety filter among the ten cylindrical obstacles (gray), colored by time. The filter steers the quadrotor around the obstacles with which the goal-directed LQR would otherwise collide. (b)~Evolution of the barrier value \(\hat{h}_{\Psi}(x(t)\mid w_j)\), which decreases toward zero during the avoidance maneuvers and remains positive throughout the run.}
  \label{fig:orn-cbf-qq-traj}
\end{figure}

\paragraph{Simulation results.}
\Cref{fig:orn-cbf-qq-results} presents the outcomes. In the training domain, ORN-CBF and ON-CBF succeed in \(91.5\%\) and \(90.5\%\) of the scenarios, respectively, compared with \(58.5\%\) for the ECBF, whose failures are mostly collisions. A plausible explanation is that the ECBF barrier, built from the observed SDF, does not match the velocity- and geometry-dependent reachability boundary uniformly, while the learned barrier represents this dependence directly. The experiment supports this interpretation but does not separate it from the effect of the gain-tuning procedure. In the out-of-domain environment, the success rates of the learned filters decrease by only one to two percentage points, to \(90.5\%\) and \(88.5\%\), while that of the ECBF decreases to \(45.0\%\). The fixed ECBF gains appear to require retuning as the obstacle scale changes, while the observation-conditioned barriers regenerate their dynamics-aware fields from each observation. The remaining collisions and timeouts may be caused by the approximation error, by signal delays, or by the finite size and resolution of the costmap. \Cref{fig:orn-cbf-qq-traj} shows a representative run. The LQR drives the quadrotor straight toward the goal, and the filter intervenes as the barrier value approaches zero and steers the quadrotor around the obstacles while the recorded value remains positive throughout.

\subsection{Hardware Experiments}
\label{subsec:orn-cbf-hardware}

\begin{figure}[tb]
  \centering
  \includegraphics[width=0.9\textwidth]{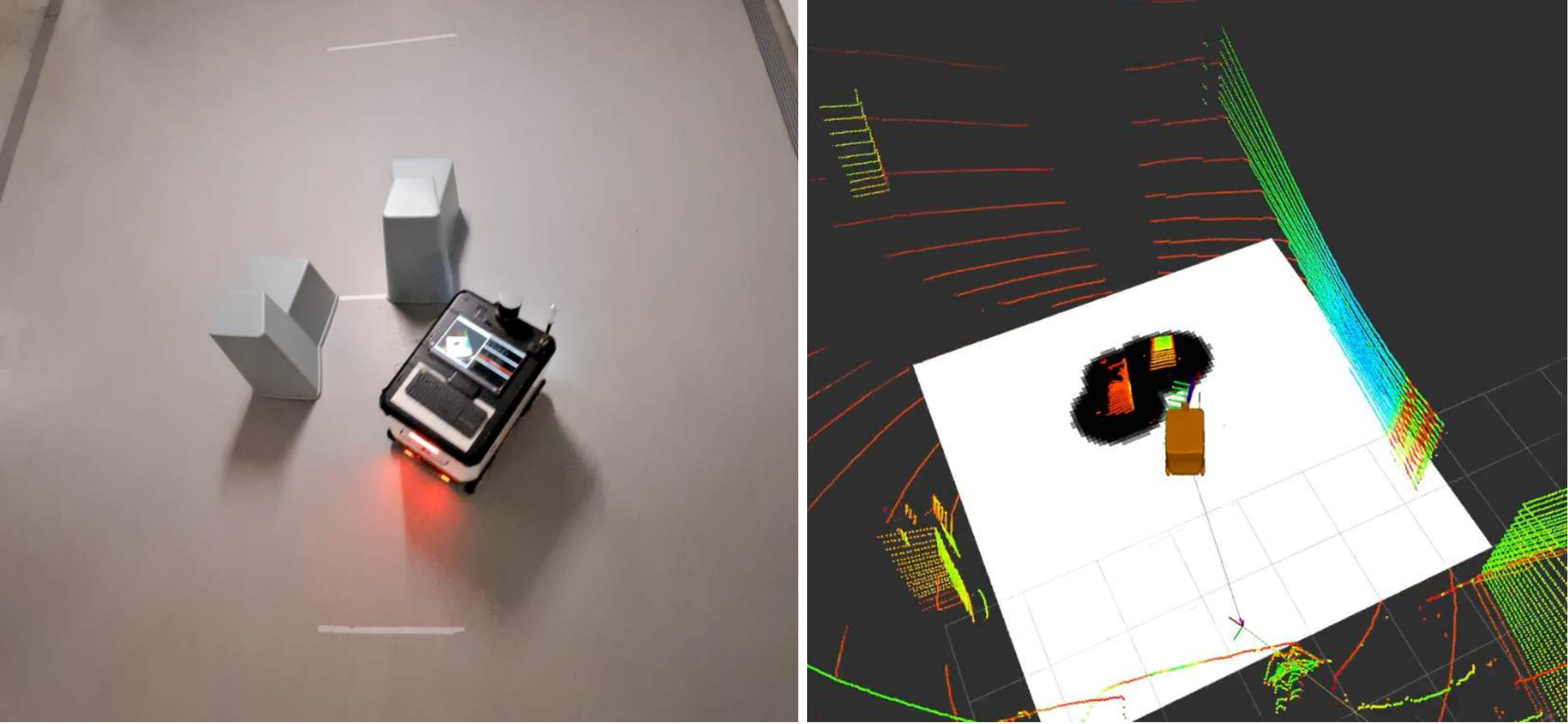}
  \caption[ORN-CBF ground-robot hardware experiments]{Ground-robot hardware experiments. Left: the differential-drive robot among the static obstacles. Right: visualization of the local observation from which the barrier is generated onboard.}
  \label{fig:orn-cbf-gr-hardware}
\end{figure}

\paragraph{Ground-robot experiments.}
The ground robot is evaluated on the delivery platform of \cref{subsec:ntc-hardware}, which operates at constant forward speed, so the Dubins model applies, and all computation runs onboard (\cref{fig:orn-cbf-gr-hardware}). ORN-CBF and ON-CBF are deployed after training on synthetic warehouse data alone, without fine-tuning. The comparison follows the simulation study, with SDF-MPC, DCBF-MPC, and NTC-MPC running at \SI{20}{\hertz} with \(N=10\) and \(\Delta t=\SI{0.1}{\second}\), and with the two filters running at \SI{200}{\hertz} downstream of SDF-MPC. Ten trials per method are performed with randomized static obstacle configurations.

\begin{table}[tb]
  \caption[ORN-CBF hardware experiment results for the ground robot]{Hardware experiment results for the ground robot over ten randomized obstacle configurations (Dubins car model, MPC horizon \(N=10\)).}
  \label{tab:orn-cbf-hardware-results}
  \centering
  \begin{tabular}{@{}lc@{}}
    \toprule
    Method   & Success rate [\%] \\
    \midrule
    SDF-MPC  & 20                \\
    DCBF-MPC & 40                \\
    NTC-MPC  & 70                \\
    ORN-CBF (ours) & \textbf{100}      \\
    ON-CBF   & \textbf{100}      \\
    \bottomrule
  \end{tabular}
\end{table}

\begin{figure}[tb]
  \centering
  \includegraphics[width=0.75\textwidth]{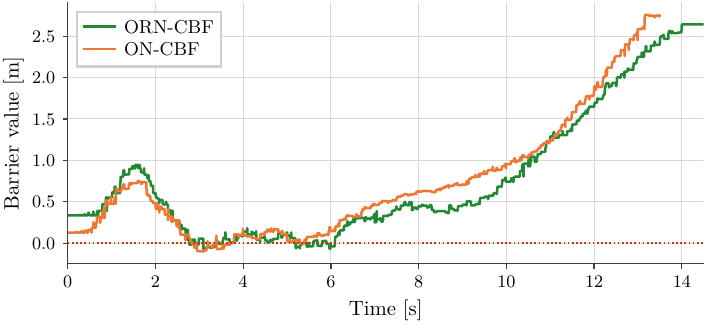}
  \caption[Barrier values during an ORN-CBF ground-robot hardware run]{Barrier values of the ORN-CBF and ON-CBF filters in one representative obstacle configuration of the ground-robot hardware experiments, with one run per filter under the same experimental setting and the zero level marked by the dotted line. Briefly negative values are followed by recovery, and no collision occurs. The physical inflation margin separates the represented failure boundary from physical contact.}
  \label{fig:orn-cbf-gr-cbf-value}
\end{figure}

\Cref{tab:orn-cbf-hardware-results} summarizes the outcomes. The robot completes all ten configurations using SDF-MPC with either ORN-CBF or ON-CBF. The best unfiltered planner, NTC-MPC, succeeds in seven, and the geometric planners in at most four, so the ordering observed in simulation also appears on hardware. \Cref{fig:orn-cbf-gr-cbf-value} shows the barrier values of both filters in one representative obstacle configuration. The values occasionally become slightly negative, which may reflect model mismatch or measurement noise, but no collision occurs, because the inflation margin in \cref{eq:orn-cbf-observed-constraint} provides a physical buffer and the filter rapidly restores a positive value. This trace illustrates the practical role of separating the learned zero level set from the physical collision boundary. While the barrier is negative, the state lies outside the predicted safe set, so no invariance guarantee applies to it.

\paragraph{Quadrotor experiments.}
The aerial evaluation uses a Crazyflie nano-quadrotor with four laser range sensors. During flight, the quadrotor rotates continuously about its vertical axis, so the four range measurements accumulate into a local costmap that updates at \SI{10}{\hertz}, and the computation itself runs offboard. With the same LQR controller and the same ORN-CBF model as in \cref{subsec:orn-cbf-sim-qq}, the quadrotor avoids the obstacles despite sparse and delayed observations. This experiment is a deployment demonstration rather than a statistical comparison, and it shows transfer to a perception configuration that is more demanding than the one used for training.

\section{Discussion}
\label{sec:orn-cbf-discussion}

The experiments show that observation-conditioned residual fields support real-time safety filtering and achieve success rates that match or exceed those of the evaluated geometric barriers and MPC planners. The following discussion separates these results from the conditions required for a formal guarantee.

\paragraph{Guarantee scope.}
The architecture guarantees failure-set exclusion, and the sensing condition \cref{eq:orn-cbf-sensing-adequacy} together with the boundary-only appearance of new information bounds how close a newly revealed obstacle can be. Under these conditions, with a static environment and an accurate model with full state feedback, the CBF theory of \cref{sec:prelim-cbf} yields forward invariance of the predicted safe set as long as the learned barrier satisfies the CBF condition and the slack of the relaxed QP is zero. The experiments support both requirements empirically rather than verifying them formally. The barrier is trained by value regression, so the CBF condition itself is not verified separately. In addition, the assembled barrier inherits the nonsmooth points of the SDF and of its bilinear interpolant. At these points, automatic differentiation returns a branch derivative that suffices for the implementation, and no observed filter failure was attributable to this property. The hardware traces of \cref{fig:orn-cbf-gr-cbf-value} show what these qualifications mean in practice, since the barrier value briefly became negative and the physical safety margin prevented a collision while the filter restored a positive value. Verifying the barrier condition on the learned field, with a smooth constraint function and interpolant where the assembled barrier is nonsmooth, would turn this empirical support into a formal statement, and \cref{ch:conclusion} names the tools available for this step.

\paragraph{Safety filtering and MPC terminal constraints.}
The ground-robot study permits a direct comparison of the two controller interfaces, because NTC-MPC and the two filters are evaluated in the same experiment, on the same robot, and over the same randomized obstacle configurations. ORN-CBF achieved higher success rates for \(N\leq20\) and the same rate at \(N=30\). This outcome is consistent with the different operation of the two interfaces. The filter evaluates safety at the \emph{current} state at \SI{200}{\hertz}, while NTC-MPC constrains a \emph{predicted} terminal state at \SI{20}{\hertz}, so the filter can react sooner and depends less directly on the prediction accuracy. The experiment does not isolate the interface completely, because ORN-CBF also uses the residual parameterization that NTC-MPC lacks and operates downstream of a separate planner. The direct-regression ablation ON-CBF is the closer counterpart, since it differs from NTC-MPC mainly in the interface, and it also outperformed NTC-MPC for \(N\leq20\). The filtering interface has drawbacks of its own. A safety filter acts on the current state only, has no task model, can modify only the control command, and depends on the nominal controller for progress. The quadrotor timeouts show that the closed loop can remain safe and still make no progress when the nominal controller performs poorly. The two interfaces are complementary, and the same learned representation can support either of them, or both, where the system architecture permits.

\paragraph{Scalability of offline supervision.}
The offline stage retains the dimensionality limit of grid-based HJ solvers, and the four-dimensional double integrator already required \SI{16.3}{\hour} of target computation. As noted in the preceding chapters, other sources of supervision could replace the grid solver. Because ORN-CBF already assumes control-affine dynamics with polytopic input constraints, the vertex-restricted search of \cref{ch:vertex-cbf} applies to it directly, although its labels would have to be generated for every training observation. Positional cropping remains compatible with any of these options, because it reduces the amount of data that is stored and fitted rather than the reachability problem that is solved.

\paragraph{Dynamic-environment limitation.}
The restriction to static environments is the main limitation of the method. The infinite-horizon value \(V_\infty(\,\cdot\mid o)\) remains stationary during an observation only when the failure set is static. Moving obstacles would require time-varying targets and temporal conditioning, as in \cref{ch:rntc-mpc}, together with a considerably more expensive data-generation process. The next chapter instead formulates reachability in \emph{relative} robot--obstacle coordinates, which makes the pairwise collision set stationary, so a single offline computation covers all obstacles of the modeled class, and a smooth composition extends the filter to a varying number of moving obstacles.

\section{Summary}
\label{sec:orn-cbf-summary}

ORN-CBF combines fixed-observation HJ learning, residual failure-set exclusion, and positional cropping, which reduced training time by roughly an order of magnitude relative to \cref{ch:ntc-mpc}. The filter matched or outperformed the evaluated MPC baselines at every horizon in ground-robot simulations. Under the change of obstacle geometry in the quadrotor study, the success rates of the learned filters decreased by at most two percentage points, compared with more than thirteen for the ECBF. The delivery robot completed all hardware trials using SDF-MPC with ORN-CBF and with its direct-regression variant ON-CBF, and the ORN-CBF model also transferred to a Crazyflie quadrotor with four-beam sensing. The method addresses real-time safety filtering at \SI{200}{\hertz} and structural exclusion (questions~\ref{rq:realtime} and~\ref{rq:structure}). The next chapter extends filtering to moving obstacles through relative coordinates and pairwise composition.

\chapter{Neural Control Barrier Functions in Dynamic Environments}
\label{ch:cn-cbf}

This chapter presents the \emph{composite neural CBF} (CN-CBF), based on the author's paper \cite{derajic_cn-cbf_2026} at the IEEE/RSJ International Conference on Intelligent Robots and Systems (IROS) 2026. The method learns one reachability-based residual barrier in relative robot--obstacle coordinates, where the collision set is stationary, and smoothly combines its evaluations over the observed obstacles. The shared model handles a varying obstacle count without retraining. CN-CBF addresses dynamic safety filtering (questions~\ref{rq:dynamic} and~\ref{rq:realtime}) and preserves failure-set exclusion through composition (question~\ref{rq:structure}).

\section{Motivation}
\label{sec:cn-cbf-motivation}

CN-CBF extends the safety-filter interface of \cref{ch:orn-cbf} from static scenes to environments with moving obstacles, whose future positions are uncertain and whose number varies as they enter and leave the sensing range. Two features challenge the preceding methods. Moving obstacles make the safe set time-varying, which requires the derivative in \cref{eq:prelim-time-varying-cbf-condition}, and prediction-conditioned learning, as in \cref{ch:rntc-mpc}, handles this variation only at the cost of many scene-specific HJ solutions. Variable obstacle counts add a second difficulty, because scene-level training must cover many joint configurations, and representations based on fixed-size obstacle lists limit the number of obstacles represented. CN-CBF addresses both through a shared pairwise model.

\begin{figure}[tbp]
  \centering
  \input{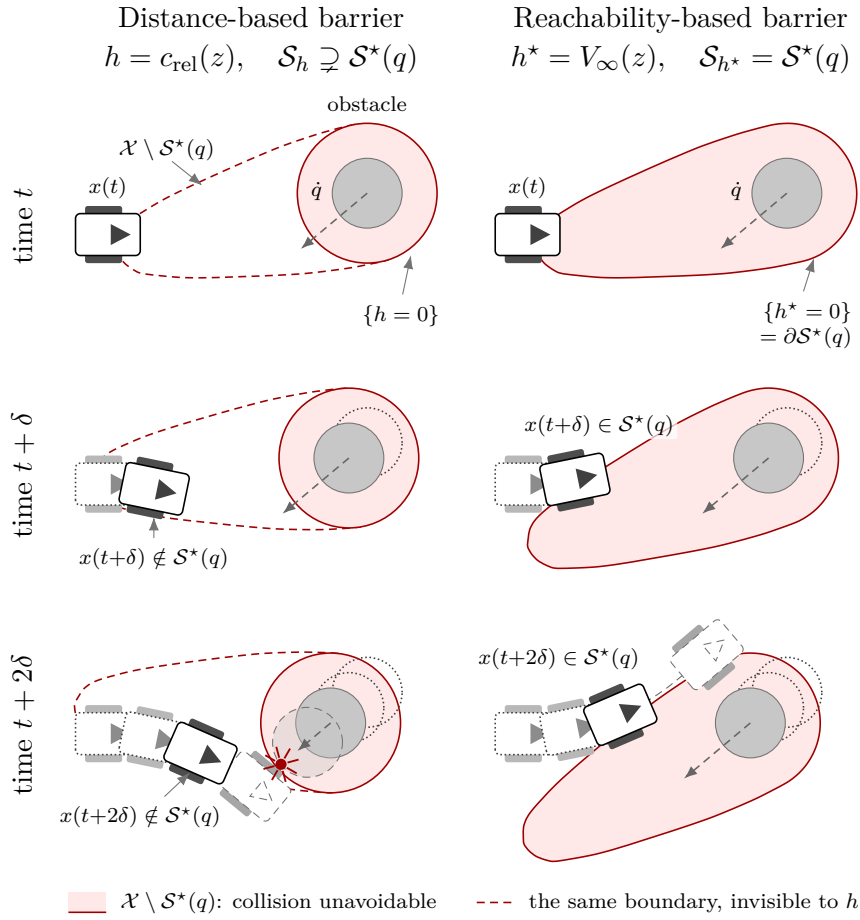}
  \caption[Effect of the barrier candidate on a CBF-QP safety filter in a dynamic environment]{Effect of the barrier candidate on a CBF-QP safety filter in a \emph{dynamic} environment, shown at three instants separated by the short interval \(\delta\). A ground robot that cannot stop is driven to the right by a goal-directed nominal controller, while an obstacle crosses its path from the front left. Dotted outlines mark the earlier positions, the dashed arrow at each obstacle shows its velocity, and the thin dashed curves show the motion under an extremal admissible input. The red region is the set \(\mathcal{X}\setminus\mathcal{S}^{\star}(q)\) of robot states from which the collision with an obstacle in state \(q\) can no longer be avoided at the current heading. Both barriers are evaluated on the relative state \(z\) of the robot--obstacle pair defined in \cref{subsec:cn-cbf-relative-dynamics}. Left: the barrier is the obstacle distance \(c_{\mathrm{rel}}(z)\), whose safe set also contains the red region. That boundary is invisible to the filter and is drawn as a dashed outline only, the filter intervenes too late, and at \(t+2\delta\) both extremal turns end in a collision. Right: the reachability-based barrier \(h^{\star}\) is the stationary HJ value function \(V_\infty(z)\), whose zero level set is the boundary of the red region, so the filter acts already at \(t\) and the robot stays safe.}
  \label{fig:cn-cbf-intro-illustration}
\end{figure}

\Cref{fig:cn-cbf-intro-illustration} shows the unsafe region for one moving obstacle. Its shape depends on relative heading and motion, and it translates with the obstacle. A distance boundary alone does not identify these dynamically unsafe states, while a pairwise reachability barrier accounts for the robot's actuation limits and the assumed obstacle motion and allows the filter to intervene before recovery becomes impossible.

In relative coordinates, collision is a fixed region around the origin, and the robot input opposes the bounded obstacle input in a stationary reachability problem. One offline solution and one compact model then describe every obstacle of the modeled class, and a smooth approximation of the minimum combines the pairwise barriers, so that the number of obstacles can vary at runtime. As in the preceding chapters, each pairwise barrier is a closed-form collision constraint minus a nonnegative learned residual, and the composite barrier inherits the exclusion property for every observed obstacle. The experiments in \cref{sec:cn-cbf-experiments} compare the filter with four MPC planners, including RNTC-MPC, and with two classical CBF constructions, in simulation and on a physical delivery robot and a quadrotor.

\section{Related Work}
\label{sec:cn-cbf-related-work}

This section reviews the work most closely related to the question of this chapter, which is how to obtain barrier candidates for a varying number of moving obstacles. The safety-filtering literature for unknown static environments is reviewed in \cref{sec:orn-cbf-related-work}, the MPC-based literature on dynamic environments in \cref{sec:rntc-related-work}, and the broader context in \cref{ch:introduction}.

Learning-based barrier design for mobile robots has produced barriers learned from expert demonstrations \cite{robey_learning_2020}, neural CBFs that respect input limits \cite{liu_safe_2022}, and real-time constructions from online Gaussian-splatting maps \cite{chen_safer-splat_2025}. As discussed in \cref{sec:orn-cbf-related-work}, such barriers typically represent either a conservative subset or an unsafe superset of the maximal safe set obtainable through HJ reachability \cite{so_how_2024}. Reachability-grounded constructions address this limitation but require the environment to be known in advance \cite{choi_robust_2021}. Observation-conditioned neural CBFs remove that restriction, either through the residual architecture of \cref{ch:orn-cbf} \cite{derajic_orn-cbf_2026} or with added uncertainty estimation \cite{lin_one_2026}, yet both condition on an instantaneous observation of static geometry and do not model obstacle motion. A neural approximation of the \emph{joint} time-varying value function, as in \cref{ch:rntc-mpc} \cite{derajic_residual_2025}, does model that motion. However, the joint problem couples all obstacles, so the supervision cost grows combinatorially with the variety of scenes, and the trained model remains restricted to the sensing range used to generate the training scenes. CN-CBF replaces the joint value function with an aggregation of \emph{pairwise} value functions, which reduces the offline effort from tens of hours to minutes at the cost of the approximation analyzed in \cref{subsec:cn-cbf-composition,sec:cn-cbf-discussion}.

The second relevant line of work composes several barrier functions into a single one, and it divides into two groups. The first group studies the composition itself. Boolean compositions of safety specifications lead to nonsmooth barriers based on pointwise minima and maxima \cite{breeden_compositions_2023}, and their smooth log-sum-exp relaxations preserve differentiability and provide explicit conditions under which the composed function is again a valid CBF \cite{lindemann_control_2019,molnar_composing_2023,rabiee_soft-minimum_2025}. The second group applies such compositions to concrete systems. A composite CBF tailored to quadrotor flight handles only static environments \cite{harms_safe_2025}, and time-varying soft-maximum barriers yield closed-form safe controllers but do not impose actuation limits \cite{safari_time-varying_2024}. Sequential neural barriers learn robot--obstacle interaction dynamics from data, but they aggregate the learned components nonsmoothly, which complicates training and produces conservative safe sets \cite{yu_sequential_2023}. CN-CBF adopts the smooth composition of the first group, while its pairwise barriers are reachability-supervised residual models rather than the geometric or data-driven barriers of the second. Each component thus approximates the maximal pairwise safe set under actuation limits and excludes its failure set by construction, and a single shared model serves every obstacle.

\section{Problem Statement}
\label{sec:cn-cbf-problem-statement}

This section formalizes navigation among moving obstacles within the safety-filtering framework of \cref{sec:prelim-cbf}. The goal is to design a barrier candidate that depends on the robot state and on a varying number of obstacle states and is inexpensive enough for the CBF-QP to run at control rate.

\paragraph{Navigation setting.}
The robot is described by the control-affine continuous-time model of \cref{sec:orn-cbf-problem-statement},
\begin{equation}
  \dot{x}
  =
  f_0(x)+g(x)u,
  \qquad
  x\in\mathcal{X},\quad
  u\in\mathcal{U},
  \label{eq:cn-cbf-dynamics}
\end{equation}
with full state feedback and a nominal controller of arbitrary design whose command \(u_{\mathrm{nom}}(t)\) is not assumed to be safe. The environment contains \(m\in\mathbb{N}_0\) moving obstacles indexed by \(i\), and this count may change as obstacles enter and leave the sensing range. Obstacle \(i\) is described by a state \(q_i\in\mathcal{Q}\subseteq\mathbb{R}^{n_q}\), which typically contains position and velocity, and follows an \emph{assumed motion model},
\begin{equation}
  \dot{q}
  =
  \nu(q,d),
  \qquad
  d\in\mathcal{D},
  \label{eq:cn-cbf-obstacle-dynamics}
\end{equation}
where \(d\) is the obstacle input, such as a turning rate or an acceleration. The compact set \(\mathcal{D}\) bounds the motion capabilities attributed to the obstacle. Model \cref{eq:cn-cbf-obstacle-dynamics} plays the role of the disturbance model in \cref{subsec:prelim-continuous-time-systems}, since the obstacle may realize any admissible input signal and safety must hold against all of them. At update instants \(t_j\), separated by the period \(\Delta t_o\), a perception module estimates every obstacle state within detection range. The observation at update \(j\) is the tuple
\begin{equation}
  o_j
  =
  \bigl(q_1,\ldots,q_{m}\bigr),
  \label{eq:cn-cbf-observation}
\end{equation}
which is an observation in the sense of \cref{eq:prelim-observation-model}, expressed as a list of obstacle states rather than as an occupancy grid. Because the obstacle states are given in a fixed frame, this observation depends on time and not on the robot state, which is the second case of \cref{subsec:prelim-observation-model}. Unlike in the preceding chapters, the perception output is used without further processing, so the conditioning input coincides with the observation, \(w=o\), and enters the learned barrier through the relative states introduced in \cref{subsec:cn-cbf-relative-dynamics}. For \(m=0\), the tuple is empty and no moving-obstacle barrier constraint is imposed. A collision occurs when the robot--obstacle distance is smaller than their combined footprint radius, and the full failure set is the union of these pairwise collision configurations.

\paragraph{Design problem.}
The goal is to construct a barrier candidate that is continuously differentiable on the collision-free operating region away from any geometric singularities and that
\begin{enumerate}
  \item enforces collision avoidance with respect to \emph{all} currently observed obstacles simultaneously, accounting for their motion,
  \item approximates, for each robot--obstacle pair, the maximal safe set obtainable by HJ reachability under the robot's actuation limits and the assumed obstacle model \cref{eq:cn-cbf-obstacle-dynamics}, rather than the geometric obstacle distance alone,
  \item satisfies the \emph{failure-set exclusion} property by construction, so that no configuration in which the robot collides with a represented obstacle is ever predicted safe, regardless of the learning error, and
  \item handles any number of obstacles, including none, without retraining and without fixed-dimension scene encodings, and is inexpensive enough to evaluate and to differentiate for the CBF-QP \cref{eq:prelim-cbf-qp} to run at control rates of hundreds of hertz.
\end{enumerate}
The second and third requirements are the same as in \cref{ch:orn-cbf}, while the first and the fourth are the dynamic-environment demands that the previous chapter explicitly excluded.

\paragraph{Standing assumptions.}
The setting rests on four assumptions. First, the perception module provides accurate states for every obstacle within detection range, while perception itself lies outside the scope of this dissertation. Second, the obstacles are non-interactive and follow the model class \cref{eq:cn-cbf-obstacle-dynamics}, which means that they do not yield to the robot and that their inputs remain within \(\mathcal{D}\). The choice of these bounds trades robustness against conservatism, and \cref{subsec:cn-cbf-models} discusses their calibration from recorded motion. Third, the static geometry of the evaluated scenarios does not restrict the motion of the robot, and \cref{sec:cn-cbf-discussion} explains how CN-CBF can be composed with a static-environment barrier such as ORN-CBF. Fourth, the detection range, the update period, and the robot and obstacle speeds satisfy the cross-update condition in \cref{subsec:cn-cbf-filter}, so that a newly detected obstacle does not make the current robot state unsafe when it enters the composition.

\section{Composite Neural Barriers in Relative Coordinates}
\label{sec:cn-cbf-method}

The method defines a stationary pairwise reachability problem, learns a shared residual barrier, and derives its smooth composition and time-varying CBF condition. The final formulation includes obstacle-update conditions and the scope of its guarantees.

\subsection{Relative Dynamics and Pairwise Reachability}
\label{subsec:cn-cbf-relative-dynamics}

Consider first a single obstacle with state \(q\) and assumed dynamics \cref{eq:cn-cbf-obstacle-dynamics}. A direct application of \cref{subsec:prelim-time-varying-hj} would track the predicted obstacle motion with a time-varying constraint \(c(x,t)\), as in \cref{ch:rntc-mpc}, and the resulting value function would require recomputation whenever the prediction changes. CN-CBF instead uses the property that a collision depends on the robot--obstacle configuration and not on the absolute location of the encounter, so the pair can be described by a relative state. Formally, let a continuously differentiable map
\begin{equation}
  \sigma:
  \mathcal{X}\times\mathcal{Q}
  \rightarrow
  \mathcal{Z}\subseteq\mathbb{R}^{n_z},
  \qquad
  z
  =
  \sigma(x,q),
  \label{eq:cn-cbf-relative-map}
\end{equation}
extract the \emph{relative state} \(z\) from the robot and obstacle states. It typically contains the relative position, the relative orientation or velocity, and any obstacle quantity that affects the future relative motion. The map \(\sigma\) is a design choice tied to the concrete robot and obstacle models, for which \cref{subsec:cn-cbf-models} gives two concrete examples. Differentiating \cref{eq:cn-cbf-relative-map} along the robot and obstacle dynamics yields the \emph{relative dynamics}
\begin{equation}
  \dot{z}
  =
  \frac{\partial\sigma}{\partial x}(x,q)\,\dot{x}
  +
  \frac{\partial\sigma}{\partial q}(x,q)\,\dot{q}
  =:
  f_{\mathrm{rel}}(z,u,d),
  \label{eq:cn-cbf-relative-dynamics}
\end{equation}
where the defining requirement on \(\sigma\) is that the right-hand side can be expressed as a function of the relative state alone.\footnote{This requirement constrains the choice of coordinates. For the ground robot of \cref{subsec:cn-cbf-models}, for example, the relative position must be expressed in the body frame of the robot, while in the world frame the same quantities would not form a self-contained system in \(z\).} The relative system \cref{eq:cn-cbf-relative-dynamics} has the two-input structure of \cref{eq:prelim-general-continuous-dynamics}, in which the robot input \(u\) drives the relative state while the obstacle input \(d\) perturbs it within the assumed motion bounds.

In relative coordinates, the collision specification becomes stationary. Let \(R_{\min}>0\) denote the sum of the robot and obstacle footprint radii, including any physical safety margin, and let \(p_{\mathrm{rel}}(z)\in\mathbb{R}^2\) denote the relative position contained in \(z\). The \emph{pairwise constraint function} and the associated failure set, following the conventions of \cref{sec:prelim-constraints-safety-conventions}, are
\begin{equation}
  c_{\mathrm{rel}}(z)
  :=
  \lVert p_{\mathrm{rel}}(z)\rVert_2
  -
  R_{\min},
  \qquad
  \mathcal{F}_{\mathrm{rel}}
  :=
  \{z\in\mathcal{Z}\mid c_{\mathrm{rel}}(z)<0\}.
  \label{eq:cn-cbf-relative-constraint}
\end{equation}
The set \(\mathcal{F}_{\mathrm{rel}}\) is a fixed cylinder around the origin of the relative-position coordinates, so that contact always corresponds to the same relative states, regardless of the absolute motion of the pair. The change of coordinates removes the time variation that prevented a direct stationary analysis.

The pairwise problem now has the robust-avoidance form of \cref{sec:prelim-hj-reachability}, which characterizes the relative states from which the robot can avoid \(\mathcal{F}_{\mathrm{rel}}\) indefinitely under every admissible obstacle behavior. The obstacle input takes the role of the adversarial player, formalized through the nonanticipative strategies of \cref{subsec:prelim-continuous-time-systems}. The stationary value function \(V_\infty:\mathcal{Z}\rightarrow\mathbb{R}\) of \cref{eq:prelim-stationary-hj-value} then solves \cref{eq:prelim-stationary-hji-vi} for \(f_{\mathrm{rel}}\) and \(c_{\mathrm{rel}}\), with the Hamiltonian \cref{eq:prelim-hji-hamiltonian} maximizing over \(u\in\mathcal{U}\) and minimizing over \(d\in\mathcal{D}\). Its nonnegative superlevel set
\begin{equation}
  \mathcal{S}^{\star}_{\mathrm{rel}}
  :=
  \{z\in\mathcal{Z}\mid V_\infty(z)\geq0\}
  \label{eq:cn-cbf-pairwise-maximal-safe-set}
\end{equation}
is the maximal robustly safe set of the pairwise problem, that is, the largest set of relative configurations from which collision with the modeled obstacle class remains avoidable within the actuation limits. Its complement is the pairwise backward reachable tube (BRT), the set of relative states from which the obstacle can force a collision, whose positional extent depends on the remaining components of the relative state, such as the relative heading and the obstacle speed. \Cref{fig:cn-cbf-slice} visualizes this set for one robot--obstacle pair.

The value function \(V_\infty\) is computed \emph{once} on a grid over the low-dimensional relative state space, and it depends only on the robot model, the obstacle model, and the two footprints, not on the number of obstacles, their trajectories, or the scene layout. \Cref{subsec:cn-cbf-training-eval} quantifies the resulting reduction in offline cost relative to \cref{ch:rntc-mpc}.

\subsection{Residual Pairwise Barrier Learning}
\label{subsec:cn-cbf-pairwise}

The numerical solver returns \(V_\infty\) as a table of values over the relative-state grid. Direct deployment of this table would require storing the complete grid, whose size is reported in \cref{subsec:cn-cbf-training-eval}, and interpolating both values and gradients for every obstacle at every query. A neural approximation is considerably more compact, and automatic differentiation supplies the gradients of the represented function wherever it is differentiable.

The residual structure of \cref{subsec:prelim-structure-preserving-residual} uses the ordering \(V_\infty(z)\leq c_{\mathrm{rel}}(z)\) to exclude collision states. The learned pairwise barrier is
\begin{equation}
  \hat{h}_{\Theta}(z)
  :=
  c_{\mathrm{rel}}(z)
  -
  \hat{r}_{\Theta}(z),
  \qquad
  \hat{r}_{\Theta}(z)\geq0,
  \label{eq:cn-cbf-pairwise-barrier}
\end{equation}
in which the residual \(\hat{r}_{\Theta}\) is a multilayer perceptron (MLP) with parameters \(\Theta\), sinusoidal hidden activations \cite{sitzmann_implicit_2020,bansal_deepreach_2021}, and the softplus output \cref{eq:orn-cbf-softplus}, which enforces the nonnegativity of the residual while keeping it everywhere differentiable. In contrast to ORN-CBF, CN-CBF obtains the relative constraint function \(c_{\mathrm{rel}}\) in closed form from \cref{eq:cn-cbf-relative-constraint}, so that neither a discretized SDF nor an interpolation is required. However, the Euclidean norm in \(c_{\mathrm{rel}}\) is not differentiable at \(p_{\mathrm{rel}}=0\). Because \(R_{\min}>0\), this singularity lies strictly inside the collision set and away from the safe-set boundary on which the filter is intended to operate. The pairwise barrier is smooth on \(p_{\mathrm{rel}}\neq0\), but a global continuous-differentiability claim would require a smooth regularization of the norm.

\paragraph{Failure-set exclusion guarantee.}
As in \cref{ch:orn-cbf}, the nonnegativity of the residual in \cref{eq:cn-cbf-pairwise-barrier} gives \(\hat{h}_{\Theta}(z)\leq c_{\mathrm{rel}}(z)\) for every \(z\) and yields, in the set notation of \cref{sec:prelim-cbf},
\begin{equation}
  \mathcal{S}_{\hat{h}}
  :=
  \bigl\{z\mid\hat{h}_{\Theta}(z)\geq0\bigr\}
  \subseteq
  \bigl\{z\mid c_{\mathrm{rel}}(z)\geq0\bigr\}
  =
  \mathcal{Z}\setminus\mathcal{F}_{\mathrm{rel}}.
  \label{eq:cn-cbf-pairwise-exclusion}
\end{equation}
The entire predicted safe set therefore has a nonnegative distance to the obstacle, regardless of the residual approximation error, while the property remains one-sided as in the preceding chapters.

\paragraph{Shared network architecture and training.}
The relative state \(z\) contains the position and motion information needed for pairwise safety, so a fixed MLP with a few thousand parameters replaces the earlier hypernetworks, and one batched forward pass applies this shared model to all \(m\) obstacles. Training is the simplest instance of the offline pipeline in \cref{subsec:prelim-offline-online-safety-learning}. A single numerical table of \(V_\infty\) over the relative-state grid supervises the assembled barrier \cref{eq:cn-cbf-pairwise-barrier} through the mean squared error, and no dataset of environment observations needs to be constructed. \Cref{subsec:cn-cbf-training-eval} states the exact configuration and timings and compares them with the cost of the joint approach in \cref{ch:rntc-mpc}.

\subsection{Smooth Barrier Composition}
\label{subsec:cn-cbf-composition}

For \(o=(q_1,\ldots,q_m)\), let \(z_i=\sigma(x,q_i)\). The pairwise requirement \(\hat{h}_{\Theta}(z_i)\geq0\) for every obstacle is represented by \(\min_i\hat{h}_{\Theta}(z_i)\) \cite{breeden_compositions_2023}. This minimum is nonsmooth where the active obstacle changes, producing abrupt gradient changes during avoidance. Pairwise conditions also need not imply joint safety, as discussed below.

CN-CBF therefore aggregates the pairwise barriers with a smooth under-approximation of the minimum, the negated log-sum-exp function \cite{lindemann_control_2019,molnar_composing_2023}. For an obstacle count \(m\geq1\) and a smoothing parameter \(\beta>0\), define \(\mu_\beta:\mathbb{R}^m\rightarrow\mathbb{R}\),
\begin{equation}
  \mu_\beta(s_1,\ldots,s_m)
  :=
  -\frac{1}{\beta}
  \ln
  \sum_{i=1}^{m}
  e^{-\beta s_i},
  \label{eq:cn-cbf-softmin}
\end{equation}
and the \emph{composite neural barrier}
\begin{equation}
  \hat{h}_{\mathrm{cn}}(x,o)
  :=
  \mu_\beta
  \Bigl(
  \hat{h}_{\Theta}\bigl(\sigma(x,q_1)\bigr),
  \ldots,
  \hat{h}_{\Theta}\bigl(\sigma(x,q_m)\bigr)
  \Bigr).
  \label{eq:cn-cbf-composite-barrier}
\end{equation}
The subscript distinguishes the composite field from the shared pairwise barrier. Obstacle states remain time-varying arguments, so \cref{subsec:cn-cbf-cbf-condition} differentiates the field as a joint function of state and observation. For \(m=0\), the composite is undefined and its constraint is omitted, and all composition formulas below assume \(m\geq1\).

\begin{figure}[tb]
  \centering
  \colorlet{obsAline}{blue!55!black}
\colorlet{obsBline}{orange!85!black}
\colorlet{softline}{green!45!black}
\colorlet{softfill}{green!55!black!20}
\colorlet{minline}{black!60}
\colorlet{minfill}{black!13}

\begin{tikzpicture}[font=\small, line cap=round, line join=round,
                    >={Latex[length=1.9mm]}]

  \begin{scope}[xscale=1.35, yscale=1.75]

    \fill[minfill] (0.8889,0) -- (2,0.5) -- (3.1111,0) -- cycle;
    \fill[softfill] (1.0845,0) --
      plot[domain=1.0845:2.9155, samples=90, variable=\s]
        ({\s},{-0.5*ln(exp(-2.8+0.9*\s)+exp(0.8-0.9*\s))}) -- cycle;

    \draw[->] (-0.12,0) -- (4.40,0) node[below, inner sep=2.5pt] {$x$};
    \draw[->] (0,-0.72) -- (0,2.05);

    \draw[obsAline, thin] (0,1.4) -- (4,-0.4);
    \draw[obsBline, thin] (0,-0.4) -- (4,1.4);
    \node[obsAline, anchor=south west, inner sep=1pt] at (0.06,1.56)
      {$\hat h_{\Theta}(z_1)$};
    \node[obsBline, anchor=south east, inner sep=1pt] at (3.94,1.56)
      {$\hat h_{\Theta}(z_2)$};

    \draw[minline, semithick, densely dashed] (0,-0.4) -- (2,0.5) -- (4,-0.4);
    \draw[softline, very thick]
      plot[domain=0:4, samples=160, variable=\s]
        ({\s},{-0.5*ln(exp(-2.8+0.9*\s)+exp(0.8-0.9*\s))});

    \draw[{Latex[length=1.5mm]}-{Latex[length=1.5mm]}, semithick]
      (2,0.5) -- (2,0.1534);
    \draw[black!60, thin] (2.04,0.327) -- (2.46,0.585);
    \node[anchor=west, font=\scriptsize, inner sep=1pt] at (2.48,0.60)
      {$\tfrac{\ln m}{\beta}$};

    \draw[minline, fill=white] (0.8889,0) circle (1.6pt);
    \draw[minline, fill=white] (3.1111,0) circle (1.6pt);
    \fill[softline] (1.0845,0) circle (1.6pt);
    \fill[softline] (2.9155,0) circle (1.6pt);

    \draw (2,0) -- (2,-0.07);
    \node[anchor=north, font=\scriptsize, inner sep=1.5pt] at (2,-0.06)
      {$\bar{x}$};

    \filldraw[fill=white, draw=black!20, rounded corners=2pt]
      (1.22,0.96) rectangle (2.98,1.50);
    \draw[minline, semithick, densely dashed] (1.30,1.35) -- (1.62,1.35);
    \node[anchor=west, font=\scriptsize, inner sep=1.5pt] at (1.66,1.35)
      {$\min_i\hat h_{\Theta}(z_i)$};
    \draw[softline, very thick] (1.30,1.11) -- (1.62,1.11);
    \node[anchor=west, font=\scriptsize, inner sep=1.5pt] at (1.66,1.11)
      {$\hat h_{\mathrm{cn}}=\mu_\beta$};

    \node[anchor=west, font=\scriptsize, inner sep=1pt] at (0.08,-0.56)
      {$\beta=2$, $m=2$};
  \end{scope}

  \begin{scope}[shift={(7.90cm,0cm)}, xscale=1.35, yscale=2.90]

    \draw[black!30, dotted] (0,0.5) -- (4.20,0.5);
    \draw[black!30, dotted] (2,0) -- (2,1.05);

    \draw[->] (-0.12,0) -- (4.40,0) node[below, inner sep=2.5pt] {$x$};
    \draw[->] (0,-0.04) -- (0,1.18);
    \foreach \l in {0.5,1}{
      \draw (0,\l) -- (-0.05,\l)
        node[anchor=east, font=\scriptsize, inner sep=2pt] {$\l$};}
    \node[anchor=north east, font=\scriptsize, inner sep=2.5pt] at (0,0) {$0$};

    \draw[obsBline, thin, densely dashed]
      plot[domain=0:4, samples=220, variable=\s]
        ({\s},{1-1/(1+exp(max(-9,min(9,8*(1.8-0.9*\s)))))});
    \draw[obsAline, thin, densely dashed]
      plot[domain=0:4, samples=220, variable=\s]
        ({\s},{1/(1+exp(max(-9,min(9,8*(1.8-0.9*\s)))))});
    \draw[obsBline, very thick]
      plot[domain=0:4, samples=160, variable=\s]
        ({\s},{1-1/(1+exp(2*(1.8-0.9*\s)))});
    \draw[obsAline, very thick]
      plot[domain=0:4, samples=160, variable=\s]
        ({\s},{1/(1+exp(2*(1.8-0.9*\s)))});

    \node[obsBline, anchor=south west, inner sep=1pt] at (0.06,0.99)
      {$\lambda_2$};
    \node[obsAline, anchor=south east, inner sep=1pt] at (3.94,0.99)
      {$\lambda_1$};

    \draw (2,0) -- (2,-0.04);
    \node[anchor=north, font=\scriptsize, inner sep=1.5pt] at (2,-0.035)
      {$\bar{x}$};

    \node[anchor=center, font=\scriptsize, align=left, inner sep=1.5pt]
      at (1.02,0.62) {solid: $\beta=2$\\dashed: $\beta=8$};
  \end{scope}

  \node[anchor=south west, inner sep=0pt] at (-0.30,3.72)
    {(a)~barrier value along the slice};
  \node[anchor=south west, inner sep=0pt] at (7.60,3.72)
    {(b)~softmax weights $\lambda_i$};
\end{tikzpicture}
  \caption[Smooth composition of two pairwise barriers]{Smooth composition of two pairwise barriers along a one-dimensional slice of the state space, for example a straight path from one obstacle to the other. (a)~The two pairwise barrier values (blue and orange) form the nonsmooth pointwise minimum (dashed), whose kink at \(\bar{x}\) marks the change of the active obstacle. The composite barrier \(\hat{h}_{\mathrm{cn}}=\mu_\beta\) (green) runs below that minimum, with a gap that is largest at the kink, where it attains the bound \(\ln m/\beta\) of \cref{eq:cn-cbf-softmin-bounds}. The composite safe interval (green) is slightly smaller than the one given by the exact minimum (gray). (b)~The softmax weights \cref{eq:cn-cbf-softmin-weights} along the same slice transfer the gradient continuously from one obstacle to the other instead of switching at \(\bar{x}\), and a larger \(\beta\) narrows this transition.}
  \label{fig:cn-cbf-softmin}
\end{figure}
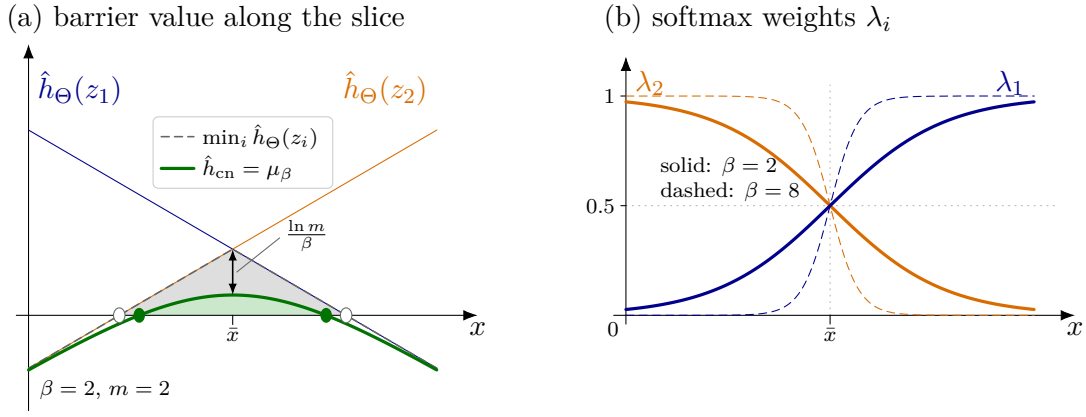

The composition has three properties on which the construction rests, and all of them follow directly from \cref{eq:cn-cbf-softmin}. First, it is an \emph{under-approximation with a bounded gap},
\begin{equation}
  \min_{i}s_i
  -
  \frac{\ln m}{\beta}
  \;\leq\;
  \mu_\beta(s_1,\ldots,s_m)
  \;\leq\;
  \min_{i}s_i.
  \label{eq:cn-cbf-softmin-bounds}
\end{equation}
The composite safe set \(\{x\mid\hat{h}_{\mathrm{cn}}(x,o)\geq0\}\) lies inside the intersection of the pairwise safe sets, so the smoothing is conservative rather than optimistic, and its largest value offset \(\ln m/\beta\) grows logarithmically with the obstacle count. Second, the composite barrier \emph{inherits failure-set exclusion}. The upper bound in \cref{eq:cn-cbf-softmin-bounds} shows that \(\hat{h}_{\mathrm{cn}}(x,o)\geq0\) implies \(\hat{h}_{\Theta}(z_i)\geq0\) for every \(i\), and \cref{eq:cn-cbf-pairwise-exclusion} then gives \(c_{\mathrm{rel}}(z_i)\geq0\) for every obstacle. Every configuration in the composite safe set consequently keeps a nonnegative distance to all represented obstacles, regardless of the learning error. Third, the composition is smooth, and its partial derivatives
\begin{equation}
  \frac{\partial\mu_\beta}{\partial s_i}
  =
  \frac{e^{-\beta s_i}}{\sum_{l=1}^{m}e^{-\beta s_l}}
  =:
  \lambda_i,
  \qquad
  \lambda_i>0,
  \quad
  \sum_{i=1}^{m}\lambda_i=1,
  \label{eq:cn-cbf-softmin-weights}
\end{equation}
are softmax weights concentrated on the smallest arguments. The composite gradient is thus a \emph{convex combination of the pairwise gradients}, dominated by the most critical obstacles, and it shifts continuously as the active obstacle changes instead of switching abruptly. The parameter \(\beta\) controls the trade-off between closeness to the exact minimum and smoothness of the composition. A large value tracks the exact minimum closely but approaches its nonsmooth switching behavior, while a small value incorporates more obstacles into the gradient and adds the conservatism bounded in \cref{eq:cn-cbf-softmin-bounds}. \Cref{fig:cn-cbf-softmin} illustrates both effects on a one-dimensional slice between two obstacles.

The composition has one structural limitation. The intersection of the pairwise \emph{maximal} safe sets is only an outer approximation of the true joint maximal safe set, because pairwise avoidability does not imply joint avoidability when the maneuvers that evade one obstacle conflict with those needed for another. The soft minimum is therefore conservative relative to the pairwise intersection by \cref{eq:cn-cbf-softmin-bounds}, while the intersection itself can be \emph{optimistic} relative to the joint reachability problem of \cref{ch:rntc-mpc}. \Cref{sec:cn-cbf-discussion} returns to this trade-off between modularity and joint optimality.

\subsection{Time-Varying Composite Barrier Condition}
\label{subsec:cn-cbf-cbf-condition}

The composite barrier \cref{eq:cn-cbf-composite-barrier} depends on both the robot state and the obstacle states, and because the obstacles move, this dependence introduces an explicit time variation. The setting is the case \(o=o(t)\) of \cref{subsec:prelim-observation-time-derivative}, in which the obstacle motion is independent of the robot state and control. The ideal continuous-time condition is derived first, assuming that the obstacle trajectories \(q_i(t)\) and their derivatives are available continuously. Along the closed-loop evolution, the barrier value
\begin{equation}
  h(x,t)
  :=
  \hat{h}_{\mathrm{cn}}
  \bigl(x,q_1(t),\ldots,q_m(t)\bigr)
  \label{eq:cn-cbf-time-varying-barrier}
\end{equation}
is a time-varying barrier in the sense of \cref{subsec:prelim-time-varying-robust-cbfs}, and its explicit time dependence originates solely from the obstacle motion. The chain rule applied through \cref{eq:cn-cbf-composite-barrier,eq:cn-cbf-relative-map}, with the weights \cref{eq:cn-cbf-softmin-weights} evaluated at \(s_i=\hat{h}_{\Theta}(z_i)\), decomposes both derivative terms over the obstacles,
\begin{subequations}
  \label{eq:cn-cbf-derivatives}
  \begin{align}
    \nabla_xh(x,t)
     & =
    \sum_{i=1}^{m}
    \lambda_i
    \frac{\partial\sigma}{\partial x}(x,q_i)^{\top}
    \nabla_z\hat{h}_{\Theta}(z_i),
    \label{eq:cn-cbf-derivatives-state}
    \\
    \frac{\partial h}{\partial t}(x,t)
     & =
    \sum_{i=1}^{m}
    \lambda_i\,
    \nabla_z\hat{h}_{\Theta}(z_i)^{\top}
    \frac{\partial\sigma}{\partial q}(x,q_i)\,
    \dot{q}_i.
    \label{eq:cn-cbf-derivatives-time}
  \end{align}
\end{subequations}
The state gradient describes changes due to robot motion, and the partial time derivative describes changes due to obstacle motion, and both weight the most critical obstacles most strongly. If all obstacles are static, \(\dot{q}_i=0\) and the condition reduces to the time-invariant form.

In practice, the obstacle states arrive at discrete observation updates rather than continuously, and the derivatives \(\dot{q}_i\) in \cref{eq:cn-cbf-derivatives-time} must be estimated. When the perception module does not supply them directly, a backward difference over the observation period gives the estimate
\begin{equation}
  \dot{q}_i(t_j)
  \approx
  \frac{q_i(t_j)-q_i(t_{j-1})}{\Delta t_o}.
  \label{eq:cn-cbf-backward-difference}
\end{equation}
The deployed filter evaluates the derivative term with the most recent state and derivative estimates.

Using the realized obstacle motion here, instead of the worst case over all admissible obstacle inputs, is a deliberate design choice. The pairwise barrier approximates the robust value function of \cref{subsec:cn-cbf-relative-dynamics}, whose Hamiltonian maximizes over \(u\in\mathcal{U}\) and minimizes over \(d\in\mathcal{D}\). As shown in \cref{sec:prelim-hj-cbf-relation}, this Hamiltonian is nonnegative, so for every relative state in the pairwise safe set \cref{eq:cn-cbf-pairwise-maximal-safe-set} at which \(V_\infty\) is differentiable, the exact value function satisfies
\begin{equation}
  \sup_{u\in\mathcal{U}}
  \left[
    \nabla_zV_\infty(z)^{\top}
    f_{\mathrm{rel}}(z,u,\bar{d})
    \right]
  \geq
  \sup_{u\in\mathcal{U}}
  \min_{d\in\mathcal{D}}
  \left[
    \nabla_zV_\infty(z)^{\top}
    f_{\mathrm{rel}}(z,u,d)
    \right]
  \geq
  -\alpha\bigl(V_\infty(z)\bigr)
  \label{eq:cn-cbf-realized-disturbance}
\end{equation}
for every admissible obstacle input \(\bar{d}\in\mathcal{D}\), because the value attained for one input is at least the minimum over all of them. The robustness is thus contained in the value function, which certifies that an admissible control exists for every obstacle behavior within the modeled bounds. The realized obstacle motion, in turn, supplies the actual time derivative of the barrier along the closed-loop trajectory, which is what the invariance argument of \cref{sec:prelim-cbf} requires. Enforcing the condition with a worst-case obstacle input would be sufficient but unnecessarily conservative, because it would constrain the robot against a behavior that the obstacle is not currently exhibiting.

\subsection{CN-CBF Safety Filter}
\label{subsec:cn-cbf-filter}

At every control instant, CN-CBF receives the current robot state \(x\), the nominal command \(u_{\mathrm{nom}}\), and the latest obstacle estimates, and it solves the CBF-QP of \cref{subsec:prelim-cbf-qp-safety-filter} with the time-varying constraint \cref{eq:prelim-time-varying-cbf-condition},
\begin{subequations}
  \label{eq:cn-cbf-qp}
  \begin{align}
    u^{\star}
    =
    \operatorname*{arg\,min}_{u\in\mathcal{U}}
    \quad               &
    \frac{1}{2}\lVert u-u_{\mathrm{nom}}\rVert_2^{2}
    \label{eq:cn-cbf-qp-objective}
    \\
    \text{ s.t. }\quad  &
    \nabla_xh(x,t)^{\top}
    \bigl(f_0(x)+g(x)u\bigr)
    +
    \frac{\partial h}{\partial t}(x,t)
    +
    \gamma\,h(x,t)
    \geq0,
    \label{eq:cn-cbf-qp-constraint}
  \end{align}
\end{subequations}
where the extended class-\(\mathcal{K}\) function is chosen linear with gain \(\gamma>0\), as in \cref{ch:orn-cbf}. For \(m\geq1\), the constraint is affine in \(u\), and linear inequalities describe \(\mathcal{U}\) for both robots considered here, so the problem is a single small convex QP regardless of the obstacle count. The entire scene enters this QP only through \(h\), \(\nabla_xh\), and \(\partial h/\partial t\). For \(m=0\), the obstacle constraint is absent and the nominal command is subject only to \(\mathcal{U}\).

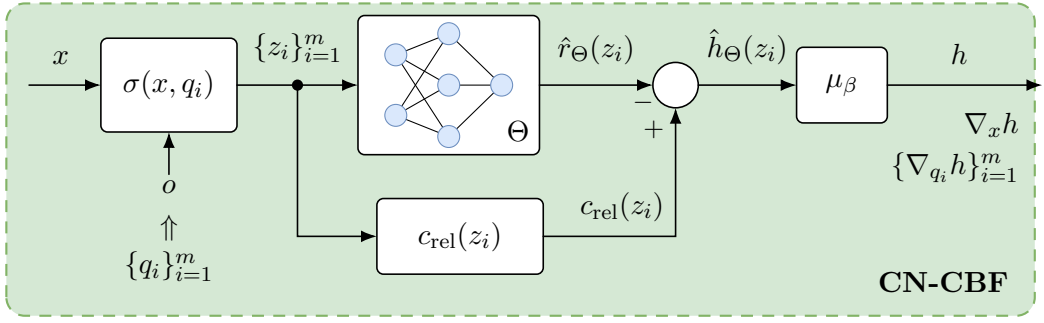
\begin{figure}[tb]
  \centering
  \definecolor{archbg}{HTML}{D5E8D4}
\definecolor{archframe}{HTML}{82B366}
\definecolor{nnfill}{HTML}{DAE8FC}
\definecolor{nndraw}{HTML}{6C8EBF}

\begin{tikzpicture}[
  font=\small,
  >={Latex[length=2.1mm]},
  sig/.style={->, semithick, line cap=round},
  blk/.style={draw, semithick, fill=white, rounded corners=2.5pt,
              inner sep=0pt, align=center},
  sum/.style={draw, semithick, fill=white, circle, minimum size=0.6cm,
              inner sep=0pt},
  slab/.style={inner sep=1.5pt},
  sign/.style={font=\footnotesize, inner sep=1pt},
  ]

  \draw[rounded corners=7pt, dash pattern=on 4.5pt off 3.5pt, thick,
        draw=archframe, fill=archbg]
    (-0.30,-3.05) rectangle (13.30,1.08);
  \node[anchor=south east, font=\small\bfseries] at (13.10,-2.88) {CN-CBF};

  \draw[sig] (0.00,0) -- (0.95,0);
  \node[slab, anchor=south] at (0.42,0.22) {$x$};

  \node[blk, minimum width=1.80cm, minimum height=1.24cm] (sigma) at (1.85,0)
    {$\sigma(x,q_i)$};

  \draw[sig] (1.85,-1.18) -- (1.85,-0.62);
  \node[slab, anchor=north] at (1.85,-1.20) {$o$};
  \node[slab, anchor=north] at (1.85,-1.66) {$\Uparrow$};
  \node[slab, anchor=north] at (1.85,-2.14) {$\{q_i\}_{i=1}^{m}$};

  \draw[sig] (2.75,0) -- (4.35,0);
  \fill (3.55,0) circle (2.1pt);
  \node[slab, anchor=south] at (3.55,0.24) {$\{z_i\}_{i=1}^{m}$};

  \node[blk, minimum width=2.40cm, minimum height=1.84cm] (net) at (5.55,0) {};
  \foreach \yi in {0.40,-0.40}{
    \foreach \yh in {0.68,0,-0.68}{
      \draw[thin] (4.85,\yi) -- (5.55,\yh);}}
  \foreach \yh in {0.68,0,-0.68}{
    \draw[thin] (5.55,\yh) -- (6.25,0);}
  \foreach \yi in {0.40,-0.40}{
    \filldraw[fill=nnfill, draw=nndraw, thin] (4.85,\yi) circle (0.145);}
  \foreach \yh in {0.68,0,-0.68}{
    \filldraw[fill=nnfill, draw=nndraw, thin] (5.55,\yh) circle (0.145);}
  \filldraw[fill=nnfill, draw=nndraw, thin] (6.25,0) circle (0.145);
  \node[slab] at (6.47,-0.60) {$\Theta$};

  \node[blk, minimum width=2.20cm, minimum height=1.00cm] (crel) at (5.70,-2.00)
    {$c_{\mathrm{rel}}(z_i)$};
  \draw[sig] (3.55,0) |- (4.60,-2.00);

  \node[sum] (add) at (8.55,0) {};
  \draw[sig] (6.75,0) -- (8.25,0);
  \node[slab, anchor=south] at (7.50,0.22) {$\hat{r}_{\Theta}(z_i)$};
  \node[sign, anchor=north east] at (8.30,-0.07) {$-$};

  \draw[sig] (6.80,-2.00) -| (8.55,-0.30);
  \node[slab, anchor=south east] at (8.46,-1.86) {$c_{\mathrm{rel}}(z_i)$};
  \node[sign, anchor=east] at (8.44,-0.58) {$+$};

  \node[blk, minimum width=1.20cm, minimum height=1.00cm] (mu) at (10.75,0)
    {$\mu_\beta$};
  \draw[sig] (8.85,0) -- (10.15,0);
  \node[slab, anchor=south] at (9.50,0.22) {$\hat{h}_{\Theta}(z_i)$};

  \draw[sig] (11.35,0) -- (13.40,0);
  \node[slab, anchor=south] at (12.30,0.22) {$h$};
  \node[slab, anchor=north east, align=right] at (13.15,-0.30)
    {$\nabla_{x}h$\\[2pt]$\{\nabla_{q_i}h\}_{i=1}^{m}$};
\end{tikzpicture}
  \caption[Structure of the CN-CBF barrier evaluation]{Structure of the CN-CBF barrier evaluation. The current obstacle states are stacked into a batch, combined with the robot state \(x\) into the batch of relative states \(z_i=\sigma(x,q_i)\), and propagated in parallel through the shared residual network \(\hat{r}_{\Theta}\) and the closed-form pairwise constraint \(c_{\mathrm{rel}}\). Their difference yields the pairwise barrier values of \cref{eq:cn-cbf-pairwise-barrier}, which the smooth aggregation \(\mu_\beta\) reduces to the composite value in \cref{eq:cn-cbf-composite-barrier}. One reverse-mode differentiation pass through this graph returns all gradients in \cref{eq:cn-cbf-derivatives}.}
  \label{fig:cn-cbf-arch}
\end{figure}

\begin{figure}[tb]
  \centering
  \definecolor{cnfill}{HTML}{D5E8D4}
\definecolor{cnframe}{HTML}{82B366}

\begin{tikzpicture}[
  font=\small,
  >={Latex[length=2.1mm]},
  sig/.style={->, semithick, line cap=round, line join=round},
  blk/.style={draw, semithick, fill=white, rounded corners=2.5pt,
              inner sep=0pt, align=center, minimum height=1.30cm},
  slab/.style={inner sep=1.5pt},
  ]

  \draw[rounded corners=8pt, dash pattern=on 4.5pt off 3.5pt, thick, draw=black]
    (0.35,-4.35) rectangle (11.70,2.55);

  \node[blk, minimum width=2.80cm] (dyn) at (2.25,0) {Robot\\dynamics};
  \node[blk, minimum width=2.60cm, fill=cnfill, draw=cnframe] (cbf) at (5.80,0)
    {CN-CBF};
  \node[blk, minimum width=3.50cm] (est) at (9.50,0)
    {Backward-difference\\estimator};
  \node[blk, minimum width=2.40cm] (qp) at (5.80,-3.30) {CBF-QP\\solver};

  \draw[sig] (5.15,2.20) -- (5.15,0.65);
  \draw[sig] (5.15,1.35) -- (2.25,1.35) -- (2.25,0.65);
  \fill (5.15,1.35) circle (2.1pt);
  \node[slab, anchor=east] at (5.02,1.95) {$x$};

  \draw[sig] (6.45,2.20) -- (6.45,0.65);
  \draw[sig] (6.45,1.35) -- (9.50,1.35) -- (9.50,0.65);
  \fill (6.45,1.35) circle (2.1pt);
  \node[slab, anchor=west] at (6.58,1.95) {$\{q_i\}_{i=1}^{m}$};

  \draw[sig] (2.25,-0.65) -- (2.25,-2.25) -- (5.15,-2.25) -- (5.15,-2.65);
  \node[slab, anchor=west] at (2.40,-1.75) {$f_0,\,g$};

  \draw[sig] (5.80,-0.65) -- (5.80,-2.65);
  \node[slab, anchor=north east] at (5.64,-0.80) {$h,\,\nabla_{x}h$};
  \node[slab, anchor=north west] at (5.96,-0.80) {$\{\nabla_{q_i}h\}_{i=1}^{m}$};

  \draw[sig] (9.50,-0.65) -- (9.50,-2.25) -- (6.45,-2.25) -- (6.45,-2.65);
  \node[slab, anchor=east] at (9.35,-1.75) {$\{\dot{q}_i\}_{i=1}^{m}$};

  \draw[sig] (3.55,-3.30) -- (4.60,-3.30);
  \node[slab, anchor=south] at (4.08,-3.12) {$u_{\mathrm{nom}}$};
  \draw[sig] (7.00,-3.30) -- (8.05,-3.30);
  \node[slab, anchor=south] at (7.52,-3.12) {$u^{\star}$};
\end{tikzpicture}
  \caption[Block diagram of the CN-CBF safety filter]{Block diagram of the CN-CBF safety filter. The robot state and the current obstacle-state estimates feed three parallel blocks. The batched CN-CBF evaluation of \cref{fig:cn-cbf-arch} returns the composite barrier value and its gradients, the robot dynamics model supplies \(f_0(x)\) and \(g(x)\), and the backward-difference estimator in \cref{eq:cn-cbf-backward-difference} supplies the obstacle state derivatives. Together they parameterize the CBF-QP \cref{eq:cn-cbf-qp}, which minimally modifies the nominal command \(u_{\mathrm{nom}}\).}
  \label{fig:cn-cbf-filter}
\end{figure}
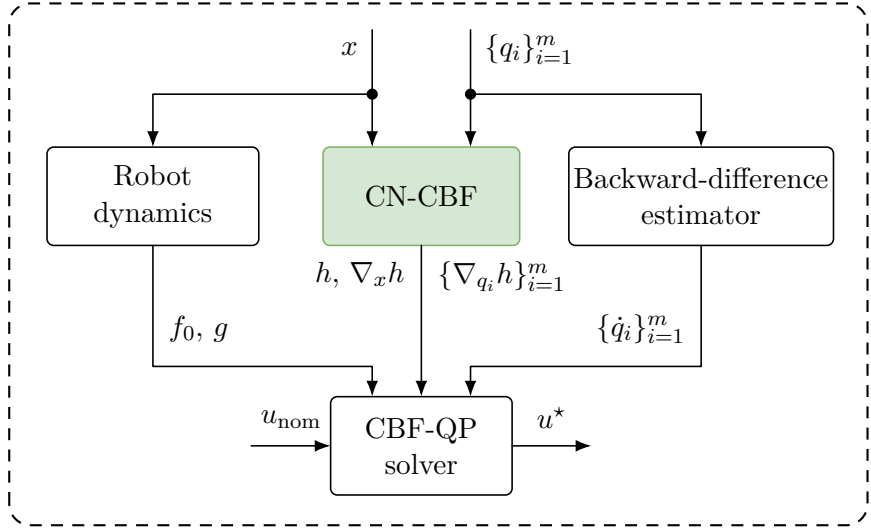

\paragraph{Batched evaluation.}
The composite barrier is assembled from elementary differentiable operations, so automatic differentiation supplies the derivatives \cref{eq:cn-cbf-derivatives}. The obstacle states are stacked into a matrix of shape \(m\times n_q\), the map \(\sigma\) and the pairwise barrier \cref{eq:cn-cbf-pairwise-barrier} operate along the batch dimension, and \cref{eq:cn-cbf-softmin} reduces the batch to a single scalar. One reverse-mode pass then returns the gradients with respect to the robot state and to every obstacle state, as visualized in \cref{fig:cn-cbf-arch}. Each control instant requires one batched forward and backward pass through a network with a few thousand parameters, followed by one QP solve whose size does not depend on \(m\). The amount of computation grows with the obstacle count, while the number of passes and the size of the QP remain fixed, which meets the fourth requirement of \cref{sec:cn-cbf-problem-statement}. \Cref{fig:cn-cbf-filter} shows the complete filter, in which the robot model and the obstacle-velocity estimation \cref{eq:cn-cbf-backward-difference} run alongside the barrier evaluation and together parameterize the QP.

\paragraph{Obstacle-count changes and cross-update safety.}
The number of terms in \cref{eq:cn-cbf-composite-barrier} follows the perception output, so that obstacles may appear or disappear between updates and change the composite barrier discontinuously. Removing an obstacle can only increase \(\mu_\beta\), and when the last obstacle disappears, the moving-obstacle constraint is bypassed entirely. Adding an obstacle, in contrast, always decreases the finite composite value, so continued safety requires the cross-update condition \cref{eq:prelim-cross-update-safety}, which demands that the current state belong to the safe set of the \emph{new} composition.

The soft minimum makes the required margin explicit. When an update both removes and adds obstacles, the terms of the disappearing obstacles are removed first, and \(h^-\) denotes the composite value over the retained terms, with the extended-real convention \(h^-=+\infty\) for an empty retained list. The update then adds \(k\geq1\) obstacles whose pairwise barrier values at the current robot state are \(s_1,\ldots,s_k\). The new composite value is
\begin{equation}
  h^+
  =
  -\frac{1}{\beta}
  \ln\!\left(
    e^{-\beta h^-}
    +
    \sum_{a=1}^{k}e^{-\beta s_a}
  \right).
  \label{eq:cn-cbf-count-update}
\end{equation}
Nonnegativity of every new pairwise barrier is consequently not sufficient, because even a distant obstacle causes a finite downward shift, which drives \(h^+\) below zero whenever \(h^-=0\). If the value before the addition satisfies \(h^-\geq\varepsilon>0\) and every new pairwise value satisfies \(s_a\geq s_{\min}\), then a sufficient condition for \(h^+\geq0\) is
\begin{equation}
  e^{-\beta\varepsilon}
  +
  k e^{-\beta s_{\min}}
  \leq 1,
  \qquad\text{equivalently}\qquad
  s_{\min}
  \geq
  \frac{1}{\beta}
  \ln\!\left(
    \frac{k}{1-e^{-\beta\varepsilon}}
  \right).
  \label{eq:cn-cbf-count-update-margin}
\end{equation}
The bound follows directly from the composition formula and introduces no empirical margin. At runtime, the most direct and least conservative test is to evaluate the new composition and to verify \(h^+\geq0\) before relying on the updated barrier. If this check fails, continued safety is not certified, and a deployment must specify a fallback response.

A sufficiently large detection range makes the newly added pairwise values large, but it does not replace the composite-margin test, because the influence of a distant obstacle decays exponentially with \(\beta\hat{h}_{\Theta}(z_i)\) according to \cref{eq:cn-cbf-softmin-weights} but is never exactly zero and can matter when the margin before the update is small. Let \(\rho_{\mathrm{det}}\) denote the detection range, let \(v_{\max}\) and \(v_{\max}^{\mathrm{o}}\) denote the maximum robot and obstacle speeds, and let \(\beta_{\mathcal{B}}^{\mathrm{rel}}\) denote the largest workspace distance by which the pairwise BRT extends beyond the collision radius \(R_{\min}\). Under boundary-only detection, absence of occlusion, the stated speed bounds, and the modeled obstacle class, a newly detected obstacle remains outside the \emph{exact} pairwise BRT whenever
\begin{equation}
  \rho_{\mathrm{det}}
  -
  \bigl(v_{\max}+v_{\max}^{\mathrm{o}}\bigr)\Delta t_o
  >
  R_{\min}+\beta_{\mathcal{B}}^{\mathrm{rel}}.
  \label{eq:cn-cbf-sensing-adequacy}
\end{equation}
Condition \cref{eq:cn-cbf-sensing-adequacy} is the concrete form of the sensing-adequacy requirement \cref{eq:prelim-sensing-adequacy} for the present configuration. The closing-speed term includes both agents, because the robot and an undetected approaching obstacle can reduce the gap simultaneously. The quantity \(\beta_{\mathcal{B}}^{\mathrm{rel}}\) can be determined from the computed pairwise BRT once \(R_{\min}\) and the numerical grid domain are specified. Occlusions, under which an obstacle can appear well inside the detection range, violate the geometric assumption.

\paragraph{Guarantee scope.}
Composite failure-set exclusion holds by construction, and the soft composition is conservative relative to the pairwise barriers and never optimistic. Validity of the learned barrier as a CBF is assessed experimentally rather than verified analytically, and because feasibility of the barrier constraint in \cref{eq:cn-cbf-qp} cannot be ensured in advance, the implementation uses the relaxed form \cref{eq:prelim-cbf-qp-slack}. The experiments assess the pairwise approximation and the closed-loop behavior empirically, and \cref{sec:cn-cbf-discussion} states the scope of these claims.

\section{Experimental Evaluation}
\label{sec:cn-cbf-experiments}

The evaluation measures offline cost, pairwise approximation, closed-loop performance against MPC and classical CBF baselines, and deployment with real perception. It uses a nonholonomic ground robot that cannot stop and a quadrotor with bounded acceleration.\footnote{The implementation is available at \url{https://github.com/bojan-derajic/CN-CBF}.}

\subsection{Robot and Obstacle Models}
\label{subsec:cn-cbf-models}

Both studies use the same type of environment. The robot moves through a planar workspace toward a goal position, and pedestrians walk through the same area without coordinating with the robot or avoiding it, and their number changes with the output of the perception system. Two robot--obstacle pairs are considered. The first pair combines a nonholonomic ground robot with a unicycle pedestrian model, and the second pair combines a constant-altitude quadrotor with a double-integrator pedestrian model. Each pair needs its own relative dynamics, its own pairwise HJ computation, and its own trained barrier, while the composition of \cref{subsec:cn-cbf-composition} and the filter of \cref{subsec:cn-cbf-filter} remain the same. In the quadrotor hardware experiment, obstacle drones take the role of the pedestrians.

\paragraph{Ground-robot model.}
The ground robot is modeled as the kinematic unicycle of \cref{eq:rntc-unicycle-dynamics}, with state \(x=(p_x,p_y,\theta)\), input \(u=(v,\omega)\), and limits \(v\in[0.25,1.0]\)~\si{\meter\per\second} and \(\omega\in[-0.8,0.8]\)~\si{\radian\per\second}. The positive lower speed bound reflects the deployment configuration and implies that the robot cannot stop, so, as for the Dubins car in \cref{ch:ntc-mpc,ch:orn-cbf}, the maximal safe set differs substantially from the geometric free space.

Each pedestrian is modeled as a unicycle as well. The state \(q=(q_x,q_y,\theta^{\mathrm{o}},v^{\mathrm{o}})\) contains the position, the heading, and the speed, and the input \(d=\omega^{\mathrm{o}}\in[-0.35,0.35]\)~\si{\radian\per\second} is the turning rate. The speed is constant in the model and covers \(v^{\mathrm{o}}\in[0,1.2]\)~\si{\meter\per\second}, so the model describes pedestrians who walk at an approximately constant speed while adjusting their direction. It is a robust version of the constant-velocity model, because the choice \(\omega^{\mathrm{o}}\equiv0\) recovers the prediction used in \cref{ch:rntc-mpc}, while the reachability analysis considers every turning behavior within the bound. The bounds themselves are a calibration choice. Recorded trajectories are used to estimate the distributions of walking speed and turning rate, and choosing \(\mathcal{D}\) and the speed range such that they cover most of these distributions trades robustness against the conservatism of wider bounds.

For this pair, the relative state is \(z=(p^{\mathrm{rel}}_x,p^{\mathrm{rel}}_y,\theta^{\mathrm{rel}},v^{\mathrm{o}})\). The first two components express the obstacle position in the body frame of the robot, and \(\theta^{\mathrm{rel}}=\theta^{\mathrm{o}}-\theta\) is the relative heading. The obstacle speed \(v^{\mathrm{o}}\) is included as well, because it affects the future relative motion. Differentiating as in \cref{eq:cn-cbf-relative-dynamics} yields the relative dynamics
\begin{equation}
  \dot{z}
  =
  f_{\mathrm{rel}}(z,u,d)
  =
  \begin{bmatrix}
    -v+v^{\mathrm{o}}\cos\theta^{\mathrm{rel}}+\omega\,p^{\mathrm{rel}}_y \\
    v^{\mathrm{o}}\sin\theta^{\mathrm{rel}}-\omega\,p^{\mathrm{rel}}_x   \\
    \omega^{\mathrm{o}}-\omega                                              \\
    0
  \end{bmatrix}.
  \label{eq:cn-cbf-relative-dynamics-gr}
\end{equation}
This four-dimensional system has the game structure of \cref{subsec:cn-cbf-relative-dynamics}, in which the robot input counters the turning input of the obstacle. The pairwise constraint is \cref{eq:cn-cbf-relative-constraint}, with \(R_{\min}=\SI{0.8}{\meter}\) as the sum of the robot and pedestrian footprint radii, and the pairwise value function is computed over relative positions \(p^{\mathrm{rel}}_x,p^{\mathrm{rel}}_y\in[-4,4]\)~\si{\meter}, all relative headings, and the pedestrian speed range above.

\paragraph{Quadrotor model.}
The constant-altitude quadrotor is modeled as the planar double integrator of \cref{eq:orn-cbf-double-integrator}, with state \(x=(p_x,p_y,v_x,v_y)\), input \(u=(a_x,a_y)\), and limits \(v_x,v_y\in[-2,2]\)~\si{\meter\per\second} and \(a_x,a_y\in[-2,2]\)~\si{\meter\per\square\second}. The pedestrians are modeled as planar double integrators as well, with state \(q=(q_x,q_y,v^{\mathrm{o}}_x,v^{\mathrm{o}}_y)\), input \(d=(a^{\mathrm{o}}_x,a^{\mathrm{o}}_y)\), velocity limits of \([-1.5,1.5]\)~\si{\meter\per\second} per axis, and acceleration limits of \([-1.0,1.0]\)~\si{\meter\per\square\second}. This model bounds the Cartesian acceleration instead of assuming an approximately constant speed, so its bounds cover a similar range of walking speeds as the unicycle model but do not describe the same set of possible motions. The pairwise constraint uses \(R_{\min}=\SI{0.7}{\meter}\). Because both models are linear and use world-frame coordinates, the relative state is the difference \(z=q-x\), which yields the double-integrator relative dynamics
\begin{equation}
  \dot{z}
  =
  f_{\mathrm{rel}}(z,u,d)
  =
  \begin{bmatrix}
    v^{\mathrm{o}}_x-v_x \\
    v^{\mathrm{o}}_y-v_y \\
    a^{\mathrm{o}}_x-a_x \\
    a^{\mathrm{o}}_y-a_y
  \end{bmatrix}.
  \label{eq:cn-cbf-relative-dynamics-qq}
\end{equation}
\bigskip

This system is again four-dimensional. The two examples illustrate the flexibility of the relative coordinates, because \(\sigma\) is a body-frame transformation for the ground robot and a difference of the two states for the quadrotor, while the learning, the composition, and the filtering remain unchanged.

\subsection{Pairwise Barrier Training and Approximation}
\label{subsec:cn-cbf-training-eval}

\begin{table}[tb]
  \caption{CN-CBF configuration shared by both robots.}
  \label{tab:cn-cbf-training-config-shared}
  \centering
  \begin{tabular}{@{}ll@{}}
    \toprule
    Residual network         & MLP, hidden layers \([48,48,24,24,12,12]\), sinusoidal \\
    Output activation        & softplus (nonnegative residual)                        \\
    Trainable parameters     & \num{4837}                                             \\
    Optimizer / epochs       & Adam / \num{10000}                                     \\
    Training time (RTX~3090) & \(\sim\)\SI{5.5}{\minute}                              \\
    \bottomrule
  \end{tabular}
\end{table}

\begin{table}[tb]
  \caption{CN-CBF configuration specific to each robot.}
  \label{tab:cn-cbf-training-config-model}
  \centering
  \begin{tabular}{@{}lll@{}}
    \toprule
                                 & Ground robot                          & Quadrotor                             \\
    \midrule
    Robot model                  & kinematic unicycle                    & planar double integrator              \\
    Obstacle model               & unicycle, bounded turn rate           & double integrator                     \\
    Relative state \(z\)         & \cref{eq:cn-cbf-relative-dynamics-gr}  & \cref{eq:cn-cbf-relative-dynamics-qq} \\
    HJ grid over \(\mathcal{Z}\) & \(80\times80\times20\times10\)        & \(80\times80\times20\times20\)        \\
    HJ computation (RTX~3090)    & \(\sim\)\SI{10}{\second}              & \(\sim\)\SI{10}{\second}              \\
    \bottomrule
  \end{tabular}
\end{table}

The offline stage is summarized in \cref{tab:cn-cbf-training-config-shared}, which lists the settings shared by both robots, and \cref{tab:cn-cbf-training-config-model}, which lists the entries that differ between them. Each stationary pairwise value function is computed with a grid-based HJ solver\footnote{\url{https://github.com/StanfordASL/hj_reachability}}, and the assembled barrier \cref{eq:cn-cbf-pairwise-barrier} is trained against the resulting grid values. For comparison, a lookup-table deployment of the ground-robot value function would store \(1.28\times10^{6}\) grid values and would interpolate values and gradients for every obstacle at every query.

The contrast with the joint approach of \cref{ch:rntc-mpc} quantifies the benefit of the relative-coordinate formulation. In a comparable dynamic setting, RNTC-MPC required about \SI{20}{\hour} to compute the value functions for \num{40000} scenes and about \SI{15}{\hour} to train the hypernetwork \cite{derajic_residual_2025,derajic_cn-cbf_2026}. CN-CBF replaces both stages with one ten-second reachability computation and a few minutes of training, because the learned function no longer depends on the scene, which reduces the combined offline time by more than two orders of magnitude. The corresponding trade-off is that the composition approximates the joint value function only through its pairwise structure, and the closed-loop experiments below evaluate the effect of this approximation.

\begin{figure}[tb]
  \centering
  \includegraphics[width=0.62\textwidth]{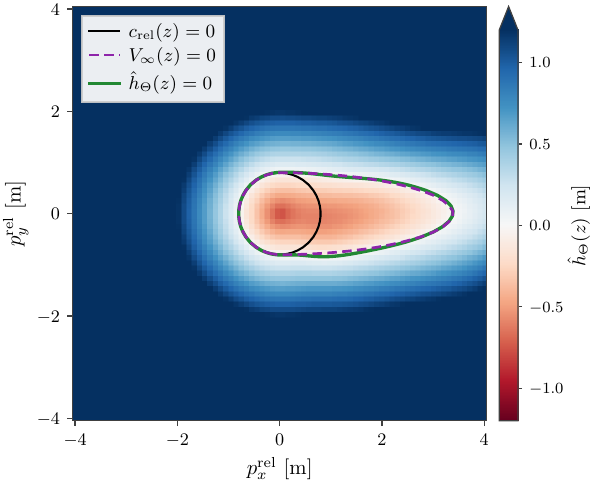}
  \caption[Positional slice of the learned pairwise barrier for the ground robot]{Positional slice of the learned pairwise barrier \(\hat{h}_{\Theta}\) for the ground robot, at relative heading \(\theta^{\mathrm{rel}}=\pi\) (obstacle moving head-on toward the robot) and obstacle speed \(v^{\mathrm{o}}=\SI{1.2}{\meter\per\second}\), in the robot body frame. The robot is located at the origin and heads along \(p^{\mathrm{rel}}_x\). The collision set (black circle, zero level of \(c_{\mathrm{rel}}\)) is the disk of relative obstacle positions within \(R_{\min}\) of the robot. The numerically computed unsafe region (red shading, zero level of \(V_\infty\) dashed) extends several meters ahead of the robot, because an oncoming obstacle in this region can force a collision on a robot that cannot stop and that turns at a limited rate. The zero level set of the learned barrier (green) closely matches the numerical HJ boundary and, by construction, never enters the collision set.}
  \label{fig:cn-cbf-slice}
\end{figure}

\Cref{fig:cn-cbf-slice} shows the head-on configuration of the ground robot, in which the unsafe region extends several meters beyond the collision disk under the actuation limits and the adversarial obstacle model. The learned boundary closely follows the numerical boundary and excludes the collision disk by construction.

\subsection{Ground-Robot Simulation}
\label{subsec:cn-cbf-sim-gr}

\paragraph{Setup.}
The closed-loop evaluation uses physics-based Gazebo simulations of pedestrian crowds, in which HuNavSim controls the pedestrians with a social-force model \cite{perez-higueras_hunavsim_2023}. The pedestrians move cyclically between random goals in a rectangular area, which keeps the effective density approximately constant, and they do \emph{not} avoid the robot, so that collision avoidance is left entirely to the evaluated methods. The task is to drive from a start position to a goal position across the crowded area without collision. Every evaluated method receives the states of the pedestrians within \SI{3.3}{\meter} of the robot, and this detection range is the same in the quadrotor study. For each crowd size of 5, 10, and 15 pedestrians, 100~randomized scenarios are generated and reused for every method. The primary metric is the collision-free success rate. The path length and the time to goal are reported as well, since safety obtained through excessively conservative motion would reduce task performance.

\paragraph{Baselines and implementation.}
CN-CBF is compared with the four MPC planners of \cref{subsec:rntc-baselines}, SDF-MPC, DCBF-MPC, VO-MPC, and RNTC-MPC, the last retrained for the unicycle speed and turning-rate limits used in this chapter. The nominal controller of CN-CBF is an MPC tracker \emph{without perception or safety constraints}, which moves straight toward the reference, so that every avoidance maneuver originates from the filter. The comparison is thus between complete navigation systems, a reachability-informed filter around a plain tracker on the one side and reachability-informed or geometric planners on the other. All planners run at \SI{20}{\hertz} with a \SI{2}{\second} horizon and a \SI{0.1}{\second} prediction interval, implemented in CasADi \cite{andersson_casadi_2019} with IPOPT and deployed as ROS~2 packages \cite{macenski_robot_2022}. The CBF-QP of CN-CBF is solved with qpOASES at \SI{250}{\hertz}, with the filter gain \(\gamma=4.0\) and the smoothing parameter \(\beta=4.0\) tuned by grid search. As in \cref{ch:orn-cbf}, the deployed filter uses the relaxed problem \cref{eq:prelim-cbf-qp-slack} rather than the hard-constrained form \cref{eq:cn-cbf-qp}, and the returned input solves \cref{eq:cn-cbf-qp} whenever the slack is zero.

\begin{figure}[tb]
  \centering
  \includegraphics[width=\textwidth]{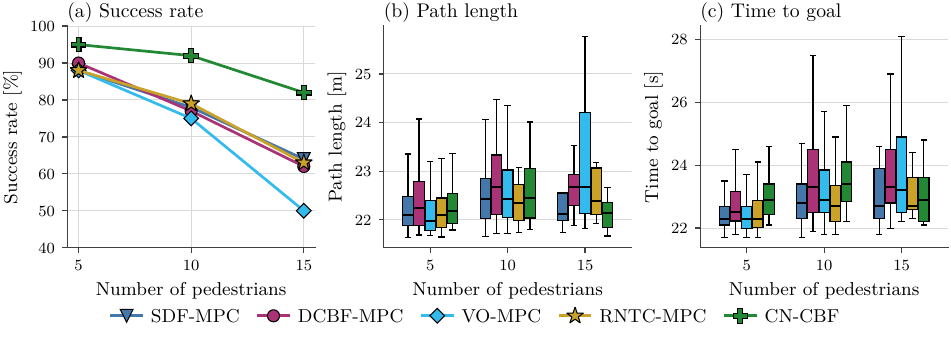}
  \caption[Results of the CN-CBF ground-robot simulation study]{Ground-robot simulation study over 100~randomized pedestrian-crowd scenarios per density. (a)~Success rates of the four MPC-based planners and of the CN-CBF safety filter. The advantage of CN-CBF increases with the crowd density. (b)~Path lengths and (c)~times to goal of the scenarios completed by all methods. Boxes span the interquartile range, lines mark the median, and whiskers extend to 1.5 times the interquartile range. The higher success rates of CN-CBF are not obtained through detours or overly conservative motion.}
  \label{fig:cn-cbf-gr-sim-results}
\end{figure}

\paragraph{Simulation results.}
\Cref{fig:cn-cbf-gr-sim-results} presents the outcomes. CN-CBF achieves the highest success rate at every density, with approximately \(95\%\), \(92\%\), and \(82\%\) for 5, 10, and 15 pedestrians. The margin over the strongest baseline grows from about 5~percentage points in the sparsest setting to 18~percentage points in the densest, where the best baseline reaches \(64\%\) and VO-MPC reaches \(50\%\). Two further findings follow. First, the path lengths and motion times of CN-CBF are comparable to those of the baselines, so the higher success rate is not obtained through detours or overly conservative motion. Second, CN-CBF outperforms RNTC-MPC even though RNTC-MPC approximates the joint value function. A plausible interpretation is that in these experiments the pairwise composition error is less consequential than the sparse coverage of multi-obstacle scenes, the enforcement at planning rate, and the horizon-limited constraint of the joint method, although the experiment does not isolate these factors.

\begin{figure}[tb]
  \centering
  \includegraphics[width=\textwidth]{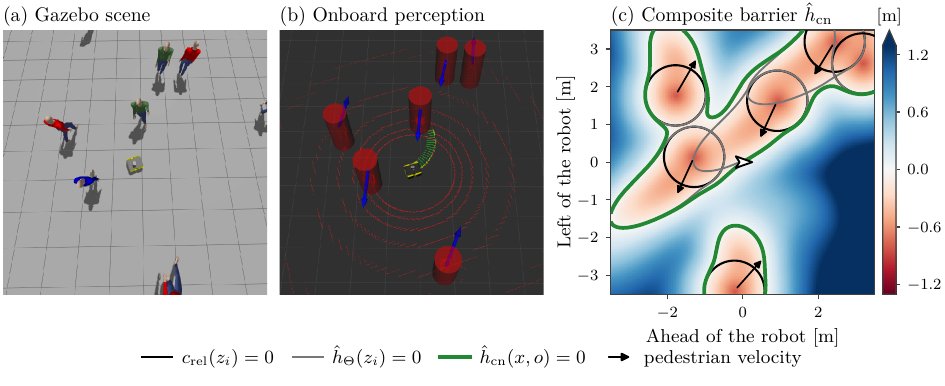}
  \caption[Example scene from the CN-CBF ground-robot simulation study]{Example scene from the ground-robot simulation study. (a)~The Gazebo environment with the robot navigating among pedestrians controlled by the social-force model. (b)~The corresponding local perception, with detected pedestrians, their estimated velocities, and the planned motion of the robot. (c)~Positional slice of the composite barrier \(\hat{h}_{\mathrm{cn}}\) in the robot frame, with the robot at the origin heading to the right (dart), evaluated over hypothetical robot positions at the current heading, with red shading marking the unsafe positions and arrows showing the pedestrian velocities. The pairwise zero level sets (gray) surround the collision disk of each pedestrian (black) and are elongated along the direction in which that pedestrian moves relative to the robot, while the composite zero level set (green) traces a smooth outline around their union without touching the collision disks.}
  \label{fig:cn-cbf-sample-scene}
\end{figure}

\Cref{fig:cn-cbf-sample-scene} visualizes one scenario together with the composite barrier in the robot frame and thereby illustrates the effect of the composition. Each pedestrian is surrounded by a teardrop-shaped unsafe region that points along the relative direction and grows with the relative speed, and the composite zero level set encloses the union of these regions smoothly. The failures observed at high density are consistent with three limitations. Several pedestrians can jointly close every escape route even when each of them remains individually avoidable (\cref{subsec:cn-cbf-composition}), pedestrians can turn more sharply than the assumed bound permits, and the finite detection range can reveal an approaching pedestrian too late. The available experiments do not quantify the contribution of each of these causes separately.

\subsection{Quadrotor Simulation}
\label{subsec:cn-cbf-sim-qq}

\paragraph{Experimental setup and baselines.}
The second study evaluates the quadrotor model in the same pedestrian environment, with crowd sizes of 5, 10, 15, and 20 and again 100~randomized scenarios per density. The nominal controller is a linear-quadratic regulator (LQR) with no obstacle awareness. The baselines are two classical CBF constructions suited to double-integrator dynamics, both of which impose one constraint per obstacle in the QP, while CN-CBF imposes a single composite constraint regardless of the obstacle count. The collision-cone CBF (C3BF) \cite{tayal_control_2024} constrains the relative velocity vector to remain outside the cone of velocities that lead to collision. The higher-order CBF (HO-CBF) \cite{xiao_high-order_2022} uses the distance function as a barrier and enforces the constraint through its second time derivative, because the distance has relative degree two for acceleration inputs (\cref{sec:prelim-cbf}). Their linear class-\(\mathcal{K}\) gains, \(\gamma=3.0\) for C3BF and \(\gamma_1=3.0\) together with \(\gamma_2=2.0\) for HO-CBF, are tuned by the same grid-search protocol as the CN-CBF parameters.

\begin{figure}[tb]
  \centering
  \includegraphics[width=\textwidth]{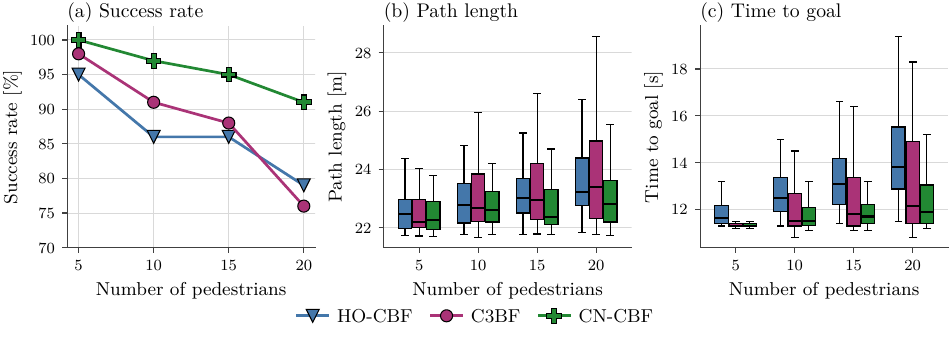}
  \caption[Results of the CN-CBF quadrotor simulation study]{Quadrotor simulation study over 100~randomized scenarios per density, comparing the CN-CBF filter against the collision-cone CBF (C3BF) and the higher-order CBF (HO-CBF), all combined with the same LQR nominal controller. (a)~Success rates. (b)~Path lengths and (c)~times to goal of the scenarios completed by all methods, with boxes as in \cref{fig:cn-cbf-gr-sim-results}. CN-CBF maintains the highest success rates at every density while matching the baselines in path length and motion time.}
  \label{fig:cn-cbf-qq-sim-results}
\end{figure}

\paragraph{Simulation results.}
\Cref{fig:cn-cbf-qq-sim-results} presents the outcomes. CN-CBF again achieves the highest success rate at every density, decreasing from full success with 5 pedestrians to approximately \(91\%\) with 20, while C3BF decreases from \(98\%\) to \(76\%\) and HO-CBF from \(95\%\) to \(79\%\). The path lengths and motion times are comparable across all three methods. All three filters solve a QP with a tuned class-\(\mathcal{K}\) gain and with the same acceleration bounds imposed on the input, so they differ in the set that each barrier represents. The HO-CBF barrier is the geometric distance, whose zero level set is the collision boundary, and the acceleration limits of the robot do not enter this barrier, so how early the robot must decelerate is governed by the two gains \(\gamma_1\) and \(\gamma_2\) alone. The C3BF barrier constrains the instantaneous relative velocity to point outside the collision cone and does not represent the future maneuvering set of the obstacle. The zero level set of the CN-CBF barrier is instead the boundary of the reachability-based safe set, which is computed from the acceleration limits of the robot and from the modeled adversarial obstacle inputs. The results are consistent with a benefit from this more accurate modeling, but they do not separate it from the effect of the particular gain values or from other implementation differences.

\subsection{Hardware Experiments}
\label{subsec:cn-cbf-hardware}

\begin{figure}[tb]
  \centering
  \includegraphics[width=0.75\textwidth]{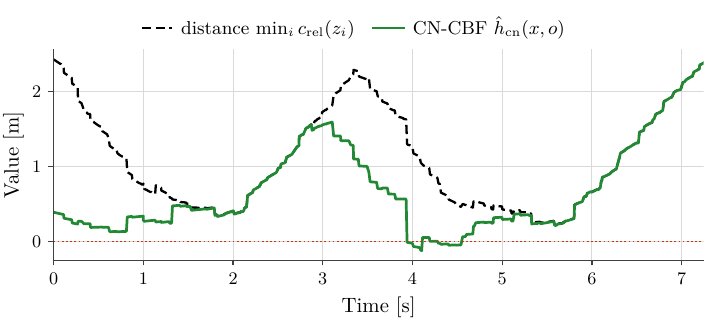}
  \caption[Composite barrier value and pedestrian distance during a CN-CBF hardware run]{Composite barrier value \(\hat{h}_{\mathrm{cn}}\) (green) and distance \(\min_i c_{\mathrm{rel}}(z_i)\) to the nearest pedestrian (dashed) over time during a successful collision-avoidance run of the CN-CBF filter on the physical delivery robot. The barrier value anticipates the encounter well before the distance becomes small and briefly becomes negative under model mismatch and measurement noise. The physical buffer contained in \(R_{\min}\) prevents a collision during this interval.}
  \label{fig:cn-cbf-gr-hw}
\end{figure}

\paragraph{Ground-robot experiments.}
The hardware evaluation of the ground robot uses the delivery platform of \cref{subsec:ntc-hardware}, whose onboard perception pipeline detects moving objects from LiDAR and stereo-camera data and estimates their positions and velocities. The robot and obstacle models, the nominal MPC tracker without perception, and all filter parameters are identical to those of the simulation study, and all computation runs onboard. \Cref{fig:cn-cbf-gr-hw} shows the composite barrier value and the distance to the nearest pedestrian along a successful pedestrian-avoidance run. As in \cref{subsec:orn-cbf-hardware}, the barrier value becomes slightly negative for a short interval under model mismatch and measurement noise. No collision occurs, because the buffer included in \(R_{\min}\) separates the learned zero level from physical contact and the filter restores a positive value within a short interval.

\begin{figure}[tb]
  \centering
  \includegraphics[width=\textwidth]{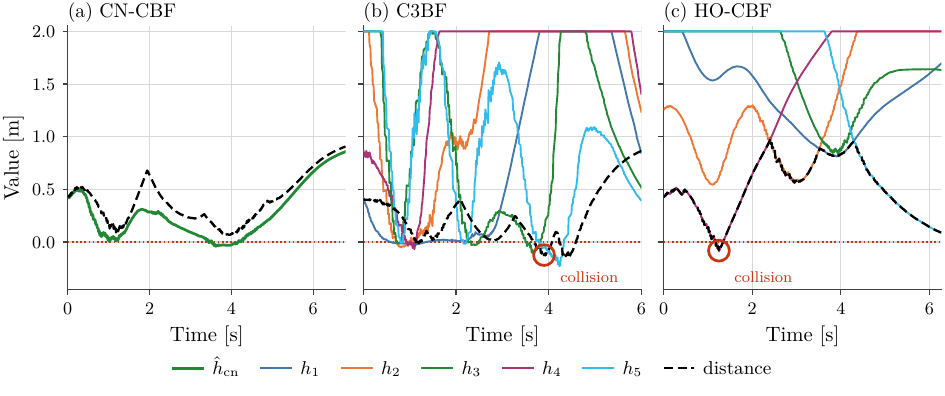}
  \caption[Barrier and distance values in the CN-CBF quadrotor hardware experiments]{Barrier values and distance to the nearest obstacle over time in the quadrotor hardware experiments with five moving obstacle drones, for (a)~the CN-CBF filter, (b)~the C3BF filter \cite{tayal_control_2024}, and (c)~the HO-CBF filter \cite{xiao_high-order_2022}, all around the same LQR nominal controller. For the two baselines, which impose one constraint per obstacle, the five pairwise barrier values \(h_1,\ldots,h_5\) are shown individually, and each HO-CBF barrier is the distance to the corresponding obstacle by definition. Only the CN-CBF filter keeps the distance positive throughout, while both baselines collide (red circles).}
  \label{fig:cn-cbf-qq-hw}
\end{figure}

\paragraph{Quadrotor experiments.}
The aerial evaluation uses a Crazyflie nano-quadrotor in a collision-avoidance task with five moving obstacles. The obstacles are themselves quadrotors flying between predefined waypoints, so that their states are known accurately and the comparison concentrates on the barrier design. The robot and obstacle models and the LQR nominal controller follow the simulation study, with the collision radius reduced to \(R_{\min}=\SI{0.3}{\meter}\) for the nano-quadrotors, CN-CBF is compared against the same C3BF and HO-CBF baselines, and the computation runs offboard as in \cref{subsec:orn-cbf-hardware}. \Cref{fig:cn-cbf-qq-hw} shows the barrier and distance traces of the three filters. With CN-CBF filtering the LQR commands, the quadrotor completes the task without collision. Its distance to the obstacles remains positive throughout, while the barrier value approaches zero during the closest encounters and briefly becomes slightly negative. Both baselines collide, and their distance traces cross zero.\footnote{The data recorded after the collision instants of the baselines correspond to the recovery behavior of the platform and are shown for completeness.} This single challenging scenario is consistent with the simulation results, but it is a demonstration rather than a statistical hardware comparison.

\section{Discussion}
\label{sec:cn-cbf-discussion}

The shared pairwise barrier and smooth composition achieved higher simulated success rates than the evaluated MPC and classical CBF methods on both robot types, and the hardware studies demonstrate deployment with a more limited comparison. The following discussion examines the structural and practical limitations.

\paragraph{Pairwise and joint safety.}
The principal limitation is structural. The bound \cref{eq:cn-cbf-softmin-bounds} places the composite safe set inside the intersection of the learned pairwise safe sets, but it does not constrain the relation to the true joint maximal safe set. Even with exact pairwise values, the intersection over-approximates the joint safe set in configurations where the individual evasion maneuvers conflict, and the smoothing offset does not correct this optimism. The composition might thus admit configurations in which several obstacles jointly close every escape route, although each obstacle remains individually evadable. The failures observed in the densest simulated crowds are consistent with this explanation, although the experiments do not isolate a unique cause for each failure. The joint method of \cref{ch:rntc-mpc} does not share the limitation in principle, but its supervision must sample the combinatorial space of multi-obstacle scenes. In the direct comparison of \cref{subsec:cn-cbf-sim-gr}, CN-CBF nevertheless achieved higher success rates at every density, so the composition was the more effective representation under the evaluated training budgets and scenarios. The study does not determine how much of the difference arose from the approximation error, the controller interface, the enforcement rate, or other implementation choices. A learnable aggregation that corrects the composition toward the joint value function in such configurations would address the structural gap directly.

\paragraph{Guarantee scope.}
As in \cref{ch:orn-cbf}, the architecture guarantees failure-set exclusion, the composition preserves it with the explicit gap \cref{eq:cn-cbf-softmin-bounds}, and the count-update margin \cref{eq:cn-cbf-count-update-margin} states what a newly added obstacle requires. Forward invariance of the composite safe set then follows from the CBF theory of \cref{sec:prelim-cbf} under accurate obstacle states, obstacle motion within \cref{eq:cn-cbf-obstacle-dynamics}, the sensing condition \cref{eq:cn-cbf-sensing-adequacy}, a learned barrier that satisfies the CBF condition, and zero slack in the relaxed QP. The experiments support the last two requirements empirically, and the hardware trace of \cref{fig:cn-cbf-gr-hw} shows the practical meaning of this qualification, since the barrier value briefly became negative and the physical buffer in \(R_{\min}\) separated that event from contact. Two points are specific to this chapter. The realized-disturbance argument \cref{eq:cn-cbf-realized-disturbance} is established for the exact pairwise value function, while the learned barrier uses the derivative estimate \cref{eq:cn-cbf-backward-difference}, which is affected by noise and latency. The robust target also holds only while the obstacle inputs remain within \(\mathcal{D}\), and the recorded failures include pedestrians who turned more sharply than the attributed bound permits, so calibrating \(\mathcal{D}\) from recorded motion data is the direct way to trade robustness against conservatism.

\paragraph{Heterogeneous obstacles and higher-dimensional models.}
The evaluated settings use one obstacle class per robot, and the offline stage inherits the dimensionality limits of grid-based HJ solvers, which here apply to the relative state space. The four-dimensional problems considered in this chapter were solved in seconds, but robot or obstacle models with more states, such as three-dimensional flight, models that include the body orientation and angular rates, and obstacle classes with internal state, increase the dimension of the relative state space beyond what grid-based solvers can handle. Both restrictions could be addressed within the framework. A heterogeneous environment would require one pairwise model per obstacle class, and all of these models can be passed to the same aggregation \cref{eq:cn-cbf-softmin}, because the composition does not require its arguments to originate from the same function. A higher-dimensional relative state, in turn, still leads to a single stationary reachability problem, which is the setting in which self-supervised neural HJ solvers \cite{bansal_deepreach_2021} and MPC-guided value approximations \cite{feng_bridging_2025} apply most directly. \Cref{ch:vertex-cbf} develops such a solver-free alternative, applies it to a robot--pedestrian pair with a different relative model, and deploys the resulting barrier on the same robot through the composition and the filter of this chapter.

\paragraph{Combining static and dynamic barriers.}
The evaluated scenarios contained only moving obstacles, and those of \cref{ch:orn-cbf} contained only static geometry, while real deployments contain both. The observation-conditioned barrier \(\hat{h}_{\Psi}(\,\cdot\mid w_j)\) of \cref{ch:orn-cbf} is itself a scalar barrier candidate and could enter the aggregation \cref{eq:cn-cbf-softmin} as one additional argument next to the pairwise terms, which would yield a single barrier and a single QP constraint for a mixed environment. Because both barriers carry the same residual failure-set exclusion property, such a composition would preserve exclusion. It would not by itself establish that the aggregate satisfies the CBF condition, and it would inherit the SDF-interpolation regularity caveat of \cref{sec:orn-cbf-discussion}. Validating this combination, together with the trainable aggregation mentioned above, is a promising direction for future work.

\section{Summary}
\label{sec:cn-cbf-summary}

CN-CBF learns one stationary pairwise reachability barrier and composes its evaluations over moving obstacles. The composite preserves failure-set exclusion, with a soft-minimum offset bounded by \(\ln(m)/\beta\). It achieved the highest simulated success rates on both robots, with a margin of 18~percentage points over the best MPC baseline at the highest ground-robot crowd density, and the hardware studies demonstrated onboard pedestrian avoidance and a collision-free quadrotor encounter with five moving obstacle drones. CN-CBF completes the static and dynamic variants of both controller interfaces, addressing questions~\ref{rq:realtime}, \ref{rq:dynamic}, and~\ref{rq:structure}. All four methods still rely on low-dimensional grid-based supervision, which the next chapter replaces with vertex-restricted search for systems with control-affine dynamics and polytopic input constraints.

\part{Grid-Free Reachability Supervision}
\chapter{Vertex-Restricted Supervision for Neural Control Barrier Functions}
\chaptermark{Vertex-Restricted Supervision for Neural CBFs}
\label{ch:vertex-cbf}

This chapter presents \emph{VertexCBF}, which replaces the grid-based supervision of the preceding methods for systems with control-affine dynamics and polytopic input constraints. It is based on the author's paper \cite{derajic_vertexcbf_2026} at the Conference on Robot Learning (CoRL) 2026. The method trains a residual neural barrier using a physics-informed loss and labels from a finite-horizon search over input-polytope vertices. For control-affine dynamics, the continuous-time HJ Hamiltonian attains its maximum at a vertex. This exact property motivates a discrete tree search that runs in parallel on a graphics processing unit (GPU), whose finite step, horizon, and width introduce the label errors analyzed in \cref{subsec:vertex-cbf-error}. The chapter answers research question~\ref{rq:supervision} on scalable supervision, extends the answer to question~\ref{rq:realtime} for the safety-filter interface, and reuses the structural property of question~\ref{rq:structure}.

\section{Motivation}
\label{sec:vertex-cbf-motivation}

Grid-based supervision limits the preceding methods to low-dimensional models \cite{bansal_hamilton-jacobi_2017}. Their experiments use three to five states, and the five-dimensional grid has only seven nodes along each heading and velocity axis (\cref{subsec:ntc-training-eval}). Quadrotors modeled with their orientation and angular rates, manipulators, and floating-trunk legged models exceed this practical range, although their dynamics make reachability-based safety especially valuable and hand-designed barriers especially hard to construct. Scalable supervision is thus needed precisely for the systems that could benefit most from a dynamics-aware safe set.

Two established alternatives address this gap only partially. The first is self-supervised neural PDE solving, which replaces the grid by a network trained to satisfy the HJ variational inequality \cite{bansal_deepreach_2021}. For the stationary problem of this dissertation, this objective alone is insufficient, because every constant function that lies below the constraint function satisfies the stationary variational inequality. Gradient-based training can converge to these trivial optima, and the experiments in \cref{sec:vertex-cbf-experiments} show that it produces uninformative barriers on many systems. The second alternative adds supervision from sampling-based trajectory optimization \cite{feng_bridging_2025}. This removes the ambiguity in principle, but without a suitable initial policy a sampler that draws inputs from the interior of the input set rarely produces the extreme maneuvers that define the boundary of the safe set, so the resulting labels underestimate the safety values.

\begin{figure}[tb]
  \centering
  \includegraphics[width=\textwidth]{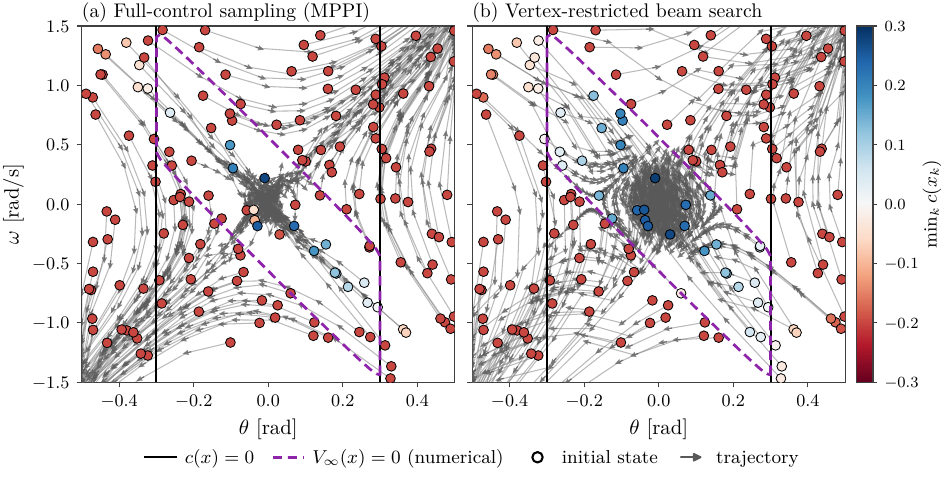}
  \caption[Finite-horizon trajectories for the inverted pendulum under full-control sampling and vertex-restricted search]{Finite-horizon trajectories for the inverted pendulum, generated by (a)~model predictive path integral (MPPI) sampling over the full control set and (b)~the proposed vertex-restricted beam search. Both methods start from the same 160 random initial states, drawn as circles and colored by the worst constraint value \(\min_{k}c(x_k)\) reached along the trajectory, so that red marks a constraint violation. The subsequent states are drawn as small arrows that point in the direction of motion. The black lines mark the constraint boundary and the purple dashed lines the boundary of the numerically computed maximal safe set. The vertex-restricted search keeps the pendulum inside the constraint set from almost every initial state in the numerically computed safe set, while the full-control sampler fails from many of them.}
  \label{fig:vertex-cbf-traj-comparison}
\end{figure}

For fixed state and costate, the control-affine Hamiltonian is linear in the input and reaches its maximum at a polytope vertex, which suggests concentrating a finite search on extreme inputs. \Cref{fig:vertex-cbf-traj-comparison} illustrates the effect on the inverted pendulum: from the same initial states, vertex-restricted search keeps the system within the constraint set far more often than full-control sampling. Because such labels can be generated independently at sparsely sampled states through a batched GPU search, the approach scales to the systems with up to 13 states evaluated in this chapter.

VertexCBF combines these labels with the physics-informed loss and with the residual parameterization of \cref{subsec:prelim-structure-preserving-residual}. The labels discourage trivial optima of the variational inequality, the physics-informed loss promotes local consistency at states that carry no label, and the residual architecture keeps the predicted safe set inside the constraint set regardless of the regression error. Unlike the previous chapters, the learned barrier is a single fixed field rather than a conditioned family, because this chapter studies the supervision rather than the environment representation.

\section{Related Work}
\label{sec:vertex-cbf-related-work}

This section reviews the work most closely related to the question of this chapter, which is how to obtain reachability-based barrier candidates without a grid-based PDE solver. The broader context is reviewed in \cref{ch:introduction} and in the related-work sections of \cref{ch:orn-cbf,ch:cn-cbf}.

Learning-based barrier construction began with barriers fitted by support vector machines \cite{srinivasan_synthesis_2020} and by neural networks trained on labeled safe and unsafe states \cite{long_learning_2021}, and recent constructions derive distance-based barriers from Gaussian-splat maps \cite{chen_safer-splat_2025}. Such barriers describe geometry rather than dynamics, so they become inadequate for higher-order systems with limited actuation \cite{hsu_safety_2024}. Formulations that account for input limits improve feasibility but typically yield a conservative subset of the maximal safe set \cite{liu_safe_2022,xiao_barriernet_2023}.

The maximal safe set itself is characterized by HJ reachability. The methods of \cref{ch:orn-cbf,ch:cn-cbf} and the observation-conditioned value network of \cite{lin_one_2026} approximate it with models trained on numerical solutions, so their scalability is bounded by the solver that produces the targets. DeepReach removes the grid by training a network to satisfy the time-dependent HJ variational inequality directly, with a curriculum that advances the horizon during training \cite{bansal_deepreach_2021}. Later work adds exact boundary conditions \cite{singh_exact_2025} and uncertainty-aware verification \cite{lin_verification_2024}. These methods target the time-dependent problem and depend on a curriculum that is difficult to tune and can be unstable.

Two approaches address the stationary problem directly and are the closest to this chapter. The first minimizes the residual of the stationary HJ variational inequality as a single self-supervised objective \cite{manda_learning_2025}. As explained in \cref{sec:vertex-cbf-motivation}, that objective admits trivial solutions, and the experiments of this chapter confirm that it fails on many systems. The second combines a PDE loss with data supervision generated by model predictive path integral (MPPI) sampling over the full control set \cite{feng_bridging_2025}. It targets the general time-dependent problem and relies on pretraining, curriculum training, and iterative refinement. VertexCBF differs from both in the problem it solves and in the source of its labels. It addresses the infinite-horizon stationary problem, which removes the time variable and the curriculum, and it exploits the control-affine dynamics and the convex polytope input set of the CBF-QP formulation, which motivate a vertex-restricted trajectory search. In addition, the residual parameterization makes pretraining unnecessary.

Restricting attention to extreme inputs is a classical idea in control. Bang-bang solutions of time-optimal control problems are extremal for the same reason, because the Hamiltonian of the minimum principle is linear in the input \cite{pontryagin_mathematical_1962}. The contribution here is not the observation itself but its use as a data-generation mechanism for learning barrier functions, together with the tree search and the training scheme that exploit it.

\section{Problem Statement}
\label{sec:vertex-cbf-problem-statement}

This section states the design problem of the chapter within the safety-filtering framework of \cref{sec:prelim-cbf}. In contrast to \cref{ch:orn-cbf,ch:cn-cbf}, the constraint function is fixed and known, and the difficulty lies in the dynamics and the actuation limits.

\paragraph{System and constraint.}
The robot is described by the control-affine model \cref{eq:prelim-control-affine-dynamics},
\begin{equation}
  \dot{x}
  =
  f_0(x)+g(x)u,
  \qquad
  x\in\mathcal{X}\subseteq\mathbb{R}^{n_x},
  \quad
  u\in\mathcal{U}\subset\mathbb{R}^{n_u},
  \label{eq:vertex-cbf-dynamics}
\end{equation}
with locally Lipschitz \(f_0\) and \(g\) and full state feedback. The state space \(\mathcal{X}\) is a bounded box that defines the operating region of interest. The dynamics and the constraint function are defined outside this box as well, so on most systems a trajectory that leaves the box is evaluated without modification. On systems where a state component is not meaningful far outside its box, such as the pendulum angle or the rocket mass, the implementation instead clamps that component to the box limits after each Euler step, which locally alters the transition model at those limits. Two assumptions are made explicit, because the method depends on them. The input set \(\mathcal{U}\) is a compact convex polytope with the vertex set
\begin{equation}
  \mathcal{V}
  :=
  \operatorname{vert}(\mathcal{U}),
  \qquad
  M
  :=
  \lvert\mathcal{V}\rvert
  <
  \infty,
  \label{eq:vertex-cbf-vertex-set}
\end{equation}
and the dynamics are affine in the input, as written in \cref{eq:vertex-cbf-dynamics}. These assumptions match the CBF-QP formulation used in this dissertation, with a barrier constraint linear in the input and a polytopic admissible input set. Box-shaped input limits, which describe actuator saturation for every robot considered in this dissertation, are the most common instance, and a box in \(\mathbb{R}^{n_u}\) has \(M=2^{n_u}\) vertices.

Safety is specified by a fixed constraint function \(c:\mathcal{X}\rightarrow\mathbb{R}\) with the constraint and failure sets of \cref{eq:prelim-constraint-and-failure-sets}. The stationary safety value function of \cref{eq:prelim-stationary-hj-value} is, for the undisturbed system \cref{eq:vertex-cbf-dynamics},
\begin{equation}
  V_\infty(x)
  =
  \sup_{u(\cdot)\in\mathbb{U}_{[0,\infty)}}
  \;
  \inf_{t\geq0}
  \;
  c\bigl(\xi_{x,0}^{u}(t)\bigr),
  \label{eq:vertex-cbf-value-function}
\end{equation}
and its nonnegative superlevel set is the maximal safe set \(\mathcal{S}^{\star}\) of \cref{eq:prelim-hj-maximal-safe-set}. The value is bounded above by \(c(x)\), because the infimum over time includes the initial instant. No disturbance is modeled in this chapter, and \cref{sec:vertex-cbf-discussion} discusses the extension to the two-player case.

\paragraph{Design problem.}
The goal is to construct a barrier candidate \(\hat{V}_{\Theta}\), represented by a neural network with parameters \(\Theta\), that
\begin{enumerate}
  \item approximates the stationary value function \cref{eq:vertex-cbf-value-function}, so that its predicted safe set \(\hat{\mathcal{S}}_{\Theta}\) approximates the maximal safe set \(\mathcal{S}^{\star}\) rather than a conservative subset of it,
  \item satisfies the \emph{failure-set exclusion} property by construction, so that \(\hat{\mathcal{S}}_{\Theta}\subseteq\mathcal{C}\) regardless of the learning error,
  \item is trained without a grid-based PDE solver, without a training curriculum, and without a nominal policy, so that the offline stage does not inherit the dimensionality limit of \cref{subsec:prelim-hj-numerical}, and
  \item is differentiable wherever the constraint function is, and inexpensive enough for the CBF-QP \cref{eq:prelim-cbf-qp} to run at control rate.
\end{enumerate}
The second and fourth requirements are the same as in \cref{ch:orn-cbf,ch:cn-cbf}, while the third is the specific demand of this chapter.

\paragraph{Standing assumptions.}
Three assumptions define the setting. First, the dynamics model \cref{eq:vertex-cbf-dynamics} and the constraint function \(c\) are known and are the same in data generation, training, and deployment. Second, the state space \(\mathcal{X}\) is a bounded box, and states outside it are not covered by any statement made here. Third, the accuracy of the learned barrier relative to the exact value function is assessed by the closed-loop validation protocol of \cref{subsec:vertex-cbf-validation}, since no error bound is derived, and \cref{sec:vertex-cbf-discussion} discusses how such a bound could be obtained.

\section{Vertex-Restricted Supervision and Physics-Informed Learning}
\label{sec:vertex-cbf-method}

The method first establishes the vertex property of the Hamiltonian and defines a discrete approximation of the vertex-restricted problem. It then constructs a finite search that produces labels, separates the resulting label errors, and specifies the residual model, the training objective, and the safety filter.

\subsection{Vertex-Restricted Reachability}
\label{subsec:vertex-cbf-hamiltonian}
\label{subsec:vertex-cbf-discrete-value}

The stationary variational inequality \cref{eq:prelim-stationary-hji-vi} for the undisturbed system \cref{eq:vertex-cbf-dynamics} reads
\begin{equation}
  \min
  \left\{
  c(x)-V_\infty(x),
  \;
  H\bigl(x,\nabla_xV_\infty(x)\bigr)
  \right\}
  =
  0,
  \label{eq:vertex-cbf-stationary-vi}
\end{equation}
with the Hamiltonian
\begin{equation}
  H(x,p)
  =
  \max_{u\in\mathcal{U}}
  \;
  p^{\top}
  \bigl(f_0(x)+g(x)u\bigr)
  =
  p^{\top}f_0(x)
  +
  \max_{u\in\mathcal{U}}
  \;
  p^{\top}g(x)u .
  \label{eq:vertex-cbf-hamiltonian}
\end{equation}
The second term is a linear function of \(u\) maximized over the compact convex polytope \(\mathcal{U}\). By the fundamental theorem of linear programming, such a maximum is attained at a vertex of the polytope \cite{bertsimas_introduction_1997}. Consequently,
\begin{equation}
  H(x,p)
  =
  \max_{u\in\mathcal{V}}
  \;
  p^{\top}
  \bigl(f_0(x)+g(x)u\bigr),
  \label{eq:vertex-cbf-hamiltonian-vertices}
\end{equation}
and the variational inequality becomes
\begin{equation}
  \min
  \left\{
  c(x)-V_\infty(x),
  \;
  \max_{u\in\mathcal{V}}
  \;
  \nabla_xV_\infty(x)^{\top}
  \bigl(f_0(x)+g(x)u\bigr)
  \right\}
  =
  0 .
  \label{eq:vertex-cbf-stationary-vi-vertices}
\end{equation}
Two consequences follow. The first is computational, since the inner maximization is now a maximum over \(M\) scalar values rather than an optimization problem, so the Hamiltonian can be evaluated exactly and at low cost at any state. The second concerns which inputs are relevant. Wherever the value function is differentiable, which is almost everywhere, a maximizer of the Hamiltonian can be chosen from \(\mathcal{V}\). This statement is exact, but it holds for a fixed pair \((x,p)\) and does not transfer by itself to a discrete-time construction in which one vertex input is held constant over a step of length \(\Delta t\). The remainder of this subsection defines such a construction and states how closely it follows the continuous-time problem, and \cref{subsec:vertex-cbf-search,subsec:vertex-cbf-error} build the label generator on it and separate the approximations it adds.

\Cref{eq:vertex-cbf-stationary-vi-vertices} could be used on its own as a self-supervised training objective, following \cite{manda_learning_2025}. That approach is insufficient, because the stationary variational inequality does not determine its solution uniquely. Let \(a\leq\min_{x\in\mathcal{X}}c(x)\) and consider the constant candidate \(V(x)\equiv a\) in place of \(V_\infty\). Its gradient vanishes, so the second term of \cref{eq:vertex-cbf-stationary-vi-vertices} is zero, and the first term \(c(x)-a\) is nonnegative by the choice of \(a\). The minimum of the two is zero at every state, so every such constant satisfies the equation, although its nonnegative superlevel set is either empty or all of \(\mathcal{X}\) and carries no information. These constants are available to gradient-based training as flat optima, and the experiments of \cref{subsec:vertex-cbf-benchmark} show that training frequently converges to them. Sparse supervision discourages these trivial solutions by anchoring the predicted values at selected states.

\paragraph{Discrete-time value function.}
Generating labels requires a computable quantity, which is obtained by discretizing the vertex-restricted problem \cref{eq:vertex-cbf-stationary-vi-vertices} in time. The order of the two steps matters. The restriction to vertices is made first, in continuous time, where the maximization inside the Hamiltonian is linear in the input and the restriction is exact, and only then are the dynamics discretized. The opposite order, which would restrict a full-control discrete-time Bellman operator to the vertices, is not exact, because the map \(u\mapsto V(F_\Delta(x,u))\) is nonlinear in \(u\). The quantity defined below approximates the vertex-restricted continuous-time problem, and not the discrete-time problem over the full control set.

For a step size \(\Delta t>0\), let \(F_\Delta\) denote the forward-Euler map of the discrete-time model \cref{eq:prelim-discrete-dynamics},
\begin{equation}
  x_{k+1}
  =
  F_\Delta(x_k,u_k)
  =
  x_k
  +
  \Delta t
  \bigl(f_0(x_k)+g(x_k)u_k\bigr).
  \label{eq:vertex-cbf-euler-map}
\end{equation}
The \emph{vertex-restricted discrete-time value function} \(V_{\Delta t}:\mathcal{X}\rightarrow\mathbb{R}\) is
\begin{equation}
  V_{\Delta t}(x)
  :=
  \sup_{\{u_k\}_{k\geq0}\subset\mathcal{V}}
  \;
  \inf_{k\geq0}
  \;
  c(x_k),
  \qquad
  x_0=x,
  \label{eq:vertex-cbf-discrete-value}
\end{equation}
where the state evolves according to \cref{eq:vertex-cbf-euler-map}. It is the infinite-horizon safety value of the Euler-discretized system when the input sequence is drawn from the vertices alone, and it satisfies the Bellman equation
\begin{equation}
  V_{\Delta t}(x)
  =
  \min
  \left\{
  c(x),
  \;
  \max_{u\in\mathcal{V}}
  \;
  V_{\Delta t}\bigl(F_\Delta(x,u)\bigr)
  \right\}.
  \label{eq:vertex-cbf-bellman}
\end{equation}
The associated Bellman operator is not a contraction \cite{fisac_bridging_2019,so_how_2024}, so the usual convergence guarantees of value iteration do not apply to \cref{eq:vertex-cbf-bellman}. A discount factor restores contraction \cite{fisac_bridging_2019,akametalu_minimum_2024}, but it reintroduces an iteration over the whole state space, which the grid-based solver already performs. Instead, rearranging \cref{eq:vertex-cbf-bellman} gives the equivalent form
\begin{equation}
  \min
  \left\{
  c(x)-V_{\Delta t}(x),
  \;
  \frac{
    \max_{u\in\mathcal{V}}V_{\Delta t}\bigl(F_\Delta(x,u)\bigr)-V_{\Delta t}(x)
  }{\Delta t}
  \right\}
  =
  0,
  \label{eq:vertex-cbf-discrete-vi}
\end{equation}
in which the second term is a finite-difference approximation of the Hamiltonian \cref{eq:vertex-cbf-hamiltonian-vertices}.

\paragraph{Local consistency.}
Safety value functions are in general nonsmooth, so the consistency of this normalized form with the continuous-time operator is verified against a smooth test function rather than against the value function itself. Let \(\mathcal{Y}\subseteq\mathcal{X}\) be a compact convex region that contains \(x\) and \(F_\Delta(x,u)\) for every \(u\in\mathcal{V}\) and every sufficiently small \(\Delta t\), and let \(\psi\in C^{1,1}(\mathcal{Y})\) have a gradient with Lipschitz constant \(L_{\nabla\psi}\) on \(\mathcal{Y}\). A Taylor expansion of \(\psi\bigl(F_\Delta(x,u)\bigr)\) about \(x\), followed by the maximum over the vertices and the vertex representation \cref{eq:vertex-cbf-hamiltonian-vertices}, bounds the local truncation error of the finite-difference term by
\begin{equation}
  \left\lvert
  \frac{
    \max_{u\in\mathcal{V}}\psi\bigl(F_\Delta(x,u)\bigr)-\psi(x)
  }{\Delta t}
  -
  H\bigl(x,\nabla_x\psi(x)\bigr)
  \right\rvert
  \leq
  \tfrac{1}{2}
  L_{\nabla\psi}
  \max_{u\in\mathcal{V}}
  \bigl\lVert f_0(x)+g(x)u\bigr\rVert^{2}
  \Delta t ,
  \label{eq:vertex-cbf-truncation-error}
\end{equation}
and substituting this bound into the discrete variational inequality \cref{eq:vertex-cbf-discrete-vi} gives
\begin{equation}
  \begin{aligned}
    \min
    \left\{
    c(x)-\psi(x),
    \;
    \frac{
      \max_{u\in\mathcal{V}}\psi\bigl(F_\Delta(x,u)\bigr)-\psi(x)
    }{\Delta t}
    \right\}
     & =                                                                    \\
     & \hspace{-6em}
    \min
    \left\{
    c(x)-\psi(x),
    \;
    H\bigl(x,\nabla_x\psi(x)\bigr)
    \right\}
    +
    O(\Delta t).
  \end{aligned}
  \label{eq:vertex-cbf-consistency}
\end{equation}
The derivation is given in \cref{sec:app-vertex-cbf-consistency}. The vertex-restricted Euler Bellman operator is a first-order consistent approximation of the continuous stationary operator \cref{eq:vertex-cbf-stationary-vi}, with an error of \(O(\Delta t^{2})\) in the one-step Bellman expansion and of \(O(\Delta t)\) in its time-normalized form.

This calculation establishes local consistency only and is not a convergence proof for the infinite-horizon value function. Such a proof would additionally require well-posedness of the limiting viscosity problem, including a comparison principle and appropriate state-constraint conditions, together with an argument that selects the dynamic-programming solution among the possible stationary viscosity solutions. The result is used here to justify the local first-order behavior of the discretization, and \cref{subsec:vertex-cbf-error} separates the label errors that the finite search adds to it. On systems where a state component is clamped to the box limits after each Euler step, or where the quaternion is renormalized (\cref{sec:vertex-cbf-problem-statement,sec:app-vertex-cbf-systems}), the calculation applies to the unmodified Euler map away from those limits.

\subsection{Search-Based Label Generation}
\label{subsec:vertex-cbf-search}
\label{subsec:vertex-cbf-tree-methods}

The infinite-horizon quantity \cref{eq:vertex-cbf-discrete-value} is still not computable, and computing it at one state should not require computing it at every other state. The label generator resolves both points by truncating the horizon. For a horizon length \(K\), define
\begin{equation}
  V^{K}_{\Delta t}(x)
  :=
  \max_{u_0,\ldots,u_{K-1}\in\mathcal{V}}
  \;
  \min_{0\leq k\leq K}
  \;
  c(x_k),
  \qquad
  x_0=x .
  \label{eq:vertex-cbf-finite-horizon-value}
\end{equation}
This is the central object of the method, and its feasible set is finite. There are \(M\) choices at each of \(K\) stages, hence \(M^{K}\) input sequences, and the maximization is a search over a tree of depth \(K\) and branching factor \(M\) rather than a continuous optimization problem.

Three properties make this search practical. First, \cref{eq:vertex-cbf-finite-horizon-value} depends only on the initial state \(x\), so labels at different states are independent. Supervision states can be sampled sparsely from \(\mathcal{X}\), and their number is chosen by the available budget instead of by a grid resolution that grows exponentially with \(n_x\). Second, the tree can be explored for many initial states simultaneously, since at each depth all initial states, all retained partial trajectories, and all \(M\) vertices are expanded and scored in a single batched GPU operation. Third, for systems with few vertices and short horizons, the tree can be enumerated completely, which makes \cref{eq:vertex-cbf-finite-horizon-value} exact.

When \(M^{K}\) is too large for complete enumeration, a width-bounded search is used. Let \(V^{K,B}_{\Delta t}(x)\) denote the value returned by such a search with budget \(B\). Since it optimizes over a subset of the input sequences,
\begin{equation}
  V^{K,B}_{\Delta t}(x)
  \leq
  V^{K}_{\Delta t}(x),
  \label{eq:vertex-cbf-beam-bound}
\end{equation}
with equality whenever the retained subset contains an optimal sequence. The inequality is one-sided, because an incomplete search lowers the search value relative to a complete search for the same discretized finite-horizon problem. Whether the learned safe set becomes smaller as a result also depends on the horizon truncation and the learning error, which \cref{subsec:vertex-cbf-error} separates.

\paragraph{Scope of the vertex restriction.}
Restricting the search to vertices does not by itself improve the finite-horizon optimum, since \cref{eq:vertex-cbf-finite-horizon-value} is a constrained version of the same problem over the full control set. The benefit lies in the effectiveness of the search under a finite computational budget. A maximizer of the Hamiltonian can be chosen from \(\mathcal{V}\) at almost every state, so the tree search operates on a small set that already contains these inputs, while a sampling-based method such as MPPI uses much of its computational budget on interior inputs that are rarely optimal. This advantage relies on control-affine dynamics, a polytope input set, and a sufficiently small step size, and the approximation can become conservative or suboptimal for a large \(\Delta t\), for a short horizon \(K\), or for a small width \(B\) when \(M\) or \(K\) is large. The claim is not that vertex-restricted search always outperforms full-control optimization, but that it does so under the stated assumptions at a comparable budget, which \cref{sec:vertex-cbf-experiments} tests.

\paragraph{Search variants.}
All search variants solve the same problem \cref{eq:vertex-cbf-finite-horizon-value} and differ only in which partial trajectories they retain, ranked by a common score. For a partial trajectory committed up to depth \(k\), the running minimum
\begin{equation}
  s_k
  :=
  \min_{0\leq j\leq k}
  \;
  c(x_j)
  \label{eq:vertex-cbf-score}
\end{equation}
is nonincreasing in \(k\), so \(s_k\) is an upper bound on the objective value of every extension of that trajectory to depth \(K\). Three variants are used \cite{russell_artificial_2016,kool_stochastic_2019,land_automatic_1960}.

\emph{Beam search} retains, after each expansion, the \(B\) partial trajectories with the largest score. At depth \(k\), every retained trajectory is extended by every vertex, which produces at most \(BM\) candidates, the scores are updated through \cref{eq:vertex-cbf-score}, and the best \(B\) candidates are kept. The label is the largest score in the final beam. The cost for \(N\) initial states is \(O(NKBM)\) transition evaluations, which is linear in each parameter.

\emph{Stochastic beam search} addresses the tendency of deterministic selection to commit the beam to one locally promising branch early in the search. It draws \(B\) candidates without replacement with a probability that increases with the score, which preserves diversity. Four sampling rules are implemented. The first is a softmax over the scores with temperature \(T\). The second is the Gumbel top-\(B\) rule, which adds independent Gumbel noise to the scaled scores and then selects the best \(B\) candidates \cite{kool_stochastic_2019}. The third is a rank-based weighting that is independent of the scale of the scores, and the fourth is an \(\varepsilon\)-greedy rule that replaces the deterministic selection by a uniform draw with probability \(\varepsilon\). Deterministic beam search is recovered as \(T\rightarrow0\) or \(\varepsilon\rightarrow0\).

\emph{Branch and bound} uses the fact that \(s_k\) upper-bounds the value of every completion, so a candidate whose score does not exceed the best value found so far cannot improve the result and may be discarded. The implementation performs several passes over the same tree. The first pass runs without pruning and behaves like beam search, establishing an initial lower bound from the best leaf. Each subsequent pass masks out candidates whose score does not exceed the current bound before selecting the best \(B\), which frees capacity in the beam for previously unexplored branches, and small tie-breaking noise on the selection scores diversifies which candidates survive across passes. Pruning never removes a completion that improves on the incumbent, so branch and bound returns a value at least as large as beam search for the same \(B\), at the cost of the additional passes.

The choice between the variants depends on the width of the beam relative to the size of the tree. When \(M\) is small, the retained beam covers a large part of the tree at every depth, the discarded subtrees are unlikely to contain the optimum, and beam search is preferred for its simplicity and its deterministic labels. As \(M\) grows, the beam becomes narrow relative to the \(M^{k}\) nodes at depth \(k\), and a deterministic selection is more likely to discard the optimal branch early. Part of the budget is then better spent on exploration, either by randomizing the selection or by running several pruned passes, at the cost of stochastic labels in the first case and of the additional passes in the second. The variant used for each system is listed in \cref{app:vertex-cbf-config}.
\subsection{Approximation Errors in the Labels}
\label{subsec:vertex-cbf-error}

The label \(y=V^{K,B}_{\Delta t}(x)\) differs from the continuous infinite-horizon value \(V_\infty(x)\) for three separate reasons, and separating them clarifies which parameter controls which effect,
\begin{equation}
  \bigl\lvert V^{K,B}_{\Delta t}(x)-V_\infty(x)\bigr\rvert
  \leq
  \underbrace{\bigl\lvert V_{\Delta t}(x)-V_\infty(x)\bigr\rvert}_{\substack{\text{discretization and}\\\text{vertex restriction}}}
  +
  \underbrace{\bigl\lvert V^{K}_{\Delta t}(x)-V_{\Delta t}(x)\bigr\rvert}_{\text{horizon truncation}}
  +
  \underbrace{\bigl\lvert V^{K,B}_{\Delta t}(x)-V^{K}_{\Delta t}(x)\bigr\rvert}_{\text{search truncation}} .
  \label{eq:vertex-cbf-error-decomposition}
\end{equation}

The first term collects two effects, the error of the forward-Euler step and the error of holding one vertex input constant over each interval of length \(\Delta t\), since the vertex restriction is exact for the continuous-time Hamiltonian but need not remain exact once an input is held over a finite step. A smaller \(\Delta t\) reduces both effects. \Cref{eq:vertex-cbf-truncation-error} bounds the local truncation error of the Bellman operator by a term of order \(\Delta t\), but that bound concerns one application of the operator to a smooth test function and does not by itself yield a rate for the difference of the two infinite-horizon value functions. First-order consistency is used only as an indication of how this term should behave as \(\Delta t\) decreases.

The second term is caused by truncating the horizon after \(K\) steps. The inequality \(V_{\Delta t}(x)\leq V^{K}_{\Delta t}(x)\) always holds, because the infinite-horizon minimum runs over more steps. The term vanishes if an optimal sequence of \cref{eq:vertex-cbf-finite-horizon-value} can be extended by further vertex inputs that keep \(c(x_k)\geq V^{K}_{\Delta t}(x)\) for all \(k>K\), so that the safety level attained within the horizon can be maintained forever, which gives \(V_{\Delta t}(x)\geq V^{K}_{\Delta t}(x)\) and hence equality. Such a continuation requires a sufficiently long horizon and a terminal region in which admissible vertex inputs can maintain the attained safety level. If later violations are unavoidable, horizon truncation overestimates the infinite-horizon discrete value.

The third term vanishes under complete enumeration and can be reduced by a larger budget \(B\) or by the more thorough search variants of \cref{subsec:vertex-cbf-tree-methods}. The three terms interact, because a smaller \(\Delta t\) requires a larger \(K\) to cover the same time horizon, and a larger \(K\) requires a larger \(B\) to retain near-optimal sequences, which \cref{subsec:vertex-cbf-ablation-search} measures. The decomposition is a conceptual account and not a computable bound, since none of the three terms is available without the exact value function.

\subsection{Network Architecture and Training}
\label{subsec:vertex-cbf-training}

The learned barrier follows the residual parameterization of \cref{subsec:prelim-structure-preserving-residual}. Because the constraint function is fixed rather than observation dependent, it specializes to
\begin{equation}
  \hat{V}_{\Theta}(x)
  :=
  c(x)-\hat{r}_{\Theta}(x),
  \qquad
  \hat{r}_{\Theta}(x)\geq0
  \quad\forall x\in\mathcal{X},
  \label{eq:vertex-cbf-residual}
\end{equation}
with the predicted safe set
\begin{equation}
  \hat{\mathcal{S}}_{\Theta}
  :=
  \left\{x\in\mathcal{X}\mid\hat{V}_{\Theta}(x)\geq0\right\}
  \subseteq
  \mathcal{C}.
  \label{eq:vertex-cbf-exclusion}
\end{equation}
This is the same structural property as in \cref{ch:rntc-mpc,ch:orn-cbf,ch:cn-cbf}, and it holds here for the same reason. In this chapter, the parameterization has a second effect that was not needed in the earlier chapters. It makes the trivial constant solutions of \cref{subsec:vertex-cbf-hamiltonian} harder for the network to represent, because reproducing a constant \(\hat{V}_{\Theta}\equiv a\) requires the output \(\hat{r}_{\Theta}(x)=c(x)-a\), which is a state-dependent function rather than a constant. \Cref{subsec:vertex-cbf-ablation-residual} measures the size of this effect.

The residual \(\hat{r}_{\Theta}\) is a multilayer perceptron with sinusoidal hidden activations \cite{sitzmann_implicit_2020} and a softplus output that enforces nonnegativity while remaining smooth. The assembled barrier inherits the nonsmooth points of the constraint function, such as the kink of an absolute value at zero. These points form sets of measure zero, and automatic differentiation returns an implementation-selected derivative there, as discussed for the interpolated SDF in \cref{sec:orn-cbf-discussion}. Non-periodic state components are rescaled linearly to \([-1,1]\) so that all inputs share a common scale, and every periodic component \(\theta\) is replaced by the pair \(\bigl(\cos\theta,\sin\theta\bigr)\), which builds the periodicity into the representation rather than leaving it to be learned.

\paragraph{Training objective.}
The parameters are obtained by minimizing a convex combination of a physics-informed term and a data term,
\begin{equation}
  \mathcal{L}(\Theta)
  =
  \lambda\,\mathcal{L}_{\mathrm{PDE}}(\Theta)
  +
  (1-\lambda)\,\mathcal{L}_{\mathrm{data}}(\Theta),
  \qquad
  \lambda\in[0,1].
  \label{eq:vertex-cbf-total-loss}
\end{equation}
The physics-informed term penalizes pointwise violations of the stationary variational inequality \cref{eq:vertex-cbf-stationary-vi-vertices} at collocation states \(\{x_i\}_{i=1}^{N_{\mathrm{PDE}}}\) drawn uniformly from \(\mathcal{X}\). Substituting the residual parameterization simplifies its first argument, because \(c(x)-\hat{V}_{\Theta}(x)=\hat{r}_{\Theta}(x)\), which gives
\begin{equation}
  \mathcal{L}_{\mathrm{PDE}}(\Theta)
  =
  \frac{1}{N_{\mathrm{PDE}}}
  \sum_{i=1}^{N_{\mathrm{PDE}}}
  \min
  \left\{
  \hat{r}_{\Theta}(x_i),
  \;
  H\bigl(x_i,\nabla_x\hat{V}_{\Theta}(x_i)\bigr)
  \right\}^{2}.
  \label{eq:vertex-cbf-pde-loss}
\end{equation}
The state gradient is obtained by automatic differentiation, and the Hamiltonian is evaluated exactly through \cref{eq:vertex-cbf-hamiltonian-vertices} as a maximum over the \(M\) vertices. The minimum in \cref{eq:vertex-cbf-pde-loss} is not differentiable where its two arguments cross, a switching set that need not coincide with the safe-set boundary. A smooth minimum was also evaluated, but it produced no consistent difference and the exact minimum caused no training instability, so all reported results use the exact minimum. The data term is a mean squared error against the generated labels at supervision states \(\{(x_j,y_j)\}_{j=1}^{N_{\mathrm{data}}}\), with \(y_j=V^{K,B}_{\Delta t}(x_j)\),
\begin{equation}
  \mathcal{L}_{\mathrm{data}}(\Theta)
  =
  \frac{1}{N_{\mathrm{data}}}
  \sum_{j=1}^{N_{\mathrm{data}}}
  \Bigl(
  c(x_j)-\hat{r}_{\Theta}(x_j)-y_j
  \Bigr)^{2}.
  \label{eq:vertex-cbf-data-loss}
\end{equation}

The physics-informed term promotes local self-consistency at densely sampled states that carry no label, which allows the supervision to remain sparse, while the data term fixes the solution at nontrivial values and moves the optimization away from the flat optima of \cref{subsec:vertex-cbf-hamiltonian}. The objective needs neither a curriculum over the horizon nor a nominal policy. The weight \(\lambda\) is chosen per system, and \cref{subsec:vertex-cbf-ablation-lambda} measures how sensitive the result is to this choice.

\subsection{VertexCBF Safety Filter}
\label{subsec:vertex-cbf-filter}

The trained function is deployed as the barrier candidate of the CBF-QP filter of \cref{subsec:prelim-cbf-qp-safety-filter}. For a nominal command \(u_{\mathrm{nom}}\) supplied by an arbitrary controller, the filtered control solves
\begin{equation}
  \begin{aligned}
    u^{\star}
     & =
    \operatorname*{arg\,min}_{u\in\mathcal{U}}
    \;
    \tfrac{1}{2}
    \lVert u-u_{\mathrm{nom}}\rVert^{2}                                                          \\
     & \;\text{s.t.}\quad
    L_{f_0}\hat{V}_{\Theta}(x)
    +
    L_{g}\hat{V}_{\Theta}(x)\,u
    +
    \alpha\bigl(\hat{V}_{\Theta}(x)\bigr)
    \geq
    0,
  \end{aligned}
  \label{eq:vertex-cbf-qp}
\end{equation}
with the Lie derivatives \cref{eq:prelim-lie-derivatives} evaluated by automatic differentiation of the network. The problem is a quadratic program with one barrier constraint and the polytope input constraint, so it has the same structure as the filters of \cref{ch:orn-cbf,ch:cn-cbf}. In this fixed-field formulation, the barrier is never replaced at runtime, so the observation-update questions of \cref{subsec:orn-cbf-observation-conditioning,subsec:cn-cbf-filter} do not arise. The hardware study in \cref{subsec:vertex-cbf-hardware} instead uses observation-dependent composition and inherits the corresponding perception and update assumptions.

The exclusion \cref{eq:vertex-cbf-exclusion} holds by construction, the approximation of \(\mathcal{S}^{\star}\) is measured rather than certified, and the CBF condition \cref{eq:prelim-cbf-condition} is assessed through the closed-loop behavior rather than verified analytically. As in the preceding chapters, the deployed QP relaxes the barrier constraint with a penalized slack variable. For these reasons, the closed-loop validation of \cref{subsec:vertex-cbf-validation} is a necessary part of the method rather than an evaluation convenience, and \cref{sec:vertex-cbf-discussion} states the scope of the claims.

\section{Experimental Evaluation}
\label{sec:vertex-cbf-experiments}

The evaluation tests performance across systems, the supervision source, search parameters, residual architecture, loss weight, and hardware deployment.\footnote{The implementation is available at \url{https://github.com/bojan-derajic/vertexcbf}.} Numerical experiments use one NVIDIA RTX~3090 GPU, as in \cref{ch:orn-cbf,ch:cn-cbf}. \Cref{app:vertex-cbf-config} gives the domains, constraints, search and training settings, validation protocol, and measured offline times.

\subsection{Systems and Baselines}
\label{subsec:vertex-cbf-setup}

\paragraph{Systems.}
The method is evaluated on 15 systems, listed in \cref{tab:vertex-cbf-results}, whose state dimension ranges from 2 to 13 and whose input dimension ranges from 1 to 6. The set covers double integrators in one, two, and three dimensions, several ground-vehicle models, systems with unstable or underactuated dynamics such as the torque-limited inverted pendulum and the cart-pole, and higher-dimensional platforms, which are a three-degree-of-freedom manipulator, a landing rocket, a quadruped trunk, a six-degree-of-freedom underwater vehicle, and a quadrotor. A broad evaluation is necessary here, because learning-based barrier methods are strongly system dependent, and a method can succeed on one system and fail on another of similar dimension whose safety value function is more difficult. Numerical references are computed with the grid-based solver for the seven systems with at most four states, and the other benchmarks are assessed by rollout validation alone. Two systems connect this chapter to the rest of the dissertation. The Dubins car is the ground-robot model of \cref{ch:ntc-mpc,ch:orn-cbf}. The relative unicycle poses the robot--pedestrian problem of \cref{ch:cn-cbf} in relative coordinates, with a second-order robot and a constant-heading pedestrian, and \cref{subsec:vertex-cbf-hardware} states how it differs from the model of that chapter.

\paragraph{Baselines.}
The comparison isolates the effect of the supervision. All three settings use the same architecture, the same hyperparameters, the same training schedule, and the same validation protocol, and they differ only in the data term of \cref{eq:vertex-cbf-total-loss}.
\begin{itemize}
  \item \emph{No data} (ND) trains with the physics-informed loss alone, following \cite{manda_learning_2025}, and corresponds to \(\lambda=1\).
  \item \emph{Full-control data} (FCD) generates labels by MPPI sampling over the full control set, following \cite{feng_bridging_2025}, with the same horizon length and step size as the tree search and with as many sampled trajectories as the beam width. Five MPPI iterations are run, and the best trajectory of each iteration seeds the next. This setting reproduces only the data-generation step of \cite{feng_bridging_2025}, and not its pretraining, curriculum, and iterative refinement, so the comparison isolates the effect of the supervision data.
  \item \emph{Vertex-restricted control data} (VRCD) is the proposed method and generates labels by the tree search of \cref{subsec:vertex-cbf-tree-methods}.
\end{itemize}
The two data-generating settings share the horizon length \(K\), the step size \(\Delta t\), and the nominal sample count, since the number of MPPI trajectories equals the beam width. Their measured computation times are not equal, because five MPPI iterations over \(B\) trajectories and a tree that expands up to \(BM\) successors at each depth evaluate different numbers of transitions. The comparison thus varies the source of the inputs at a comparable nominal budget rather than at an equal computational cost. Every configuration is trained with five random seeds, and all metrics are reported as mean and standard deviation over the seeds.

\subsection{Closed-Loop Validation}
\label{subsec:vertex-cbf-validation}

Because the learned barrier approximates the value function without a derived error bound, it is assessed empirically before it is used, and this subsection defines that assessment. For a trained barrier \(\hat{V}_{\Theta}\), a set of \(N_{\mathrm{valid}}\) initial states is drawn from the state-space box \(\mathcal{X}\) by stratified rejection sampling, so that roughly half of them lie in the predicted safe set \(\hat{\mathcal{S}}_{\Theta}\) and the other half lie outside it. Balancing the two strata reduces the variance of the metrics when the predicted safe set is much smaller than its complement. For the quadrotor, this sampling is uniform over the non-orientation coordinates and over unit quaternions rather than over a 13-dimensional box (\cref{app:vertex-cbf-config}). From each initial state, the closed loop is simulated for a horizon \(T_{\mathrm{valid}}\) under the Hamiltonian-maximizing policy
\begin{equation}
  \kappa_{\Theta}(x)
  =
  \operatorname*{arg\,max}_{u\in\mathcal{V}}
  \;
  \nabla_x\hat{V}_{\Theta}(x)^{\top}
  \bigl(f_0(x)+g(x)u\bigr),
  \label{eq:vertex-cbf-validation-policy}
\end{equation}
which is the policy that the learned barrier itself prescribes and is computable exactly by \cref{eq:vertex-cbf-hamiltonian-vertices}. An initial state is called \emph{validated safe} if the whole simulated trajectory remains in \(\mathcal{C}\). Two properties separate this measurement from forward invariance. It is a finite-horizon property, so a violation that occurs after \(T_{\mathrm{valid}}\) is not observed, and it is evaluated under the policy \cref{eq:vertex-cbf-validation-policy} rather than under the slack-relaxed CBF-QP \cref{eq:vertex-cbf-qp} that is deployed. The protocol assesses the closed-loop behavior under the policy that the learned barrier itself prescribes, and it is not a statement about the deployed filter.

Three metrics are computed from these rollouts. The \emph{false-safe rate} \(\rho_{\mathrm{FS}}\) is the fraction of states in \(\hat{\mathcal{S}}_{\Theta}\) whose trajectories leave \(\mathcal{C}\), and it measures whether the predicted safe set is reliable. The \emph{false-unsafe rate} \(\rho_{\mathrm{FU}}\) is the fraction of states outside \(\hat{\mathcal{S}}_{\Theta}\) whose trajectories remain in \(\mathcal{C}\), and it measures conservatism. Because the rollouts follow the policy of the learned barrier, this rate is also low when that policy fails from most predicted-unsafe states, so a low value alone does not show that the predicted set is close to maximal. The \emph{effective safe volume}
\begin{equation}
  \eta_{\mathrm{eff}}
  :=
  \frac{\operatorname{vol}\bigl(\hat{\mathcal{S}}_{\Theta}\bigr)}{\operatorname{vol}(\mathcal{X})}
  \bigl(1-\rho_{\mathrm{FS}}\bigr)
  \label{eq:vertex-cbf-eta-eff}
\end{equation}
is the fraction of the sampled state space that the barrier both predicts and validates as safe over the rollout horizon. The volume ratio is a Monte Carlo estimate, obtained as the fraction of the uniform proposals of the rejection sampling at which \(\hat{V}_{\Theta}\geq0\). It measures the predicted-safe fraction of the chosen box and is comparable across the three supervision settings of one system, for which \(\mathcal{X}\) is identical, but not across systems. A useful barrier has a small \(\rho_{\mathrm{FS}}\) and a large \(\eta_{\mathrm{eff}}\). Reporting them together is essential, because a barrier with an empty predicted safe set leaves \(\rho_{\mathrm{FS}}\) undefined with \(\eta_{\mathrm{eff}}=0\), and a barrier that predicts the whole box as safe achieves a large predicted volume with a large \(\rho_{\mathrm{FS}}\). The false-unsafe rate is secondary and indicates how much safe behavior is rejected unnecessarily. Seeds with an empty predicted safe set are excluded from the average of \(\rho_{\mathrm{FS}}\) and contribute \(\eta_{\mathrm{eff}}=0\).

Unlike the evaluations of the preceding chapters, this protocol requires no reference value function, so it applies to systems for which no tractable numerical reference exists, and it yields an empirical result on a finite sample that sample-based certification methods can later strengthen (\cref{sec:vertex-cbf-discussion}).

\subsection{Benchmark Results}
\label{subsec:vertex-cbf-benchmark}

\begin{table}[tb]
  \caption[Validation results for the learned barriers on the 15 evaluated systems]{Validation results for the learned barriers on the 15 evaluated systems, comparing the three supervision settings of \cref{subsec:vertex-cbf-setup}. Reported are the false-safe rate \(\rho_{\mathrm{FS}}\), the false-unsafe rate \(\rho_{\mathrm{FU}}\), and the effective safe volume \(\eta_{\mathrm{eff}}\) of \cref{eq:vertex-cbf-eta-eff}, as mean and standard deviation over five training seeds. A dash marks a setting whose predicted safe set was empty for every seed, for which the false-safe rate is undefined, and a missing standard deviation indicates that the false-safe rate is defined for a single seed only. Best mean values in each group are printed in bold. Settings with an empty predicted safe set are not ranked, because their false-unsafe rate reflects only that every state is predicted unsafe.}
  \label{tab:vertex-cbf-results}
  \centering
  \scriptsize
  \providecommand{\stdvs}[1]{\,\scalebox{0.68}{$\pm$#1}}
  \setlength{\tabcolsep}{1pt}
  \begin{tabular*}{\textwidth}{@{\extracolsep{\fill}}l ccc ccc ccc@{}}
    \toprule
    \multirow{2}{*}{System (\(n_x\), \(n_u\))} & \multicolumn{3}{c}{\(\rho_{\mathrm{FS}}\) (\%) \(\downarrow\)} & \multicolumn{3}{c}{\(\rho_{\mathrm{FU}}\) (\%) \(\downarrow\)} & \multicolumn{3}{c}{\(\eta_{\mathrm{eff}}\) \(\uparrow\)} \\
    \cmidrule(lr){2-4} \cmidrule(lr){5-7} \cmidrule(lr){8-10}
                            & ND & FCD & \shortstack{VRCD\\(ours)} & ND & FCD & \shortstack{VRCD\\(ours)} & ND & FCD & \shortstack{VRCD\\(ours)} \\
    \midrule
    Inverted Pendulum (2, 1) & 100.00                     & \textbf{0.00}\stdvs{0.00}  & \textbf{0.00}\stdvs{0.00}  & \textbf{0.00}\stdvs{0.00}  & 15.96\stdvs{0.53}          & 1.11\stdvs{0.30}           & 0.00\stdvs{0.00}           & 0.08\stdvs{0.00}           & \textbf{0.22}\stdvs{0.00} \\
    1D Double Integrator (2, 1) & \textbf{0.00}\stdvs{0.00}  & \textbf{0.00}\stdvs{0.00}  & \textbf{0.00}\stdvs{0.00}  & \textbf{0.95}\stdvs{0.56}  & 4.01\stdvs{0.42}           & 1.18\stdvs{0.31}           & \textbf{0.41}\stdvs{0.00}  & 0.39\stdvs{0.00}           & \textbf{0.41}\stdvs{0.00} \\
    2D Vertical Drone (2, 1) & 36.80\stdvs{5.39}          & --                        & \textbf{0.00}\stdvs{0.00}  & 4.73\stdvs{6.81}           & 4.72\stdvs{6.81}           & \textbf{1.88}\stdvs{0.46}  & 0.35\stdvs{0.20}           & 0.00\stdvs{0.00}           & \textbf{0.58}\stdvs{0.01} \\
    Dubins Car (3, 1) & 0.06\stdvs{0.04}           & \textbf{0.01}\stdvs{0.01}  & 0.04\stdvs{0.01}           & 4.75\stdvs{1.28}           & 4.33\stdvs{0.31}           & \textbf{2.25}\stdvs{0.28}  & 0.92\stdvs{0.00}           & 0.92\stdvs{0.00}           & \textbf{0.93}\stdvs{0.00} \\
    2D Double Integrator (4, 2) & 0.03\stdvs{0.01}           & \textbf{0.00}\stdvs{0.00}  & 0.03\stdvs{0.01}           & 4.68\stdvs{0.25}           & 6.84\stdvs{0.30}           & \textbf{3.42}\stdvs{0.22}  & \textbf{0.93}\stdvs{0.00}  & 0.92\stdvs{0.00}           & \textbf{0.93}\stdvs{0.00} \\
    Kinematic Bicycle (4, 2) & 4.68\stdvs{0.91}           & \textbf{0.52}\stdvs{0.38}  & 1.16\stdvs{0.51}           & 15.49\stdvs{1.07}          & \textbf{3.16}\stdvs{0.24}  & 5.03\stdvs{0.15}           & 0.86\stdvs{0.01}           & \textbf{0.91}\stdvs{0.00}  & \textbf{0.91}\stdvs{0.00} \\
    Cart-Pole (4, 1) & 88.14\stdvs{8.15}          & --                        & \textbf{0.16}\stdvs{0.19}  & \textbf{0.65}\stdvs{0.48}  & 4.06\stdvs{2.46}           & 2.64\stdvs{0.34}           & 0.06\stdvs{0.04}           & 0.00\stdvs{0.00}           & \textbf{0.32}\stdvs{0.00} \\
    Dynamic Unicycle (5, 2) & 4.17\stdvs{0.89}           & \textbf{0.09}\stdvs{0.09}  & 0.23\stdvs{0.26}           & \textbf{2.93}\stdvs{0.38}  & 5.50\stdvs{0.42}           & 6.49\stdvs{0.44}           & 0.90\stdvs{0.01}           & \textbf{0.93}\stdvs{0.00}  & \textbf{0.93}\stdvs{0.00} \\
    Relative Unicycle (5, 2) & 6.06\stdvs{0.10}           & 0.26\stdvs{0.51}           & \textbf{0.16}\stdvs{0.02}  & 4.49\stdvs{0.33}           & 37.07\stdvs{3.49}          & \textbf{2.35}\stdvs{0.13}  & 0.87\stdvs{0.00}           & 0.87\stdvs{0.01}           & \textbf{0.93}\stdvs{0.00} \\
    3D Double Integrator (6, 3) & 0.01\stdvs{0.00}           & \textbf{0.00}\stdvs{0.00}  & \textbf{0.00}\stdvs{0.00}  & \textbf{9.12}\stdvs{0.37}  & 11.20\stdvs{0.43}          & 10.60\stdvs{0.27}          & \textbf{0.97}\stdvs{0.00}  & \textbf{0.97}\stdvs{0.00}  & \textbf{0.97}\stdvs{0.00} \\
    3-DOF Manipulator (6, 3) & 13.52\stdvs{12.09}         & 0.55\stdvs{0.36}           & \textbf{0.43}\stdvs{0.49}  & \textbf{3.69}\stdvs{0.78}  & 6.74\stdvs{0.16}           & 5.68\stdvs{0.12}           & 0.58\stdvs{0.08}           & 0.65\stdvs{0.00}           & \textbf{0.66}\stdvs{0.00} \\
    Landing Rocket (7, 2) & 76.45\stdvs{14.68}         & 56.41\stdvs{13.54}         & \textbf{1.17}\stdvs{0.78}  & \textbf{0.13}\stdvs{0.07}  & 2.95\stdvs{0.91}           & 5.95\stdvs{0.20}           & 0.09\stdvs{0.06}           & 0.02\stdvs{0.01}           & \textbf{0.24}\stdvs{0.00} \\
    Quadruped Trunk (9, 4) & 99.98\stdvs{0.05}          & --                        & \textbf{0.07}\stdvs{0.09}  & \textbf{0.00}\stdvs{0.00}  & 0.00\stdvs{0.00}           & 3.14\stdvs{0.11}           & 0.00\stdvs{0.00}           & 0.00\stdvs{0.00}           & \textbf{0.07}\stdvs{0.00} \\
    6-DOF Underw. Veh. (12, 6) & 22.24\stdvs{0.93}          & \textbf{0.61}\stdvs{0.07}  & 1.00\stdvs{0.25}           & \textbf{0.80}\stdvs{0.14}  & 5.15\stdvs{0.06}           & 2.27\stdvs{0.09}           & 0.33\stdvs{0.00}           & 0.34\stdvs{0.00}           & \textbf{0.36}\stdvs{0.00} \\
    Quadrotor (13, 4) & 5.75\stdvs{0.20}           & 1.21\stdvs{0.08}           & \textbf{1.07}\stdvs{0.12}  & 11.18\stdvs{0.35}          & \textbf{9.50}\stdvs{0.42}  & 10.10\stdvs{0.42}          & 0.88\stdvs{0.00}           & \textbf{0.91}\stdvs{0.00}  & \textbf{0.91}\stdvs{0.00} \\
    \bottomrule
  \end{tabular*}
\end{table}

\Cref{tab:vertex-cbf-results} reports the validation results, and three findings follow from it.

The first finding concerns the physics-informed loss alone. Training without supervision is unreliable, and the state dimension does not predict on which systems it fails. It succeeds on the double integrators, whose value functions are simple, but it fails on the inverted pendulum, which has only two states, where it produces either an empty predicted safe set or one that is entirely unsafe. Similar failures appear on the cart-pole, the landing rocket, and the quadruped trunk. These outcomes are the trivial optima predicted in \cref{subsec:vertex-cbf-hamiltonian}, and they confirm that the stationary variational inequality alone is not a sufficient training signal.

The second finding concerns the supervision source. Adding sampling-based supervision improves the result in several cases and reduces the false-safe rate on most of the higher-dimensional systems, but it is often either conservative or ineffective. On three systems, the vertical drone, the cart-pole, and the quadruped trunk, its predicted safe set collapsed to the empty set, which is safe but provides no usable safe set. On the inverted pendulum, it recovered only a small part of the safe set. On the landing rocket, more than half of the states it predicted safe produced a constraint violation under the validation policy, and on the relative unicycle its false-unsafe rate of \SI{37.07}{\percent} shows that many of the states it predicted unsafe could in fact be kept safe over the validation horizon. The proposed vertex-restricted supervision avoided both failure modes. Its mean false-safe rate over seeds is below \SI{1.2}{\percent} on every system, its false-unsafe rate is competitive, and it achieves the highest or tied-highest effective safe volume on all 15 systems. On the systems that all three settings solve well, such as the Dubins car and the double integrators, the differences are small, and the sampling-based supervision reaches a slightly lower false-safe rate on some of them.

The third finding concerns the systems on which the differences are largest. They are not the highest-dimensional ones but those with unstable or underactuated dynamics, the inverted pendulum, the cart-pole, the landing rocket, and the quadruped trunk. On these systems, safety depends on committing to an extreme input early, so the safe set has a thin, elongated shape that supervision from interior inputs does not reveal. This is the setting in which the vertex restriction matters most, and it is also the setting in which hand-designed barriers are hardest to construct.

\paragraph{Comparison with numerical references.}
\begin{table}[tb]
  \caption[Intersection over union with the numerical maximal safe set]{Intersection over union (\%) between the predicted safe set \(\hat{\mathcal{S}}_{\Theta}\) and the numerically computed maximal safe set, evaluated on the grid of the numerical solution for the seven systems with at most four states. Entries are mean and standard deviation over five training seeds.}
  \label{tab:vertex-cbf-iou}
  \centering
  \small
  \providecommand{\stdv}[1]{\,\scalebox{0.75}{$\pm$#1}}
  \begin{tabular}{@{}lccc@{}}
    \toprule
    System & ND & FCD & VRCD (ours) \\
    \midrule
    Inverted Pendulum      & 1.54\stdv{3.07}           & 33.53\stdv{0.61}          & \textbf{96.39}\stdv{1.16} \\
    1D Double Integrator & \textbf{98.37}\stdv{0.72} & 94.02\stdv{0.43}          & 98.10\stdv{0.35} \\
    2D Vertical Drone    & 64.09\stdv{32.05}         & 0.00\stdv{0.00}           & \textbf{98.48}\stdv{0.20} \\
    Dubins Car             & 99.56\stdv{0.15}          & 99.66\stdv{0.04}          & \textbf{99.80}\stdv{0.03} \\
    2D Double Integrator & 99.52\stdv{0.03}          & 99.39\stdv{0.02}          & \textbf{99.64}\stdv{0.02} \\
    Kinematic Bicycle      & 97.71\stdv{0.24}          & \textbf{99.65}\stdv{0.00} & 99.59\stdv{0.00} \\
    Cart-Pole              & 65.07\stdv{0.22}          & 0.00\stdv{0.00}           & \textbf{93.38}\stdv{0.35} \\
    \bottomrule
  \end{tabular}
\end{table}

The rollout metrics apply to every system, but they do not compare the predicted safe set with the maximal safe set directly. For the seven systems with at most four states, the grid-based solver of \cref{subsec:prelim-hj-numerical} provides that reference, and \cref{tab:vertex-cbf-iou} reports the intersection over union (IoU) between the predicted safe set and the numerical maximal safe set on the grid of the reference. The IoU penalizes a predicted safe set that is too large and one that is too small, so it complements \(\rho_{\mathrm{FS}}\) and \(\eta_{\mathrm{eff}}\), which separate the two effects. Vertex-restricted supervision reaches the highest IoU on five of the seven systems and remains within \(0.3\) percentage points of the best value on the other two. The three settings are nearly identical on the systems that all of them solve well, while the differences are large on the inverted pendulum, the vertical drone, and the cart-pole. There, the sampling-based supervision predicts no state as safe on the vertical drone and the cart-pole and only a small subset of the true safe set on the inverted pendulum, and the physics-informed loss alone recovers only a part of the true safe set. On these seven systems, the learned barriers agree with the numerical reference to between \SI{93}{\percent} and \SI{99.8}{\percent}, which is comparable to the accuracy that the preceding chapters report for their grid-supervised models.

\paragraph{Effect of the supervision source.}

\begin{figure}[tb]
  \centering
  \includegraphics[width=\textwidth]{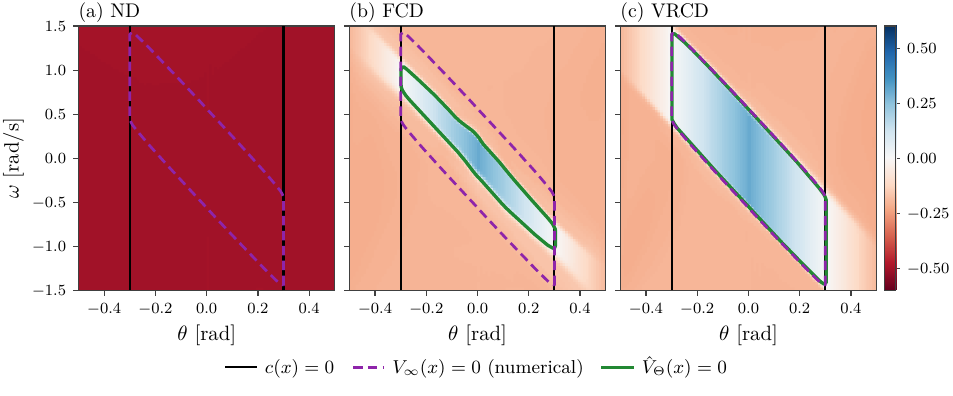}
  \caption[Learned barriers for the inverted pendulum under the three supervision settings]{Learned barriers \(\hat{V}_{\Theta}\) for the inverted pendulum under the three supervision settings, (a)~no data (ND), (b)~full-control data (FCD), and (c)~vertex-restricted control data (VRCD), shown in the \((\theta,\omega)\) plane with red shading marking the states predicted to be unsafe. The black lines mark the constraint boundary, the purple dashed lines the boundary of the numerically computed maximal safe set, and the green lines the boundary of the predicted safe set. The physics-informed loss alone (ND) collapses to a trivial constant solution with an empty predicted safe set. Full-control supervision (FCD) produces a barrier that is safe but strongly conservative. Vertex-restricted supervision (VRCD) approximately recovers the maximal safe set while keeping its predicted safe set inside the constraint set.}
  \label{fig:vertex-cbf-comparison-ip}
\end{figure}

The inverted pendulum shows why the three supervision settings produce such different results. It is chosen because its value function can be computed numerically and its complete state space can be plotted, while the effects shown here also occur in higher dimensions. \Cref{fig:vertex-cbf-traj-comparison} compares the two data generators at the trajectory level. Sampling over the full control set fails from many initial states in the numerical safe set, because recovering the pendulum requires sustained maximal torque and a sampler rarely draws such a sequence. Vertex-restricted beam search keeps the pendulum safe from almost all of these states, and the resulting labels follow the true safety boundary.

\Cref{fig:vertex-cbf-comparison-ip} shows the consequence for the trained barriers. The physics-informed loss alone collapses to a trivial constant solution with an empty predicted safe set, although the system has only two states, and the only seed that avoids this collapse yields a barrier that is unsafe everywhere it predicts safety (\cref{tab:vertex-cbf-results}). Sampling-based supervision corrects the sign but leaves a strongly conservative barrier, because the labels systematically underestimate the achievable safety value. Vertex-restricted supervision approximately recovers the maximal safe set. The comparison supports the interpretation that the benefit comes from the vertex restriction rather than from the presence of supervision by itself.

\subsection{Search Parameters and Label Accuracy}
\label{subsec:vertex-cbf-ablation-search}

\begin{figure}[tb]
  \centering
  \includegraphics[width=\textwidth]{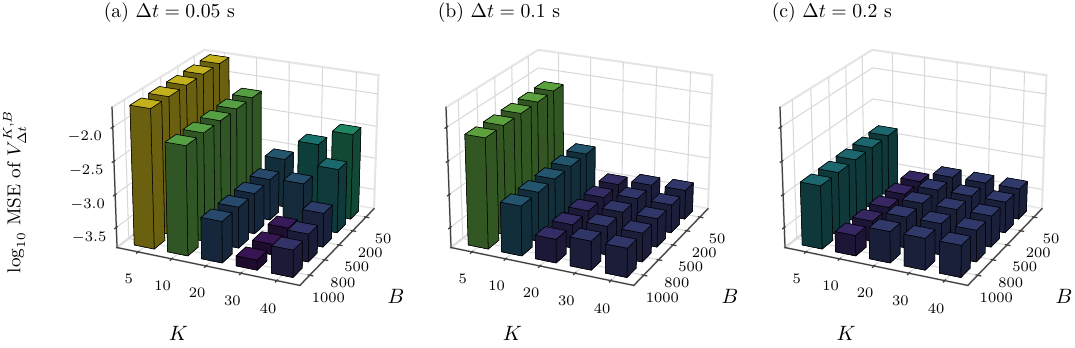}
  \caption[Label error on the inverted pendulum as a function of the search parameters]{Mean squared error of the generated labels \(V^{K,B}_{\Delta t}\) against the numerically computed value function on the inverted pendulum over a uniform lattice of \(100\times100\) states, as a function of the beam width \(B\) for the horizon lengths \(K\) and the step sizes (a)~\(\Delta t=\SI{0.05}{\second}\), (b)~\(\SI{0.1}{\second}\), and (c)~\(\SI{0.2}{\second}\).}
  \label{fig:vertex-cbf-ablation-mse}
\end{figure}

\begin{figure}[tb]
  \centering
  \includegraphics[width=0.5\textwidth]{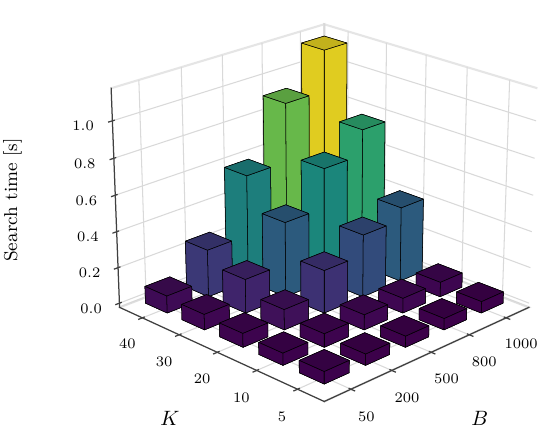}
  \caption[Execution time of the vertex-restricted beam search]{Execution time of the vertex-restricted beam search on the inverted pendulum, as a function of the beam width \(B\) for the horizon lengths \(K\). With a single input, the search tree has at most \(2^{k}\) nodes at depth \(k\), so the beam width takes effect only once the tree is wider than the beam, and for long horizons the runtime grows approximately in proportion to \(BK\). It does not depend on the step size \(\Delta t\).}
  \label{fig:vertex-cbf-ablation-runtime}
\end{figure}

The label quality is governed by the three parameters of \cref{eq:vertex-cbf-error-decomposition}, the beam width \(B\), the horizon length \(K\), and the step size \(\Delta t\). This subsection measures their effect first on the inverted pendulum, whose numerically computed value function provides a reference, and then on a high-dimensional system, where no reference exists.

\paragraph{Beam width and horizon length.}
The sweep on the inverted pendulum covers all combinations of \(B\in\{50,200,500,800,1000\}\), \(K\in\{5,10,20,30,40\}\), and \(\Delta t\in\{0.05,0.1,0.2\}\)~\si{\second}. \Cref{fig:vertex-cbf-ablation-mse} reports the mean squared error of the labels against the reference over a uniform lattice, and three trends are visible, each of which corresponds to one term of the decomposition. First, if the time horizon \(K\Delta t\) is too short to reveal an unavoidable violation, the error is large and the labels overestimate the safety value, which is the horizon-truncation term. Second, as \(K\) grows, the beam width must grow with it, because a deeper tree offers more opportunities for the beam to discard the branch that leads to the optimal sequence, which is the search-truncation term. Third, a larger step size covers the same time horizon with fewer stages and needs a smaller \(K\), but increasing it further degrades the discretization itself, which is the first term and makes the choice of \(\Delta t\) a compromise. \Cref{fig:vertex-cbf-ablation-runtime} reports the corresponding generation time. It grows approximately in proportion to \(BK\) and is independent of \(\Delta t\), which matches the complexity \(O(NKBM)\) of \cref{subsec:vertex-cbf-tree-methods}. Label quality is thus obtained at a predictable and linear cost, while the cost of a grid-based solution grows exponentially in the state dimension.

\paragraph{Step size on a high-dimensional system.}
\begin{table}[tb]
  \caption[Influence of the step size on the underwater vehicle]{Influence of the step size \(\Delta t\) on the six-degree-of-freedom underwater vehicle. The horizon length \(K=40\) and the beam width \(B=800\) are fixed, so the time horizon \(K\Delta t\) ranges from \SI{0.8}{\second} to \SI{16}{\second}. Entries are mean and standard deviation over three training seeds.}
  \label{tab:vertex-cbf-ablation-dt}
  \centering
  \small
  \providecommand{\stdv}[1]{\,\scalebox{0.75}{$\pm$#1}}
  \begin{tabular}{@{}cccc@{}}
    \toprule
    \(\Delta t\)~[\si{\second}] & \(\rho_{\mathrm{FS}}\) (\%) \(\downarrow\) & \(\rho_{\mathrm{FU}}\) (\%) \(\downarrow\) & \(\eta_{\mathrm{eff}}\) \(\uparrow\) \\
    \midrule
    0.02 & 3.94\stdv{0.24} & 2.88\stdv{0.03} & 0.340\stdv{0.001} \\
    0.05 & 0.63\stdv{0.03} & 3.12\stdv{0.02} & 0.354\stdv{0.000} \\
    0.1  & 1.15\stdv{0.06} & 2.16\stdv{0.02} & 0.356\stdv{0.000} \\
    0.2  & 1.18\stdv{0.13} & 3.22\stdv{0.06} & 0.348\stdv{0.000} \\
    0.4  & 0.51\stdv{0.13} & 7.02\stdv{0.14} & 0.317\stdv{0.000} \\
    \bottomrule
  \end{tabular}
\end{table}

\begin{figure}[tb]
  \centering
  \includegraphics[width=\textwidth]{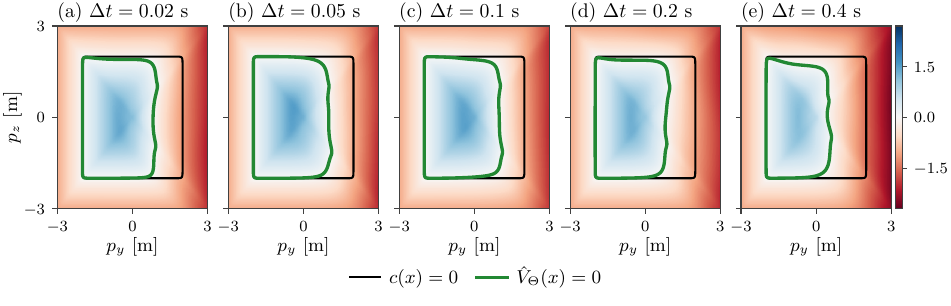}
  \caption[Learned barrier slices for the underwater vehicle at different step sizes]{Slices of the learned barrier \(\hat{V}_{\Theta}\) for the six-degree-of-freedom underwater vehicle in the \((p_y,p_z)\) plane, one for each step size of \cref{tab:vertex-cbf-ablation-dt}, with red shading marking the states predicted to be unsafe. The black line is the constraint boundary and the green line is the boundary of the predicted safe set. The predicted safe set becomes visibly smaller at the largest step size.}
  \label{fig:vertex-cbf-ablation-dt}
\end{figure}
The same trade-off is examined on the underwater vehicle, which has twelve states and no numerical reference. The supervision data are regenerated for \(\Delta t\in\{0.02,0.05,0.1,0.2,0.4\}\)~\si{\second}, and a model is trained from scratch for each value with three seeds. All other settings are those of \cref{app:vertex-cbf-config}, including \(K=40\) and \(B=800\), so the time horizon ranges from \SI{0.8}{\second} to \SI{16}{\second} while the search cost remains the same. \Cref{tab:vertex-cbf-ablation-dt} and \cref{fig:vertex-cbf-ablation-dt} report the result, in which the two extremes fail for different reasons. At \(\Delta t=\SI{0.02}{\second}\), the time horizon of \SI{0.8}{\second} is shorter than the braking time of the vehicle, so the labels do not detect later violations and the false-safe rate increases to \SI{3.94}{\percent}. At \(\Delta t=\SI{0.4}{\second}\), the discretization itself becomes inaccurate, the predicted safe set becomes visibly smaller, the false-unsafe rate increases to \SI{7.02}{\percent}, and the effective safe volume decreases to \(0.317\). Between these extremes the method is stable, since for \(\Delta t\) between \SI{0.05}{\second} and \SI{0.2}{\second} the false-safe rate remains at or below \SI{1.18}{\percent} and the effective safe volume varies only between \(0.348\) and \(0.356\), with a small spread over the seeds. The best result is obtained at \(\Delta t=\SI{0.1}{\second}\), which is the value used in \cref{tab:vertex-cbf-results}. The resulting guideline is to choose \(\Delta t\) small enough for the Euler step to remain accurate and then to choose \(K\) so that \(K\Delta t\) covers the time the system needs to reach a state in which the attained safety level can be maintained, for example a stopped state. The beam width must then grow with \(K\).

\subsection{Architecture and Training Ablations}
\label{subsec:vertex-cbf-ablations}
\label{subsec:vertex-cbf-ablation-residual}

Two ablations examine the two components that the training combines with the supervision, the residual parameterization and the physics-informed loss term.

\paragraph{Residual architecture.}

\begin{table}[tb]
  \caption[Validation results without the residual architecture]{Validation results without the residual architecture, in which the network predicts the value function directly through a linear output instead of a nonnegative residual subtracted from \(c\). All other settings and reporting conventions are those of \cref{tab:vertex-cbf-results}. A dash marks a predicted safe set that was empty for every seed.}
  \label{tab:vertex-cbf-ablation-residual}
  \centering
  \scriptsize
  \providecommand{\stdvs}[1]{\,\scalebox{0.68}{$\pm$#1}}
  \setlength{\tabcolsep}{1pt}
  \begin{tabular*}{\textwidth}{@{\extracolsep{\fill}}l ccc ccc ccc@{}}
    \toprule
    \multirow{2}{*}{System (\(n_x\), \(n_u\))} & \multicolumn{3}{c}{\(\rho_{\mathrm{FS}}\) (\%) \(\downarrow\)} & \multicolumn{3}{c}{\(\rho_{\mathrm{FU}}\) (\%) \(\downarrow\)} & \multicolumn{3}{c}{\(\eta_{\mathrm{eff}}\) \(\uparrow\)} \\
    \cmidrule(lr){2-4} \cmidrule(lr){5-7} \cmidrule(lr){8-10}
                            & ND & FCD & \shortstack{VRCD\\(ours)} & ND & FCD & \shortstack{VRCD\\(ours)} & ND & FCD & \shortstack{VRCD\\(ours)} \\
    \midrule
    Inverted Pendulum (2, 1) & --                & \textbf{1.06}\stdvs{0.19}  & 6.49\stdvs{3.23}            & 0.00\stdvs{0.01}           & 14.91\stdvs{0.76}          & \textbf{0.29}\stdvs{0.17}           & 0.00\stdvs{0.00}          & 0.08\stdvs{0.00}           & \textbf{0.21}\stdvs{0.01} \\
    1D Double Integrator (2, 1) & --                & 0.16\stdvs{0.07}           & \textbf{0.15}\stdvs{0.07}   & 26.15\stdvs{15.13}         & 4.35\stdvs{0.28}           & \textbf{1.20}\stdvs{0.10}  & 0.00\stdvs{0.00}          & 0.39\stdvs{0.00}           & \textbf{0.41}\stdvs{0.00} \\
    2D Vertical Drone (2, 1) & --                & --                        & \textbf{0.14}\stdvs{0.06}   & 3.95\stdvs{6.12}           & 4.06\stdvs{7.96}           & \textbf{1.80}\stdvs{0.14}  & 0.00\stdvs{0.00}          & 0.00\stdvs{0.00}           & \textbf{0.58}\stdvs{0.00} \\
    Dubins Car (3, 1) & --                & \textbf{0.08}\stdvs{0.04}  & 0.18\stdvs{0.07}            & 81.06\stdvs{4.98}          & 4.78\stdvs{1.16}           & \textbf{2.51}\stdvs{0.49}  & 0.00\stdvs{0.00}          & \textbf{0.92}\stdvs{0.00}  & \textbf{0.92}\stdvs{0.00} \\
    2D Double Integrator (4, 2) & --                & 1.60\stdvs{0.55}           & \textbf{1.58}\stdvs{0.83}   & 67.67\stdvs{10.10}         & \textbf{3.69}\stdvs{0.39}  & 3.87\stdvs{0.43}           & 0.00\stdvs{0.00}          & \textbf{0.91}\stdvs{0.01}  & \textbf{0.91}\stdvs{0.01} \\
    Kinematic Bicycle (4, 2) & --                & \textbf{0.19}\stdvs{0.12}  & 0.73\stdvs{0.81}            & 71.98\stdvs{5.65}          & \textbf{4.91}\stdvs{0.25}  & 5.77\stdvs{0.45}           & 0.00\stdvs{0.00}          & \textbf{0.91}\stdvs{0.00}  & \textbf{0.91}\stdvs{0.01} \\
    Cart-Pole (4, 1) & --                & --                        & \textbf{20.39}\stdvs{20.37} & 1.66\stdvs{2.98}           & 6.00\stdvs{10.21}          & \textbf{1.96}\stdvs{0.62}           & 0.00\stdvs{0.00}          & 0.00\stdvs{0.00}           & \textbf{0.26}\stdvs{0.07} \\
    Dynamic Unicycle (5, 2) & --                & \textbf{0.15}\stdvs{0.05}  & 0.26\stdvs{0.05}            & 79.18\stdvs{1.55}          & \textbf{8.41}\stdvs{1.10}  & 10.61\stdvs{0.69}          & 0.00\stdvs{0.00}          & \textbf{0.93}\stdvs{0.00}  & 0.92\stdvs{0.00} \\
    Relative Unicycle (5, 2) & --                & \textbf{0.17}\stdvs{0.11}  & 0.36\stdvs{0.07}            & 83.25\stdvs{2.48}          & 39.64\stdvs{0.67}          & \textbf{4.67}\stdvs{1.05}  & 0.00\stdvs{0.00}          & 0.88\stdvs{0.00}           & \textbf{0.93}\stdvs{0.00} \\
    3D Double Integrator (6, 3) & --                & 0.79\stdvs{0.71}           & \textbf{0.75}\stdvs{0.87}   & 84.35\stdvs{3.09}          & \textbf{4.62}\stdvs{0.77}  & 4.97\stdvs{1.04}           & 0.00\stdvs{0.00}          & \textbf{0.96}\stdvs{0.01}  & \textbf{0.96}\stdvs{0.01} \\
    3-DOF Manipulator (6, 3) & 40.01\stdvs{32.95} & 0.48\stdvs{0.10}           & \textbf{0.44}\stdvs{0.05}   & \textbf{0.56}\stdvs{0.29}  & 7.31\stdvs{0.18}           & 5.88\stdvs{0.05}           & 0.43\stdvs{0.24}          & 0.65\stdvs{0.00}           & \textbf{0.66}\stdvs{0.00} \\
    Landing Rocket (7, 2) & --                & 80.40\stdvs{9.66}          & \textbf{1.10}\stdvs{0.41}   & 0.00\stdvs{0.00}           & \textbf{2.22}\stdvs{1.40}  & 6.01\stdvs{0.14}           & 0.00\stdvs{0.00}          & 0.01\stdvs{0.01}           & \textbf{0.24}\stdvs{0.00} \\
    Quadruped Trunk (9, 4) & --                & --                        & \textbf{15.62}\stdvs{14.32} & 0.00\stdvs{0.00}           & 1.96\stdvs{1.79}           & \textbf{2.47}\stdvs{0.28}           & 0.00\stdvs{0.00}          & 0.00\stdvs{0.00}           & \textbf{0.06}\stdvs{0.01} \\
    6-DOF Underw. Veh. (12, 6) & --                & \textbf{1.92}\stdvs{0.15}  & 3.24\stdvs{0.16}            & 1.65\stdvs{0.32}           & 5.38\stdvs{0.03}           & \textbf{2.49}\stdvs{0.04}           & 0.00\stdvs{0.00}          & 0.33\stdvs{0.00}           & \textbf{0.35}\stdvs{0.00} \\
    Quadrotor (13, 4) & --                & 6.09\stdvs{2.48}           & \textbf{6.04}\stdvs{4.17}   & 82.36\stdvs{0.62}          & \textbf{6.93}\stdvs{0.50}  & 7.78\stdvs{0.57}           & 0.00\stdvs{0.00}          & \textbf{0.87}\stdvs{0.02}  & \textbf{0.87}\stdvs{0.04} \\
    \bottomrule
  \end{tabular*}
\end{table}

All three settings of \cref{tab:vertex-cbf-results} use the residual parameterization. To isolate its contribution, every experiment is repeated with a network that predicts the value function directly, by replacing the softplus output of the residual with a linear output. Architecture, hyperparameters, supervision data, and validation protocol are otherwise unchanged, and each configuration is again trained with five seeds, with the results reported in \cref{tab:vertex-cbf-ablation-residual}.

The residual parameterization proves essential for the physics-informed loss alone. Without it, training converges to an empty predicted safe set on 14 of the 15 systems, so the false-safe rate is undefined, the effective safe volume is zero, and the large false-unsafe rates only reflect that every state is predicted unsafe. Only the manipulator retains a non-empty predicted safe set, with a high and strongly seed-dependent false-safe rate. This behavior is expected, because a sufficiently negative constant makes the loss \cref{eq:vertex-cbf-pde-loss} vanish everywhere, and a direct network represents such a constant with constant outputs. The residual network, in contrast, would have to reproduce the shape of \(c\), as anticipated in \cref{subsec:vertex-cbf-training}.

The supervised settings are considerably less sensitive, because the data term already fixes the solution at nontrivial values. Removing the residual parameterization nevertheless degrades most results. For vertex-restricted supervision, the false-safe rate increases on 13 of the 15 systems, most strongly on the cart-pole and the quadruped trunk, where it increases to \SI{20.39}{\percent} and \SI{15.62}{\percent} with a large spread across seeds. The effective safe volume does not increase on any system and decreases most on the cart-pole and the quadrotor. The two components are complementary. The residual parameterization reduced the tendency toward degenerate solutions in the evaluated settings and guarantees \(\hat{\mathcal{S}}_{\Theta}\subseteq\mathcal{C}\) by construction, while the supervision data determine how closely \(\hat{\mathcal{S}}_{\Theta}\) approaches the maximal safe set.

\paragraph{Loss weight.}
\label{subsec:vertex-cbf-ablation-lambda}

\begin{figure}[tb]
  \centering
  \includegraphics[width=\textwidth]{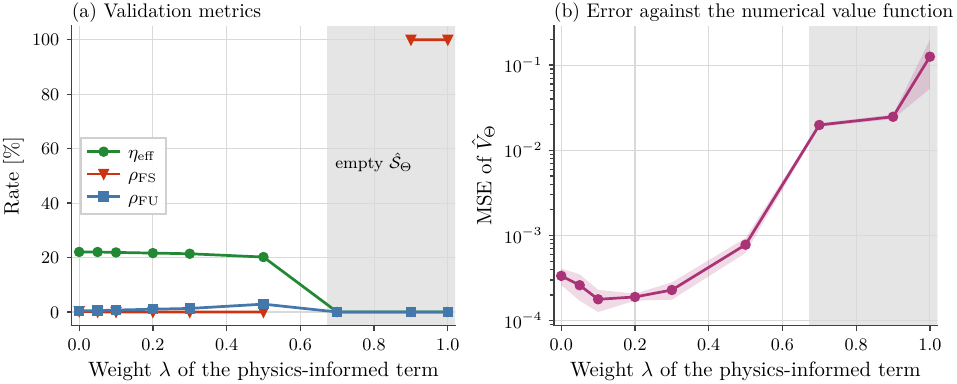}
  \caption[Sensitivity to the weight of the physics-informed term]{Sensitivity to the weight \(\lambda\) of the physics-informed term on the inverted pendulum. (a)~Validation metrics. (b)~Mean squared error against the numerically computed value function. Curves are means over three seeds and the bands are one standard deviation. In the shaded region the predicted safe set is empty for \(\lambda=0.7\), where \(\eta_{\mathrm{eff}}\) is zero and \(\rho_{\mathrm{FS}}\) is undefined, and every state predicted safe is unsafe for \(\lambda\geq0.9\).}
  \label{fig:vertex-cbf-ablation-lambda}
\end{figure}

The weight \(\lambda\) in \cref{eq:vertex-cbf-total-loss} balances the physics-informed term and the data term. It is varied on the inverted pendulum with every other setting of \cref{app:vertex-cbf-config} unchanged and three seeds per value, where \(\lambda=0\) trains on the labels alone and \(\lambda=1\) is the physics-informed setting of \cref{tab:vertex-cbf-results}.

\Cref{fig:vertex-cbf-ablation-lambda} shows the result. The method is insensitive to \(\lambda\) over the whole range from \(0\) to \(0.5\). The effective safe volume remains between \SI{20.2}{\percent} and \SI{22.1}{\percent}, the false-safe rate remains below \SI{0.3}{\percent}, and the error against the reference varies between \(1.8\cdot10^{-4}\) and \(7.8\cdot10^{-4}\), with its minimum near \(\lambda=0.1\). The behavior changes abruptly above \(\lambda=0.5\). At \(\lambda=0.7\) the predicted safe set is empty for every seed, and at \(\lambda=0.9\) and \(\lambda=1\) it is non-empty but every state predicted safe is unsafe, while the error against the reference grows by more than two orders of magnitude. The data term is what keeps the optimization away from the trivial solutions of the physics-informed term, and a moderate weight on that term is enough to enforce local consistency without dominating the objective. The value \(\lambda=0.2\) used for this system lies inside the flat region, and \(\lambda\) was selected for the other systems in the same way, by choosing a value inside the range in which the validation metrics are flat. On four systems this value is zero, as \cref{app:vertex-cbf-config} records.

\subsection{Hardware Experiments}
\label{subsec:vertex-cbf-hardware}

\begin{figure}[tb]
  \centering
  \includegraphics[width=\textwidth]{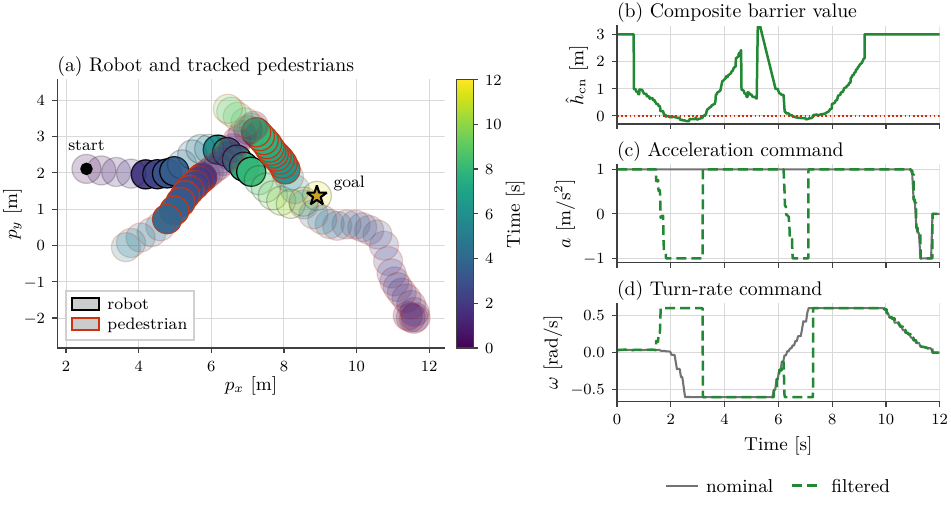}
  \caption[VertexCBF hardware experiment with the delivery robot]{Hardware experiment with the delivery robot. (a)~The recorded robot motion and the tracked pedestrians in the workspace, drawn as circles colored by time. Opaque circles mark the two encounters in which a pedestrian crosses the path of the robot. (b)~The composite barrier value \(\hat{h}_{\mathrm{cn}}\) over the run, with the zero level marked by the dotted line. While no pedestrian is tracked, the barrier is undefined and its value is drawn at \SI{3}{\meter}. (c),~(d)~The nominal and the filtered acceleration and turning-rate commands. The filter deviates from the nominal command only while a pedestrian is on an intercepting course, and the barrier value becomes slightly negative for a short period while the pedestrians pass the robot.}
  \label{fig:vertex-cbf-hardware}
\end{figure}

\begin{figure}[tb]
  \centering
  \includegraphics[width=\textwidth]{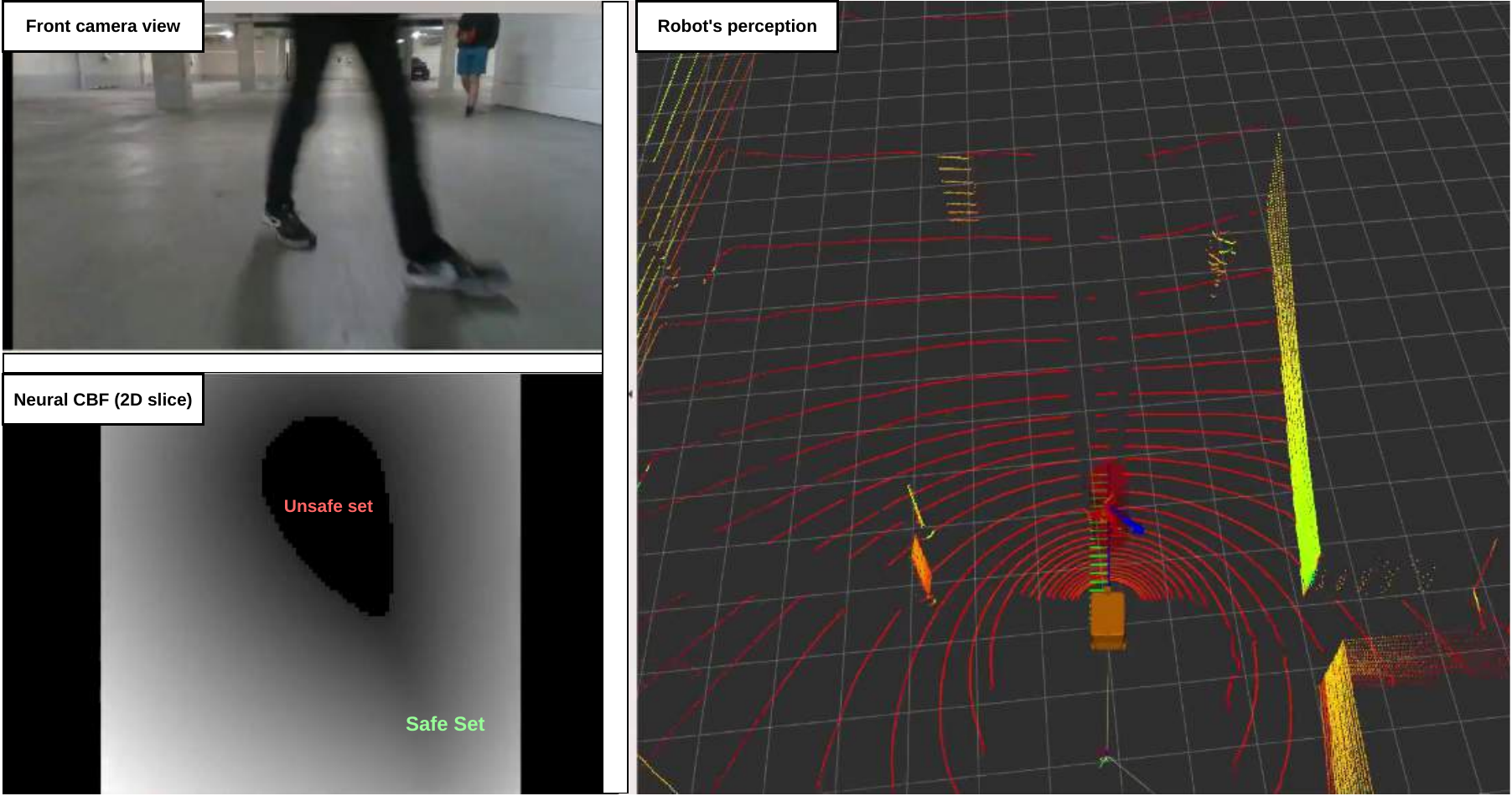}
  \caption[Onboard view during a VertexCBF hardware trial]{View from the onboard computer of the delivery robot during a trial, showing the front camera image, a positional slice of the learned barrier, and the output of the perception module.}
  \label{fig:vertex-cbf-hardware-screen}
\end{figure}

The final study asks whether a barrier trained by the proposed method operates on a physical robot under real perception. The platform, the task, and the deployment pipeline are those of \cref{subsec:cn-cbf-hardware}, so the two chapters run the same composition and the same filter on the same robot. Object detection and tracking, localization, and the QP solution all run on the onboard computer. An MPC planner without safety constraints supplies the nominal command that drives the robot from its start position to its goal. The barrier is evaluated for the relative state between the robot and each tracked pedestrian, the pairwise values are aggregated into one composite constraint as in \cref{subsec:cn-cbf-composition}, and the time-varying CBF-QP filter \cref{eq:cn-cbf-qp} modifies the nominal command, with \(\alpha(s)=\gamma s\) and \(\gamma=2.8\). As in the preceding chapters, the constraint is relaxed by a penalized slack variable, so the QP always returns an input, even when the perception module reports a new pedestrian at a state where the hard constraint cannot be satisfied.

The barrier is the one trained for the relative unicycle in \cref{tab:vertex-cbf-results}, and its pairwise model differs from that of \cref{ch:cn-cbf} in two respects. The robot is a dynamic unicycle with inputs \((a,\omega)\), so its speed is a state that the filter influences only through an acceleration limit, while \cref{ch:cn-cbf} commands the speed directly. For this reason, the relative state has five components, the pedestrian position and heading in the robot frame, the robot speed, and the pedestrian speed. The pedestrian moves at a constant heading, while \cref{ch:cn-cbf} treats its turning rate as a bounded adversarial disturbance and solves a two-player game. The barrier deployed here assumes more about how the pedestrian moves, so the trials are a second navigation deployment that reuses the composition and the filter, and not a comparison of one barrier obtained from two supervision sources.

Ten trials were conducted in which one or two pedestrians deliberately crossed the path of the robot, and the robot avoided collision in all of them. \Cref{fig:vertex-cbf-hardware} shows a representative run and \cref{fig:vertex-cbf-hardware-screen} the corresponding view from the onboard computer. The filter departs from the nominal command only while a pedestrian is on an intercepting course. The barrier value remains close to zero during the encounter and becomes slightly negative for a short period while the pedestrians pass the robot. Possible causes include mismatch between the training model and the real system, estimation error, and sampled control. The barrier is learned for an ideal kinematic model, while the skid-steer robot slips during turns and its velocity controller has its own dynamics, the pedestrian states estimated from LiDAR and camera data are noisy and delayed, and the input is held constant between two QP solutions. To compensate, the robot and pedestrian radii used during training are \SI{5}{\centi\meter} larger than the physical dimensions, so the predicted safe set is slightly smaller than the collision-free set of the real system and a small negative barrier value does not correspond to a collision. The study is a deployment demonstration rather than a controlled comparison, since it does not evaluate baselines under the same conditions. Its purpose is to establish that a grid-free barrier is deployable in the same pipeline, at the same rate, and on the same hardware as the barriers of the preceding chapters.

\section{Discussion}
\label{sec:vertex-cbf-discussion}

Across 15 systems with up to 13 states, vertex-restricted supervision produced low false-safe rates and the largest validated safe regions, while physics-informed training alone was unreliable and full-control sampling was often conservative or produced empty sets. These results support grid-free barrier learning within the following limits.

\paragraph{Guarantee scope.}
Three kinds of statements are made about the method. The exclusion property \cref{eq:vertex-cbf-exclusion} holds by construction, and the vertex representation \cref{eq:vertex-cbf-hamiltonian-vertices} and the local consistency result \cref{eq:vertex-cbf-consistency} are derived properties of the Hamiltonian and its discretization. The approximation of the maximal safe set and the closed-loop behavior are measured by the validation protocol of \cref{subsec:vertex-cbf-validation}, because the consistency result is local and the label errors of \cref{eq:vertex-cbf-error-decomposition} are controlled by the search parameters rather than bounded in closed form. This measurement is the evidence that remains available for systems beyond the reach of numerical solvers, and sample-based certification methods such as scenario optimization or conformal prediction could convert it into a probabilistic statement with finite-sample coverage \cite{lin_verification_2024,feng_bridging_2025}. Verification of the CBF condition \cref{eq:prelim-cbf-condition} on the learned field is the same step as for the barriers of \cref{ch:orn-cbf,ch:cn-cbf}.

\paragraph{Scalability of the data generation.}
The vertex restriction removes the exponential dependence of the grid on the state dimension, but it introduces a dependence on the number of input vertices. For box-shaped input limits, \(M=2^{n_u}\), so the branching factor grows exponentially with the input dimension. The tree search cost \(O(NKBM)\) is linear in \(M\) for a fixed beam width, but a larger \(M\) requires a larger beam to retain near-optimal sequences, so the effective cost grows faster than the stated complexity suggests. The evaluated systems reached \(n_u=6\), which corresponds to 64 vertices and remained practical. Whole-body models of humanoid robots, with tens of actuated joints, would exceed this range, and for such systems the method is better applied to reduced-order models or to safety-critical subsystems, or extended with a search over a subset of the vertices, as \cref{ch:conclusion} outlines. The combinatorial bottleneck thus shifts from the state dimension to the number of input vertices, while the state dimension still affects the evaluation cost and the learning difficulty.

\paragraph{Relation to the preceding chapters.}
The formulation has three restrictions. First, it requires control-affine dynamics with polytopic input constraints. The CBF-based methods of \cref{ch:orn-cbf,ch:cn-cbf} already assume this structure, but the MPC methods of \cref{ch:ntc-mpc,ch:rntc-mpc} do not assume it, so the search cannot supervise them in general. Second, it learns a fixed field, so the hardware study composes a relative-state field as in \cref{ch:cn-cbf}, while an ORN-CBF extension would generate labels for each conditioning input. Third, it models no disturbance, although a two-player extension is possible in principle for dynamics \(f_0(x)+g(x)u+g_d(x)d\) and polytopic input sets, since the control and disturbance optimizations separate and both attain their extrema at vertices. The branching factor of the resulting game tree is the product of the two vertex counts. Extending the search and analysis to the robust targets of \cref{ch:cn-cbf} remains future work.

\paragraph{Choice of the search parameters and loss weight.}
The studies of \cref{subsec:vertex-cbf-ablation-search} give the same guidance on a system with a reference and on one without, and within the stated limits there is a broad range of values in which the result is stable. On a new system, however, none of the parameters can be verified against a reference, and the available check is the validation protocol itself. Labels that are too optimistic because of a short horizon increase the false-safe rate, and labels that are too conservative because of a coarse step or a narrow beam increase the false-unsafe rate. Parameter selection is thus an iteration between generation and validation rather than a direct computation, which is a practical limitation of the method. A similar remark applies to the loss weight, since the experiments establish different things about the two loss terms. The data term was necessary on the evaluated systems, since the physics-informed setting failed on many of them and, on the inverted pendulum, the result collapsed once the weight of the physics-informed term exceeded 0.5. The physics-informed term constrains the field at states that carry no label and permits a sparser set of supervision states, but four of the evaluated systems were trained on the labels alone, so its contribution depends on the system. Its weight did not require fine tuning within the flat region, but it remains a per-system choice, and a systematic study of its interaction with the number of supervision and collocation states remains to be done.

\section{Summary}
\label{sec:vertex-cbf-summary}

VertexCBF generates labels through GPU-parallel search over input vertices and trains a residual barrier with data and physics-informed losses. The Hamiltonian vertex property is exact, while the labels incur discretization, horizon, and search errors, and their generation avoids a state-space grid but still depends on the number of input vertices. Across 15 systems with 2--13 states and 1--6 inputs, vertex-restricted supervision achieved mean false-safe rates below \SI{1.2}{\percent} and reached the highest or tied-highest effective safe volume, with IoU against numerical safe sets between \SI{93}{\percent} and \SI{99.8}{\percent} on the seven reference systems. Ablations supported residual learning and showed stable results over tested ranges of step sizes and loss weights, and a composed relative-state barrier avoided collisions in all ten delivery-robot trials. These results answer question~\ref{rq:supervision}, since grid-free supervision extends the approach beyond the practical state dimensions of the earlier solvers, while finite-horizon rollout validation supplies empirical evidence where no numerical reference exists. \Cref{ch:conclusion} compares this contribution with the other four.

\bookmarksetup{startatroot}
\addtocontents{toc}{\protect\addvspace{2.25em plus 1pt}}
\chapter{Conclusions and Future Work}
\label{ch:conclusion}

This dissertation developed five methods that make reachability-based safety information available to robot controllers in unknown and dynamic environments. Neural implicit representations replace repeated online Hamilton--Jacobi (HJ) computation with fast value and gradient queries for model predictive control (MPC) and control barrier function (CBF) safety filters. For control-affine systems with polytopic input constraints, the fifth method replaces grid-based supervision with labels from a search over the input vertices, which extends the evaluated state dimensions beyond the practical range of grid-based solvers.

This chapter first answers the research questions. It then compares the five methods, separates the safety properties that hold by theory or by construction from those that were demonstrated experimentally, and closes with an outlook.

\section{Answers to the Research Questions}
\label{sec:conclusion-rq}

The introduction posed four research questions in \cref{sec:problem-statement}. This section answers them in order, and the following two sections give the detailed comparison and the safety properties of the methods together with their assumptions.

\paragraph{Question~\ref{rq:realtime}. Real-time safety functions.}
Reachability-based safety functions can be made available in real time by solving the reachability problems offline and replacing the online computation with the evaluation of a network trained on their solutions. For the observation-conditioned methods, the hypernetwork factorization matches how optimization-based controllers use the function, since they process each observation once and then query the value and gradient many times. CN-CBF reaches the same goal without scene conditioning, because one pairwise model in relative coordinates can be evaluated in batches for any obstacle list. Through the MPC interface, the learned function enters as one scalar terminal constraint on the final predicted state. At the shortest evaluated horizon, the planners with this constraint achieved considerably higher success rates than the distance-based formulations. Through the safety-filter interface, the learned barriers ran in CBF-QP filters at several hundred hertz in the evaluated configurations, achieved higher success rates than the evaluated classical barriers, and matched or exceeded the success rates of the MPC planners. The experiments also provide evidence of generalization to new observations and environments, including out-of-domain obstacle geometry for ORN-CBF.

\paragraph{Question~\ref{rq:dynamic}. Moving obstacles and changing obstacle counts.}
The dissertation developed two approaches. RNTC-MPC conditions the terminal constraint on the \emph{predicted} obstacle geometry, and because its input consists of two SDF snapshots taken a fixed time apart, one trained model can be used with any prediction horizon that is at least as long as this time difference. CN-CBF makes each pairwise target stationary in relative coordinates and composes the learned pairwise barriers for a varying obstacle count. The smooth minimum underestimates the exact minimum by at most \(\ln(m)/\beta\), and a newly detected obstacle can be added safely if the composite value had a sufficient margin before the update or if the new composite value is checked directly. The intersection of the pairwise safe sets can be larger than the true joint safe set when the maneuvers that avoid different obstacles conflict, which \cref{subsec:conclusion-representations} discusses. The direct comparison favored CN-CBF at every tested density, although the data do not establish how often such conflicting configurations occurred.

\paragraph{Question~\ref{rq:structure}. Structural safety properties.}
The residual parameterization represents the learned safety function as a directly computed constraint function minus a nonnegative network output. The predicted safe set therefore excludes the represented failure set regardless of the approximation error of the network. Because the exact value function itself has this form, the parameterization does not restrict what the network can represent. It applies whether the constraint geometry is observed, predicted, or expressed in relative coordinates, and it is preserved under smooth composition (\cref{eq:rntc-failure-set-exclusion,eq:orn-cbf-exclusion,eq:cn-cbf-pairwise-exclusion,eq:cn-cbf-softmin-bounds}). The structure also improved learning in the controlled ablations of \cref{ch:rntc-mpc,ch:vertex-cbf}. The remaining safety-relevant properties are either inherited from HJ theory or established empirically, as \cref{sec:conclusion-guarantees} details. Verifying the conditions for closed-loop safety, recursive feasibility of the MPC problem and validity of the CBF condition, on the learned fields is the natural next step, and \cref{sec:conclusion-guarantees} shows that the structural properties established here are what such a verification builds on.

\paragraph{Question~\ref{rq:supervision}. Scalable reachability supervision.}
Reachability supervision can be generated without a grid-based solver by exploiting the control-affine dynamics and polytopic input sets that the safety filters here already use. Because the Hamiltonian is maximized at a vertex, restricting the inner maximization to the vertices is exact in continuous time. Discretizing the trajectories then turns the label computation into a finite tree search over vertex inputs. The labels at different states are independent, and for a fixed beam width the number of dynamics evaluations grows linearly with the number of sampled states, the horizon, the beam width, and the number of vertices. Three qualifications accompany this answer. The vertex statement holds in continuous time, while the labels come from a finite Euler step over a truncated horizon with a bounded search width, and the relation between the two is established locally through the consistency result rather than through a convergence theorem. The results demonstrate scaling to 13 states empirically and not independence from the state dimension, and the exponential growth moves from the state dimension to the input dimension, because the number of vertices of a box-shaped input set doubles with every additional input. Where no numerical reference exists, the accuracy is assessed by closed-loop rollouts.

\section{Comparison of the Five Methods}
\label{sec:comparative-discussion}

\begin{figure}[!htb]
  \centering
  \newcommand{\desc}[1]{\parbox{\linewidth}{\centering\scriptsize #1}}
\begin{tikzpicture}[
    font=\small,
    method/.style={draw, rounded corners=3pt, align=center, inner sep=5pt,
      text width=52mm, minimum height=21mm, fill=black!4},
    axis/.style={font=\small\bfseries, align=center},
    group/.style={draw=black!55, thin, dashed, rounded corners=7pt},
  ]

  \draw[group] (-2.95,-1.95) rectangle (2.95,8.15);
  \draw[group] ( 4.75,-1.95) rectangle (10.65,8.15);
  \draw[group] (-4.45, 3.40) rectangle (10.95,6.60);
  \draw[group] (-4.45,-1.60) rectangle (10.95,1.60);

  \node[method] (ntc) at (0,5)
    {\textbf{NTC-MPC} (\cref{ch:ntc-mpc})\\[4pt]
     \desc{direct value regression,\\
           conditioned on the observed SDF}};
  \node[method] (orn) at (7.7,5)
    {\textbf{ORN-CBF} (\cref{ch:orn-cbf})\\[4pt]
     \desc{residual barrier,\\
           conditioned on the observed SDF}};
  \node[method] (rntc) at (0,0)
    {\textbf{RNTC-MPC} (\cref{ch:rntc-mpc})\\[4pt]
     \desc{residual value regression, conditioned\\
           on predicted SDF sequences}};
  \node[method] (cn) at (7.7,0)
    {\textbf{CN-CBF} (\cref{ch:cn-cbf})\\[4pt]
     \desc{residual pairwise barriers in relative\\
           coordinates, smooth composition}};

  \begin{scope}[on background layer]
    \draw[group] (-4.45,-5.88) rectangle (10.95,-2.68);
  \end{scope}

  \node[method] (vertex) at (7.7,-4.28)
    {\textbf{VertexCBF} (\cref{ch:vertex-cbf})\\[4pt]
     \desc{residual barrier from vertex-restricted\\
           search and physics-informed learning}};

  \node[axis] at (0,7.4) {MPC terminal\\constraint (Part~I)};
  \node[axis] at (7.7,7.4) {CBF safety\\filter (Part~II)};
  \node[axis, rotate=90] at (-3.7,5) {Static\\environments};
  \node[axis, rotate=90] at (-3.7,0) {Dynamic\\environments};
  \node[axis] at (0,-4.28) {Grid-free supervision\\(Part~III)};
\end{tikzpicture}
  \caption[The five methods organized by their design choices]{The five methods organized by the design choices that separate them. The columns of the grid give the controller interface, its rows give the environment class for which the reachability targets are computed, and each box names the learned representation and its conditioning input. The band below the grid holds the third design choice, the source of the supervision, which VertexCBF generates from the vertex structure of the input set instead of taking it from a grid-based solver.}
  \label{fig:conclusion-method-map}
\end{figure}
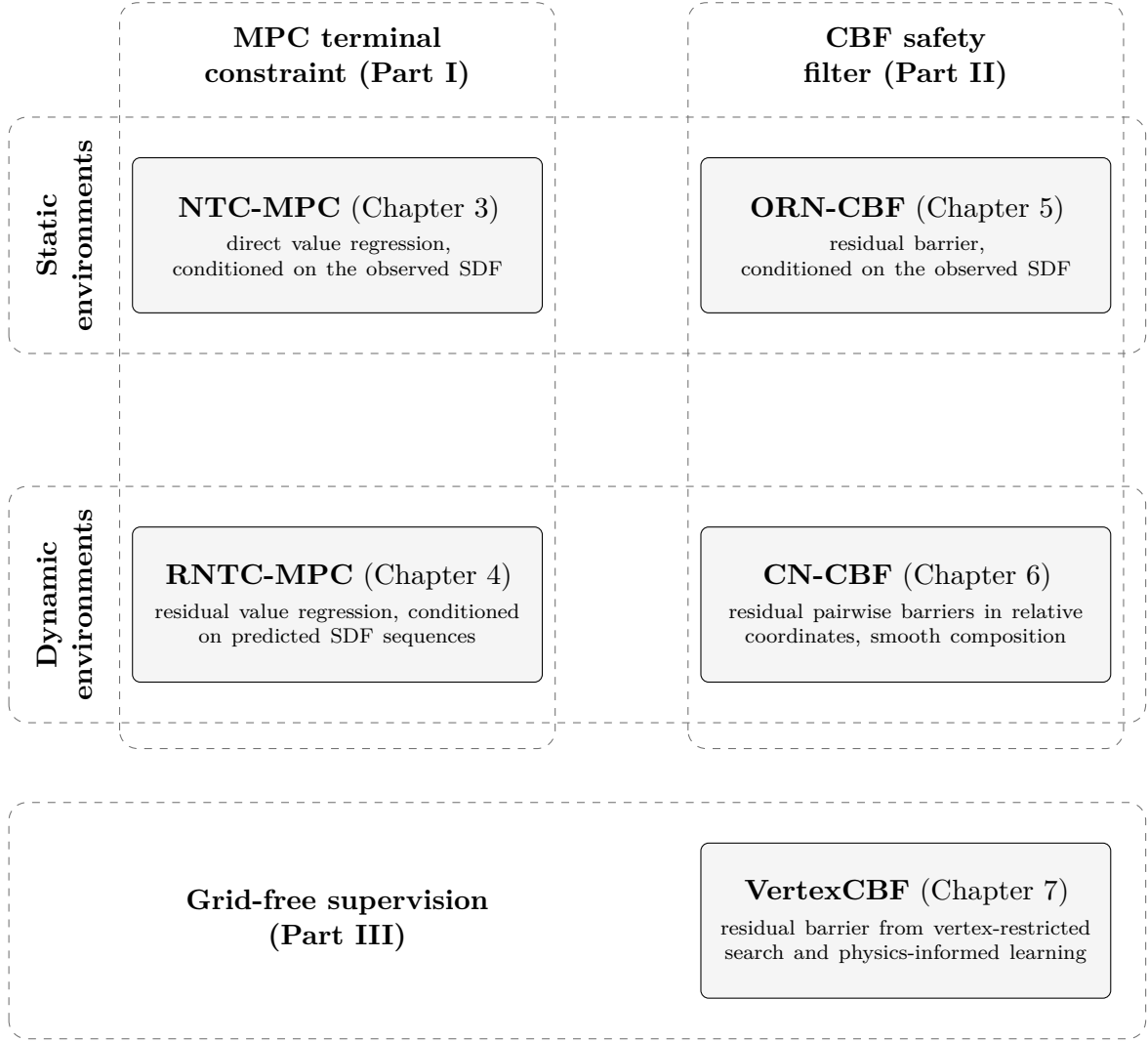

The five methods differ in three design choices, which \cref{fig:conclusion-method-map} summarizes. The first four combine a static or dynamic environment with an MPC or safety-filter interface, and the fifth changes the source of the supervision. \Cref{tab:method-comparison} lists further properties of the five methods. The following subsections compare the controller interfaces, the safety representations and their training, and the sources of supervision, and the section closes with guidance for selecting a method. Wherever possible, the comparison draws on experiments in which the methods were tested on the same robots, in the same environments, and against the same baselines. ORN-CBF was compared with NTC-MPC in static environments, and CN-CBF with RNTC-MPC in pedestrian crowds. Direct-regression ablations isolate residual learning in \cref{subsec:rntc-training-eval,sec:orn-cbf-experiments,subsec:vertex-cbf-ablations}, and VertexCBF varies the supervision source under one architecture and one validation protocol (\cref{subsec:vertex-cbf-benchmark}).

\begin{table}[!htb]
  \caption[Overview of the five methods]{Overview of the five methods, summarizing for each one the controller interface and setting, the learned representation, the source of the reachability supervision, and the computation performed online.}
  \label{tab:method-comparison}
  \centering
  \footnotesize
  \setlength{\tabcolsep}{4pt}
  \renewcommand{\arraystretch}{1.1}
  \begin{tabular}{@{}R{0.125\textwidth}R{0.18\textwidth}R{0.22\textwidth}R{0.2\textwidth}R{0.205\textwidth}@{}}
    \toprule
    Method & Interface and setting & Learned representation & Offline supervision & Online operation \\
    \midrule
    \mbox{NTC-MPC}\newline(\cref{ch:ntc-mpc})
     & MPC terminal constraint for a static environment observed locally
     & Value function regressed directly by a hypernetwork-generated field conditioned on the observed SDF
     & One stationary grid-based HJ solution per training observation
     & One hypernetwork pass per planning cycle, then value and gradient queries of the compact field inside the NLP \\
    \addlinespace[5pt]
    \mbox{RNTC-MPC}\newline(\cref{ch:rntc-mpc})
     & MPC terminal constraint for moving obstacles under a nominal motion prediction
     & Predicted terminal SDF minus a nonnegative residual, conditioned on snapshots of the predicted SDF sequence
     & One time-varying grid-based HJ solution per training scene
     & Motion prediction and SDF sequence, one hypernetwork pass per cycle, then queries of the compact field inside the NLP \\
    \addlinespace[5pt]
    \mbox{ORN-CBF}\newline(\cref{ch:orn-cbf})
     & CBF-QP safety filter for a static environment observed locally
     & Observed SDF minus a nonnegative residual generated by a hypernetwork conditioned on the observed SDF
     & One stationary grid-based HJ solution per training observation, cropped to a central positional patch
     & One hypernetwork pass per observation update, then value and gradient at the current state and one QP per control step \\
    \addlinespace[5pt]
    \mbox{CN-CBF}\newline(\cref{ch:cn-cbf})
     & CBF-QP safety filter for a varying number of moving obstacles with bounded inputs
     & Relative obstacle distance minus a nonnegative residual for one robot--obstacle pair, shared over obstacles and combined by a smooth minimum
     & One robust stationary grid-based HJ solution per robot--obstacle class
     & One batched pass over the obstacle list and one QP with a single composite constraint per control step \\
    \addlinespace[5pt]
    \mbox{VertexCBF}\newline(\cref{ch:vertex-cbf})
     & CBF-QP safety filter for a fixed known constraint function
     & Constraint function minus a nonnegative residual, represented by a single fixed network
     & Finite-horizon search over input vertices at sampled states, combined with a physics-informed loss
     & Value and gradient at the current state and one QP per control step, composed over obstacles as in CN-CBF when deployed \\
    \bottomrule
  \end{tabular}
\end{table}

\subsection{Controller Interfaces}
\label{subsec:conclusion-interfaces}

The MPC terminal constraint restricts where the planned trajectory may end, which allows the planner to consider task progress, actuation limits, and safety beyond the horizon in one optimization. The constraint is enforced only once per planning cycle and only on the \emph{predicted} final state, so its protection depends on the MPC problem remaining feasible from cycle to cycle and on the accuracy of the predicted trajectory and geometry. A CBF safety filter instead checks the applied input at the \emph{measured} state at every control step. Because a QP with one constraint is considerably less expensive to solve than a nonlinear program (NLP) over a horizon, the filter runs at a much higher rate than the planner, and it can be combined with any nominal controller. Its advantage is that a valid barrier and a feasible filtered control directly imply invariance of the safe set, and its disadvantage is that it has no look-ahead and relies on the nominal controller for task progress.

Two studies compare the interfaces directly. In the static-environment study of \cref{subsec:orn-cbf-sim-gr}, NTC-MPC, ORN-CBF, and the direct-regression variant ON-CBF were evaluated in the same experiment, and both filters achieved higher success rates than NTC-MPC at the three shorter horizons and similar rates at the longest one. The comparison does not isolate the interface completely, because ORN-CBF adds the residual parameterization and operates downstream of a separate planner. However, the ON-CBF method differs from NTC-MPC mainly in the interface, and it also outperformed NTC-MPC at the three shorter horizons. In the crowd study of \cref{subsec:cn-cbf-sim-gr}, CN-CBF achieved higher success rates than RNTC-MPC at every density, although the two methods also differ in their scene representation. Both studies point in the same direction, but neither can attribute the difference to the interface alone. The quadrotor timeouts of \cref{subsec:orn-cbf-sim-qq} illustrate the corresponding disadvantage of the filter, because a collision-free closed loop that makes no progress can arise when the nominal controller performs poorly, while the MPC interface resolves avoidance and task progress within one optimization, as the small lateral deviations of RNTC-MPC among pedestrians show. Neither interface is superior in every respect.

\subsection{Safety Representations and Training}
\label{subsec:conclusion-representations}

\paragraph{Direct and residual learning.}
Direct regression, as used in NTC-MPC, trains the network to reproduce the HJ value function itself, so a sufficiently large approximation error allows the model to classify a state inside an observed obstacle as safe. The residual parameterization introduced in \cref{ch:rntc-mpc} and reused by the three filtering methods exploits the HJ ordering \cref{eq:prelim-hj-value-constraint-ordering} and makes failure-set exclusion a property of the architecture rather than an outcome of training. The ablations show that the structure also simplifies learning, because the network represents only the dynamics-induced correction rather than the scene geometry. In the controlled comparison of \cref{subsec:rntc-training-eval}, the residual models reached the validation safe-set intersection over union (IoU) of direct regression roughly five times faster and ended higher for every tested model size, and the closed-loop ablations of \cref{ch:rntc-mpc,ch:orn-cbf} are consistent with this advantage. \Cref{ch:vertex-cbf} adds a second mechanism. Trained without the residual structure, the purely physics-informed objective collapsed to an empty predicted safe set on 14 of the 15 evaluated systems (\cref{subsec:vertex-cbf-ablations}). The reason is that any constant below the constraint function satisfies the stationary variational inequality, and a network without the residual structure can represent such a constant directly. The parameterization does not remove these solutions from the equation, but it makes them harder for gradient descent to reach. The property has two limits. It is one-sided, because \(\hat{V}\leq c\) does not imply \(\hat{V}\leq V_\infty\), so a state outside the obstacles but inside the true backward reachable tube can still be predicted safe if the residual is too small there. It also holds only relative to the geometry supplied to the representation, so it protects against residual-regression errors but not against perception or prediction errors.

\paragraph{Conditioning and composition.}
Observation conditioning, used by the first three methods, encodes the entire local scene in one high-dimensional conditioning input, which a hypernetwork translates into a scene-specific safety function. Its advantage is its generality, since it handles obstacles of any shape that the sensing resolution and the training data cover. Its cost is that the training data must cover the expected scenes and that every training scene needs its own reachability solution, which for RNTC-MPC is a time-varying one. Composition, used by CN-CBF, replaces the scene by a list of robot--obstacle pairs in relative coordinates and rebuilds a scene-level barrier at runtime. Its offline stage consists of one stationary reachability solution and one training run per robot--obstacle class, independent of the scene distribution, which reduced the offline time from hours to minutes relative to RNTC-MPC (\cref{subsec:cn-cbf-training-eval}). Its limitation is structural. Each obstacle may be avoidable on its own while the maneuvers needed to avoid different obstacles conflict, so the intersection of the pairwise safe sets can be larger than the true joint safe set. One plausible explanation for the higher success rates of CN-CBF in the crowd comparison is that the more accurate pairwise approximation outweighed the composition error. VertexCBF needs no conditioning, because its constraint function is fixed, which allows its experiments to isolate the effect of the supervision. It is not by itself a method for unknown environments, but for control-affine systems with polytopic input constraints its supervision can be combined with either representation (\cref{subsec:conclusion-guidance}).

\paragraph{Training objectives.}
The predicted safe set is the nonnegative superlevel set of the learned function, so its accuracy matters most near the zero level set, where a displaced boundary either admits unsafe states or restricts the controller unnecessarily. The radially weighted mean squared error (RWMSE) of \cref{subsec:ntc-training} assigns a larger weight to the squared error where the target is close to zero, and it improved the safe-set IoU and reduced the safety-critical misclassifications. The combined MSE--exponential (CME) loss of \cref{subsec:rntc-training} penalizes sign disagreement wherever it occurs, and its pointwise optimum preserves the zero level set of the target. In the ablation of \cref{subsec:rntc-training-eval}, the network converged faster and reached a higher final IoU with CME than with plain MSE. The objective of \cref{ch:vertex-cbf} differs, because its physics-informed term penalizes violations of the stationary variational inequality at unlabeled states. The labels discourage the trivial solutions of this inequality, and the inequality propagates the label information to the unlabeled states, which allows sparse supervision. In the evaluated systems the labels were necessary (\cref{subsec:vertex-cbf-ablations}).

\subsection{Supervision and Computational Requirements}
\label{subsec:conclusion-supervision}

The first four methods use grid-based HJ targets, which provide a numerical reference for learning under the chosen model. Storage and computation of these targets grow exponentially with the state dimension, which restricts grid-based solvers to roughly five or six state variables in practice, while the evaluated models have three to five. For control-affine dynamics with polytopic input sets, VertexCBF replaces the solver by labels obtained from a search over input sequences restricted to the vertices of the input polytope, together with a physics-informed loss. For a fixed beam width, the number of dynamics evaluations is then linear in the number of supervision states, the horizon length, the beam width, and the vertex count, and the supervision states can be sampled sparsely instead of covering a grid. The state dimension still affects the cost of every dynamics evaluation and the number of samples needed to cover the state space. Moreover, the number of vertices of a box-shaped input set grows exponentially with the number of inputs, so the exponential dependence is moved to the input dimension rather than removed. The two sources also have different error characteristics. The errors of the numerical target come from the grid resolution, the finite domain, and the convergence tolerance. The errors of the generated target come from the time discretization, the horizon, and the search width (\cref{eq:vertex-cbf-error-decomposition}), and their size cannot be measured without a reference solution. For this reason, \cref{ch:vertex-cbf} assesses its barriers through the closed-loop protocol of \cref{subsec:vertex-cbf-validation}.

Offline, the cost is determined by the number of reachability solutions that a method needs and by the cost of each solution, as listed in \cref{tab:method-comparison}, and the time-varying solutions of RNTC-MPC are the most expensive per target. For ORN-CBF, positional cropping reduced the training time by roughly an order of magnitude, because the network is trained only on the central part of the sensing window that the safety filter queries. Online, all methods query a compact network whose size is independent of the scene complexity, and the obstacle count enters only through the batched pairwise pass of CN-CBF. For the scene-conditioned methods, the perception, the SDF construction, the prediction, and the hypernetwork pass still have to run at the observation rate before the compact network can be queried. The model requirements also differ. The MPC methods need a dynamics model that can be discretized for the planner and used for the reachability targets in a consistent way. RNTC-MPC additionally needs an obstacle prediction that follows the same motion model as the training targets, of which the evaluated constant-velocity prediction is one instance. The filters need control-affine dynamics for the QP formulation. CN-CBF also needs relative dynamics that can be written in the relative state alone together with a bounded obstacle input model, and VertexCBF additionally needs a polytopic input set and, in the formulation developed here, no disturbance.

\subsection{Method Selection}
\label{subsec:conclusion-guidance}

The comparison yields concrete guidance for assembling a navigation system. The supervision should be chosen first, because it determines which representations are available. For small state spaces, a grid-based solver provides both the targets and a numerical reference. With five or six states, the grid becomes very large, so a solution is feasible only for a small state domain or a coarse grid and can take a long time to compute. When grid-based computation is impractical, the grid-free supervision of \cref{ch:vertex-cbf} provides an alternative.

However, it is not a general replacement for grid-based solvers. It requires control-affine dynamics with polytopic input constraints, which the MPC methods of Part~I do not assume in general. Its cost also grows with the number of reachability solutions that a method requires. NTC-MPC, RNTC-MPC, and ORN-CBF require one solution for every training observation, and generating grid-free labels for tens of thousands of observations is time-intensive for higher-dimensional models. CN-CBF requires only one solution per obstacle class, which makes grid-free supervision most practical for compositional methods, although the search must first be extended to bounded obstacle inputs (\cref{sec:conclusion-outlook}).

If collisions are to be avoided primarily with static geometry discovered at runtime, the observation-conditioned methods apply. ORN-CBF suits systems in which a safety filter can run at the control rate alongside an independent nominal controller, and NTC-MPC suits settings in which the reachability information must be integrated into an MPC planner.

If collisions are to be avoided primarily with moving agents, the choice depends on how well a simple model describes their motion. If the motion of each agent relative to the robot can be modeled by relative dynamics with bounded agent inputs, as for a mobile robot in open space or a vehicle on a highway, CN-CBF is the preferred method. It is simpler than RNTC-MPC and requires only one reachability solution per obstacle class offline and a single batched network pass over the observed obstacles online. It also achieved the highest success rates in the tested dense scenes. In most real environments, CN-CBF also requires an additional barrier for static geometry such as road boundaries, because it represents only obstacles of the modeled classes. ORN-CBF could provide this barrier within the same aggregation (\cref{sec:conclusion-outlook}). For more complex motion, such as that of road users at urban intersections, bounds on the agent inputs are either overly conservative or potentially unsafe, and the composite barrier may differ considerably from the true joint value function. In such settings, the approach of RNTC-MPC is more suitable, because it learns a time-varying HJ value function for the predicted geometry of all obstacles jointly. This also applies to complex but predictable motion, such as several obstacles that move together or machine parts that follow a known cycle. \Cref{ch:rntc-mpc} uses a constant-velocity model for simplicity, but any motion model or simulator that generates realistic obstacle trajectories can be used, provided that it supplies both the online prediction and the time-varying failure sets for the training targets.

\section{Safety Properties and Assumptions}
\label{sec:conclusion-guarantees}

The safety-relevant properties of the five methods have different origins. Some follow from existing HJ theory, some are derived in this dissertation and hold by construction of the representations, and others are demonstrated experimentally under stated model and sensing assumptions. Separating them makes clear what the dissertation contributes and where existing verification methods could extend it.

\paragraph{Properties inherited from HJ theory.}
All five methods build on existing HJ reachability theory \cite{mitchell_time-dependent_2005,bansal_hamilton-jacobi_2017}. Under the model, constraint, and horizon assumptions of \cref{sec:prelim-hj-reachability}, the nonnegative superlevel set of the exact value function is the maximal safe set. The grid-based and search-based targets approximate this value function, so this property carries over to the learned models only up to their approximation error.

\paragraph{Properties derived in this dissertation.}
These properties result from the design of the representations and training objectives developed here. The residual parameterization excludes the represented failure set from the predicted safe set for observed, predicted, and relative geometry (\cref{eq:rntc-failure-set-exclusion,eq:orn-cbf-exclusion,eq:cn-cbf-pairwise-exclusion}), and the log-sum-exp composition preserves this property while its gap to the exact minimum is bounded by \cref{eq:cn-cbf-softmin-bounds}. The pointwise optimum of the CME loss preserves the sign of the target \cref{eq:rntc-cme-optimum}, so the loss is consistent with the goal of classifying states correctly as safe or unsafe. For VertexCBF, restricting the Hamiltonian to the vertices of a polytopic input set is exact for control-affine dynamics \cref{eq:vertex-cbf-hamiltonian-vertices}, and the Euler discretization of the resulting problem is first-order consistent \cref{eq:vertex-cbf-consistency}. The finite search that produces the labels remains an approximation, whose three error terms \cref{eq:vertex-cbf-error-decomposition} are controlled by the step size, the horizon, and the beam width. At observation updates, the sensing inequalities \cref{eq:orn-cbf-sensing-adequacy,eq:cn-cbf-sensing-adequacy} guarantee sufficient separation from newly revealed obstacles for the exact targets under boundary-only appearance.

\paragraph{Model and sensing assumptions.}
The methods share a common set of assumptions. The dynamics model used for planning, filtering, and supervision is assumed accurate and the required state estimates are assumed available, so robot-model disturbances and estimation errors are not modeled explicitly, although inflation margins provided practical buffers in the hardware runs. Perception is assumed to deliver sufficiently accurate occupancy or obstacle-state estimates within the sensing range, with new information appearing only at the boundary of the sensing window, so an occlusion that reveals an obstacle directly next to the robot violates this assumption. The dynamic methods rely on their obstacle-motion models, the constant-velocity prediction of RNTC-MPC and the bounded-input class of CN-CBF, and the observed failures are consistent with violations of these models. The learned components assume that the deployment conditions remain within the training domains or that the model generalizes beyond them, which the out-of-domain experiments of \cref{ch:orn-cbf} test with favorable results. \Cref{ch:vertex-cbf} uses control-affine dynamics and a convex polytope input set, matching the CBF-QP formulation, and models no disturbance.

\paragraph{Conditions for closed-loop safety and experimental evidence.}
Classical MPC and CBF theory provides invariance of the predicted safe set under the conditions stated in \cref{subsec:ntc-ocp,subsec:rntc-ocp,subsec:orn-cbf-filter,subsec:cn-cbf-filter,subsec:vertex-cbf-filter}, which are recursive feasibility of the MPC problem and a valid barrier with a feasible filtered control of CBF-QP. For the learned fields, the dissertation supports these conditions in two ways. The residual parameterization excludes one important class of learning errors by construction, and the boundary-aware losses concentrate the accuracy at the safe-set boundary. On this basis, the reported studies demonstrated collision avoidance in randomized simulations and on hardware, with the complete pipeline running onboard the delivery robot in all five studies. What separates this evidence from a formal closed-loop certificate is a verification step common to all learned safety functions. The predicted safe set would have to be certified as a subset of the exact safe set. Failure-set exclusion already secures one direction of this inclusion, while the other direction requires additional techniques, such as one-sided training objectives or reachability-based patching, which have not yet been combined with these representations. In addition, the barrier condition and the feasibility of the deployed QP would have to be verified on the learned field, for which robust CBF theory and neural-network verification offer suitable methods. The methods state their assumptions explicitly and provide the structural properties on which such a verification can build.

\section{Outlook}
\label{sec:conclusion-outlook}

\paragraph{Broader implications.}
Beyond the individual methods, the dissertation shows how learning and control theory can be combined in a safety-critical system. Reachability analysis defines the safe set, optimization-based control enforces it, and learning makes the offline reachability solutions available in real time as differentiable functions. The neural networks complement the control-theoretic framework rather than replace it. The same problem arises in other applications in which an autonomous system operates in unstructured space and the safe set depends on high-dimensional information available only at runtime. Examples include manipulation near people, legged locomotion over perceived terrain, and automated driving. These applications require their own models, constraint functions, and perception interfaces, but the offline--online pipeline, the hypernetwork factorization, the residual parameterization, and the boundary-aware losses should transfer whenever the safety condition can be determined offline. For applications with control-affine dynamics and polytopic input constraints, the grid-free supervision of \cref{ch:vertex-cbf} is relevant, because their models often have more states than grid-based solvers can handle.

\paragraph{Future research.}
Three directions follow most directly from the results. The first is certification. The structural properties established here narrow the certification of a learned safety function to one question, since the predicted safe set already excludes the failure set and what remains to be bounded is how far it may extend into the unsafe region. One-sided training objectives, error bounds subtracted from the learned value \cite{taheri_statistical_2021,lin_one_2026}, and reachability-based patching of the trained network \cite{tonkens_patching_2024} are candidate approaches. The closed-loop rollouts of the validation protocol also provide the data from which scenario optimization or conformal prediction compute margins with finite-sample coverage \cite{lin_verification_2024,feng_bridging_2025}. The same methods could be applied to perception and prediction uncertainty, which can be propagated into the safety function by inflating the failure set with a calibrated estimation error, by replacing the nominal obstacle prediction with a distribution over futures, or by calibrating the disturbance set of CN-CBF from motion data \cite{fridovich-keil_confidence-aware_2020}.

The second direction is extending the methods to a wider range of systems and environments. On the supervision side, the grid-free labels of \cref{ch:vertex-cbf} could be extended to the two-player targets of \cref{ch:cn-cbf}, because control and disturbance both attain their extrema at vertices. Since CN-CBF requires only one set of labels per obstacle class, this extension would make grid-free supervision practical for higher-dimensional robot--obstacle models and would also allow grid-free and grid-based targets to be compared within the same method. For systems with many inputs, the number of vertices to be searched could be reduced by selecting a subset of vertices with the current value estimate. On the representation side, the ORN-CBF barrier could enter the CN-CBF aggregation as an additional argument, which would yield one barrier for mixed static and dynamic environments while preserving failure-set exclusion. A trainable aggregation could further correct the composition where pairwise reasoning is optimistic, and graph-structured barrier certificates suggest suitable architectures for it \cite{zhang_gcbf_2025-2}.

The third direction is integration into complete navigation systems. The methods implement the local planner and safety filter blocks of \cref{fig:nav-architecture}, and some of the failure modes observed in the experiments originate in two other blocks, global planning and motion prediction. Failures in which the robot entered a dead end that was not yet visible, or in which an occlusion revealed an obstacle at close range, originate in global planning. Avoiding them requires a global planner that uses the partial map to choose between locally equivalent directions and that treats unobserved regions as potentially occupied. The learned safety functions can support such a planner, because they can be evaluated at any candidate waypoint in real time and indicate whether the robot can remain collision-free from there. Collisions with pedestrians whose motion deviated from the constant-velocity model are failures of motion prediction. Since RNTC-MPC accepts any predicted obstacle geometry as input, a learned predictor \cite{korbmacher_review_2022} can replace its constant-velocity model, and interaction-aware models \cite{trautman_unfreezing_2010} can account for how pedestrians react to the robot.

\paragraph{Concluding remarks.}
HJ reachability analysis is one of the most comprehensive model-based frameworks for characterizing the states from which a robot can remain safe. This dissertation investigated how it can be transformed from an offline tool into an online safety representation for robot navigation in unknown and dynamic environments. The five methods show that this transformation is practical in the evaluated settings when learning and structure are combined. Networks trained offline on reachability solutions replace the expensive online computation. The residual parameterization ensures that approximation errors of these networks cannot place occupied states in the predicted safe set. The final method shows that, for control-affine systems with polytopic input constraints, the reachability targets themselves can be generated from the structure of the problem rather than solved on a grid. The methods state explicitly which properties hold by construction, which are inherited from HJ theory, and which are demonstrated experimentally, and this separation defines what a future certification has to establish. As mobile robots are deployed more widely in spaces shared with people, safety representations that are fast enough to react, structured enough to be analyzed, and explicit about their limits will be an important component of trustworthy autonomy.

\appendix
\chapter{Publications and Author Contributions}
\label{app:publications}

\Cref{ch:ntc-mpc,ch:rntc-mpc,ch:orn-cbf,ch:cn-cbf,ch:vertex-cbf} are each based on one of the five peer-reviewed publications listed below. Parts of the text, figures, and tables of these chapters are adapted from the corresponding publication and were revised and extended for this dissertation. The author of this dissertation is the first author of all five publications, and the contributions of all authors are stated below each publication.

\paragraph{\Cref{ch:ntc-mpc}, NTC-MPC.}
\AtNextCite{\defcounter{maxnames}{99}}\fullcite{derajic_learning_2025}

\noindent\emph{Author contributions.} Bojan Derajić developed the concept and the method, implemented the method, designed and conducted the experiments, analyzed the results, and wrote the manuscript. Mohamed-Khalil Bouzidi contributed to the discussions of the method and the experiments and revised the manuscript. Sebastian Bernhard and Wolfgang Hönig supervised the work and revised the manuscript.

\paragraph{\Cref{ch:rntc-mpc}, RNTC-MPC.}
\AtNextCite{\defcounter{maxnames}{99}}\fullcite{derajic_residual_2025}

\noindent\emph{Author contributions.} Bojan Derajić developed the concept and the method, implemented the method, designed and conducted the experiments, analyzed the results, and wrote the manuscript. Mohamed-Khalil Bouzidi contributed to the discussions of the method and the experiments and revised the manuscript. Sebastian Bernhard and Wolfgang Hönig supervised the work and revised the manuscript.

\paragraph{\Cref{ch:orn-cbf}, ORN-CBF.}
\AtNextCite{\defcounter{maxnames}{99}}\fullcite{derajic_orn-cbf_2026}

\noindent\emph{Author contributions.} Bojan Derajić developed the concept and the method, implemented the method, designed and conducted the experiments, analyzed the results, and wrote the manuscript. Sebastian Bernhard and Wolfgang Hönig supervised the work and revised the manuscript.

\paragraph{\Cref{ch:cn-cbf}, CN-CBF.}
\AtNextCite{\defcounter{maxnames}{99}}\fullcite{derajic_cn-cbf_2026}

\noindent\emph{Author contributions.} Bojan Derajić developed the concept and the method, implemented the method, designed and conducted the experiments, analyzed the results, and wrote the manuscript. Sebastian Bernhard and Wolfgang Hönig supervised the work and revised the manuscript.

\paragraph{\Cref{ch:vertex-cbf}, VertexCBF.}
\AtNextCite{\defcounter{maxnames}{99}}\fullcite{derajic_vertexcbf_2026}

\noindent\emph{Author contributions.} Bojan Derajić developed the concept and the method, implemented the method, designed and conducted the experiments, analyzed the results, and wrote the manuscript. Sebastian Bernhard and Wolfgang Hönig supervised the work and revised the manuscript.

\chapter{Supporting Details for VertexCBF}
\label{app:vertex-cbf-config}

This appendix collects the derivation of the consistency result of \cref{subsec:vertex-cbf-hamiltonian} and the experimental configuration of \cref{ch:vertex-cbf}. The configuration part records the domains, constraints, dynamics, search settings, network architectures, training schedules, and validation settings for the 15 VertexCBF systems, followed by the hardware configuration. Values follow the underlying publication \cite{derajic_vertexcbf_2026}, and the reported offline times were measured on one NVIDIA RTX~3090 GPU.\footnote{The implementation is available at \url{https://github.com/bojan-derajic/vertexcbf}.}

Three conventions apply to the configuration tables of \cref{sec:app-vertex-cbf-systems,sec:app-vertex-cbf-data,sec:app-vertex-cbf-training,sec:app-vertex-cbf-validation}. The systems appear in the order of \cref{tab:vertex-cbf-results}, that is by increasing state dimension. Angles are given in radians, lengths in meters, and times in seconds unless stated otherwise. Every configuration listed here is shared by the three supervision settings of \cref{subsec:vertex-cbf-setup}, since those settings differ only in the data term of \cref{eq:vertex-cbf-total-loss}. The label generator is not run by the \emph{no data} setting and is replaced by MPPI sampling in the \emph{full-control data} setting, at the same horizon length, step size, and nominal sample count. The loss weight \(\lambda\) listed in \cref{tab:app-vertex-cbf-training} is the value used by the two supervised settings, while the \emph{no data} setting has no data term and corresponds to \(\lambda=1\).

\section{Consistency of the Discretization}
\label{sec:app-vertex-cbf-consistency}

This section derives the truncation bound \cref{eq:vertex-cbf-truncation-error} and the consistency statement \cref{eq:vertex-cbf-consistency}. Let \(\mathcal{Y}\subseteq\mathcal{X}\) be a compact convex region that contains \(x\) and \(F_\Delta(x,u)\) for every \(u\in\mathcal{V}\) and every sufficiently small \(\Delta t\), and let \(\psi\in C^{1,1}(\mathcal{Y})\) have a gradient with Lipschitz constant \(L_{\nabla\psi}\) on \(\mathcal{Y}\). The function \(\psi\) serves as a smooth local substitute for the value function, which is in general nonsmooth. For any vertex \(u\in\mathcal{V}\), Taylor's theorem applied to \(\psi\) along the Euler step \cref{eq:vertex-cbf-euler-map} gives
\begin{equation}
  \psi\bigl(F_\Delta(x,u)\bigr)
  =
  \psi(x)
  +
  \Delta t\,
  \nabla_x\psi(x)^{\top}
  \bigl(f_0(x)+g(x)u\bigr)
  +
  R_{\Delta t}(x,u),
  \label{eq:vertex-cbf-taylor}
\end{equation}
with a remainder bounded by
\begin{equation}
  \bigl\lvert R_{\Delta t}(x,u)\bigr\rvert
  \leq
  \tfrac{1}{2}
  L_{\nabla\psi}
  \bigl\lVert f_0(x)+g(x)u\bigr\rVert^{2}
  \Delta t^{2},
  \label{eq:vertex-cbf-taylor-remainder}
\end{equation}
because the increment of the Euler step has norm \(\Delta t\,\lVert f_0(x)+g(x)u\rVert\) and the segment between \(x\) and \(F_\Delta(x,u)\) lies in \(\mathcal{Y}\) by convexity. Taking the maximum over the vertices on both sides of \cref{eq:vertex-cbf-taylor}, dividing by \(\Delta t\), and using the vertex representation \cref{eq:vertex-cbf-hamiltonian-vertices} of the Hamiltonian yields
\begin{equation}
  \frac{
    \max_{u\in\mathcal{V}}\psi\bigl(F_\Delta(x,u)\bigr)-\psi(x)
  }{\Delta t}
  =
  H\bigl(x,\nabla_x\psi(x)\bigr)
  +
  O(\Delta t),
  \label{eq:vertex-cbf-normalized-consistency}
\end{equation}
where the maximum of a sum is bounded by the maximum of the linear term plus the maximum of the remainder, so the difference between the left-hand side and the Hamiltonian is bounded by the maximum over the vertices of \(\lvert R_{\Delta t}(x,u)\rvert/\Delta t\). Inserting \cref{eq:vertex-cbf-taylor-remainder} gives the explicit bound \cref{eq:vertex-cbf-truncation-error}. Substituting \cref{eq:vertex-cbf-normalized-consistency} for the second argument of the discrete variational inequality \cref{eq:vertex-cbf-discrete-vi} and using that the minimum of two arguments is Lipschitz continuous with constant one in each argument gives \cref{eq:vertex-cbf-consistency}. The one-step Bellman expansion \cref{eq:vertex-cbf-taylor} has error \(O(\Delta t^{2})\), and its time-normalized form has error \(O(\Delta t)\), which is the first-order consistency stated in \cref{subsec:vertex-cbf-hamiltonian}.

The calculation concerns the unmodified Euler map. On the systems where the implementation clamps a state component to its box limits after each step or renormalizes the quaternion (\cref{sec:vertex-cbf-problem-statement,sec:app-vertex-cbf-systems}), the transition differs from \cref{eq:vertex-cbf-euler-map} at those limits, and the statement applies to states whose Euler successors remain inside the box.

\section{Systems, State Domains, and Constraints}
\label{sec:app-vertex-cbf-systems}

\Cref{tab:app-vertex-cbf-domains} lists the state box \(\mathcal{X}\) and the input polytope \(\mathcal{U}\) of each system. Every input set is a box, so the vertex count is \(M=2^{n_u}\) throughout, ranging from \(M=2\) for the single-input systems to \(M=64\) for the underwater vehicle. Since \(\mathcal{X}\) is the domain over which the predicted-safe volume of \cref{eq:vertex-cbf-eta-eff} is measured, the reported \(\eta_{\mathrm{eff}}\) values are comparable across the three supervision settings of one system but not across systems. For the quadrotor, the four quaternion components of \(\mathcal{X}\) are not sampled or integrated as an independent box. The non-quaternion coordinates are sampled uniformly from their bounds and the orientation is sampled uniformly over unit quaternions, and the trainer, beam search, and validator all renormalize the quaternion to the unit sphere after every Euler step. Its effective safe volume therefore denotes probability under this sampling distribution rather than a fraction of 13-dimensional box volume, and the four quaternion components remain a valid state representation without being four independent physical coordinates.

\begin{table}[tb]
  \caption[State and input domains of the 15 evaluated systems]{State box \(\mathcal{X}\) and input box \(\mathcal{U}\) for the 15 evaluated systems, together with the model parameters. All input sets are boxes, so the vertex count is \(M=2^{n_u}\).}
  \label{tab:app-vertex-cbf-domains}
  \centering
  \scriptsize
  \setlength{\tabcolsep}{3pt}
  \begin{tabular*}{\textwidth}{@{\extracolsep{\fill}}l c R{0.40\textwidth} R{0.25\textwidth}@{}}
    \toprule
    System & \((n_x,n_u)\) & State box \(\mathcal{X}\) & Input box \(\mathcal{U}\), parameters \\
    \midrule
    Inverted Pendulum & (2, 1) & \(\theta\in[-0.5,0.5]\), \(\omega\in[-1.5,1.5]\) & \(\tau\in[-3.5,3.5]\); \(m{=}2.0\), \(l{=}1.0\), \(g{=}9.81\) \\
    \addlinespace[2pt]
    1D Double Integrator & (2, 1) & \(p\in[-1.5,1.5]\), \(v\in[-1.5,1.5]\) & \(a\in[-0.5,0.5]\) \\
    \addlinespace[2pt]
    2D Vertical Drone & (2, 1) & \(z\in[-0.5,3.5]\), \(v_z\in[-4,4]\) & \(a\in[-1,1]\); \(K{=}12.0\), \(g{=}9.81\) \\
    \addlinespace[2pt]
    Dubins Car & (3, 1) & \(p_x,p_y\in[-4,4]\), \(\theta\in[-3.14,3.14]\) & \(\omega\in[-0.5,0.5]\); \(v{=}1.0\) \\
    \addlinespace[2pt]
    2D Double Integrator & (4, 2) & \(p_x,p_y\in[-4,4]\), \(v_x,v_y\in[-2,2]\) & \(a_x,a_y\in[-1,1]\) \\
    \addlinespace[2pt]
    Kinematic Bicycle & (4, 2) & \(p_x,p_y\in[-25,25]\), \(\psi\in[-3.14,3.14]\), \(v\in[0,20]\) & \(\tan\delta\in[-0.3,0.3]\), \(a\in[-3,3]\); \(L{=}2.8\) \\
    \addlinespace[2pt]
    Cart-Pole & (4, 1) & \(p\in[-1.5,1.5]\), \(\theta\in[-0.35,0.35]\), \(\dot{p},\dot{\theta}\in[-1,1]\) & \(F\in[-5,5]\); \(m_c{=}2.0\), \(m_p{=}0.5\), \(l{=}0.5\), \(g{=}9.81\) \\
    \addlinespace[2pt]
    Dynamic Unicycle & (5, 2) & \(p_x,p_y\in[-4,4]\), \(\theta\in[-3.14,3.14]\), \(v\in[-2,2]\), \(\omega\in[-1,1]\) & \(a,\alpha\in[-1,1]\) \\
    \addlinespace[2pt]
    Relative Unicycle & (5, 2) & \(p_x,p_y\in[-4,4]\), \(\psi\in[-3.14,3.14]\), \(v_{\mathrm{r}},v_{\mathrm{p}}\in[0,1.5]\) & \(a\in[-1,1]\), \(\omega\in[-0.6,0.6]\) \\
    \addlinespace[2pt]
    3D Double Integrator & (6, 3) & \(p_x,p_y,p_z\in[-3,3]\), \(v_x,v_y,v_z\in[-2,2]\) & \(a_x,a_y,a_z\in[-1,1]\) \\
    \addlinespace[2pt]
    3-DOF Manipulator & (6, 3) & \(q_1,q_2,q_3\in[-1,1]\), \(\dot{q}_1\in[-3,3]\), \(\dot{q}_2,\dot{q}_3\in[-6,6]\) & \(\tau_1,\tau_3\in[-10,10]\), \(\tau_2\in[-20,20]\); \(m_1{=}m_2{=}2.0\), \(l_1{=}l_2{=}0.5\), \(l_{c1}{=}l_{c2}{=}0.25\), \(I_1{=}I_2{=}I_3{=}0.0417\), \(g{=}9.81\) \\
    \addlinespace[2pt]
    Landing Rocket & (7, 2) & \(p_x\in[-6,6]\), \(p_z\in[0,10]\), \(v_x\in[-3,3]\), \(v_z\in[-6,3]\), \(\theta\in[-1,1]\), \(\omega\in[-3,3]\), \(m\in[0.6,1.0]\) & \(T\in[0,15]\), \(\tau\in[-1,1]\); \(I{=}0.25\), \(\alpha{=}0.01\), \(g{=}9.81\) \\
    \addlinespace[2pt]
    Quadruped Trunk & (9, 4) & \(p_z\in[0.05,0.55]\), \(\phi,\theta\in[-0.6,0.6]\), \(v_x,v_y\in[-1.5,1.5]\), \(v_z\in[-2,2]\), \(p,q\in[-5,5]\), \(r\in[-2,2]\) & \(F_1,\ldots,F_4\in[0,60]\); \(m{=}12.0\), \(I_x{=}0.10\), \(I_y{=}0.20\), \(I_z{=}0.25\), \(a{=}0.20\), \(b{=}0.15\), \(g{=}9.81\) \\
    \addlinespace[2pt]
    6-DOF Underwater Veh. & (12, 6) & \(p_x,p_y,p_z\in[-3,3]\), \(\phi,\psi\in[-\pi,\pi]\), \(\theta\in[-1,1]\), \(u_b,v_b,w_b\in[-2,2]\), \(p,q,r\in[-1.5,1.5]\) & \(F_x\in[-80,80]\), \(F_y,F_z\in[-10,10]\), \(\tau_x,\tau_y,\tau_z\in[-20,20]\); \(m_t{=}30.0\), \(I_x{=}0.5\), \(I_y{=}I_z{=}2.0\), \(h_{\mathrm{cb}}{=}0.02\), \(g{=}9.81\) \\
    \addlinespace[2pt]
    Quadrotor & (13, 4) & \(p\in[-3,3]^{3}\), \(q\in[-1,1]^{4}\), \(v\in[-5,5]^{3}\), \(\omega\in[-5,5]^{3}\) & \(F\in[0,20]\), \(\alpha_x,\alpha_y\in[-8,8]\), \(\alpha_z\in[-4,4]\); \(m{=}1.0\), \(C_T{=}1.0\), \(g{=}9.81\) \\
    \bottomrule
  \end{tabular*}
\end{table}

\Cref{tab:app-vertex-cbf-constraints} gives the constraint function \(c\) of each system and its parameters. The constraint functions take four forms. For seven systems, the constraint function is the Euclidean distance to a circular, spherical, or cylindrical obstacle. For three systems, it is the signed distance to an interval that bounds one state component. For four systems, it is a smooth minimum of several such terms,
\begin{equation}
  c(x)
  =
  -\frac{1}{\alpha}
  \log
  \sum_{i}
  e^{-\alpha\,d_i(x)},
  \label{eq:app-vertex-cbf-softmin}
\end{equation}
which under-approximates \(\min_i d_i(x)\) by at most \(\ln(n)/\alpha\) for \(n\) terms and is the same construction that \cref{subsec:cn-cbf-composition} uses for the composition over obstacles. For the landing rocket, the constraint function is a smooth funnel that couples the lateral position to the altitude and penalizes speed and tilt. Every constraint function is evaluated in closed form at runtime, which the residual parameterization \cref{eq:vertex-cbf-residual} requires.

\begin{table}[tb]
  \caption[Constraint functions and parameters of the 15 evaluated systems]{Constraint function \(c\) of each system and its parameters. The smooth minimum is \cref{eq:app-vertex-cbf-softmin}, and \(\alpha\) is its sharpness parameter.}
  \label{tab:app-vertex-cbf-constraints}
  \centering
  \scriptsize
  \setlength{\tabcolsep}{3pt}
  \begin{tabular*}{\textwidth}{@{\extracolsep{\fill}}l R{0.44\textwidth} R{0.30\textwidth}@{}}
    \toprule
    System & Constraint function \(c(x)\) & Parameters \\
    \midrule
    Inverted Pendulum & \(d-\lvert\theta\rvert\) (pole-angle bound) & \(d{=}0.3\) \\
    \addlinespace[2pt]
    1D Double Integrator & \(d-\lvert p-p_{\mathrm{c}}\rvert\) (position band) & \(p_{\mathrm{c}}{=}0.0\), \(d{=}1.0\) \\
    \addlinespace[2pt]
    2D Vertical Drone & \(d-\lvert z-z_{\mathrm{c}}\rvert\) (altitude band) & \(z_{\mathrm{c}}{=}1.5\), \(d{=}1.5\) \\
    \addlinespace[2pt]
    Dubins Car & \(\sqrt{p_x^{2}+p_y^{2}}-r\) (circular obstacle) & \(r{=}1.0\) \\
    \addlinespace[2pt]
    2D Double Integrator & \(\sqrt{p_x^{2}+p_y^{2}}-r\) & \(r{=}1.0\) \\
    \addlinespace[2pt]
    Kinematic Bicycle & \(\sqrt{p_x^{2}+p_y^{2}}-r\) & \(r{=}7.0\) \\
    \addlinespace[2pt]
    Cart-Pole & smooth minimum of \(p_{\max}-\lvert p\rvert\) and \(\theta_{\max}-\lvert\theta\rvert\) & \(p_{\max}{=}1.2\), \(\theta_{\max}{=}0.25\), \(\alpha{=}30.0\) \\
    \addlinespace[2pt]
    Dynamic Unicycle & \(\sqrt{p_x^{2}+p_y^{2}}-r\) & \(r{=}1.0\) \\
    \addlinespace[2pt]
    Relative Unicycle & \(\sqrt{p_x^{2}+p_y^{2}}-(r_{\mathrm{r}}+r_{\mathrm{o}})\) (distance between two disks in the body frame) & \(r_{\mathrm{r}}{=}0.5\), \(r_{\mathrm{o}}{=}0.5\) \\
    \addlinespace[2pt]
    3D Double Integrator & \(\sqrt{p_x^{2}+p_y^{2}+p_z^{2}}-r\) (spherical obstacle) & \(r{=}1.0\) \\
    \addlinespace[2pt]
    3-DOF Manipulator & smooth minimum over the two links of the capsule-to-sphere distance \(\operatorname{dist}(c_{\mathrm{obs}},\mathrm{link}_i(q))-r-r_{\ell}\) & \(c_{\mathrm{obs}}{=}(0.6,0.0,-0.2)\), \(r{=}0.2\), \(r_{\ell}{=}0.1\), \(\alpha{=}20.0\) \\
    \addlinespace[2pt]
    Landing Rocket & \((\rho+s\,p_z)-\sqrt{p_x^{2}+\varepsilon}-w_v(v_x^{2}+v_z^{2})-w_\theta\,\theta^{2}\) (landing funnel) & \(\rho{=}0.5\), \(s{=}0.5\), \(w_v{=}0.02\), \(w_\theta{=}0.1\), \(\varepsilon{=}10^{-4}\) \\
    \addlinespace[2pt]
    Quadruped Trunk & smooth minimum of the height band and the roll and pitch bounds & \(p_z\in[0.18,0.42]\), \(\phi_{\max}{=}\theta_{\max}{=}0.4\), \(\alpha{=}20.0\) \\
    \addlinespace[2pt]
    6-DOF Underwater Veh. & smooth minimum of the four half-space distances of a tunnel of half-width \(r_t\) in the \((p_y,p_z)\) plane & \(r_t{=}2.0\), \(\alpha{=}20.0\) \\
    \addlinespace[2pt]
    Quadrotor & \(\sqrt{p_x^{2}+p_y^{2}}-r\) (vertical cylinder) & \(r{=}0.67\) \\
    \bottomrule
  \end{tabular*}
\end{table}

\section{Benchmark Dynamics}
\label{sec:app-vertex-cbf-dynamics}

The following equations specify the continuous-time models of the 15 benchmark systems of \cite{derajic_vertexcbf_2026}. They use the state and input order of \cref{tab:app-vertex-cbf-domains}. The scalar \(g_0\) denotes the gravitational acceleration, listed as \(g\) in that table. All models are affine in the stated inputs, as \cref{eq:vertex-cbf-dynamics} requires. The label generator integrates every model with the forward-Euler map \cref{eq:vertex-cbf-euler-map}, and the validation rollouts integrate the same models with the smaller step sizes of \cref{tab:app-vertex-cbf-validation}. Periodic angles enter the residual network as the pair \((\cos\theta,\sin\theta)\), as stated in \cref{subsec:vertex-cbf-training}.

\paragraph{Inverted pendulum.}
For pole angle \(\theta\) measured from upright, angular velocity \(\omega\), and input torque \(\tau\),
\begin{equation}
  \dot\theta=\omega,
  \qquad
  \dot\omega=\frac{g_0}{l}\sin\theta+\frac{\tau}{ml^2}.
\end{equation}

\paragraph{Double integrators in one, two, and three dimensions.}
For \(p,v,a\in\mathbb R^d\), with \(d\in\{1,2,3\}\), the state is \(x=(p^\top,v^\top)^\top\), the input is \(a\), and
\begin{equation}
  \dot p=v,\qquad \dot v=a.
\end{equation}
The one-dimensional benchmark uses a position band. The two- and three-dimensional benchmarks use circular and spherical obstacles, respectively.

\paragraph{Vertical drone.}
This two-state model describes altitude \(z\) and vertical velocity \(v_z\), with a scalar actuator input \(a\),
\begin{equation}
  \dot z=v_z,\qquad \dot v_z=-g_0+Ka.
\end{equation}
The signed input interval in \cref{tab:app-vertex-cbf-domains} defines an idealized vertical-actuation model. It is not a model of individual nonnegative rotor thrusts.

\paragraph{Dubins car.}
For position \((p_x,p_y)\), heading \(\theta\), fixed speed \(v\), and turning-rate input \(\omega\),
\begin{equation}
  \dot p_x=v\cos\theta,\qquad
  \dot p_y=v\sin\theta,\qquad
  \dot\theta=\omega.
\end{equation}

\paragraph{Kinematic bicycle.}
The state is \(x=(p_x,p_y,\psi,v)^\top\). The input is \(u=(s_\delta,a)^\top\), where \(s_\delta=\tan\delta\) and \(\delta\) is the steering angle. For wheelbase \(L\),
\begin{equation}
  \dot p_x=v\cos\psi,\qquad
  \dot p_y=v\sin\psi,\qquad
  \dot\psi=\frac{v}{L}s_\delta,\qquad
  \dot v=a.
\end{equation}
The transformed steering input makes the model control affine.

\paragraph{Cart-pole.}
The state is \(x=(p,\theta,v,\omega)^\top\), where \(\theta\) is measured from upright. The input is the horizontal cart force \(F\). Define \(D(\theta)=m_c+m_p\sin^2\theta\). Then
\begin{align}
  \dot p&=v,\qquad \dot\theta=\omega,\\
  \dot v&=\frac{F+m_p\sin\theta\left(l\omega^2-g_0\cos\theta\right)}{D(\theta)},\\
  \dot\omega&=
  \frac{(m_c+m_p)g_0\sin\theta-m_p l\omega^2\cos\theta\sin\theta-F\cos\theta}
       {lD(\theta)}.
\end{align}

\paragraph{Dynamic unicycle.}
For \(x=(p_x,p_y,\theta,v,\omega)^\top\) and acceleration inputs \(u=(a,\alpha)^\top\),
\begin{equation}
  \dot p_x=v\cos\theta,\quad
  \dot p_y=v\sin\theta,\quad
  \dot\theta=\omega,\quad
  \dot v=a,\quad
  \dot\omega=\alpha.
\end{equation}

\paragraph{Relative unicycle.}
The state is \(x=(p_x,p_y,\psi,v_{\mathrm r},v_{\mathrm p})^\top\). The position is that of the obstacle in the robot frame, and \(\psi\) is the obstacle heading relative to the robot. The inputs are robot acceleration \(a\) and turning rate \(\omega\). A constant-speed, constant-heading obstacle gives
\begin{align}
  \dot p_x&=v_{\mathrm p}\cos\psi-v_{\mathrm r}+\omega p_y,
  &\dot p_y&=v_{\mathrm p}\sin\psi-\omega p_x,\\
  \dot\psi&=-\omega,
  &\dot v_{\mathrm r}&=a,\qquad \dot v_{\mathrm p}=0.
\end{align}
The obstacle speed is a state parameter, not an adversarial input. This differs from the game formulation in \cref{ch:cn-cbf}.

\paragraph{Three-degree-of-freedom manipulator.}
The arm has a yaw joint and two pitch joints. Its two moving links have lengths \(l_1,l_2\), point masses \(m_1,m_2\), and center-of-mass distances \(l_{c1},l_{c2}\). Let \(q=(q_1,q_2,q_3)^\top\), with state \(x=(q^\top,\dot q^\top)^\top\) and torque input \(\tau\). The link endpoints are
\begin{align}
  P_0&=0,\\
  P_1&=l_1(\cos q_1\cos q_2,\ \sin q_1\cos q_2,\ \sin q_2)^\top,\\
  P_2&=P_1+l_2(\cos q_1\cos(q_2+q_3),\
                   \sin q_1\cos(q_2+q_3),\ \sin(q_2+q_3))^\top.
\end{align}
These endpoints define the link segments used in the collision constraint. The joint dynamics are
\begin{equation}
  \ddot q=M(q)^{-1}\bigl(\tau-b(q,\dot q)-G(q)\bigr).
\end{equation}
Writing \(r_c=l_1\cos q_2+l_{c2}\cos(q_2+q_3)\), the inertia matrix is
\begin{equation}
  M(q)=
  \begin{bmatrix}
    M_{11}&0&0\\
    0&M_{22}&M_{23}\\
    0&M_{23}&M_{33}
  \end{bmatrix},
\end{equation}
with
\begin{align}
  M_{11}&=m_1l_{c1}^2\cos^2q_2+m_2r_c^2+I_1,\\
  M_{22}&=m_1l_{c1}^2+m_2(l_1^2+l_{c2}^2+2l_1l_{c2}\cos q_3)+I_2,\\
  M_{23}&=m_2(l_{c2}^2+l_1l_{c2}\cos q_3),\qquad
  M_{33}=m_2l_{c2}^2+I_3.
\end{align}
The Coriolis and centrifugal vector is specified by
\begin{equation}
  b_i(q,\dot q)=\frac12\sum_{j,k=1}^{3}
  \left(
    \frac{\partial M_{ij}}{\partial q_k}
    +\frac{\partial M_{ik}}{\partial q_j}
    -\frac{\partial M_{jk}}{\partial q_i}
  \right)\dot q_j\dot q_k.
\end{equation}
The gravity terms are
\begin{align}
  G_1&=0,\\
  G_2&=g_0\left[(m_1l_{c1}+m_2l_1)\cos q_2
                     +m_2l_{c2}\cos(q_2+q_3)\right],\\
  G_3&=m_2g_0l_{c2}\cos(q_2+q_3).
\end{align}
The block structure of \(M\) does not remove the velocity coupling in \(b\).

\paragraph{Landing rocket.}
The state is \(x=(p_x,p_z,v_x,v_z,\theta,\omega,m)^\top\). The inputs are thrust \(T\) and torque \(\tau\). For inertia \(I\) and mass-flow coefficient \(\alpha_m\),
\begin{align}
  \dot p_x&=v_x,& \dot p_z&=v_z,\\
  \dot v_x&=-\frac{T}{m}\sin\theta,&
  \dot v_z&=-g_0+\frac{T}{m}\cos\theta,\\
  \dot\theta&=\omega,&
  \dot\omega&=\frac{\tau}{I},\qquad \dot m=-\alpha_m T.
\end{align}
The coefficient \(\alpha_m\) is listed as \(\alpha\) in \cref{tab:app-vertex-cbf-domains}. The differential equation requires \(m>0\). The benchmark constraint describes a landing funnel, but the model has no touchdown event or fuel-exhaustion mode.

\paragraph{Quadruped trunk.}
This reduced model represents trunk motion under four body-normal forces \(F_1,\ldots,F_4\). It does not model leg motion, changes in contact, or a walking gait. The state is \(x=(p_z,\phi,\theta,v_x,v_y,v_z,p,q,r)^\top\), where \((p,q,r)\) are body angular rates. Let \(F_\Sigma=\sum_iF_i\). Then
\begin{align}
  \dot p_z&=v_z,\\
  \dot\phi&=p+\sin\phi\tan\theta\,q+\cos\phi\tan\theta\,r,\\
  \dot\theta&=\cos\phi\,q-\sin\phi\,r,\\
  \dot v_x&=\frac{F_\Sigma}{m}\sin\theta\cos\phi,\qquad
  \dot v_y=-\frac{F_\Sigma}{m}\sin\phi,\\
  \dot v_z&=-g_0+\frac{F_\Sigma}{m}\cos\theta\cos\phi,\\
  I_x\dot p&=-(I_z-I_y)qr+b(F_1-F_2+F_3-F_4),\\
  I_y\dot q&=-(I_x-I_z)pr+a(-F_1-F_2+F_3+F_4),\\
  I_z\dot r&=-(I_y-I_x)pq.
\end{align}
The translational equations use the zero-yaw rotation, while yaw rate is retained through gyroscopic coupling. This is a reduced benchmark assumption. Euler-angle kinematics require \(\cos\theta\ne0\).

\paragraph{Underwater vehicle.}
The state is \(x=(p_x,p_y,p_z,\phi,\theta,\psi,u_b,v_b,w_b,p,q,r)^\top\). Let \(v_b^{\mathrm{vec}}=(u_b,v_b,w_b)^\top\), \(\omega=(p,q,r)^\top\), and \(\nu=((v_b^{\mathrm{vec}})^\top,\omega^\top)^\top\). The inputs are body forces \(F=(F_x,F_y,F_z)^\top\) and moments \(\tau=(\tau_x,\tau_y,\tau_z)^\top\). In a world frame with upward \(z\),
\begin{align}
  \dot p^{\mathrm{vec}}&=R_z(\psi)R_y(\theta)R_x(\phi)v_b^{\mathrm{vec}},\\
  \begin{bmatrix}\dot\phi\\\dot\theta\\\dot\psi\end{bmatrix}
  &=
  \begin{bmatrix}
    1&\sin\phi\tan\theta&\cos\phi\tan\theta\\
    0&\cos\phi&-\sin\phi\\
    0&\sin\phi/\cos\theta&\cos\phi/\cos\theta
  \end{bmatrix}\omega.
\end{align}
Here \(R_x,R_y,R_z\) are right-handed rotations about their indicated axes. Define \(D_i(\nu_i)=(d_i^\ell+d_i^q|\nu_i|)\nu_i\), and let \(D_v,D_\omega\) collect its first and last three entries. For \(I=\operatorname{diag}(I_x,I_y,I_z)\) and buoyancy \(B=m_tg_0\),
\begin{align}
  m_t\dot v_b^{\mathrm{vec}}
     &=F-m_t\,\omega\times v_b^{\mathrm{vec}}-D_v,\\
  I\dot\omega
     &=\tau-\omega\times(I\omega)-D_\omega
       -h_{\mathrm{cb}}B
       \begin{bmatrix}\cos\theta\sin\phi\\\sin\theta\\0\end{bmatrix}.
\end{align}
The published damping vectors are \(d^\ell=(3,3,3,0.3,0.5,0.5)\) and \(d^q=(5,5,5,0.3,1,1)\). Their entries use the corresponding SI force and moment units. The model assumes neutral buoyancy, isotropic translational mass, and principal-axis inertia. It excludes the Euler-angle singularity \(\cos\theta=0\).

\paragraph{Quadrotor.}
The state is \(x=(p^\top,q^\top,v^\top,\omega^\top)^\top\), with scalar-first unit quaternion \(q=(q_w,q_x,q_y,q_z)^\top\). The inputs are total thrust \(F\) and the three angular-acceleration contributions \((\alpha_x,\alpha_y,\alpha_z)\). Define
\begin{equation}
  R(q)e_3=
  \begin{bmatrix}
    2(q_wq_y+q_xq_z)\\
    2(-q_wq_x+q_yq_z)\\
    1-2(q_x^2+q_y^2)
  \end{bmatrix}.
\end{equation}
The dynamics are
\begin{align}
  \dot p&=v,\qquad
  \dot v=-g_0e_3+\frac{C_TF}{m}R(q)e_3,\\
  \dot q&=\frac12
  \begin{bmatrix}
    -q_x&-q_y&-q_z\\
    q_w&-q_z&q_y\\
    q_z&q_w&-q_x\\
    -q_y&q_x&q_w
  \end{bmatrix}\omega,\\
  \dot\omega_x&=\alpha_x-\frac59\omega_y\omega_z,\qquad
  \dot\omega_y=\alpha_y+\frac59\omega_x\omega_z,\qquad
  \dot\omega_z=\alpha_z.
\end{align}
The angular inputs describe the controlled contributions after inertia scaling. The full angular acceleration also includes the displayed gyroscopic terms. The benchmark uses independent bounds on total thrust and these inputs, without a rotor-allocation constraint.

\section{Label-Generation Settings}
\label{sec:app-vertex-cbf-data}

\Cref{tab:app-vertex-cbf-data} gives the settings of the finite-horizon tree search of \cref{subsec:vertex-cbf-search}. The columns correspond to the three terms of the error decomposition \cref{eq:vertex-cbf-error-decomposition}. The step size \(\Delta t\) and the horizon length \(K\) determine the first two terms, and the time horizon \(K\Delta t\) is listed separately because it is the quantity that must be long enough for an unavoidable violation to appear. The beam width \(B\) and the search variant determine the search-truncation term. No system was solved by complete enumeration, so \cref{eq:vertex-cbf-beam-bound} applies throughout, and every label is a lower bound on the finite-horizon vertex-restricted value.

The supervision states \(\{x_j\}_{j=1}^{N_{\mathrm{data}}}\) are drawn uniformly from \(\mathcal{X}\), except on the three two-dimensional systems, where a regular lattice was used because it also serves the visualizations of \cref{fig:vertex-cbf-comparison-ip,fig:vertex-cbf-ablation-mse}. The generation time \(T_{\mathrm{data}}\) covers all supervision states, which are processed in GPU batches.

\begin{table}[tb]
  \caption[Label-generation settings for the 15 evaluated systems]{Label-generation settings. \(N_{\mathrm{data}}\) is the number of supervision states, \(K\) the horizon length in steps, \(\Delta t\) the step size, \(K\Delta t\) the resulting time horizon, and \(B\) the beam width. \(T_{\mathrm{data}}\) is the wall-clock time for the whole batch. \enquote{Extra} lists the parameters of the stochastic and branch-and-bound variants of \cref{subsec:vertex-cbf-tree-methods}.}
  \label{tab:app-vertex-cbf-data}
  \centering
  \scriptsize
  \setlength{\tabcolsep}{2pt}
  \begin{tabular*}{\textwidth}{@{\extracolsep{\fill}}l l r r r r r R{0.14\textwidth} r@{}}
    \toprule
    System & Search variant & \(N_{\mathrm{data}}\) & \(K\) & \(\Delta t\)~[\si{\second}] & \(K\Delta t\)~[\si{\second}] & \(B\) & Extra & \(T_{\mathrm{data}}\) \\
    \midrule
    Inverted Pendulum      & Beam            & \num{3600} (\(60{\times}60\))  & 20  & 0.10  & 2.0 & \num{500}  & ---                          & \SI{1}{\second} \\
    1D Double Integrator & Stochastic beam & \num{2500} (\(50{\times}50\))  & 40  & 0.10  & 4.0 & \num{1500} & softmax, \(T{=}0.05\)        & \SI{1}{\second} \\
    2D Vertical Drone    & Branch and bound& \num{6400} (\(80{\times}80\))  & 60  & 0.05  & 3.0 & \num{2000} & 2 passes                     & \SI{5}{\second} \\
    Dubins Car             & Beam            & \num{50000}                    & 150 & 0.05  & 7.5 & \num{1000} & ---                          & \SI{27}{\second} \\
    2D Double Integrator & Branch and bound& \num{300000}                   & 100 & 0.05  & 5.0 & \num{1000} & 3 passes                     & \SI{8}{\minute}\,\SI{2}{\second} \\
    Kinematic Bicycle      & Beam            & \num{200000}                   & 80  & 0.10  & 8.0 & \num{1000} & ---                          & \SI{2}{\minute}\,\SI{24}{\second} \\
    Cart-Pole              & Branch and bound& \num{400000}                   & 50  & 0.05  & 2.5 & \num{500}  & 2 passes                     & \SI{1}{\minute}\,\SI{40}{\second} \\
    Dynamic Unicycle       & Beam            & \num{600000}                   & 50  & 0.10  & 5.0 & \num{1000} & ---                          & \SI{3}{\second} \\
    Relative Unicycle      & Beam            & \num{1000000}                  & 50  & 0.10  & 5.0 & \num{1000} & ---                          & \SI{6}{\minute}\,\SI{26}{\second} \\
    3D Double Integrator & Stochastic beam & \num{400000}                   & 80  & 0.05  & 4.0 & \num{600}  & Gumbel top-\(B\), \(T{=}0.05\) & \SI{4}{\minute}\,\SI{12}{\second} \\
    3-DOF Manipulator      & Beam            & \num{600000}                   & 60  & 0.05  & 3.0 & \num{1000} & ---                          & \SI{28}{\minute}\,\SI{49}{\second} \\
    Landing Rocket         & Stochastic beam & \num{600000}                   & 40  & 0.10  & 4.0 & \num{1500} & Gumbel top-\(B\), \(T{=}0.2\)  & \SI{6}{\minute}\,\SI{55}{\second} \\
    Quadruped Trunk        & Stochastic beam & \num{600000}                   & 100 & 0.02  & 2.0 & \num{1000} & Gumbel top-\(B\), \(T{=}0.05\) & \SI{1}{\hour}\,\SI{41}{\minute}\,\SI{32}{\second} \\
    6-DOF Underwater Veh.  & Stochastic beam & \num{800000}                   & 40  & 0.10  & 4.0 & \num{800}  & softmax, \(T{=}0.08\)        & \SI{3}{\hour}\,\SI{7}{\minute}\,\SI{6}{\second} \\
    Quadrotor              & Branch and bound& \num{1000000}                  & 30  & 0.10  & 3.0 & \num{150}  & 2 passes                     & \SI{24}{\minute}\,\SI{41}{\second} \\
    \bottomrule
  \end{tabular*}
\end{table}

Two observations follow from the table and support the discussion of \cref{subsec:vertex-cbf-ablation-search}. The first is that the generation time is not ordered by state dimension. The quadruped trunk, with nine states, is the most expensive entry after the underwater vehicle, and the 13-state quadrotor is cheaper than the 6-state manipulator. The cost is driven by the product \(N_{\mathrm{data}}KBM\) of \cref{subsec:vertex-cbf-tree-methods}, in which the vertex count \(M=2^{n_u}\) enters through the input dimension. The second is that the beam width can be reduced to control the cost of a large vertex count. Several systems with \(n_u\leq3\) use \(B=\num{1000}\), compared with \(B=\num{800}\) for the underwater vehicle and \(B=\num{150}\) for the quadrotor. This is the practical form of the trade-off described in \cref{sec:vertex-cbf-discussion}.

\section{Network Architecture and Training Settings}
\label{sec:app-vertex-cbf-training}

\Cref{tab:app-vertex-cbf-training} gives the network and the training schedule. All systems share the residual parameterization \cref{eq:vertex-cbf-residual}, sinusoidal hidden activations, a softplus output, the Adam optimizer, an initial learning rate of \(10^{-3}\), and \num{10000} epochs. They differ in the width and depth of the residual network, in the epoch at which the learning rate is reduced by a factor of ten, in the number of collocation states \(N_{\mathrm{PDE}}\), and in the loss weight \(\lambda\) of \cref{eq:vertex-cbf-total-loss}. Collocation states are drawn uniformly from \(\mathcal{X}\) and are independent of the supervision states, which is why \(N_{\mathrm{PDE}}\) and \(N_{\mathrm{data}}\) differ on most systems. The queried networks are small, between about \num{3000} and \num{70000} parameters, which allows the CBF-QP to evaluate the barrier and its gradient at \SI{250}{\hertz}.

\begin{table}[tb]
  \caption[Architecture and training configuration for the 15 evaluated systems]{Architecture and training configuration. \enquote{Hidden} lists the hidden-layer widths of the residual network \(\hat{r}_{\Theta}\), all with sinusoidal activations, and \(\beta\) is the softplus sharpness. \(N_{\mathrm{PDE}}\) is the number of collocation states, and \(\lambda\) is the weight of the physics-informed term in \cref{eq:vertex-cbf-total-loss}. All runs use Adam, an initial learning rate of \(10^{-3}\), and \num{10000} epochs, with the learning rate reduced by a factor of ten at the listed epoch.}
  \label{tab:app-vertex-cbf-training}
  \centering
  \footnotesize
  \setlength{\tabcolsep}{3pt}
  \begin{tabular*}{\textwidth}{@{\extracolsep{\fill}}l l c r c r r@{}}
    \toprule
    System & Hidden & \(\beta\) & \(N_{\mathrm{PDE}}\) & \(\lambda\) & LR drop & \(T_{\mathrm{train}}\) \\
    \midrule
    Inverted Pendulum      & \([32,32,32,32,32]\)                 & 10 & \num{10000}   & 0.20   & \num{7000} & \SI{1}{\minute}\,\SI{0}{\second} \\
    1D Double Integrator & \([32,32,32,32]\)                 & 1  & \num{10000}   & 0.90   & \num{8000} & \SI{51}{\second} \\
    2D Vertical Drone    & \([32,32,32,32]\)                 & 10 & \num{10000}   & 0.55   & \num{7000} & \SI{52}{\second} \\
    Dubins Car             & \([32,32,32,32]\)                 & 10 & \num{200000}  & 0.80   & \num{8000} & \SI{2}{\minute}\,\SI{15}{\second} \\
    2D Double Integrator & \([32,32,32,32,32]\)                 & 10 & \num{600000}  & norm.\ & \num{7000} & \SI{6}{\minute}\,\SI{46}{\second} \\
    Kinematic Bicycle      & \([32,64,64,32]\)                   & 10 & \num{500000}  & 0.15   & \num{7000} & \SI{6}{\minute}\,\SI{3}{\second} \\
    Cart-Pole              & \([32,64,64,64,32]\)                & 10 & \num{800000}  & 0.00   & \num{8000} & \SI{2}{\minute}\,\SI{14}{\second} \\
    Dynamic Unicycle       & \([64,64,64,64]\)                 & 10 & \num{1000000} & 0.60   & \num{7000} & \SI{15}{\minute}\,\SI{44}{\second} \\
    Relative Unicycle      & \([32,32,32,32,32]\)                 & 10 & \num{2000000} & 0.30   & \num{8000} & \SI{22}{\minute}\,\SI{2}{\second} \\
    3D Double Integrator & \([64,64,64,64]\)                 & 10 & \num{600000}  & norm.\ & \num{8000} & \SI{9}{\minute}\,\SI{40}{\second} \\
    3-DOF Manipulator      & \([64,64,64,64]\)                 & 10 & \num{600000}  & 0.00   & \num{8000} & \SI{3}{\minute}\,\SI{17}{\second} \\
    Landing Rocket         & \([84,84,84,84]\)                 & 10 & \num{800000}  & 0.20   & \num{8000} & \SI{18}{\minute}\,\SI{2}{\second} \\
    Quadruped Trunk        & \([84,84,84,84,84]\)                 & 10 & \num{600000}  & 0.00   & \num{8000} & \SI{5}{\minute}\,\SI{28}{\second} \\
    6-DOF Underwater Veh.  & \([128,128,128]\)                & 10 & \num{800000}  & 0.00   & \num{8000} & \SI{4}{\minute}\,\SI{58}{\second} \\
    Quadrotor              & \([128,128,128,128]\)                & 10 & \num{2000000} & 0.15   & \num{9000} & \SI{54}{\minute}\,\SI{8}{\second} \\
    \bottomrule
  \end{tabular*}
\end{table}

Two kinds of entry in the \(\lambda\) column require comment. The entry \enquote{norm.} marks the two runs in which \(\lambda\) was not chosen by hand. Both loss terms are evaluated over one epoch, and the two measured magnitudes are then used as fixed scale factors for the remainder of training, to balance their loss scales. Equal loss scales do not necessarily imply equal gradient magnitudes. The four entries with \(\lambda=0\) mark the systems on which the reported models were trained on the data term alone, which are the cart-pole, the manipulator, the quadruped trunk, and the underwater vehicle. This does not affect the comparison of \cref{tab:vertex-cbf-results}, since the two supervised settings share the listed \(\lambda\) within each system and the third setting has no data term at all. The remaining values were selected per system inside the range in which the validation metrics are flat, as \cref{subsec:vertex-cbf-ablation-lambda} describes for the inverted pendulum.

\section{Validation Settings}
\label{sec:app-vertex-cbf-validation}

\Cref{tab:app-vertex-cbf-validation} gives the settings of the protocol of \cref{subsec:vertex-cbf-validation}. Initial states are drawn from \(\mathcal{X}\) by stratified rejection sampling so that roughly half lie in \(\hat{\mathcal{S}}_{\Theta}\), and each is simulated for the horizon \(T_{\mathrm{valid}}\) under the Hamiltonian-maximizing policy \cref{eq:vertex-cbf-validation-policy}. The trajectory is integrated at the listed step size, which is between five and twenty-five times smaller than the step size used for label generation. The reported metrics are means and standard deviations over the training seeds.

\begin{table}[tb]
  \caption[Validation configuration for the 15 evaluated systems]{Validation configuration. \(N_{\mathrm{valid}}\) is the number of sampled initial states, \(T_{\mathrm{valid}}\) the rollout horizon, and \(\Delta t_{\mathrm{valid}}\) the step size of the rollouts. The last column is the wall-clock time of the whole validation batch.}
  \label{tab:app-vertex-cbf-validation}
  \centering
  \footnotesize
  \setlength{\tabcolsep}{3pt}
  \begin{tabular*}{\textwidth}{@{\extracolsep{\fill}}l r r r r@{}}
    \toprule
    System & \(N_{\mathrm{valid}}\) & \(T_{\mathrm{valid}}\)~[\si{\second}] & \(\Delta t_{\mathrm{valid}}\)~[\si{\second}] & Wall-clock \\
    \midrule
    Inverted Pendulum      & \num{20000}   & 5.0 & 0.010 & \SI{1}{\second} \\
    1D Double Integrator & \num{20000}   & 4.0 & 0.010 & \SI{1}{\second} \\
    2D Vertical Drone    & \num{20000}   & 5.0 & 0.010 & \SI{1}{\second} \\
    Dubins Car             & \num{200000}  & 5.0 & 0.010 & \SI{3}{\second} \\
    2D Double Integrator & \num{1000000} & 5.0 & 0.010 & \SI{12}{\second} \\
    Kinematic Bicycle      & \num{500000}  & 5.0 & 0.010 & \SI{6}{\second} \\
    Cart-Pole              & \num{1000000} & 3.0 & 0.002 & \SI{51}{\second} \\
    Dynamic Unicycle       & \num{1000000} & 5.0 & 0.010 & \SI{17}{\second} \\
    Relative Unicycle      & \num{1000000} & 5.0 & 0.005 & \SI{28}{\second} \\
    3D Double Integrator & \num{1000000} & 5.0 & 0.005 & \SI{33}{\second} \\
    3-DOF Manipulator      & \num{1000000} & 4.0 & 0.005 & \SI{35}{\second} \\
    Landing Rocket         & \num{1000000} & 5.0 & 0.005 & \SI{47}{\second} \\
    Quadruped Trunk        & \num{1000000} & 3.0 & 0.002 & \SI{1}{\minute}\,\SI{38}{\second} \\
    6-DOF Underwater Veh.  & \num{1000000} & 4.0 & 0.005 & \SI{52}{\second} \\
    Quadrotor              & \num{2000000} & 4.0 & 0.010 & \SI{55}{\second} \\
    \bottomrule
  \end{tabular*}
\end{table}

\paragraph{Estimating the predicted-safe volume.}
The ratio \(\operatorname{vol}(\hat{\mathcal{S}}_{\Theta})/\operatorname{vol}(\mathcal{X})\) in \cref{eq:vertex-cbf-eta-eff} is a Monte Carlo estimate and not a geometric computation. The stratified rejection sampling that produces the validation states draws uniform proposals from the box \(\mathcal{X}\) and classifies each by the sign of \(\hat{V}_{\Theta}\), so the acceptance rate of the predicted-safe stratum is the required fraction and the estimate is available at no additional cost. For the quadrotor, the proposals instead come from the product sampling of \cref{sec:app-vertex-cbf-systems}. The number of proposals is at least \(N_{\mathrm{valid}}\), so the standard error of the estimate is below \(0.4\) percentage points for the smallest batch in \cref{tab:app-vertex-cbf-validation} and below \(0.05\) percentage points for the largest. Two consequences follow. The ratio is a fraction of the chosen box rather than an intrinsic quantity, so changing the box changes the ratio without changing the barrier, which partly explains the small values of the underactuated systems. The factor \(1-\rho_{\mathrm{FS}}\) is measured only over the finite horizon \(T_{\mathrm{valid}}\), so \(\eta_{\mathrm{eff}}\) inherits the finite-horizon caveat stated in \cref{subsec:vertex-cbf-validation}.

\begin{figure}[p]
  \centering
  \includegraphics[width=\textwidth]{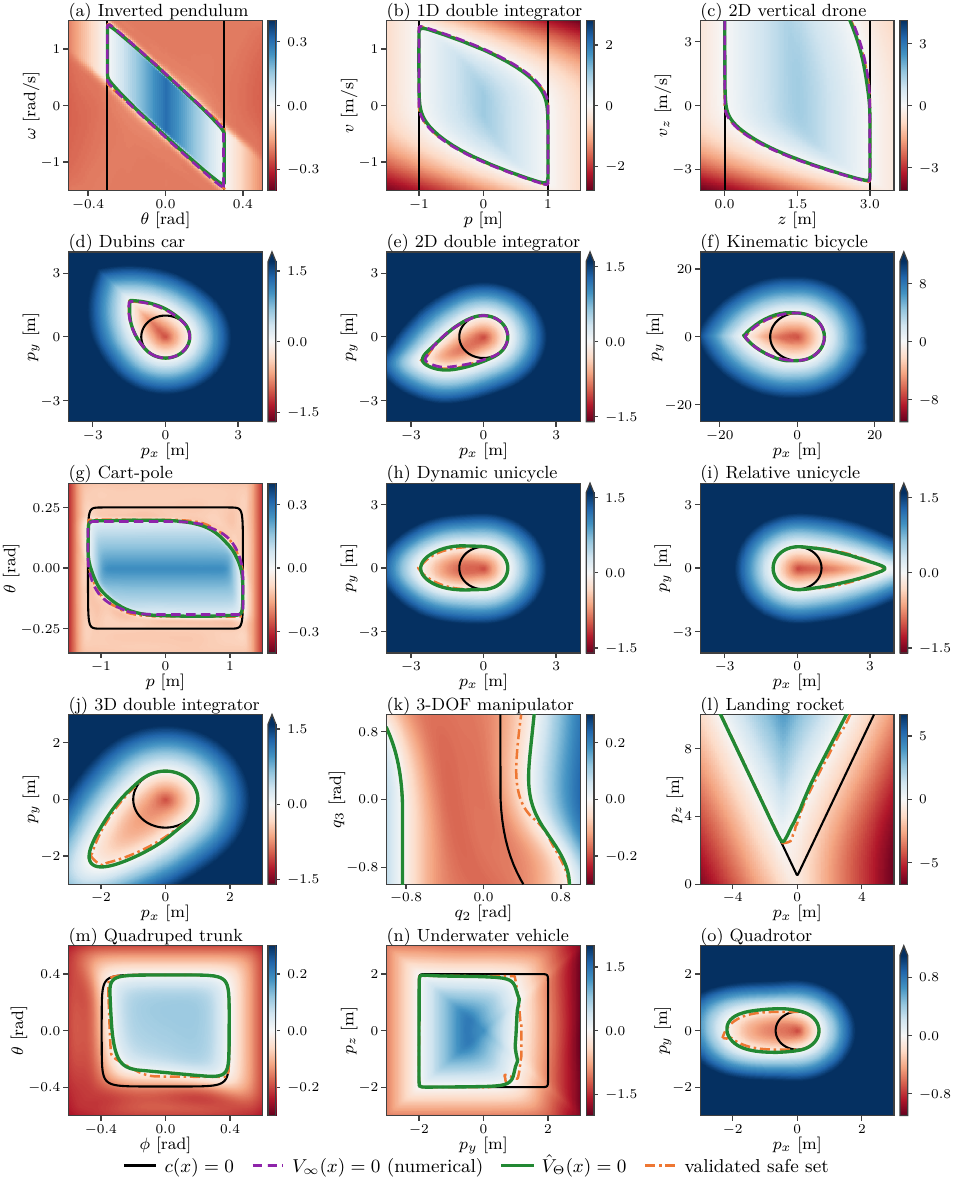}
  \caption[Two-dimensional slices of the learned barriers for the 15 evaluated systems]{Two-dimensional slices of the learned barriers \(\hat{V}_{\Theta}\) for the 15 evaluated systems under vertex-restricted supervision (first seed), in the order of \cref{tab:vertex-cbf-results}, with red shading marking the states predicted to be unsafe. The black line is the constraint boundary, the green line the boundary of the predicted safe set, the orange dash-dotted line the boundary of the set that the rollouts of \cref{subsec:vertex-cbf-validation} validate as safe within the slice, and the purple dashed line the boundary of the numerically computed maximal safe set where a numerical solution is available.}
  \label{fig:app-vertex-cbf-slices}
\end{figure}

\paragraph{Learned barriers.}
\Cref{fig:app-vertex-cbf-slices} shows a two-dimensional slice of the learned barrier for every system, obtained with vertex-restricted supervision. The plotted model is the first of the five seeded runs that enter \cref{tab:vertex-cbf-results}. Each slice plots the barrier over two state components and fixes the remaining ones. The Dubins car is shown at \(\theta=-0.84\), the 2D double integrator at \((v_x,v_y)=(2,1)\), the kinematic bicycle at \(\psi=0\) and \(v=20\), the cart-pole at \(\dot{p}=\dot{\theta}=0\), and the dynamic unicycle at \(\theta=0\), \(v=2\), and \(\omega=0\). The relative unicycle is shown at \(\psi=\pi\) and \(v_{\mathrm{r}}=v_{\mathrm{p}}=1\), the 3D double integrator at \(p_z=0\) and \(v=(2,2,0)\), the manipulator at \(q_1=0.25\) and \(\dot{q}=(0,-5,0)\), and the landing rocket at \(v=(1,-6)\), \(\theta=-0.2\), \(\omega=0\), and \(m=0.9\). The quadruped trunk is shown at \(p_z=0.3\), \(v=0\), and \((p,q,r)=(-4,-4,0)\), the underwater vehicle at \(p_x=0\), \((\phi,\theta,\psi)=(0,-1,1)\), and \(\nu=(1.8,0,-1.8,0,0,0)\), and the quadrotor at \(p_z=0\), \(q=(0,0,0,1)\) (a yaw of \(\pi\) under the scalar-first convention), \(v=(4,0,0)\), and \(\omega=0\).
The black line is the constraint boundary, the green line is the boundary of the predicted safe set, and the orange dash-dotted line is the boundary of the set that the rollouts of \cref{subsec:vertex-cbf-validation} validate as safe within the slice. The proximity of the last two lines is the visual counterpart of the low false-safe and false-unsafe rates in \cref{tab:vertex-cbf-results}, and the predicted safe set lies inside the constraint set on every slice, as \cref{eq:vertex-cbf-exclusion} requires.

\section{Hardware Deployment}
\label{sec:app-vertex-cbf-hardware}

The hardware study of \cref{subsec:vertex-cbf-hardware} deploys the relative-unicycle barrier without retraining, so its offline configuration is the relative-unicycle row of every table in this appendix. That subsection describes the platform, the filter, and the composition, and it states how the deployed pairwise model differs from the one used in \cref{ch:cn-cbf}. Two quantities of the deployed model are given here. The robot and obstacle radii are \(r_{\mathrm{r}}=r_{\mathrm{o}}=\SI{0.5}{\meter}\), which is \SI{5}{\centi\meter} more than the physical dimensions of the robot and of a pedestrian, and their sum plays the role of \(R_{\min}\) in \cref{ch:cn-cbf}. The pedestrian speed \(v_{\mathrm{p}}\) is carried as a state over \([0,1.5]\)~\si{\meter\per\second} rather than fixed at deployment.

\backmatter

\printbibliography

\end{document}